%% file: main.tex
\documentclass[11pt]{article}

\usepackage[margin=1in]{geometry}
\usepackage{graphicx}
\usepackage{booktabs}
\usepackage{array}
\usepackage{amsmath}
\usepackage{amssymb}
\usepackage{microtype}
\usepackage{xcolor}
\usepackage[hidelinks]{hyperref}
\usepackage{natbib}

\makeatletter
\newcommand\FloatBarrier{\par\begingroup\let\@elt\relax
  \edef\@tempa{\@botlist\@deferlist\@dbldeferlist}%
  \ifx\@tempa\@empty\else\clearpage\fi\endgroup}
\makeatother

\graphicspath{{figures/}}

\input{numbers}

\definecolor{revcolor}{rgb}{0.0,0.25,0.75}

\newcommand{\condA}{A}
\newcommand{\condBn}{B-neutral}
\newcommand{\condBw}{B-walls}
\newcommand{\condAp}{A$'$}
\newcommand{\condApp}{A$''$}

\title{Evaluating Language Models as Spatial Priors\\
for Object-Goal Navigation}
\author{Jun Miura\\
Department of Computer Science and Engineering,\\
Toyohashi University of Technology, Japan\\
\texttt{jun.miura@tut.jp}}
\date{}

\begin{document}
\maketitle

\begin{abstract}
Semantic exploration methods for object-goal navigation use large language models (LLMs) to score how strongly a target object is associated with a room type or with a nearby object. This paper considers obtaining from the LLM a spatial prior instead: a distribution over the position of the target in a room. We evaluate whether two open-weight LLMs, Qwen3.5-27B and Gemma-4-31B-it, can provide such a prior. Using real room layouts from CubiCasa5K and ScanNet, we measure two outputs of the models: the likelihood they assign to real and perturbed layouts, and the layouts they generate. The object frequencies and room sizes of the generated rooms follow the room type, as in real rooms, but the proximity of the objects to the walls does not follow the differences between room types found in real rooms. The likelihood distinguishes real layouts from perturbed ones partly through their geometry and partly through the agreement between the coordinates and a field that states the wall each object touches; for the movable furniture of ScanNet, it is higher for layouts with all objects piled near the room center than for the real layouts in most rooms. When an object is given in the prompt, the models place an object related to it relative to it and move other objects away from it. When the wall of each object is given in the prompt, the models place most objects against that wall, but they do not keep all objects inside a given room rectangle. As a prior for object search, the generated placements reduce the searched area relative to a uniform prior, but by less than a third of the reduction obtained with a prior built from real layouts; when both priors are spatially smoothed, the fraction remains below two thirds, and the real-layout prior gives the lower search cost in every room type. These results support the use of LLMs for object--room associations and for writing coordinates under constraints determined by another component, followed by a geometric check, but not as a source of placement priors where real layouts are available.
\end{abstract}

\input{sections/introduction}

\input{sections/related}
\input{sections/setup}

\input{sections/discrimination}
\input{sections/generation}
\input{sections/mechanism}
\input{sections/search}
\input{sections/design}
\input{sections/limitations}
\input{sections/conclusion}

\bibliographystyle{plainnat}
\bibliography{refs}

\appendix
\setcounter{figure}{0}\setcounter{table}{0}
\renewcommand\thefigure{A\arabic{figure}}
\renewcommand\thetable{A\arabic{table}}
\renewcommand\theHfigure{A\arabic{figure}}%
\renewcommand\theHtable{A\arabic{table}}
\input{sections/appendix_measurement}

\input{prompts}
\input{sections/appendix_schemas}

\end{document}

%% file: numbers.tex
\newcommand{\AdsAdsVsOrigBathGemma}{0.075}

\newcommand{\AdsAdsVsOrigBath}{0.015}

\newcommand{\AdsAdsVsOrigKitchenGemma}{0.025}
\newcommand{\AdsAdsVsOrigKitchenInv}{0.915}
\newcommand{\AdsAdsVsOrigKitchen}{0.085}
\newcommand{\AdsAdsVsRandBathGemma}{0.990}
\newcommand{\AdsAdsVsRandBath}{0.995}
\newcommand{\AdsAdsVsRandKitchenGemma}{1.000}
\newcommand{\AdsAdsVsRandKitchen}{1.000}
\newcommand{\AdsAdsVsWrongWallBathGemma}{0.980}
\newcommand{\AdsAdsVsWrongWallBath}{0.910}
\newcommand{\AdsAdsVsWrongWallKitchenGemma}{1.000}
\newcommand{\AdsAdsVsWrongWallKitchen}{0.940}
\newcommand{\AdsDAdsVsOrigBathGemma}{$-$23.929}
\newcommand{\AdsDAdsVsOrigBath}{$-$14.282}
\newcommand{\AdsDAdsVsOrigKitchenGemma}{$-$21.955}
\newcommand{\AdsDAdsVsOrigKitchen}{$-$13.180}
\newcommand{\AdsDAdsVsRandBathGemma}{$+$45.675}
\newcommand{\AdsDAdsVsRandBath}{$+$26.285}
\newcommand{\AdsDAdsVsRandKitchenGemma}{$+$66.236}
\newcommand{\AdsDAdsVsRandKitchen}{$+$38.766}
\newcommand{\AdsDAdsVsWrongWallBathGemma}{$+$62.376}
\newcommand{\AdsDAdsVsWrongWallBath}{$+$13.849}
\newcommand{\AdsDAdsVsWrongWallKitchenGemma}{$+$102.123}
\newcommand{\AdsDAdsVsWrongWallKitchen}{$+$21.969}
\newcommand{\AdsDRandVsOrigBathGemma}{$-$69.604}
\newcommand{\AdsDRandVsOrigBath}{$-$40.567}
\newcommand{\AdsDRandVsOrigKitchenGemma}{$-$88.191}
\newcommand{\AdsDRandVsOrigKitchen}{$-$51.947}
\newcommand{\AdsDWrongWallVsRandBathGemma}{$-$16.701}
\newcommand{\AdsDWrongWallVsRandBath}{$+$12.436}
\newcommand{\AdsDWrongWallVsRandKitchenGemma}{$-$35.887}
\newcommand{\AdsDWrongWallVsRandKitchen}{$+$16.797}
\newcommand{\AdsDeltaAdsVsOrigBathGemma}{23.9}
\newcommand{\AdsDeltaAdsVsOrigBath}{14.3}
\newcommand{\AdsDeltaAdsVsOrigKitchenGemma}{22.0}
\newcommand{\AdsDeltaAdsVsOrigKitchen}{13.2}

\newcommand{\AdsDeltaAdsVsRandKitchenGemma}{66.2}
\newcommand{\AdsDeltaAdsVsRandKitchen}{38.8}

\newcommand{\AdsDeltaRandVsOrigKitchenGemma}{88.2}
\newcommand{\AdsDeltaRandVsOrigKitchen}{51.9}

\newcommand{\AdsDistinctAdsBathGemma}{49}
\newcommand{\AdsDistinctAdsBath}{126}
\newcommand{\AdsDistinctAdsKitchenGemma}{64}
\newcommand{\AdsDistinctAdsKitchen}{154}
\newcommand{\AdsDistinctOrigBathGemma}{37}
\newcommand{\AdsDistinctOrigBath}{90}
\newcommand{\AdsDistinctOrigKitchenGemma}{51}
\newcommand{\AdsDistinctOrigKitchen}{107}
\newcommand{\AdsDistinctRandBathGemma}{60}

\newcommand{\AdsDistinctRandKitchenGemma}{110}
\newcommand{\AdsDistinctRandKitchen}{180}

\newcommand{\AdsDistinctWrongKitchenGemma}{70}
\newcommand{\AdsDistinctWrongKitchen}{149}
\newcommand{\AdsGrid}{0.05}

\newcommand{\AdsMoveMeanKitchenGemma}{0.19}
\newcommand{\AdsMoveMeanKitchen}{1.36}

\newcommand{\AdsMoveMedianKitchenGemma}{0.20}
\newcommand{\AdsMoveMedianKitchen}{0.75}

\newcommand{\AdsNValuesKitchenGemma}{6,304}
\newcommand{\AdsNValuesKitchen}{7,036}

\newcommand{\AdsNtokKitchenGemma}{393.0}
\newcommand{\AdsNtokKitchen}{372.5}
\newcommand{\AdsRandVsOrigBathGemma}{0.000}
\newcommand{\AdsRandVsOrigBath}{0.000}
\newcommand{\AdsRandVsOrigKitchenGemma}{0.000}
\newcommand{\AdsRandVsOrigKitchen}{0.000}
\newcommand{\AdsWrongWallVsRandBathGemma}{0.305}
\newcommand{\AdsWrongWallVsRandBath}{0.860}
\newcommand{\AdsWrongWallVsRandKitchenGemma}{0.160}
\newcommand{\AdsWrongWallVsRandKitchen}{0.870}

\newcommand{\AppFeasInfeasibleQwen}{1}

\newcommand{\AppFeasObjectsQwen}{3,549}

\newcommand{\AppFeasRoomsQwen}{565}

\newcommand{\AppInEligibleBed}{237}

\newcommand{\AppInEligibleTotal}{602}
\newcommand{\AppInMaxObj}{14}
\newcommand{\AppInMax}{200}
\newcommand{\AppInMinObj}{2}
\newcommand{\AppInNoWallPct}{48}

\newcommand{\AppInTol}{0.15}

\newcommand{\AppInUsedBed}{200}

\newcommand{\AppInUsedTotal}{565}
\newcommand{\AreaPearsonGemma}{0.86}
\newcommand{\AreaPearsonQwen}{0.86}
\newcommand{\AreaSDRatioGemmaA}{1.09}

\newcommand{\AreaSDRatioQwenA}{1.50}

\newcommand{\AreaSlopeGemma}{0.93}
\newcommand{\AreaSlopeQwen}{1.30}
\newcommand{\BandCovBathFiveZero}{0.71}

\newcommand{\BandDiffBathFiveZero}{$-$0.081}
\newcommand{\BandDiffBathOneFive}{$+$0.149}

\newcommand{\BandDiffBedFiveZero}{$-$0.021}
\newcommand{\BandDiffBedOneFive}{$+$0.070}

\newcommand{\BandExcessBathFiveZero}{$+$0.163}
\newcommand{\BandExcessBathOneFive}{$-$0.298}

\newcommand{\BandExcessBedFiveZero}{$+$0.043}

\newcommand{\BandExcessLivingFiveZero}{$-$0.086}

\newcommand{\BandMaxErr}{0.000}
\newcommand{\BandM}{0.15}
\newcommand{\CanonRateBath}{0.90}
\newcommand{\CanonRateBed}{0.88}
\newcommand{\CanonRateGenMin}{0.99}
\newcommand{\CanonRateLiving}{0.42}
\newcommand{\CanonRateOffice}{0.84}
\newcommand{\CatOriMinObjects}{60}

\newcommand{\CatOriNQwen}{16}

\newcommand{\CatOriWinQwen}{3}

\newcommand{\ClaimFalseDetachBath}{96.9}

\newcommand{\ClaimFalseDetachKitchen}{95.1}

\newcommand{\ClaimFalseNormalOutBath}{1.7}

\newcommand{\ClaimFalseNormalOutKitchen}{1.3}

\newcommand{\ClaimFalsePileBath}{95.6}

\newcommand{\ClaimFalsePileKitchen}{95.2}
\newcommand{\ClaimFalsePileLiving}{99.7}

\newcommand{\ClaimFalseRefBath}{0.0}

\newcommand{\ClaimFalseRefKitchen}{0.0}

\newcommand{\ClaimFalseTangentKitchen}{0.0}

\newcommand{\ClaimTol}{0.15}
\newcommand{\CondOverRealGemmaAp}{4}

\newcommand{\CondRhoGemmaAp}{$+$0.400}

\newcommand{\CondRhoQwenAp}{$+$1.000}

\newcommand{\CondSlopeGemmaApCI}{[$+$0.435, $+$0.636]}
\newcommand{\CondSlopeGemmaAp}{$+$0.531}
\newcommand{\CondSlopeGemmaA}{$+$0.227}

\newcommand{\CondSlopeQwenApCI}{[$+$0.167, $+$0.381]}
\newcommand{\CondSlopeQwenAp}{$+$0.272}
\newcommand{\CondSlopeQwenA}{$+$0.029}

\newcommand{\CondSpanPctQwenAp}{28}

\newcommand{\ConsGapMedQwenA}{0.55}

\newcommand{\ConsGemmaAppi}{0.349}

\newcommand{\ConsGemmaApp}{0.989}
\newcommand{\ConsGemmaAp}{0.905}
\newcommand{\ConsGemmaA}{0.467}

\newcommand{\ConsNaturalQwen}{0.891}
\newcommand{\ConsOutsideGemmaAppc}{12}
\newcommand{\ConsOutsideGemmaAppi}{11}
\newcommand{\ConsOutsideGemmaAppper}{12}
\newcommand{\ConsOutsideGemmaAppr}{12}
\newcommand{\ConsOutsideGemmaApp}{13}

\newcommand{\ConsOutsideQwenAppc}{30}
\newcommand{\ConsOutsideQwenAppi}{26}
\newcommand{\ConsOutsideQwenAppper}{28}
\newcommand{\ConsOutsideQwenAppr}{31}
\newcommand{\ConsOutsideQwenApp}{32}
\newcommand{\ConsOutsideQwenAp}{15}
\newcommand{\ConsOutsideQwenA}{10}

\newcommand{\ConsQwenAppi}{0.225}

\newcommand{\ConsQwenApp}{0.896}
\newcommand{\ConsQwenAp}{0.429}
\newcommand{\ConsQwenA}{0.244}
\newcommand{\ConsRandGemmaAppi}{0.298}

\newcommand{\ConsRandGemmaA}{0.161}
\newcommand{\ConsRandQwenAppi}{0.193}

\newcommand{\ConsRandQwenApp}{0.300}
\newcommand{\ConsRandQwenAp}{0.125}
\newcommand{\ConsRandQwenA}{0.093}

\newcommand{\ConsUnnaturalQwen}{0.911}

\newcommand{\CoordDistinctModelKitchenGemma}{49}
\newcommand{\CoordDistinctModelKitchen}{109}

\newcommand{\CoordDistinctRealKitchen}{605}

\newcommand{\CoordGridModelKitchenGemma}{100}
\newcommand{\CoordGridModelKitchen}{100}

\newcommand{\CoordGridRealKitchen}{29}
\newcommand{\CostGemmaMABath}{0.433}
\newcommand{\CostGemmaMABed}{0.459}
\newcommand{\CostGemmaMALiving}{0.487}
\newcommand{\CostGemmaMAOffice}{0.463}
\newcommand{\CostGemmaMAcatBath}{0.396}
\newcommand{\CostGemmaMAcatBed}{0.429}
\newcommand{\CostGemmaMAcatLiving}{0.503}
\newcommand{\CostGemmaMAcatOffice}{0.501}
\newcommand{\CostGemmaMAroomBath}{0.437}
\newcommand{\CostGemmaMAroomBed}{0.442}
\newcommand{\CostGemmaMAroomLiving}{0.453}
\newcommand{\CostGemmaMAroomOffice}{0.425}
\newcommand{\CostGemmaMBwBath}{0.489}
\newcommand{\CostGemmaMBwBed}{0.493}
\newcommand{\CostGemmaMBwLiving}{0.512}
\newcommand{\CostGemmaMBwOffice}{0.485}

\newcommand{\CostHomogPGemmaMA}{\ensuremath{=0.002}}

\newcommand{\CostHomogPQwenMA}{\ensuremath{=0.063}}

\newcommand{\CostHomogPQwenRroom}{\ensuremath{<0.001}}

\newcommand{\CostMAGemmaMax}{0.487}
\newcommand{\CostMAGemmaMin}{0.433}
\newcommand{\CostMAQwenMax}{0.466}
\newcommand{\CostMAQwenMin}{0.425}

\newcommand{\CostQwenMABath}{0.425}
\newcommand{\CostQwenMABed}{0.466}
\newcommand{\CostQwenMALiving}{0.453}
\newcommand{\CostQwenMAOffice}{0.444}
\newcommand{\CostQwenMAcatBath}{0.390}
\newcommand{\CostQwenMAcatBed}{0.454}
\newcommand{\CostQwenMAcatLiving}{0.506}
\newcommand{\CostQwenMAcatOffice}{0.476}
\newcommand{\CostQwenMAroomBath}{0.417}
\newcommand{\CostQwenMAroomBed}{0.465}
\newcommand{\CostQwenMAroomLiving}{0.424}
\newcommand{\CostQwenMAroomOffice}{0.419}
\newcommand{\CostQwenMBwBath}{0.418}
\newcommand{\CostQwenMBwBed}{0.482}
\newcommand{\CostQwenMBwLiving}{0.502}
\newcommand{\CostQwenMBwOffice}{0.484}
\newcommand{\CostQwenRBath}{0.209}
\newcommand{\CostQwenRBed}{0.250}
\newcommand{\CostQwenRLiving}{0.347}
\newcommand{\CostQwenROffice}{0.278}
\newcommand{\CostQwenRcatBath}{0.228}
\newcommand{\CostQwenRcatBed}{0.318}
\newcommand{\CostQwenRcatLiving}{0.388}
\newcommand{\CostQwenRcatOffice}{0.321}
\newcommand{\CostQwenRroomBath}{0.208}
\newcommand{\CostQwenRroomBed}{0.263}
\newcommand{\CostQwenRroomLiving}{0.352}
\newcommand{\CostQwenRroomOffice}{0.286}

\newcommand{\CostQwenWBath}{0.651}
\newcommand{\CostQwenWBed}{0.572}
\newcommand{\CostQwenWLiving}{0.595}
\newcommand{\CostQwenWOffice}{0.579}
\newcommand{\CostRMax}{0.347}
\newcommand{\CostRMin}{0.209}

\newcommand{\CostUExpected}{0.502}
\newcommand{\CoverageMax}{0.96}
\newcommand{\CoverageMin}{0.81}
\newcommand{\CoverageTarget}{90}

\newcommand{\CubiDeltaDetachKitchen}{6.47}

\newcommand{\CubiDetachBath}{0.813}
\newcommand{\CubiDetachKitchen}{0.945}

\newcommand{\CubiGemmaDeltaDetachKitchen}{15.54}

\newcommand{\CubiGemmaDetachBath}{0.671}
\newcommand{\CubiGemmaDetachKitchen}{0.892}
\newcommand{\CubiGemmaPileBath}{0.862}
\newcommand{\CubiGemmaPileKitchen}{0.928}
\newcommand{\CubiGemmaSwapPosBath}{0.950}
\newcommand{\CubiGemmaSwapPosKitchen}{0.982}
\newcommand{\CubiGemmaSwapSizeBath}{0.879}
\newcommand{\CubiGemmaSwapSizeKitchen}{0.976}
\newcommand{\CubiOriAMaxS}{0.289}
\newcommand{\CubiOriAMinS}{0.247}
\newcommand{\CubiOriJSDMax}{0.057}
\newcommand{\CubiOriJSDMin}{0.039}
\newcommand{\CubiOriRealBathE}{0.424}

\newcommand{\CubiOriRealBathS}{0.080}

\newcommand{\CubiPileBath}{0.955}
\newcommand{\CubiPileKitchen}{0.955}
\newcommand{\CubiSwapPosBath}{0.977}
\newcommand{\CubiSwapPosKitchen}{0.994}
\newcommand{\CubiSwapSizeBath}{0.945}
\newcommand{\CubiSwapSizeKitchen}{0.996}

\newcommand{\DeclAgreeGemmaBw}{97}
\newcommand{\DeclAgreeQwenA}{64}

\newcommand{\DeclChanceQwenA}{26}

\newcommand{\DeclJSDQwenA}{0.0005}

\newcommand{\DetachFrac}{0.20}

\newcommand{\DiffMAUGemmaMax}{0.069}
\newcommand{\DiffMAUGemmaMin}{0.015}

\newcommand{\DiffMAUQwenMax}{0.077}
\newcommand{\DiffMAUQwenMin}{0.036}

\newcommand{\DiffMAcatMAGemmaBath}{$-$0.037}
\newcommand{\DiffMAcatMAGemmaBed}{$-$0.030}
\newcommand{\DiffMAcatMAGemmaLiving}{$+$0.016}
\newcommand{\DiffMAcatMAGemmaOffice}{$+$0.039}
\newcommand{\DiffMAcatMAQwenBath}{$-$0.035}
\newcommand{\DiffMAcatMAQwenBed}{$-$0.012}
\newcommand{\DiffMAcatMAQwenLiving}{$+$0.053}
\newcommand{\DiffMAcatMAQwenOffice}{$+$0.032}

\newcommand{\DiffMAroomMAGemmaBath}{$+$0.003}

\newcommand{\DiffMAroomMAGemmaLiving}{$-$0.034}
\newcommand{\DiffMAroomMAGemmaOffice}{$-$0.037}

\newcommand{\DiffMAroomMAQwenLiving}{$-$0.028}
\newcommand{\DiffMAroomMAQwenOffice}{$-$0.024}

\newcommand{\DiffMBwMAQwenBed}{$+$0.017}
\newcommand{\DiffMBwMAQwenLiving}{$+$0.049}
\newcommand{\DiffMBwMAQwenOffice}{$+$0.040}

\newcommand{\DiffRcatRQwenBath}{$+$0.019}
\newcommand{\DiffRcatRQwenBed}{$+$0.068}

\newcommand{\DiffRroomRQwenBath}{$-$0.001}
\newcommand{\DiffRroomRQwenBed}{$+$0.013}

\newcommand{\DirBathTwoZeroGap}{$+$0.040}

\newcommand{\DoorFoundBath}{189}
\newcommand{\DoorFoundBed}{240}
\newcommand{\DoorFoundLiving}{95}
\newcommand{\DoorFoundOffice}{146}
\newcommand{\DoorInferredBath}{65}
\newcommand{\DoorInferredBed}{84}
\newcommand{\DoorInferredLiving}{20}
\newcommand{\DoorInferredOffice}{39}
\newcommand{\EmptyShareMAGemma}{0.731}
\newcommand{\EmptyShareMAQwen}{0.522}
\newcommand{\EmptyShareMAroomGemma}{0.557}
\newcommand{\EmptyShareMAroomQwen}{0.285}
\newcommand{\EntGapQFiveBathGemma}{132.5}
\newcommand{\EntGapQFiveBath}{26.6}
\newcommand{\EntGapQFiveKitchenGemma}{217.4}
\newcommand{\EntGapQFiveKitchen}{51.6}
\newcommand{\EntGapQTenBathGemma}{97.4}
\newcommand{\EntGapQTenBath}{18.0}
\newcommand{\EntGapQTenKitchenGemma}{174.4}
\newcommand{\EntGapQTenKitchen}{35.4}
\newcommand{\EntGapQTwentyFiveBathGemma}{65.1}
\newcommand{\EntGapQTwentyFiveBath}{-2.3}
\newcommand{\EntGapQTwentyFiveKitchenGemma}{113.8}
\newcommand{\EntGapQTwentyFiveKitchen}{7.1}
\newcommand{\EntGapRawBathGemma}{170.1}
\newcommand{\EntGapRawBath}{56.9}
\newcommand{\EntGapRawKitchenGemma}{252.8}
\newcommand{\EntGapRawKitchen}{86.1}
\newcommand{\EntGrid}{0.05}
\newcommand{\ExcessGemmaABath}{$-$0.035}
\newcommand{\ExcessGemmaABed}{$-$0.190}
\newcommand{\ExcessGemmaALiving}{$-$0.185}
\newcommand{\ExcessGemmaAMaxAll}{$+$0.023}
\newcommand{\ExcessGemmaAMinAll}{$-$0.190}
\newcommand{\ExcessGemmaAOffice}{$+$0.023}
\newcommand{\ExcessGemmaBnBath}{$-$0.053}
\newcommand{\ExcessGemmaBnBed}{$-$0.313}
\newcommand{\ExcessGemmaBnLiving}{$-$0.257}

\newcommand{\ExcessGemmaBnOffice}{$-$0.155}
\newcommand{\ExcessGemmaBwBath}{$+$0.365}
\newcommand{\ExcessGemmaBwBed}{$+$0.051}
\newcommand{\ExcessGemmaBwLiving}{$+$0.443}

\newcommand{\ExcessGemmaBwOffice}{$+$0.326}

\newcommand{\ExcessQwenABath}{$-$0.121}
\newcommand{\ExcessQwenABed}{$-$0.093}
\newcommand{\ExcessQwenALiving}{$-$0.119}
\newcommand{\ExcessQwenAMaxAll}{$-$0.093}
\newcommand{\ExcessQwenAMinAll}{$-$0.121}
\newcommand{\ExcessQwenAOffice}{$-$0.113}
\newcommand{\ExcessQwenBnBath}{$-$0.017}
\newcommand{\ExcessQwenBnBed}{$-$0.145}
\newcommand{\ExcessQwenBnLiving}{$-$0.261}

\newcommand{\ExcessQwenBnOffice}{$-$0.153}
\newcommand{\ExcessQwenBwBath}{$+$0.182}
\newcommand{\ExcessQwenBwBed}{$+$0.059}
\newcommand{\ExcessQwenBwLiving}{$+$0.193}

\newcommand{\ExcessQwenBwOffice}{$+$0.148}
\newcommand{\ExcessRealBath}{$+$0.146}
\newcommand{\ExcessRealBed}{$+$0.092}
\newcommand{\ExcessRealLiving}{$-$0.163}
\newcommand{\ExcessRealOffice}{$-$0.018}
\newcommand{\FloorZMax}{0.30}

\newcommand{\GeoGapBathGemma}{$+$0.013}
\newcommand{\GeoGapBath}{$-$0.013}
\newcommand{\GeoGapKitchenGemma}{$+$0.188}
\newcommand{\GeoGapKitchen}{$+$0.181}

\newcommand{\GeoNormalInBathGemma}{0.746}

\newcommand{\GeoNormalInBath}{0.738}

\newcommand{\GeoNormalInKitchenGemma}{0.838}

\newcommand{\GeoNormalInKitchen}{0.910}

\newcommand{\GeoNormalOutBathGemma}{0.724}

\newcommand{\GeoNormalOutBath}{0.729}

\newcommand{\GeoNormalOutKitchenGemma}{0.832}

\newcommand{\GeoNormalOutKitchen}{0.893}

\newcommand{\GeoTangentBathGemma}{0.711}

\newcommand{\GeoTangentBath}{0.742}

\newcommand{\GeoTangentKitchenGemma}{0.644}

\newcommand{\GeoTangentKitchen}{0.712}
\newcommand{\GridN}{16}
\newcommand{\HomogPGemmaA}{\ensuremath{<0.001}}

\newcommand{\HomogPQwenA}{\ensuremath{=0.274}}

\newcommand{\IdentMismatchGemma}{$-$0.09}
\newcommand{\IdentMismatchQwen}{$-$0.03}
\newcommand{\IdentRhoGemmaA}{0.67}

\newcommand{\IdentRhoQwenA}{0.75}

\newcommand{\IsoClear}{0.30}
\newcommand{\IsoGapBathGemma}{$+$0.007}
\newcommand{\IsoGapBath}{$+$0.020}
\newcommand{\IsoGapKitchenGemma}{$+$0.262}
\newcommand{\IsoGapKitchen}{$+$0.251}
\newcommand{\IsoNBath}{1069}
\newcommand{\IsoNKitchen}{1268}

\newcommand{\IsoNormalBathGemma}{0.783}
\newcommand{\IsoNormalBath}{0.832}
\newcommand{\IsoNormalKitchenGemma}{0.857}
\newcommand{\IsoNormalKitchen}{0.907}
\newcommand{\IsoTangentBathGemma}{0.775}
\newcommand{\IsoTangentBath}{0.812}
\newcommand{\IsoTangentKitchenGemma}{0.595}
\newcommand{\IsoTangentKitchen}{0.656}

\newcommand{\MinOcc}{0.10}

\newcommand{\MirrorRawAccKitchen}{0.437}

\newcommand{\MirrorRawDeltaKitchen}{$-$0.833}
\newcommand{\MixLivingAllCI}{[$-$0.188, $-$0.137]}
\newcommand{\MixLivingAllN}{221}
\newcommand{\MixLivingAll}{$-$0.163}

\newcommand{\MixLivingDomCI}{[$-$0.161, $-$0.033]}
\newcommand{\MixLivingDomN}{54}
\newcommand{\MixLivingDom}{$-$0.095}

\newcommand{\MixSlopeGemmaAll}{$+$0.227}
\newcommand{\MixSlopeGemmaDom}{$+$0.189}
\newcommand{\MixSlopeQwenAll}{$+$0.029}
\newcommand{\MixSlopeQwenDom}{$+$0.031}
\newcommand{\ModelSDGemmaA}{0.108}

\newcommand{\ModelSDQwenA}{0.013}

\newcommand{\NABedChairCIGemma}{[$-$0.612, $-$0.480]}
\newcommand{\NABedChairCIQwen}{[$-$0.489, $-$0.326]}
\newcommand{\NABedChairGemma}{$-$0.544}
\newcommand{\NABedChairQwen}{$-$0.410}
\newcommand{\NABedNightstandCIGemma}{[$+$0.876, $+$0.955]}
\newcommand{\NABedNightstandCIQwen}{[$+$0.742, $+$0.842]}
\newcommand{\NABedNightstandGemma}{$+$0.915}
\newcommand{\NABedNightstandQwen}{$+$0.794}
\newcommand{\NABothBedChair}{183}
\newcommand{\NABothBedNightstand}{129}
\newcommand{\NABothDeskChair}{146}

\newcommand{\NADeskChairCIGemma}{[$+$0.795, $+$0.840]}
\newcommand{\NADeskChairCIQwen}{[$+$0.655, $+$0.733]}
\newcommand{\NADeskChairGemma}{$+$0.817}
\newcommand{\NADeskChairQwen}{$+$0.694}
\newcommand{\NADiffBedNightstandCIGemma}{[$+$1.390, $+$1.533]}
\newcommand{\NADiffBedNightstandCIQwen}{[$+$1.104, $+$1.303]}
\newcommand{\NADiffBedNightstandGemma}{$+$1.459}
\newcommand{\NADiffBedNightstandQwen}{$+$1.204}
\newcommand{\NADiffDeskChairCIGemma}{[$+$1.292, $+$1.434]}
\newcommand{\NADiffDeskChairCIQwen}{[$+$1.016, $+$1.194]}
\newcommand{\NADiffDeskChairGemma}{$+$1.362}
\newcommand{\NADiffDeskChairQwen}{$+$1.104}
\newcommand{\NADropBedChair}{55}
\newcommand{\NADropBedNightstand}{37}
\newcommand{\NADropDeskChair}{10}

\newcommand{\NANBedChair}{128}
\newcommand{\NANBedNightstand}{92}
\newcommand{\NANDeskChair}{136}

\newcommand{\NAOtherBedNightstandCIGemma}{[$-$0.355, $-$0.260]}
\newcommand{\NAOtherBedNightstandCIQwen}{[$-$0.334, $-$0.213]}
\newcommand{\NAOtherBedNightstandGemma}{$-$0.310}
\newcommand{\NAOtherBedNightstandQwen}{$-$0.275}
\newcommand{\NAOtherDeskChairCIGemma}{[$-$0.214, $-$0.179]}
\newcommand{\NAOtherDeskChairCIQwen}{[$-$0.160, $-$0.108]}
\newcommand{\NAOtherDeskChairGemma}{$-$0.196}
\newcommand{\NAOtherDeskChairQwen}{$-$0.134}

\newcommand{\NARevCIGemma}{[$+$0.009, $+$0.059]}
\newcommand{\NARevCIQwen}{[$-$0.048, $-$0.006]}
\newcommand{\NARevDiffCIGemma}{[$+$0.835, $+$0.926]}
\newcommand{\NARevDiffCIQwen}{[$+$0.760, $+$0.879]}
\newcommand{\NARevDiffGemma}{$+$0.880}
\newcommand{\NARevDiffQwen}{$+$0.821}
\newcommand{\NARevGemma}{$+$0.035}
\newcommand{\NARevN}{126}
\newcommand{\NARevQwen}{$-$0.027}
\newcommand{\NCubiBath}{4,258}
\newcommand{\NCubiKitchen}{2,848}
\newcommand{\NGenPerRoom}{200}
\newcommand{\NScanBath}{210}
\newcommand{\NScanBed}{273}
\newcommand{\NScanLiving}{224}
\newcommand{\NScanOffice}{173}

\newcommand{\NoneBaselineQwen}{0.406}
\newcommand{\NoneNearWallGemma}{0.056}
\newcommand{\NoneNearWallQwen}{0.163}
\newcommand{\NonzeroChairBedMAroom}{218}
\newcommand{\NonzeroChairBedMA}{111}

\newcommand{\NonzeroMAQwen}{87}

\newcommand{\NonzeroMAroomQwen}{144}
\newcommand{\NonzeroToiletBathMAroom}{126}
\newcommand{\NonzeroToiletBathMA}{116}
\newcommand{\NormRadiusBathOneZeroZero}{0.86}

\newcommand{\NormRadiusLivingOneZeroZero}{0.42}

\newcommand{\ObsMoveTries}{10}

\newcommand{\OrderLexAccBathGemma}{0.543}
\newcommand{\OrderLexAccBath}{0.516}

\newcommand{\OrderLexAccKitchenGemma}{0.330}
\newcommand{\OrderLexAccKitchen}{0.191}

\newcommand{\OrderLexDeltaKitchenGemma}{$-$9.608}
\newcommand{\OrderLexDelta}{$-$9.089}

\newcommand{\OriAllGemmaBathS}{0.119}

\newcommand{\OriAllGemmaCentredCI}{[$-$1.631, $-$0.261]}
\newcommand{\OriAllGemmaCentred}{$-$1.124}

\newcommand{\OriAllGemmaE}{$-$0.708}

\newcommand{\OriAllGemmaN}{$-$1.198}

\newcommand{\OriAllGemmaS}{$-$1.126}

\newcommand{\OriAllGemmaW}{$-$1.364}

\newcommand{\OriAllQwenBathS}{0.227}

\newcommand{\OriAllQwenCentredCI}{[$-$0.636, $+$0.069]}
\newcommand{\OriAllQwenCentred}{$-$0.320}

\newcommand{\OriAllQwenE}{$+$0.182}

\newcommand{\OriAllQwenN}{$-$0.437}

\newcommand{\OriAllQwenS}{$-$0.232}

\newcommand{\OriAllQwenW}{$-$0.772}

\newcommand{\OriAllRealBathS}{0.185}

\newcommand{\OriAllRealBedS}{0.190}

\newcommand{\OriAllRealOfficeS}{0.180}

\newcommand{\OriRealUnifMaxAll}{0.0061}

\newcommand{\OriRealUnifMinAll}{0.0010}

\newcommand{\OriSpreadAllGemma}{0.31}

\newcommand{\OriSpreadAllQwen}{2.22}

\newcommand{\OutsideGemmaAMax}{17}
\newcommand{\OutsideGemmaAMin}{2}
\newcommand{\OutsideGemmaBnMax}{14}
\newcommand{\OutsideGemmaBnMin}{1}
\newcommand{\OutsideGemmaBwMax}{82}
\newcommand{\OutsideGemmaBwMin}{36}
\newcommand{\OutsideQwenAMax}{12}
\newcommand{\OutsideQwenAMin}{8}
\newcommand{\OutsideQwenBnMax}{27}
\newcommand{\OutsideQwenBnMin}{8}
\newcommand{\OutsideQwenBwMax}{43}
\newcommand{\OutsideQwenBwMin}{34}
\newcommand{\OverlapGemmaAMax}{0.135}
\newcommand{\OverlapGemmaAMin}{0.057}
\newcommand{\OverlapGemmaBnMax}{0.112}
\newcommand{\OverlapGemmaBnMin}{0.021}
\newcommand{\OverlapGemmaBwMax}{0.020}
\newcommand{\OverlapGemmaBwMin}{0.000}
\newcommand{\OverlapQwenAMax}{0.075}
\newcommand{\OverlapQwenAMin}{0.051}
\newcommand{\OverlapQwenBnMax}{0.147}
\newcommand{\OverlapQwenBnMin}{0.046}
\newcommand{\OverlapQwenBwMax}{0.077}
\newcommand{\OverlapQwenBwMin}{0.043}

\newcommand{\PilePctQwenLivingThreeFive}{91}

\newcommand{\PilePctQwenOfficeThreeFive}{97}

\newcommand{\PileQwenBedOneZeroZero}{0.735}

\newcommand{\PileRadius}{0.35}
\newcommand{\PlaceRhoGemmaA}{0.00}

\newcommand{\PrioBaseGemma}{18}
\newcommand{\PrioBaseQwen}{35}

\newcommand{\PrioDAppcQwenCI}{[$-$0.032, $+$0.011]}
\newcommand{\PrioDAppcQwen}{$-$0.010}
\newcommand{\PrioDAppiGemmaCI}{[$-$0.089, $-$0.044]}
\newcommand{\PrioDAppiGemma}{$-$0.066}
\newcommand{\PrioDAppiQwenCI}{[$-$0.113, $-$0.061]}
\newcommand{\PrioDAppiQwen}{$-$0.087}

\newcommand{\PrioDAppperGemma}{$-$0.018}
\newcommand{\PrioDAppperQwenCI}{[$-$0.060, $-$0.015]}
\newcommand{\PrioDAppperQwen}{$-$0.038}

\newcommand{\PrioDApprGemma}{$-$0.013}
\newcommand{\PrioDApprQwenCI}{[$-$0.030, $+$0.014]}
\newcommand{\PrioDApprQwen}{$-$0.008}

\newcommand{\PrioShareAppiGemma}{38}
\newcommand{\PrioShareAppiQwen}{25}

\newcommand{\RealAreaSD}{8.23}
\newcommand{\RealHomogP}{\ensuremath{<0.001}}

\newcommand{\RealSD}{0.136}
\newcommand{\RectMin}{0.95}

\newcommand{\RelAccCtrlGemmaCubi}{0.798}
\newcommand{\RelAccCtrlGemmaScan}{0.599}
\newcommand{\RelAccCtrlQwenCubi}{0.832}
\newcommand{\RelAccCtrlQwenScan}{0.625}
\newcommand{\RelAccRepairGemmaCubi}{0.811}
\newcommand{\RelAccRepairGemmaScan}{0.593}
\newcommand{\RelAccRepairQwenCubi}{0.844}
\newcommand{\RelAccRepairQwenScan}{0.638}
\newcommand{\RelAccUnpairGemmaCubi}{0.866}
\newcommand{\RelAccUnpairGemmaScan}{0.758}
\newcommand{\RelAccUnpairQwenCubi}{0.925}
\newcommand{\RelAccUnpairQwenScan}{0.712}
\newcommand{\RelDTol}{10}
\newcommand{\RelMinGap}{0.30}
\newcommand{\RelMinRooms}{40}

\newcommand{\RelNAccUnpairCubi}{506}
\newcommand{\RelNAccUnpairScan}{66}

\newcommand{\RelNCubi}{4,980}
\newcommand{\RelNScan}{312}
\newcommand{\RelNear}{0.30}
\newcommand{\RelNegNBed}{3}
\newcommand{\RelNegNLiving}{17}

\newcommand{\RelNegPairingCIGemma}{[$-$0.150, $+$0.150]}
\newcommand{\RelNegPairingCIQwen}{[$-$0.150, $+$0.000]}
\newcommand{\RelNegPairingGemma}{$+$0.000}
\newcommand{\RelNegPairingQwen}{$-$0.050}
\newcommand{\RelNullDraws}{200}

\newcommand{\RelPairGapBedDeskChair}{1.55}

\newcommand{\RelPairNearBedBedBed}{9}
\newcommand{\RelPairNearBedBedNightstand}{97}
\newcommand{\RelPairNearBedDeskChair}{76}
\newcommand{\RelPairNearLivingSofaChair}{23}
\newcommand{\RelPairNullBedBedBed}{47}

\newcommand{\RelPairNullBedDeskChair}{24}
\newcommand{\RelPairNullLivingSofaChair}{33}

\newcommand{\RelPairingCIGemmaBed}{[$-$0.045, $+$0.109]}
\newcommand{\RelPairingCIGemmaCubi}{[$+$0.004, $+$0.022]}
\newcommand{\RelPairingCIGemmaKitchen}{[$+$0.003, $+$0.031]}

\newcommand{\RelPairingCIGemmaScan}{[$-$0.061, $+$0.051]}

\newcommand{\RelPairingCIQwenBed}{[$+$0.006, $+$0.147]}
\newcommand{\RelPairingCIQwenCubi}{[$+$0.003, $+$0.020]}
\newcommand{\RelPairingCIQwenKitchen}{[$+$0.013, $+$0.041]}

\newcommand{\RelPairingCIQwenScan}{[$-$0.038, $+$0.064]}
\newcommand{\RelPairingCleanCIGemmaCubi}{[$+$0.004, $+$0.023]}

\newcommand{\RelPairingCleanCIQwenCubi}{[$+$0.004, $+$0.020]}

\newcommand{\RelPairingCleanGemmaCubi}{$+$0.013}

\newcommand{\RelPairingCleanQwenCubi}{$+$0.012}

\newcommand{\RelPairingGemmaBed}{$+$0.032}
\newcommand{\RelPairingGemmaCubi}{$+$0.013}
\newcommand{\RelPairingGemmaKitchen}{$+$0.017}

\newcommand{\RelPairingGemmaScan}{$-$0.006}

\newcommand{\RelPairingQwenBed}{$+$0.077}
\newcommand{\RelPairingQwenCubi}{$+$0.012}
\newcommand{\RelPairingQwenKitchen}{$+$0.027}

\newcommand{\RelPairingQwenScan}{$+$0.013}
\newcommand{\RelProximityCIGemmaCubi}{[$-$0.032, $+$0.008]}
\newcommand{\RelProximityCIGemmaScan}{[$+$0.000, $+$0.182]}
\newcommand{\RelProximityCIQwenCubi}{[$+$0.000, $+$0.034]}
\newcommand{\RelProximityCIQwenScan}{[$-$0.076, $+$0.076]}
\newcommand{\RelProximityGemmaCubi}{$-$0.012}
\newcommand{\RelProximityGemmaScan}{$+$0.091}
\newcommand{\RelProximityQwenCubi}{$+$0.016}
\newcommand{\RelProximityQwenScan}{$+$0.000}
\newcommand{\RelSameCatN}{74}
\newcommand{\RelSameCatPct}{1.4}
\newcommand{\RelSameCatTotal}{5,292}
\newcommand{\RelStep}{0.01}
\newcommand{\ResOneCmBath}{0.668}
\newcommand{\ResOneCmKitchen}{0.729}
\newcommand{\SAtTenMAGemmaMax}{0.162}
\newcommand{\SAtTenMAGemmaMin}{0.114}
\newcommand{\SAtTenMAQwenMax}{0.139}
\newcommand{\SAtTenMAQwenMin}{0.107}
\newcommand{\SAtTenRMax}{0.539}
\newcommand{\SAtTenRMin}{0.251}

\newcommand{\SAtTenUMin}{0.098}
\newcommand{\SDRatioGemmaACI}{[1.05, 1.53]}
\newcommand{\SDRatioGemmaA}{1.27}

\newcommand{\SDRatioQwenACI}{[4.88, 23.59]}
\newcommand{\SDRatioQwenA}{10.58}

\newcommand{\ScanCovExcessBathAll}{$+$0.146}

\newcommand{\ScanCovExcessBathHalf}{$+$0.102}

\newcommand{\ScanCovExcessBathQuarter}{$+$0.145}

\newcommand{\ScanCovExcessBathThreeQ}{$+$0.018}

\newcommand{\ScanCovExcessBedAll}{$+$0.092}

\newcommand{\ScanCovExcessBedHalf}{$+$0.117}

\newcommand{\ScanCovExcessBedQuarter}{$+$0.127}

\newcommand{\ScanCovExcessBedThreeQ}{$+$0.105}

\newcommand{\ScanCovExcessLivingAll}{$-$0.163}

\newcommand{\ScanCovExcessLivingHalf}{$-$0.109}

\newcommand{\ScanCovExcessLivingQuarter}{$-$0.110}

\newcommand{\ScanCovExcessLivingThreeQ}{$-$0.167}

\newcommand{\ScanCovExcessOfficeAll}{$-$0.018}

\newcommand{\ScanCovExcessOfficeHalf}{$+$0.049}

\newcommand{\ScanCovExcessOfficeQuarter}{$+$0.041}

\newcommand{\ScanCovExcessOfficeThreeQ}{$+$0.073}
\newcommand{\ScanCovMedianBath}{0.11}
\newcommand{\ScanCovMedianBed}{0.13}
\newcommand{\ScanCovMedianLiving}{0.00}
\newcommand{\ScanCovMedianOffice}{0.11}
\newcommand{\ScanCovNAll}{864}
\newcommand{\ScanCovNHalf}{153}
\newcommand{\ScanCovNQuarter}{283}
\newcommand{\ScanCovNThreeQ}{63}
\newcommand{\ScanCovRoomsBathAll}{197}
\newcommand{\ScanCovRoomsBathHalf}{51}
\newcommand{\ScanCovRoomsBathQuarter}{80}
\newcommand{\ScanCovRoomsBathThreeQ}{15}
\newcommand{\ScanCovRoomsBedAll}{273}
\newcommand{\ScanCovRoomsBedHalf}{47}
\newcommand{\ScanCovRoomsBedQuarter}{98}
\newcommand{\ScanCovRoomsBedThreeQ}{16}
\newcommand{\ScanCovRoomsLivingAll}{221}
\newcommand{\ScanCovRoomsLivingHalf}{15}
\newcommand{\ScanCovRoomsLivingQuarter}{32}
\newcommand{\ScanCovRoomsLivingThreeQ}{8}
\newcommand{\ScanCovRoomsOfficeAll}{173}
\newcommand{\ScanCovRoomsOfficeHalf}{40}
\newcommand{\ScanCovRoomsOfficeQuarter}{73}
\newcommand{\ScanCovRoomsOfficeThreeQ}{24}

\newcommand{\ScanCovShareBath}{26}
\newcommand{\ScanCovShareBed}{17}
\newcommand{\ScanCovShareLiving}{7}
\newcommand{\ScanCovShareOffice}{23}
\newcommand{\ScanCovSlopeGemmaAll}{$+$0.227}
\newcommand{\ScanCovSlopeGemmaHalf}{$+$0.340}
\newcommand{\ScanCovSlopeGemmaQuarter}{$+$0.287}
\newcommand{\ScanCovSlopeGemmaThreeQ}{$+$0.333}
\newcommand{\ScanCovSlopeQwenAll}{$+$0.029}
\newcommand{\ScanCovSlopeQwenHalf}{$+$0.062}
\newcommand{\ScanCovSlopeQwenQuarter}{$+$0.044}
\newcommand{\ScanCovSlopeQwenThreeQ}{$+$0.067}
\newcommand{\ScanCovTol}{0.30}

\newcommand{\ScanGemmaPileBath}{0.503}
\newcommand{\ScanGemmaPileBed}{0.239}
\newcommand{\ScanGemmaPileLiving}{0.288}

\newcommand{\ScanQwenDetachBed}{0.853}
\newcommand{\ScanQwenDetachLiving}{0.442}

\newcommand{\ScanQwenPileLiving}{0.088}
\newcommand{\ScanQwenPileOffice}{0.029}

\newcommand{\ScanQwenSwapPosBed}{0.963}
\newcommand{\ScanQwenSwapPosLiving}{0.870}

\newcommand{\ScanQwenSwapSizeBed}{0.993}
\newcommand{\ScanQwenSwapSizeLiving}{0.935}

\newcommand{\SearchCatsBath}{5}
\newcommand{\SearchCatsBed}{14}
\newcommand{\SearchCatsLiving}{7}
\newcommand{\SearchCatsOffice}{9}
\newcommand{\SearchRealAllBath}{197}
\newcommand{\SearchRealAllBed}{273}
\newcommand{\SearchRealAllLiving}{221}
\newcommand{\SearchRealAllOffice}{173}
\newcommand{\SearchRealCanonBath}{189}
\newcommand{\SearchRealCanonBed}{240}
\newcommand{\SearchRealCanonLiving}{95}
\newcommand{\SearchRealCanonOffice}{146}

\newcommand{\SearchSkippedGemma}{18}
\newcommand{\SearchSkippedQwen}{1}
\newcommand{\ShareMAGemmaMax}{23}
\newcommand{\ShareMAGemmaMin}{10}
\newcommand{\ShareMAQwenMax}{32}
\newcommand{\ShareMAQwenMin}{14}

\newcommand{\SlopeGemmaACI}{[$+$0.097, $+$0.354]}
\newcommand{\SlopeGemmaA}{$+$0.227}
\newcommand{\SlopeGemmaBnCI}{[$+$0.201, $+$0.489]}
\newcommand{\SlopeGemmaBn}{$+$0.345}
\newcommand{\SlopeGemmaBwCI}{[$-$0.878, $-$0.473]}
\newcommand{\SlopeGemmaBw}{$-$0.661}
\newcommand{\SlopeQwenACI}{[$-$0.065, $+$0.120]}
\newcommand{\SlopeQwenA}{$+$0.029}
\newcommand{\SlopeQwenBnCI}{[$+$0.524, $+$0.832]}
\newcommand{\SlopeQwenBn}{$+$0.677}
\newcommand{\SlopeQwenBwCI}{[$-$0.358, $-$0.022]}
\newcommand{\SlopeQwenBw}{$-$0.188}

\newcommand{\SmBestGapGemmaMax}{0.243}
\newcommand{\SmBestGapGemmaMin}{0.082}

\newcommand{\SmBestGapQwenMax}{0.240}
\newcommand{\SmBestGapQwenMin}{0.103}

\newcommand{\SmBestGemmaMin}{0.380}

\newcommand{\SmBestQwenMin}{0.393}

\newcommand{\SmBestRMax}{0.305}
\newcommand{\SmBestRMin}{0.153}

\newcommand{\SmCostGemmaBathHalf}{0.396}
\newcommand{\SmCostGemmaBathOne}{0.430}
\newcommand{\SmCostGemmaBathTwo}{0.457}
\newcommand{\SmCostGemmaBathZero}{0.433}
\newcommand{\SmCostGemmaBedHalf}{0.387}
\newcommand{\SmCostGemmaBedOne}{0.380}
\newcommand{\SmCostGemmaBedTwo}{0.406}
\newcommand{\SmCostGemmaBedZero}{0.459}
\newcommand{\SmCostGemmaLivingHalf}{0.427}
\newcommand{\SmCostGemmaLivingOne}{0.401}
\newcommand{\SmCostGemmaLivingTwo}{0.387}
\newcommand{\SmCostGemmaLivingZero}{0.487}
\newcommand{\SmCostGemmaOfficeHalf}{0.411}
\newcommand{\SmCostGemmaOfficeOne}{0.413}
\newcommand{\SmCostGemmaOfficeTwo}{0.428}
\newcommand{\SmCostGemmaOfficeZero}{0.463}

\newcommand{\SmCostQwenBathHalf}{0.393}
\newcommand{\SmCostQwenBathOne}{0.394}
\newcommand{\SmCostQwenBathTwo}{0.409}
\newcommand{\SmCostQwenBathZero}{0.425}
\newcommand{\SmCostQwenBedHalf}{0.421}
\newcommand{\SmCostQwenBedOne}{0.427}
\newcommand{\SmCostQwenBedTwo}{0.453}
\newcommand{\SmCostQwenBedZero}{0.466}
\newcommand{\SmCostQwenLivingHalf}{0.412}
\newcommand{\SmCostQwenLivingOne}{0.408}
\newcommand{\SmCostQwenLivingTwo}{0.410}
\newcommand{\SmCostQwenLivingZero}{0.453}
\newcommand{\SmCostQwenOfficeHalf}{0.403}
\newcommand{\SmCostQwenOfficeOne}{0.404}
\newcommand{\SmCostQwenOfficeTwo}{0.424}
\newcommand{\SmCostQwenOfficeZero}{0.444}

\newcommand{\SmCostRBathHalf}{0.153}
\newcommand{\SmCostRBathOne}{0.179}
\newcommand{\SmCostRBathTwo}{0.262}
\newcommand{\SmCostRBathZero}{0.209}
\newcommand{\SmCostRBedHalf}{0.220}
\newcommand{\SmCostRBedOne}{0.246}
\newcommand{\SmCostRBedTwo}{0.297}
\newcommand{\SmCostRBedZero}{0.250}
\newcommand{\SmCostRLivingHalf}{0.305}
\newcommand{\SmCostRLivingOne}{0.314}
\newcommand{\SmCostRLivingTwo}{0.352}
\newcommand{\SmCostRLivingZero}{0.347}
\newcommand{\SmCostROfficeHalf}{0.241}
\newcommand{\SmCostROfficeOne}{0.259}
\newcommand{\SmCostROfficeTwo}{0.310}
\newcommand{\SmCostROfficeZero}{0.278}
\newcommand{\SmGainGemmaBathHalfCI}{[$-$0.057, $-$0.018]}
\newcommand{\SmGainGemmaBathHalf}{$-$0.038}
\newcommand{\SmGainGemmaBathOneCI}{[$-$0.024, $+$0.020]}
\newcommand{\SmGainGemmaBathOne}{$-$0.003}
\newcommand{\SmGainGemmaBathTwoCI}{[$+$0.002, $+$0.046]}
\newcommand{\SmGainGemmaBathTwo}{$+$0.024}
\newcommand{\SmGainGemmaBedHalfCI}{[$-$0.083, $-$0.060]}
\newcommand{\SmGainGemmaBedHalf}{$-$0.071}
\newcommand{\SmGainGemmaBedOneCI}{[$-$0.093, $-$0.065]}
\newcommand{\SmGainGemmaBedOne}{$-$0.078}
\newcommand{\SmGainGemmaBedTwoCI}{[$-$0.069, $-$0.037]}
\newcommand{\SmGainGemmaBedTwo}{$-$0.053}
\newcommand{\SmGainGemmaLivingHalfCI}{[$-$0.079, $-$0.041]}
\newcommand{\SmGainGemmaLivingHalf}{$-$0.060}
\newcommand{\SmGainGemmaLivingOneCI}{[$-$0.108, $-$0.064]}
\newcommand{\SmGainGemmaLivingOne}{$-$0.086}
\newcommand{\SmGainGemmaLivingTwoCI}{[$-$0.124, $-$0.076]}
\newcommand{\SmGainGemmaLivingTwo}{$-$0.100}
\newcommand{\SmGainGemmaOfficeHalfCI}{[$-$0.065, $-$0.040]}
\newcommand{\SmGainGemmaOfficeHalf}{$-$0.052}
\newcommand{\SmGainGemmaOfficeOneCI}{[$-$0.068, $-$0.032]}
\newcommand{\SmGainGemmaOfficeOne}{$-$0.050}
\newcommand{\SmGainGemmaOfficeTwoCI}{[$-$0.054, $-$0.016]}
\newcommand{\SmGainGemmaOfficeTwo}{$-$0.035}

\newcommand{\SmGainQwenBathHalfCI}{[$-$0.042, $-$0.022]}
\newcommand{\SmGainQwenBathHalf}{$-$0.032}
\newcommand{\SmGainQwenBathOneCI}{[$-$0.054, $-$0.010]}
\newcommand{\SmGainQwenBathOne}{$-$0.031}
\newcommand{\SmGainQwenBathTwoCI}{[$-$0.043, $+$0.011]}
\newcommand{\SmGainQwenBathTwo}{$-$0.016}
\newcommand{\SmGainQwenBedHalfCI}{[$-$0.054, $-$0.035]}
\newcommand{\SmGainQwenBedHalf}{$-$0.044}
\newcommand{\SmGainQwenBedOneCI}{[$-$0.052, $-$0.026]}
\newcommand{\SmGainQwenBedOne}{$-$0.038}
\newcommand{\SmGainQwenBedTwoCI}{[$-$0.027, $+$0.004]}
\newcommand{\SmGainQwenBedTwo}{$-$0.012}
\newcommand{\SmGainQwenLivingHalfCI}{[$-$0.055, $-$0.026]}
\newcommand{\SmGainQwenLivingHalf}{$-$0.040}
\newcommand{\SmGainQwenLivingOneCI}{[$-$0.065, $-$0.022]}
\newcommand{\SmGainQwenLivingOne}{$-$0.044}
\newcommand{\SmGainQwenLivingTwoCI}{[$-$0.067, $-$0.016]}
\newcommand{\SmGainQwenLivingTwo}{$-$0.043}
\newcommand{\SmGainQwenOfficeHalfCI}{[$-$0.049, $-$0.033]}
\newcommand{\SmGainQwenOfficeHalf}{$-$0.041}
\newcommand{\SmGainQwenOfficeOneCI}{[$-$0.054, $-$0.026]}
\newcommand{\SmGainQwenOfficeOne}{$-$0.039}
\newcommand{\SmGainQwenOfficeTwoCI}{[$-$0.035, $-$0.005]}
\newcommand{\SmGainQwenOfficeTwo}{$-$0.020}

\newcommand{\SmGainRBathHalfCI}{[$-$0.070, $-$0.044]}
\newcommand{\SmGainRBathHalf}{$-$0.057}
\newcommand{\SmGainRBathOneCI}{[$-$0.051, $-$0.011]}
\newcommand{\SmGainRBathOne}{$-$0.030}
\newcommand{\SmGainRBathTwoCI}{[$+$0.026, $+$0.081]}
\newcommand{\SmGainRBathTwo}{$+$0.053}
\newcommand{\SmGainRBedHalfCI}{[$-$0.037, $-$0.024]}
\newcommand{\SmGainRBedHalf}{$-$0.030}
\newcommand{\SmGainRBedOneCI}{[$-$0.015, $+$0.007]}
\newcommand{\SmGainRBedOne}{$-$0.004}
\newcommand{\SmGainRBedTwoCI}{[$+$0.032, $+$0.061]}
\newcommand{\SmGainRBedTwo}{$+$0.047}
\newcommand{\SmGainRLivingHalfCI}{[$-$0.054, $-$0.031]}
\newcommand{\SmGainRLivingHalf}{$-$0.042}
\newcommand{\SmGainRLivingOneCI}{[$-$0.054, $-$0.016]}
\newcommand{\SmGainRLivingOne}{$-$0.034}
\newcommand{\SmGainRLivingTwoCI}{[$-$0.017, $+$0.029]}
\newcommand{\SmGainRLivingTwo}{$+$0.005}
\newcommand{\SmGainROfficeHalfCI}{[$-$0.046, $-$0.029]}
\newcommand{\SmGainROfficeHalf}{$-$0.037}
\newcommand{\SmGainROfficeOneCI}{[$-$0.034, $-$0.004]}
\newcommand{\SmGainROfficeOne}{$-$0.019}
\newcommand{\SmGainROfficeTwoCI}{[$+$0.013, $+$0.050]}
\newcommand{\SmGainROfficeTwo}{$+$0.032}
\newcommand{\SmGapZeroGemmaMax}{0.224}
\newcommand{\SmGapZeroGemmaMin}{0.140}
\newcommand{\SmGapZeroQwenMax}{0.216}
\newcommand{\SmGapZeroQwenMin}{0.105}
\newcommand{\SmGemQwenBathHalfCI}{[$-$0.025, $+$0.014]}
\newcommand{\SmGemQwenBathHalf}{$-$0.005}
\newcommand{\SmGemQwenBathOneCI}{[$+$0.005, $+$0.052]}
\newcommand{\SmGemQwenBathOne}{$+$0.028}
\newcommand{\SmGemQwenBathTwoCI}{[$+$0.011, $+$0.069]}
\newcommand{\SmGemQwenBathTwo}{$+$0.040}
\newcommand{\SmGemQwenBedHalfCI}{[$-$0.036, $-$0.009]}
\newcommand{\SmGemQwenBedHalf}{$-$0.022}
\newcommand{\SmGemQwenBedOneCI}{[$-$0.048, $-$0.015]}
\newcommand{\SmGemQwenBedOne}{$-$0.032}
\newcommand{\SmGemQwenBedTwoCI}{[$-$0.053, $-$0.016]}
\newcommand{\SmGemQwenBedTwo}{$-$0.035}
\newcommand{\SmGemQwenLivingHalfCI}{[$-$0.043, $+$0.003]}
\newcommand{\SmGemQwenLivingHalf}{$-$0.020}
\newcommand{\SmGemQwenLivingOneCI}{[$-$0.071, $-$0.014]}
\newcommand{\SmGemQwenLivingOne}{$-$0.042}
\newcommand{\SmGemQwenLivingTwoCI}{[$-$0.089, $-$0.028]}
\newcommand{\SmGemQwenLivingTwo}{$-$0.057}
\newcommand{\SmGemQwenOfficeHalfCI}{[$-$0.024, $+$0.001]}
\newcommand{\SmGemQwenOfficeHalf}{$-$0.011}
\newcommand{\SmGemQwenOfficeOneCI}{[$-$0.029, $+$0.008]}
\newcommand{\SmGemQwenOfficeOne}{$-$0.011}
\newcommand{\SmGemQwenOfficeTwoCI}{[$-$0.037, $+$0.005]}
\newcommand{\SmGemQwenOfficeTwo}{$-$0.015}

\newcommand{\SmInGemmaOne}{0.200}

\newcommand{\SmInGemmaZero}{0.116}

\newcommand{\SmInQwenOne}{0.279}

\newcommand{\SmInQwenZero}{0.216}

\newcommand{\SmInROne}{0.188}

\newcommand{\SmInRZero}{0.143}

\newcommand{\SmOutGemmaOne}{0.481}

\newcommand{\SmOutGemmaZero}{0.586}

\newcommand{\SmOutQwenOne}{0.539}

\newcommand{\SmOutQwenZero}{0.667}

\newcommand{\SmOutROne}{0.459}

\newcommand{\SmOutRZero}{0.641}
\newcommand{\SmRoomGemmaBathHalfCI}{[$-$0.006, $+$0.010]}
\newcommand{\SmRoomGemmaBathHalf}{$+$0.002}
\newcommand{\SmRoomGemmaBathOneCI}{[$-$0.019, $-$0.002]}
\newcommand{\SmRoomGemmaBathOne}{$-$0.010}
\newcommand{\SmRoomGemmaBathTwoCI}{[$-$0.027, $-$0.011]}
\newcommand{\SmRoomGemmaBathTwo}{$-$0.019}

\newcommand{\SmRoomGemmaBedHalfCI}{[$-$0.017, $-$0.002]}
\newcommand{\SmRoomGemmaBedHalf}{$-$0.010}
\newcommand{\SmRoomGemmaBedOneCI}{[$-$0.016, $-$0.001]}
\newcommand{\SmRoomGemmaBedOne}{$-$0.009}
\newcommand{\SmRoomGemmaBedTwoCI}{[$-$0.031, $-$0.011]}
\newcommand{\SmRoomGemmaBedTwo}{$-$0.021}

\newcommand{\SmRoomGemmaLivingHalfCI}{[$-$0.030, $+$0.025]}
\newcommand{\SmRoomGemmaLivingHalf}{$-$0.002}
\newcommand{\SmRoomGemmaLivingOneCI}{[$-$0.008, $+$0.031]}
\newcommand{\SmRoomGemmaLivingOne}{$+$0.011}
\newcommand{\SmRoomGemmaLivingTwoCI}{[$-$0.001, $+$0.022]}
\newcommand{\SmRoomGemmaLivingTwo}{$+$0.010}

\newcommand{\SmRoomGemmaOfficeHalfCI}{[$-$0.026, $-$0.002]}
\newcommand{\SmRoomGemmaOfficeHalf}{$-$0.013}
\newcommand{\SmRoomGemmaOfficeOneCI}{[$-$0.030, $-$0.004]}
\newcommand{\SmRoomGemmaOfficeOne}{$-$0.017}
\newcommand{\SmRoomGemmaOfficeTwoCI}{[$-$0.041, $-$0.016]}
\newcommand{\SmRoomGemmaOfficeTwo}{$-$0.028}

\newcommand{\SmRoomQwenBathHalfCI}{[$-$0.010, $+$0.017]}
\newcommand{\SmRoomQwenBathHalf}{$+$0.004}
\newcommand{\SmRoomQwenBathOneCI}{[$-$0.009, $+$0.012]}
\newcommand{\SmRoomQwenBathOne}{$+$0.002}
\newcommand{\SmRoomQwenBathTwoCI}{[$-$0.012, $+$0.008]}
\newcommand{\SmRoomQwenBathTwo}{$-$0.002}

\newcommand{\SmRoomQwenBedHalfCI}{[$+$0.010, $+$0.033]}
\newcommand{\SmRoomQwenBedHalf}{$+$0.022}
\newcommand{\SmRoomQwenBedOneCI}{[$+$0.019, $+$0.043]}
\newcommand{\SmRoomQwenBedOne}{$+$0.031}
\newcommand{\SmRoomQwenBedTwoCI}{[$+$0.003, $+$0.030]}
\newcommand{\SmRoomQwenBedTwo}{$+$0.016}

\newcommand{\SmRoomQwenLivingHalfCI}{[$-$0.032, $+$0.004]}
\newcommand{\SmRoomQwenLivingHalf}{$-$0.014}
\newcommand{\SmRoomQwenLivingOneCI}{[$-$0.032, $-$0.000]}
\newcommand{\SmRoomQwenLivingOne}{$-$0.016}
\newcommand{\SmRoomQwenLivingTwoCI}{[$-$0.026, $+$0.003]}
\newcommand{\SmRoomQwenLivingTwo}{$-$0.011}

\newcommand{\SmRoomQwenLivingZero}{$-$0.028}
\newcommand{\SmRoomQwenOfficeHalfCI}{[$-$0.013, $+$0.012]}
\newcommand{\SmRoomQwenOfficeHalf}{$-$0.001}
\newcommand{\SmRoomQwenOfficeOneCI}{[$-$0.017, $+$0.005]}
\newcommand{\SmRoomQwenOfficeOne}{$-$0.006}
\newcommand{\SmRoomQwenOfficeTwoCI}{[$-$0.017, $+$0.003]}
\newcommand{\SmRoomQwenOfficeTwo}{$-$0.006}

\newcommand{\SmRoomQwenOfficeZero}{$-$0.024}
\newcommand{\SmShareBestGemmaMax}{58}
\newcommand{\SmShareBestGemmaMin}{30}
\newcommand{\SmShareBestQwenMax}{48}
\newcommand{\SmShareBestQwenMin}{29}

\newcommand{\SmShareOutR}{0.272}
\newcommand{\SmVsRGemmaBathHalfCI}{[$+$0.208, $+$0.277]}
\newcommand{\SmVsRGemmaBathHalf}{$+$0.243}
\newcommand{\SmVsRGemmaBathOneCI}{[$+$0.217, $+$0.285]}
\newcommand{\SmVsRGemmaBathOne}{$+$0.251}
\newcommand{\SmVsRGemmaBathTwoCI}{[$+$0.160, $+$0.232]}
\newcommand{\SmVsRGemmaBathTwo}{$+$0.195}

\newcommand{\SmVsRGemmaBedHalfCI}{[$+$0.146, $+$0.193]}
\newcommand{\SmVsRGemmaBedHalf}{$+$0.170}
\newcommand{\SmVsRGemmaBedOneCI}{[$+$0.118, $+$0.162]}
\newcommand{\SmVsRGemmaBedOne}{$+$0.140}
\newcommand{\SmVsRGemmaBedTwoCI}{[$+$0.087, $+$0.139]}
\newcommand{\SmVsRGemmaBedTwo}{$+$0.113}

\newcommand{\SmVsRGemmaLivingHalfCI}{[$+$0.090, $+$0.152]}
\newcommand{\SmVsRGemmaLivingHalf}{$+$0.122}
\newcommand{\SmVsRGemmaLivingOneCI}{[$+$0.061, $+$0.113]}
\newcommand{\SmVsRGemmaLivingOne}{$+$0.087}
\newcommand{\SmVsRGemmaLivingTwoCI}{[$+$0.014, $+$0.056]}
\newcommand{\SmVsRGemmaLivingTwo}{$+$0.035}

\newcommand{\SmVsRGemmaOfficeHalfCI}{[$+$0.154, $+$0.186]}
\newcommand{\SmVsRGemmaOfficeHalf}{$+$0.170}
\newcommand{\SmVsRGemmaOfficeOneCI}{[$+$0.137, $+$0.170]}
\newcommand{\SmVsRGemmaOfficeOne}{$+$0.154}
\newcommand{\SmVsRGemmaOfficeTwoCI}{[$+$0.102, $+$0.135]}
\newcommand{\SmVsRGemmaOfficeTwo}{$+$0.118}

\newcommand{\SmVsRQwenBathHalfCI}{[$+$0.211, $+$0.271]}
\newcommand{\SmVsRQwenBathHalf}{$+$0.240}
\newcommand{\SmVsRQwenBathOneCI}{[$+$0.183, $+$0.244]}
\newcommand{\SmVsRQwenBathOne}{$+$0.215}
\newcommand{\SmVsRQwenBathTwoCI}{[$+$0.117, $+$0.178]}
\newcommand{\SmVsRQwenBathTwo}{$+$0.147}

\newcommand{\SmVsRQwenBedHalfCI}{[$+$0.181, $+$0.222]}
\newcommand{\SmVsRQwenBedHalf}{$+$0.201}
\newcommand{\SmVsRQwenBedOneCI}{[$+$0.158, $+$0.201]}
\newcommand{\SmVsRQwenBedOne}{$+$0.181}
\newcommand{\SmVsRQwenBedTwoCI}{[$+$0.131, $+$0.179]}
\newcommand{\SmVsRQwenBedTwo}{$+$0.156}

\newcommand{\SmVsRQwenLivingHalfCI}{[$+$0.076, $+$0.140]}
\newcommand{\SmVsRQwenLivingHalf}{$+$0.107}
\newcommand{\SmVsRQwenLivingOneCI}{[$+$0.066, $+$0.124]}
\newcommand{\SmVsRQwenLivingOne}{$+$0.095}
\newcommand{\SmVsRQwenLivingTwoCI}{[$+$0.032, $+$0.081]}
\newcommand{\SmVsRQwenLivingTwo}{$+$0.058}

\newcommand{\SmVsRQwenOfficeHalfCI}{[$+$0.143, $+$0.182]}
\newcommand{\SmVsRQwenOfficeHalf}{$+$0.162}
\newcommand{\SmVsRQwenOfficeOneCI}{[$+$0.128, $+$0.161]}
\newcommand{\SmVsRQwenOfficeOne}{$+$0.145}
\newcommand{\SmVsRQwenOfficeTwoCI}{[$+$0.098, $+$0.131]}
\newcommand{\SmVsRQwenOfficeTwo}{$+$0.114}

\newcommand{\SpreadCellsAtNGemma}{31}
\newcommand{\SpreadCellsAtNQwen}{65}
\newcommand{\SpreadCellsAtNReal}{51}
\newcommand{\SpreadCellsGemma}{46}
\newcommand{\SpreadCellsQwen}{106}
\newcommand{\SpreadCellsReal}{63}
\newcommand{\SpreadDimsMaxGemma}{18}
\newcommand{\SpreadDimsMaxQwen}{58}
\newcommand{\SpreadDimsMaxReal}{240}
\newcommand{\SpreadDimsMinGemma}{7}
\newcommand{\SpreadDimsMinQwen}{49}
\newcommand{\SpreadDimsMinReal}{95}
\newcommand{\SpreadDimsTopMaxGemma}{68}

\newcommand{\SpreadDimsTopMinGemma}{48}

\newcommand{\SpreadListsMaxGemma}{22}
\newcommand{\SpreadListsMaxQwen}{104}

\newcommand{\SpreadListsMinGemma}{5}
\newcommand{\SpreadListsMinQwen}{48}

\newcommand{\SpreadNGemma}{200}
\newcommand{\SpreadNPriors}{26}
\newcommand{\SpreadNQwen}{200}
\newcommand{\SpreadNReal}{98}
\newcommand{\SpreadNarrowerGemma}{25}

\newcommand{\SpreadPosDistinctGemma}{0.45}
\newcommand{\SpreadPosDistinctQwen}{0.93}
\newcommand{\SpreadPosDistinctReal}{0.94}
\newcommand{\SpreadTopOneGemma}{0.12}
\newcommand{\SpreadTopOneQwen}{0.02}
\newcommand{\SpreadTopOneReal}{0.02}

\newcommand{\SpreadWiderQwen}{20}
\newcommand{\SpreadXYMaxGemma}{55}
\newcommand{\SpreadXYMaxQwen}{111}
\newcommand{\SpreadXYMaxReal}{668}
\newcommand{\SpreadXYMinGemma}{37}
\newcommand{\SpreadXYMinQwen}{81}
\newcommand{\SpreadXYMinReal}{330}

\newcommand{\WcClaimGapGemmaBath}{$-$0.183}
\newcommand{\WcClaimGapGemmaKitchen}{$+$0.064}
\newcommand{\WcClaimGapQwenBath}{$-$0.043}
\newcommand{\WcClaimGapQwenKitchen}{$+$0.078}

\newcommand{\WcClaimGemmaBath}{0.855}

\newcommand{\WcClaimGemmaKitchen}{0.828}

\newcommand{\WcClaimQwenBath}{0.857}

\newcommand{\WcClaimQwenKitchen}{0.866}

\newcommand{\WcDropGemmaBath}{0.581}

\newcommand{\WcDropGemmaKitchen}{0.849}

\newcommand{\WcDropQwenBath}{0.669}

\newcommand{\WcDropQwenKitchen}{0.875}

\newcommand{\WcGeoGapGemmaBath}{$+$0.090}
\newcommand{\WcGeoGapGemmaKitchen}{$+$0.042}
\newcommand{\WcGeoGapQwenBath}{$+$0.145}
\newcommand{\WcGeoGapQwenKitchen}{$+$0.070}

\newcommand{\WcKeepGemmaBath}{0.671}

\newcommand{\WcKeepGemmaKitchen}{0.892}

\newcommand{\WcKeepQwenBath}{0.814}

\newcommand{\WcKeepQwenKitchen}{0.945}

\newcommand{\WcNQwenBath}{4255}
\newcommand{\WcNQwenKitchen}{2829}
\newcommand{\nLadderMax}{272}
\newcommand{\nLadderMin}{163}

\newcommand{\nRoomsMax}{273}
\newcommand{\nRoomsMin}{173}

\newcommand{\nVocabCubi}{14}
\newcommand{\nVocab}{18}

%% file: sections/introduction.tex
\section{Introduction}
\label{sec:introduction}

Object-goal navigation is a fundamental task for mobile service
robots. Semantic exploration is a strategy for object-goal navigation
and sometimes utilizes large language models (LLMs) for selecting next
viewpoints. Systems such as ESC, L3MVN and their successors query the
model for a \emph{scalar affinity}, which indicates how strongly a
target category is associated with a room type or a nearby object, and
use it to rank frontiers
\citep{zhou2023esc,yu2023l3mvn,ji2024diffusion,jia2026hro}. The model
supplies a number per candidate frontier, and the geometry is obtained
from a separate mapping process.

In this paper, we investigate whether LLMs can provide a \emph{placement} rather than an affinity. If the layouts generated by a model gave the usual positions of objects in a room type, a robot could enter an unmapped room with a spatial prior over the position of the target instead of a ranking of frontiers. Language-driven scene generation systems use LLMs to generate metric layouts that look plausible
\citep{feng2023layoutgpt,yang2024holodeck,celen2024idesign}.
Their evaluations, however, measure the plausibility of the generated scenes or their agreement with the prompt; they do not compare the distribution of generated layouts with that of real rooms, which is the property that a spatial prior requires.

\begin{figure}[tbp]
  \centering
  \includegraphics[width=\linewidth]{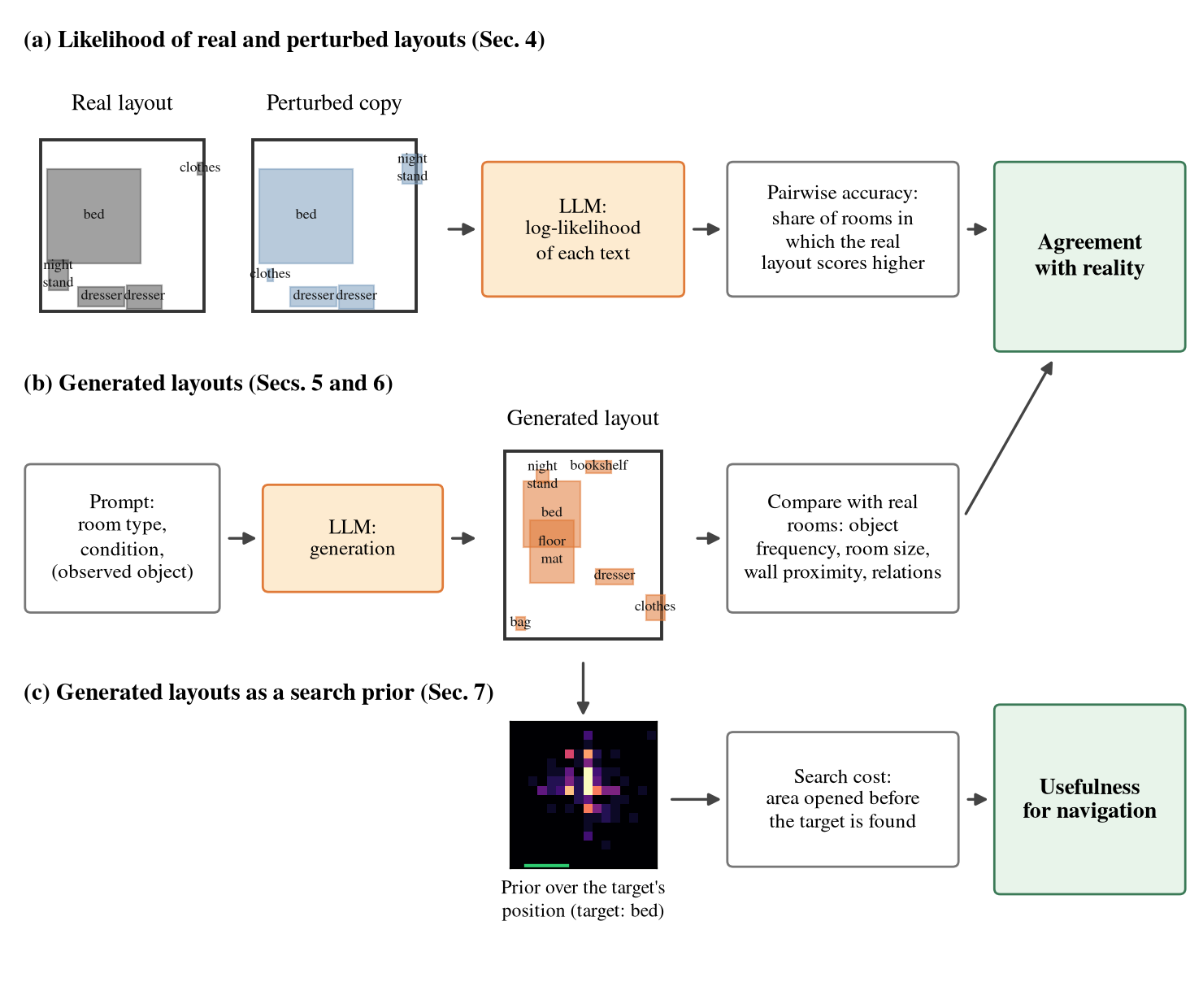}
  \caption{Overview of the evaluation. (a)~The likelihood that an LLM assigns to a real layout, written as text, is compared with that of a perturbed copy (here, two objects exchanged); the pairwise accuracy is the fraction of rooms in which the real layout scores higher. (b)~Layouts generated for a room type are compared with real rooms of that type. (c)~A prior over the target's position built from the generated layouts is evaluated by the area searched before the target is found. The real layout is a ScanNet bedroom; the generated layout and the prior (target: bed) are Qwen's under condition~\condA{}. In the prior, the door is on the south wall (green) and brighter cells are more probable.}
  \label{fig:overview}
\end{figure}

We evaluate the placement information that two open-weight LLMs, Qwen3.5-27B and Gemma-4-31B-it (Qwen and Gemma below), provide along two axes: \textbf{agreement with reality} and \textbf{usefulness for navigation}. Fig.~\ref{fig:overview} shows the overview of the evaluation. We use two outputs of the models, both on real rooms from two datasets, ScanNet and CubiCasa5K. The first output is the likelihood that a model assigns to a layout written as text (Fig.~\ref{fig:overview}(a)). We compare the likelihood of each real layout with that of perturbed copies, in which objects are moved, exchanged or piled up, and measure the fraction of rooms in which the real layout scores higher. The second output is the set of layouts that a model generates for a given room type (Fig.~\ref{fig:overview}(b)). We compare their object frequencies, room sizes and wall proximity with those of real rooms of the same type, and examine whether a generated object moves with an object stated in the prompt. These two comparisons measure the agreement with reality. For the usefulness for navigation, we build from the generated layouts a prior over the position of a target object (Fig.~\ref{fig:overview}(c)), and compute the fraction of the area of a real room that a search guided by the prior examines before it finds the target, compared with a uniform prior and with a prior built from real layouts.

The contributions of this paper are as follows:
\begin{enumerate}
  \item We propose a method for evaluating LLMs as spatial priors. The method uses two outputs of the models, the likelihood of real and perturbed layouts and the generated layouts, on real rooms from two datasets, and controls for the properties of the serialized text that affect the likelihood (Secs.~\ref{sec:setup}, \ref{sec:discrimination} and~\ref{sec:likelihooddesign}).
  \item We show that object frequencies and room sizes in the generated rooms follow the room type but object placement does not. We also show that the generated coordinates reflect regularities in how coordinates are written rather than the geometric constraints that determine them (Secs.~\ref{sec:generation} and~\ref{sec:mechanism}).
  \item We show that, when an observed object is given in the prompt, the models place a related object relative to it and move unrelated objects away from it (Sec.~\ref{sec:observed}).
  \item We evaluate the generated placements as a prior for object search and show that they reduce the search cost relative to a uniform prior by \DiffMAUQwenMin--\DiffMAUQwenMax{} of the room area for Qwen and \DiffMAUGemmaMin--\DiffMAUGemmaMax{} for Gemma, which is \ShareMAQwenMin--\ShareMAQwenMax\% and \ShareMAGemmaMin--\ShareMAGemmaMax\%, respectively, of the reduction obtained with a prior built from real layouts (\SmShareBestQwenMin--\SmShareBestQwenMax\% and \SmShareBestGemmaMin--\SmShareBestGemmaMax\% when each prior is spatially smoothed). Based on all the results, we derive design guidelines for using LLMs in object-goal navigation (Secs.~\ref{sec:search} and~\ref{sec:design}).
\end{enumerate} 

The remainder of the paper is organized as follows. Sec.~\ref{sec:related} surveys related work. Sec.~\ref{sec:setup} describes the datasets, the serialization, the placement statistic and the generation conditions. Sec.~\ref{sec:discrimination} measures whether the likelihood of a real layout is higher than that of a perturbed copy, separates the effects of the geometry and of the text, and measures whether the likelihood changes when the neighbor of an object is changed. Sec.~\ref{sec:generation} measures which properties of generated layouts depend on the room type, and whether placement follows an object stated in the prompt. Sec.~\ref{sec:mechanism} analyzes the generated coordinates. Sec.~\ref{sec:search} evaluates the generated placements as a search prior. Sec.~\ref{sec:design} states the design guidelines, Sec.~\ref{sec:limitations} the limitations, and Sec.~\ref{sec:conclusion} concludes. Appendix~\ref{app:measurement} defines every quantity, Appendix~\ref{app:prompts} reproduces the prompts, and Appendix~\ref{app:schemas} the output schemas. The tables of Appendix~\ref{app:measurement} give every number the main text summarizes, each placed after the definition it depends on.

%% file: sections/related.tex
\section{Related work}
\label{sec:related}

\paragraph{Language priors for object-goal navigation.}
A line of work starting with commonsense-constrained exploration uses a
language model to score candidate frontiers by their semantic relevance to the
goal. ESC converts commonsense relations between the goal, the objects already
seen, and the inferred room type into soft logic predicates over frontiers
\citep{zhou2023esc}. L3MVN queries a language model for the frontier most
likely to lead to the target, in both zero-shot and feed-forward forms
\citep{yu2023l3mvn}. More recent systems make the room level explicit;
HRO reasons first about which room type a frontier belongs to, then scores
target--room affinity \citep{jia2026hro}, or use language-derived
object--room knowledge to condition a learned predictor of the target's
location \citep{ji2024diffusion}. All of these follow the modular
semantic-exploration strategy established by goal-oriented semantic exploration
\citep{chaplot2020semexp}. In these works, geometric information such
as frontier locations comes from usual perception and mapping
process. In this paper, we evaluate whether such geometric information can be obtained from language models.

\paragraph{Priors for object search.}
Before language models, priors for object search were obtained from data. Co-occurrence statistics between objects and between objects and scene types, collected from annotated images, have been used to predict where an unseen object is and to plan the search \citep{kollar2009context}, and probabilistic models that combine such semantic knowledge with the spatial structure of a partially known environment have been used for active visual search \citep{aydemir2013active}. These priors, like the affinities obtained from language models, relate an object to a room type or to other objects; they do not give a distribution over metric positions in a room, which is the quantity we evaluate.

\paragraph{Language-driven scene generation.}
LayoutGPT prompts a language model to emit numeric layouts as structured text,
including 3D indoor scenes \citep{feng2023layoutgpt}. Holodeck builds whole
environments for embodied AI from a text prompt, using the language model for
commonsense and a solver for spatial constraints \citep{yang2024holodeck}.
I-Design has a team of language-model agents negotiate a scene graph, then
places objects with a separate algorithm \citep{celen2024idesign}.
These systems are evaluated on whether the produced scene is plausible
or matches the prompt rather than on whether the distribution of
layouts matches that of real rooms. In addition, Holodeck and I-Design compute the final coordinates with a solver or a separate placement algorithm rather than taking them from the language model. In this paper, we investigate the possibility of
generating metric information only from LLMs.

Distributions over metric layouts can also be learned from collections of layouts. ATISS, for example, is an autoregressive transformer trained on furnished rooms that generates the category, position, size and orientation of each object \citep{paschalidou2021atiss}. Such a model requires layouts of the target domain for training. The prior built from real layouts in Sec.~\ref{sec:search} is the simplest model of this kind, a histogram of object positions, and serves as the reference with which the generated placements are compared.

\paragraph{Spatial reasoning benchmarks.}

\paragraph{Spatial reasoning benchmarks.}

A separate literature asks whether vision--language models can answer questions
about spatial configurations, for instance by building a mental model of a
scene from limited views \citep{wang2025mindcube}. These benchmarks measure
accuracy against an external ground truth on posed questions. In this
paper, we investigate whether language models can generate the
distribution of an object which can be used as its spatial prior.

\paragraph{Likelihood and the form of the text.}
The probability that a language model assigns to a string depends on the form of the string as well as on its meaning. In multiple-choice scoring, different strings that express the same answer share the probability mass, so the string with the highest probability is not always the correct answer \citep{holtzman2021surface}. For likelihood-based generative models of images, the likelihood depends strongly on low-level statistics of the input \citep{nalisnick2019know,serra2020input}. Our likelihood comparisons are affected in the same way, and Secs.~\ref{sec:discrimination} and~\ref{sec:likelihooddesign} measure and control these effects for serialized layouts.

%% file: sections/setup.tex
\section{Setup for Analysis}
\label{sec:setup}

\subsection{Two datasets with different room and object types}
\label{sec:data}

\subsubsection{Datasets}

We use the following two datasets of real rooms:

\textbf{CubiCasa5K}~\citep{kalervo2019cubicasa} is a corpus of annotated
architectural floor plans. Its
objects are \emph{fixtures}: sinks, bathtubs, counters, stoves. Fixtures are attached to walls by construction, and their positions are drawn rather than measured. We use the kitchens and bathrooms, and keep the rooms whose polygon fills at least \RectMin{} of its oriented bounding box, so that a rectangle describes them: \NCubiKitchen{} kitchens and \NCubiBath{} bathrooms, with a closed vocabulary of \nVocabCubi{} fixture categories, such as sink, counter, stove, dishwasher, toilet and bathtub.

\textbf{ScanNet\,v2}~\citep{dai2017scannet} is a corpus of RGB-D scans of real
rooms. Its objects are
the \emph{movable furniture} a robot has to navigate around: chairs, tables,
sofas, boxes. We use four room types: bathrooms, bedrooms, offices and living rooms (\NScanBath, \NScanBed, \NScanOffice{} and \NScanLiving{} rooms). We align each scan with its axis-alignment matrix, take the extent of the floor and wall geometry as the room rectangle, and keep the instances whose lower bound lies below \FloorZMax~m above the floor, so that a sensor at floor level would detect them.

\subsubsection{Canonical coordinates and serialization}
\label{sec:canon}

\begin{figure}[tbp]
\begin{center}
   \includegraphics[width=0.3\linewidth]{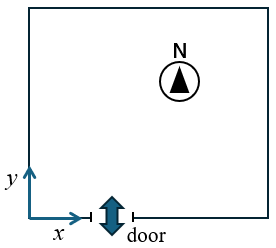}
   \caption{Canonical coordinates.}
   \label{fig:CanonicalCoordinates}
\end{center}
\end{figure}

Each room is expressed in meters from its south-west corner, with $x$ running
east and $y$ north. To remove the arbitrary choice of orientation, we rotate the
room so that a door lies on the south wall and mirror it, if necessary, so that
the door is left of center (see Fig. \ref{fig:CanonicalCoordinates}).

Layouts are serialized as JSON with one entry per object, giving its category,
center, and footprint size. All numbers are written with exactly two decimal
places, so that the string length does not vary with the value. The room type
is \emph{not} part of the scored string: it appears only in the prompt.

A room and its mirror image describe the same arrangement, so every score used for comparing layouts is averaged over the layout and its mirrored copy (Sec.~\ref{sec:scoring}). Appendix~\ref{app:canon} states the rotation and mirroring rule and reports the fraction of rooms that can be oriented in this way.

\subsubsection{Closed vocabulary}
\label{sec:vocab}

Both the model and the reference data are restricted to a closed vocabulary, so
that a comparison of $p(o \mid c)$ is a comparison of the same quantity on both
sides. For ScanNet we take the \nVocab{} categories needed to cover
\CoverageTarget\% of floor-standing instances in the room types we study,
excluding categories that are not floor-standing in purpose (curtains). The
resulting per-room-type coverage runs from \CoverageMin{} to \CoverageMax.

\subsection{Wall proximity excess as statistic and its baseline}
\label{sec:excess}

Scene-generation systems treat wall contact as a binary relation or as a constraint to be satisfied~\citep{yang2024holodeck,celen2024idesign}. A binary measure is adequate when the room is fixed, but the models choose their own room dimensions under condition~\condA{}, and a large object in a small room is near a wall because of its size alone. We therefore use, as the placement statistic, the \emph{wall-proximity excess}: the fraction of objects near a wall minus the fraction expected from the sizes of the objects and the rooms.

Let object $i$ have footprint $w_i \times d_i$ and center $(x_i, y_i)$ in a
room $W_j \times D_j$, and let $\beta = \BandM$~m. The object is
\emph{near a wall} when its nearest face is within $\beta$:
\begin{equation}
  n_i = \mathbf{1}\Bigl[\min\bigl(x_i - \tfrac{w_i}{2},\;
    W_j - x_i - \tfrac{w_i}{2},\; y_i - \tfrac{d_i}{2},\;
    D_j - y_i - \tfrac{d_i}{2}\bigr) \le \beta\Bigr].
  \label{eq:near}
\end{equation}

To consider the size of an object with respect to the room size, we
calculate the baseline $b_i$, which is the probability that an object
would be near a wall if its center is uniform over the admissible
region, given by:
\begin{equation}
  b_i = 1 - \Bigl(1 - \frac{w_i + 2\beta}{W_j}\Bigr)
	  \Bigl(1 - \frac{d_i + 2\beta}{D_j}\Bigr).
  \label{eq:baseline}
\end{equation}
Then, the wall-proximity excess of a set of objects is calculated as:
\begin{equation}
  \mathrm{excess} = \frac{\sum_i n_i - \sum_i b_i}{\sum_i 1}.
  \label{eq:excess}
\end{equation}
The objects are pooled, not averaged per room first, so that each room contributes in proportion to the number of its objects and rooms with few objects are not given a higher weight. The unit of resampling for
statistical analysis is nevertheless the room. For calculating a
confidence interval (see Appendix~\ref{app:excess}), we resample room with replacement,
recompute these three sums over the resampled multiset of rooms, and
take the 2.5 and 97.5 percentiles over 10,000 draws. Resampling rooms rather than objects accounts for the dependence between the objects of one room.

\subsection{Scoring with serialization and symmetrization}
\label{sec:scoring}

For the comparisons of Sec.~\ref{sec:discrimination}, the LLM computes the likelihood of a given layout (the object placements in a room). A layout is first serialized as JSON with
the object list in the order given by the source, where each object is
rendered as:
\begin{quote}\ttfamily\footnotesize\raggedright
\{"category": "bed",\\
\hspace*{1em}"center\_m": \{"x": 2.35, "y": 0.76\},\\
\hspace*{1em}"size\_m": \{"w": 2.03, "d": 1.52\},\\
\hspace*{1em}"wall\_contact": "N", "stated\_probability": 0.90\}.
\end{quote}
The field \texttt{stated\_probability} is the model's stated probability that a room of the type contains the object; it is the same in the layouts that are compared. All numbers, including the stated probability, are written with exactly two decimal places, so that the number of tokens does not depend on the value.

We obtain log-likelihoods of layouts from the completions endpoint with
\texttt{echo=true}, which returns per-token log-probabilities for a
supplied string; the prompt with the JSON description of a layout is
input to the LLM, and the score of the layout is calculated as the sum
over the tokens of the serialized JSON body. Requests are issued one at a time, because with continuous batching the returned log-probabilities depend on the other requests in the batch, and no KV-cache quantization is used. We used two models, Qwen3.5-27B and Gemma-4-31B-it, in bfloat16 with
eager execution. The sampling parameters shipped with the models (temperature, top-$k$ and top-$p$) are disabled on the server; generation (Sec.~\ref{sec:conditions}) samples at temperature $1.0$ from the full distribution for both models. 

As described in Sec.~\ref{sec:canon}, every
score used for layout evaluation is symmetrized:
\begin{equation}
  \mathrm{score}(L) = \tfrac{1}{2}\bigl[\log p(L)
    + \log p(\mathrm{mirror}(L))\bigr],
  \label{eq:sym2}
\end{equation}
where $\mathrm{mirror}$ reflects the layout in $x$ and swaps $E$ and
$W$ wall annotations. This symmetrization is applied to every likelihood comparison, but not to generation: the statistics of the generated layouts are computed on the layouts as generated. 

\subsection{Layout generation conditions}
\label{sec:conditions}

For Sec.~\ref{sec:generation}, the models generate layouts as JSON constrained by a schema (Appendix~\ref{app:schemas}), at temperature $1.0$, for each of the four ScanNet room types; for the analyses of Sec.~\ref{sec:adsorption}, both models also generate CubiCasa kitchens and bathrooms. Each model generates \NGenPerRoom{} layouts per room type under each of three prompts:

\begin{description}
  \item[\condA{} (survey)] asks for the whole layout at once, choosing the
    room's dimensions itself.
  \item[\condBn{} (neutral route)] asks the model to walk through the room and
    report the objects it comes across, in order (the prompts of Appendix~\ref{app:prompts} call all objects fixtures).
  \item[\condBw{} (wall route)] asks it to walk \emph{along the walls},
    starting to its left and going clockwise.
\end{description}

Appendix~\ref{app:prompts} gives the prompts. Further conditions used in individual analyses are defined where they are used: \condAp, \condApp{} and its variants in Sec.~\ref{sec:symbols} and Sec.~\ref{sec:containment}, and O-anchor and O-move in Sec.~\ref{sec:observed}.

\begin{figure}[tbp]
  \centering
  \includegraphics[width=\linewidth]{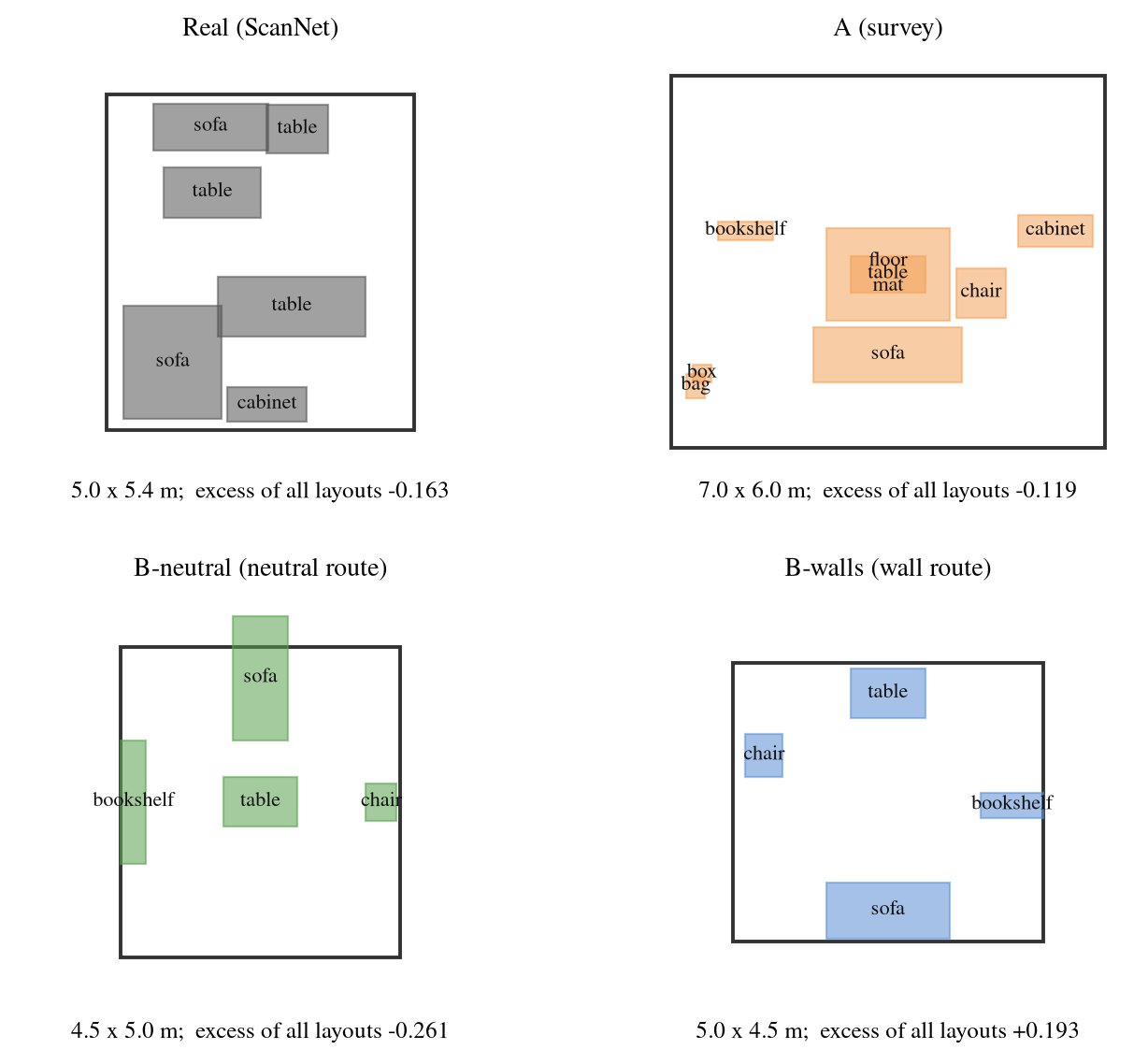}
  \caption{Examples of layouts under the three generation conditions. Top left: a real ScanNet living room. Others: one layout generated by Qwen for the living room under each condition. Each panel is the layout of its set closest to the median object count and then to the median area. Below each panel are the room size and the wall-proximity excess (Sec.~\ref{sec:excess}) of all layouts of the set; the panels are examples, and the numbers are not.}
  \label{fig:condexamples}
\end{figure}

Fig.~\ref{fig:condexamples} shows examples of the generated layouts: one ScanNet living room and one layout that Qwen generated for the living room under each condition. Under conditions~\condA{} and~\condBn{} the objects of the examples stand mostly away from the walls, and under condition~\condBw{} they stand against the walls. Below each panel is the wall-proximity excess (Sec.~\ref{sec:excess}) computed over all layouts of the set, not over the example. It is negative for the real living rooms and for conditions~\condA{} and~\condBn{}, that is, fewer objects are near a wall than expected from their sizes, and positive for condition~\condBw{}, that is, more objects are near a wall than expected. The signs agree with the examples: objects away from the walls under \condA{} and \condBn{}, and against the walls under \condBw{}. Sec.~\ref{sec:generation} compares these values across room types.

%% file: sections/discrimination.tex
\section{Discrimination of Real and Perturbed Layouts}
\label{sec:discrimination}

This section evaluates the likelihood that the models assign to real layouts and to perturbed copies of them. For each real room, we compare the likelihood of the real layout with that of a perturbed copy and measure the fraction of rooms in which the real layout has the higher likelihood. Sec.~\ref{sec:ladder} applies the perturbations to CubiCasa rooms and separates the effect of the geometry from that of the text, Sec.~\ref{sec:reversal} applies them to ScanNet rooms, and Sec.~\ref{sec:relladder} perturbs the relation between two objects. 

\subsection{Perturbation strategies and results for CubiCasa kitchens and bathrooms}
\label{sec:ladder}

\subsubsection{Perturbation and Results}
\label{sec:perturbresults}

Each perturbation changes one property of a real (reference) layout and keeps the others. The object list, the object
footprints, and the room rectangle are preserved unless stated. We
score both versions with the same serialization and report \emph{pairwise accuracy}: the fraction of rooms in which the reference scores above
the perturbed copy. Chance is $0.5$. Appendix~\ref{app:perturbations}
defines each perturbation by formula, together with the handling of
ties, perturbations that leave the two-decimal string unchanged, which
are excluded and counted rather than scored.

\begin{figure}[tbp]
  \centering
  \includegraphics[width=\linewidth]{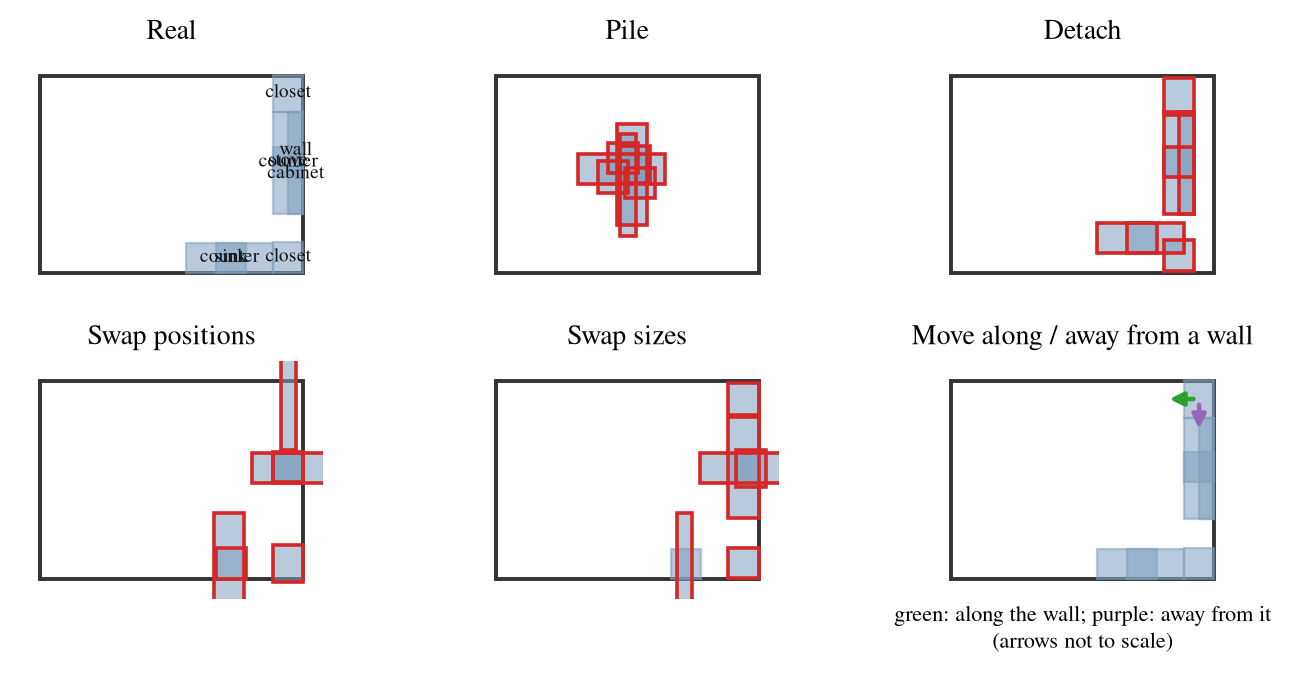}
  \caption{The perturbations applied to one real CubiCasa kitchen. Objects whose position or size the perturbation changes are outlined in red. The bottom-right panel marks the two single-object moves of Sec.~\ref{sec:geometric}, along the nearest wall and away from it; the arrows are not to scale (the moves are $0.10$~m).}
  \label{fig:perturb}
\end{figure}

\begin{figure}[tbp]
  \centering
  \includegraphics[width=\linewidth]{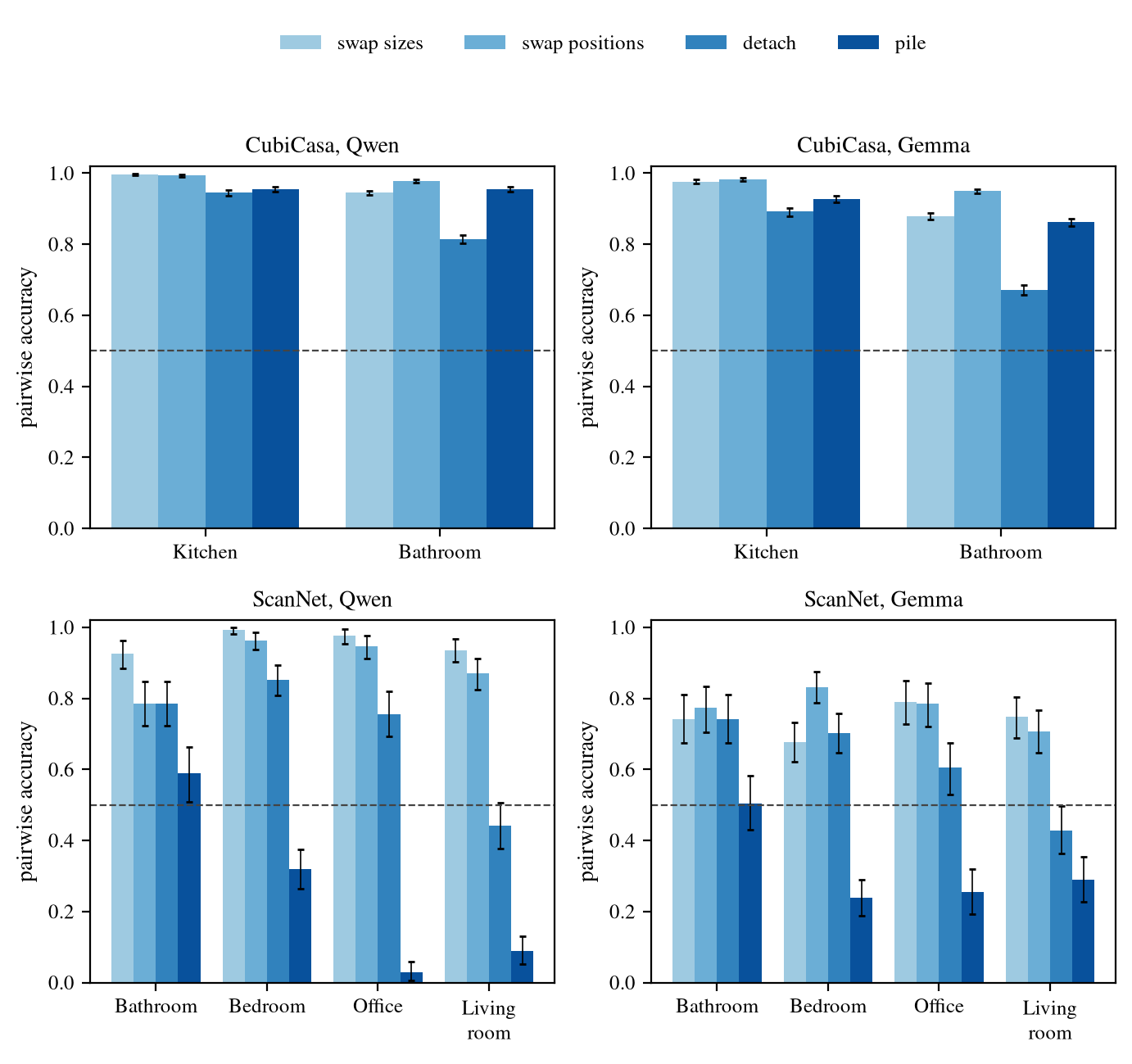}
  \caption{Pairwise accuracy of the real layout against each perturbation, with 95\% intervals over rooms. Top: CubiCasa; bottom: ScanNet; left: Qwen; right: Gemma. Below the dashed line the perturbed layout is preferred. Detach moves objects away from their walls but leaves their \texttt{wall\_contact} field unchanged, so the field contradicts the coordinates; its accuracy combines this contradiction with the geometric change (Sec.~\ref{sec:selfconsistency}). Tables~\ref{tab:app-ladder-cubi}, \ref{tab:app-ladder-cubi-gemma}, \ref{tab:app-ladder-qwen} and~\ref{tab:app-ladder-gemma} give the values.}
  \label{fig:ladder}
\end{figure}

We applied four perturbations to the CubiCasa kitchens and bathrooms: \emph{pile}, \emph{detach}, \emph{swap positions} and \emph{swap sizes} (Fig.~\ref{fig:perturb}). Fig.~\ref{fig:ladder} (top) and Tables~\ref{tab:app-ladder-cubi} and~\ref{tab:app-ladder-cubi-gemma} give the results. Both models assign the higher likelihood to the real layout in most rooms for all four perturbations. For Qwen, the pairwise accuracy in kitchens and bathrooms is \CubiPileKitchen{} and \CubiPileBath{} for pile, \CubiDetachKitchen{} and \CubiDetachBath{} for detach, \CubiSwapPosKitchen{} and \CubiSwapPosBath{} for swap positions, and \CubiSwapSizeKitchen{} and \CubiSwapSizeBath{} for swap sizes. For Gemma, the corresponding values are \CubiGemmaPileKitchen{} and \CubiGemmaPileBath{}, \CubiGemmaDetachKitchen{} and \CubiGemmaDetachBath{}, \CubiGemmaSwapPosKitchen{} and \CubiGemmaSwapPosBath{}, and \CubiGemmaSwapSizeKitchen{} and \CubiGemmaSwapSizeBath{}. Gemma's accuracy is lower than Qwen's for every perturbation, and the lowest value for both models is detach in bathrooms.

The serialization states, in the \texttt{wall\_contact} field, which wall each object touches. Detach changes the coordinates but not this field, so the field of a detached object contradicts its coordinates. Sec.~\ref{sec:selfconsistency} separates the effect of this contradiction from that of the geometric change.

A high pairwise accuracy does not by itself show that the likelihood depends on the geometry, because a perturbation changes the text as well as the geometry and the likelihood is a function of the text. A \emph{reordering} perturbation, which sorts the object list lexicographically and moves no object, gives a pairwise accuracy of \OrderLexAccKitchen{} in Qwen's kitchens, so that the sorted copy has the higher likelihood in most rooms; the mean log-likelihood difference (real minus sorted) is \OrderLexDelta{} nats, larger in magnitude than the \CubiDeltaDetachKitchen{} nats of detach (Tables~\ref{tab:app-direction} and~\ref{tab:app-ladder-cubi}). For Gemma, the pairwise accuracy of the same perturbation is \OrderLexAccKitchenGemma{} and the mean difference is \OrderLexDeltaKitchenGemma{} nats, which is smaller in magnitude than the \CubiGemmaDeltaDetachKitchen{} nats of detach for that model. In bathrooms, the sorted copy does not have the higher likelihood in most rooms for either model (\OrderLexAccBath{} for Qwen and \OrderLexAccBathGemma{} for Gemma). The size of a likelihood difference therefore does not measure the size of a geometric change: for both models, an edit that moves no object changes the likelihood in kitchens by an amount comparable to that of a geometric perturbation.

\subsubsection{Moves along and away from a wall}
\label{sec:geometric}

The perturbations of Sec.~\ref{sec:perturbresults} change the coordinate values in the string, and thereby both the geometry and the text that the likelihood is computed from. To measure the effect of the geometry with the change of the text held fixed, we used two perturbations that rewrite the same number of coordinate values and differ only in the \emph{direction} of a move. We moved a single object that touches a wall by a fixed distance, either along that wall (tangential move) or away from it (normal move), in the CubiCasa kitchens and bathrooms, and scored both moves with both models. Both changes rewrite one coordinate pair, so
the serialization is perturbed equally; only the geometric consequence
differs, because the normal move breaks wall contact and the
tangential move does not.
The normal move also falsifies the \texttt{wall\_contact}
description for the moved object, which the edit leaves in place.
However, because only one object is moved per room, this affects
only \ClaimFalseNormalOutKitchen\% of declarations in kitchens and
\ClaimFalseNormalOutBath\% in bathrooms at $0.10$~m, against
\ClaimFalseTangentKitchen\% for the tangential move. Table~\ref{tab:direction} summarizes the results.

\begin{table}[tbp]
  \centering
  \caption{Pairwise accuracy when one object is moved by $0.10$~m in three
    directions (CubiCasa, both models, symmetrized, ties excluded). The tangential
    and the normal move rewrite the same coordinate pair; only the geometric
    consequence differs. ``Isolated objects'' restricts to rooms whose moved
    object has no neighbor within \IsoClear~m ($n = \IsoNKitchen$ kitchens,
    \IsoNBath{} bathrooms). The last row moves one object by $0.01$~m and was scored with Qwen only.
    Intervals, sample sizes and the distance sweep for Qwen are in
    Table~\ref{tab:app-direction}.}
  \label{tab:direction}
  \begin{tabular}{lcccc}
    \toprule
    & \multicolumn{2}{c}{Qwen} & \multicolumn{2}{c}{Gemma} \\
    \cmidrule(lr){2-3}\cmidrule(lr){4-5}
    Move & Kitchens & Bathrooms & Kitchens & Bathrooms \\
    \midrule
    tangential, along the wall & \GeoTangentKitchen & \GeoTangentBath & \GeoTangentKitchenGemma & \GeoTangentBathGemma \\
    normal, away from the wall & \GeoNormalOutKitchen & \GeoNormalOutBath & \GeoNormalOutKitchenGemma & \GeoNormalOutBathGemma \\
    normal, into the wall & \GeoNormalInKitchen & \GeoNormalInBath & \GeoNormalInKitchenGemma & \GeoNormalInBathGemma \\
    gap, normal outward $-$ tangential & \GeoGapKitchen & \GeoGapBath & \GeoGapKitchenGemma & \GeoGapBathGemma \\
    \midrule
    tangential, isolated objects & \IsoTangentKitchen & \IsoTangentBath & \IsoTangentKitchenGemma & \IsoTangentBathGemma \\
    normal outward, isolated objects & \IsoNormalKitchen & \IsoNormalBath & \IsoNormalKitchenGemma & \IsoNormalBathGemma \\
    gap, isolated objects & \IsoGapKitchen & \IsoGapBath & \IsoGapKitchenGemma & \IsoGapBathGemma \\
    \midrule
    one object by $0.01$~m & \ResOneCmKitchen & \ResOneCmBath & --- & --- \\
    \bottomrule
  \end{tabular}
\end{table}

At $0.10$~m in kitchens, for Qwen, the tangential move gives
\GeoTangentKitchen{} and the normal move \GeoNormalOutKitchen{}, a gap
of \GeoGapKitchen{}. Restricting to rooms where the moved object has
no neighbor within \IsoClear~m, so that a tangential move cannot
create an overlap and confound the comparison, widens the gap to
\IsoGapKitchen{} (\IsoTangentKitchen{} against \IsoNormalKitchen{}, $n
= \IsoNKitchen$). Gemma gives the same result in kitchens: a gap of \GeoGapKitchenGemma{} (\GeoTangentKitchenGemma{} against \GeoNormalOutKitchenGemma), which widens to \IsoGapKitchenGemma{} for the isolated objects. In kitchens, therefore, part of the pairwise accuracy depends on the geometry for both models. In bathrooms, the gap at $0.10$~m is \GeoGapBath{} for Qwen and \GeoGapBathGemma{} for Gemma, and for Qwen, the only model for which we varied the displacement, a positive gap appears only at larger displacements (\DirBathTwoZeroGap{} at $0.20$~m; Fig.~\ref{fig:dirsweep}). A likely reason is that bathroom fixtures are large relative to the room, so that after a $0.10$~m normal move most of them are still near the wall. The dependence on the geometry is therefore not fixed but depends on the displacement relative to the size of the objects. A third move pushes the object $0.10$~m into the wall, which keeps the \texttt{wall\_contact} field true. It gives the highest pairwise accuracy of the three moves in kitchens (\GeoNormalInKitchen, against \GeoNormalOutKitchen{} for the outward and \GeoTangentKitchen{} for the tangential move; for Gemma, \GeoNormalInKitchenGemma, \GeoNormalOutKitchenGemma{} and \GeoTangentKitchenGemma), so the order of the three moves is not explained by the contradiction of the field.

Moving one object by $0.01$~m gives a pairwise accuracy of \ResOneCmKitchen{} in kitchens and \ResOneCmBath{} in bathrooms, so the differences are not caused by rounding of the coordinates. Table~\ref{tab:app-direction} lists every displacement control of this subsection with its sample size, tie rate and interval.

\subsubsection{Analysis of self-consistency in serialization}
\label{sec:selfconsistency}

Besides the coordinates, each object in the serialization has a \texttt{wall\_contact} field that \emph{states} which wall it touches, and every displacement perturbation changes the coordinates but not this field. After detaching, \ClaimFalseDetachKitchen\% of declarations in kitchens
and \ClaimFalseDetachBath\% in bathrooms contradict the coordinates. The real layouts contain no such contradiction, because the declarations of the CubiCasa rooms are derived from their geometry when the sets of real rooms are built. The pairwise accuracy of such a perturbation therefore includes the effect of the contradiction as well as that of the geometric change.

We separated the two effects with the following three versions of the same
comparison, scored over the same rooms ($n = \WcNQwenKitchen$
kitchens, $\WcNQwenBath$ bathrooms; ties: none):
\begin{description}
  \item[Keep] the perturbation as run above: coordinates detached,
    wall\_contact state left unchanged, so that it still names the wall the object touched before the move and contradicts the coordinates. This is the row
    in Table~\ref{tab:app-ladder-cubi}.
  \item[Claim only] Every object stays exactly where the real room puts it,
    and one character of its wall\_contact state is changed to name a
    wall it does not touch; the opposite wall where that wall is
    genuinely far, otherwise the farthest wall. Object order, token
    count and the number of digits are all preserved.
  \item[Drop field] the \texttt{wall\_contact} field is deleted from the reference
    \emph{and} from the detached copy, leaving coordinates and footprints.
    This measures the geometric information alone.
\end{description}

Table~\ref{tab:ThreeChannels} summarizes the results; Table~\ref{tab:app-wallclaim} gives the intervals and Table~\ref{tab:app-wallclaim-diff} the paired differences. The results show the following three points for both models.
\begin{itemize}
\item
The difference Keep minus Drop field is \WcGeoGapQwenKitchen{} in kitchens and \WcGeoGapQwenBath{} in bathrooms for Qwen (\WcGeoGapGemmaKitchen{} and \WcGeoGapGemmaBath{} for Gemma), and all four intervals exclude zero. Therefore, part of the detection of the detached layout comes from the contradiction between the wall\_contact state and the coordinates. 
\item
With the field removed (Drop field), the pairwise accuracy of detach in bathrooms is \WcDropQwenBath{} for Qwen and \WcDropGemmaBath{} for Gemma, against \WcDropQwenKitchen{} and \WcDropGemmaKitchen{} in kitchens. The geometric change of detach is therefore detected less often in bathrooms than in kitchens, as in the comparison of move directions in Sec.~\ref{sec:geometric}, which uses a different perturbation.
\item
A change of one character of the field, with every object at its real position (Claim only), gives a pairwise accuracy of \WcClaimQwenKitchen{} and \WcClaimQwenBath{} for Qwen and \WcClaimGemmaKitchen{} and \WcClaimGemmaBath{} for Gemma. This is lower than the accuracy of detach (Keep) in kitchens and higher in bathrooms: the difference Keep minus Claim only is \WcClaimGapQwenKitchen{} in kitchens and \WcClaimGapQwenBath{} in bathrooms for Qwen, and \WcClaimGapGemmaKitchen{} and \WcClaimGapGemmaBath{} for Gemma.
\end{itemize}

\begin{table}[htbp]
\begin{center}
  \caption{Pairwise accuracy of the three versions of the detach comparison (CubiCasa, symmetrized, no ties). Keep: detach with the \texttt{wall\_contact} field unchanged. Claim only: one character of the field changed, no object moved. Drop field: the field removed from both layouts. Intervals are in Table~\ref{tab:app-wallclaim}.}
  \label{tab:ThreeChannels}
\begin{tabular}{llll}
  \toprule
  & Keep & Claim only & Drop field \\
  \midrule
  Qwen, kitchens & \WcKeepQwenKitchen{} & \WcClaimQwenKitchen{} & \WcDropQwenKitchen{} \\
  Qwen, bathrooms & \WcKeepQwenBath{} & \WcClaimQwenBath{} & \WcDropQwenBath{} \\
  Gemma, kitchens & \WcKeepGemmaKitchen{} & \WcClaimGemmaKitchen{} & \WcDropGemmaKitchen{} \\
  Gemma, bathrooms & \WcKeepGemmaBath{} & \WcClaimGemmaBath{} & \WcDropGemmaBath{} \\
  \bottomrule
\end{tabular}
\end{center}
\end{table}

The Claim only accuracy differs little between kitchens and bathrooms (\WcClaimQwenKitchen{} and \WcClaimQwenBath{} for Qwen), whereas Keep and Drop field differ between them (Drop field \WcDropQwenKitchen{} and \WcDropQwenBath). The effect of the contradiction therefore depends little on the room type, and that of the geometric change depends on it. The reordering of Sec.~\ref{sec:perturbresults} changes the whole list; Claim only keeps the order of the objects, the number of tokens and the number of digits, changes a single character, and has an effect comparable to that of moving every object by the detach distance of $0.10$--$0.40$~m (Appendix~\ref{app:perturbations}). For a perturbation that moves objects and makes their declarations false, the pairwise accuracy thus combines two effects, and in bathrooms the effect of the geometry (Drop field) is the smaller one.

The comparison of move directions in Sec.~\ref{sec:geometric} is not affected, because at $0.10$~m the normal move makes only \ClaimFalseNormalOutKitchen\% of the declarations false, and the inward move, which makes none false, has the highest pairwise accuracy.

\subsubsection{Summary of the CubiCasa results}
\label{sec:cubisummary}

For the fixtures of CubiCasa, both models assign a higher likelihood to the real layout than to each of the four perturbed copies in most rooms (Sec.~\ref{sec:perturbresults}). Part of this pairwise accuracy depends on the geometry. In kitchens, moving an object $0.10$~m away from its wall is detected more often than moving it the same distance along the wall (gap \GeoGapKitchen{} for Qwen and \GeoGapKitchenGemma{} for Gemma; Sec.~\ref{sec:geometric}), and with the \texttt{wall\_contact} field removed from both layouts, detach is still detected in \WcDropQwenKitchen{} (Qwen) and \WcDropGemmaKitchen{} (Gemma) of the kitchens (Sec.~\ref{sec:selfconsistency}). The other part depends on the text. Sorting the object list, which moves no object, changes the likelihood by more than detach does in Qwen's kitchens and by less than detach does in Gemma's (Sec.~\ref{sec:perturbresults}), and a contradiction between the \texttt{wall\_contact} field and the coordinates lowers the likelihood by itself (Claim only, pairwise accuracy \WcClaimQwenBath--\WcClaimQwenKitchen{} for Qwen and \WcClaimGemmaKitchen--\WcClaimGemmaBath{} for Gemma; Sec.~\ref{sec:selfconsistency}). The geometric part is smaller in bathrooms than in kitchens for both models (Drop field \WcDropQwenBath{} and \WcDropGemmaBath{} in bathrooms). A high pairwise accuracy against a perturbed copy therefore shows that the likelihood depends on the layout, but not by itself that it depends on the geometry.

\subsection{Perturbation strategies and results for ScanNet}
\label{sec:reversal}
\label{sec:pilemech}

The objects of CubiCasa are fixtures attached to the walls, whereas a robot moves among movable furniture. We therefore applied the same four perturbations, with the same scoring, to the ScanNet rooms of the four room types, over the closed vocabulary of Sec.~\ref{sec:vocab} and \nLadderMin--\nLadderMax{} rooms per room type (Fig.~\ref{fig:ladder}, bottom; Tables~\ref{tab:app-ladder-qwen} and~\ref{tab:app-ladder-gemma}). For Qwen, the pairwise accuracy is \ScanQwenSwapPosLiving--\ScanQwenSwapPosBed{} for swap positions, \ScanQwenSwapSizeLiving--\ScanQwenSwapSizeBed{} for swap sizes and \ScanQwenDetachLiving--\ScanQwenDetachBed{} for detach. For pile, with radius \PileRadius~m, it is below $0.5$ in three room types: the piled layout has the higher likelihood in \PilePctQwenOfficeThreeFive\% of offices and \PilePctQwenLivingThreeFive\% of living rooms (pairwise accuracy \ScanQwenPileOffice{} and \ScanQwenPileLiving). For Gemma, the pairwise accuracy of pile is \ScanGemmaPileBed--\ScanGemmaPileLiving{} in bedrooms, offices and living rooms and \ScanGemmaPileBath{} in bathrooms.

The piled layouts contradict most of their \texttt{wall\_contact} declarations (\ClaimFalsePileLiving\% in living rooms), which lowers the likelihood by itself (Sec.~\ref{sec:selfconsistency}); their likelihood is nevertheless higher than that of the real layouts in most offices and living rooms, so this result is not explained by the contradiction. Two properties of the data are consistent with the result. First, the likelihood increases as the coordinate values become less diverse: the likelihood gap between Qwen's generated CubiCasa kitchens and real ones falls from \EntGapRawKitchen{} to \EntGapQTwentyFiveKitchen{} nats when both are quantized to a $0.25$~m grid (Sec.~\ref{sec:adsorption}; Appendix~\ref{app:granularity}), and piling makes the coordinates of all objects similar. Second, in CubiCasa, piling moves objects away from the walls they touch, which lowers the likelihood (Sec.~\ref{sec:geometric}), whereas in ScanNet living rooms and offices the real objects are already away from the walls (wall-proximity excess \ExcessRealLiving{} and \ExcessRealOffice).

We then varied the pile radius over $0.2$, $0.35$, $0.6$ and $1.0$~m (Table~\ref{tab:app-pile}; Fig.~\ref{fig:pilesweep}); a smaller radius places the objects closer to the room center, and a larger radius gives more diverse coordinates (column ``Spread'' of Table~\ref{tab:app-pile}). In bedrooms, offices and living rooms, the pairwise accuracy increases with the radius for both models, consistent with the dependence of the likelihood on the diversity of the coordinates; in bathrooms it stays near $0.5$ for both models. At $1.0$~m the radius is \NormRadiusLivingOneZeroZero{} of half the shorter side of the room in living rooms and \NormRadiusBathOneZeroZero{} in bathrooms, so the piled layout differs less from the real one, and an accuracy near $0.5$ at this radius does not indicate detection. In Qwen's bedrooms, the accuracy at $1.0$~m is \PileQwenBedOneZeroZero{}, above $0.5$ (Table~\ref{tab:app-pile}).

\subsection{Perturbing relations between objects}
\label{sec:relladder}

The perturbations above change the position of objects relative to the room. This subsection measures whether the likelihood also depends on the position of an object relative to another object, for example whether a layout with a chair beside a wrong piece of furniture has a lower likelihood than one with the chair beside its own desk.

\paragraph{Method.}
For each room of both datasets we took a pair of categories that real rooms
keep adjacent, an \emph{anchor} and a \emph{target}, such as desk and chair
or counter and sink (Appendix~\ref{app:relations} gives how the pairs are
chosen and lists them), with the two objects within \RelNear~m of each other,
edge to edge. The target was slid along its nearest wall, keeping its distance
to that wall, inside the room and without increasing the overlap of the
layout. Three versions move it by the same distance $d$, to within
$\pm\RelDTol$\%, and differ only in where it ends up. In the \emph{wrong-neighbor}
version it ends beside an object of another category, in the
\emph{same-neighbor} version it stays beside its own anchor, and in the
\emph{no-neighbor} version it is beside nothing. Each version was scored against
the real layout as in Sec.~\ref{sec:ladder}. The difference
$\mathrm{acc}(\text{wrong-neighbor}) - \mathrm{acc}(\text{same-neighbor})$ is the \emph{neighbor effect}: the two versions move the same object by the same distance and differ only in the category of its neighbor. The
difference $\mathrm{acc}(\text{no-neighbor}) - \mathrm{acc}(\text{wrong-neighbor})$
is the \emph{proximity effect}: it compares a target beside no object with a target beside an object. A positive proximity effect would mean that the likelihood favors objects placed beside other objects, which is one possible cause of the pile result on ScanNet (Sec.~\ref{sec:reversal}). Both are paired over rooms. The analysis pools
each dataset (\RelNCubi{} CubiCasa rooms and \RelNScan{} ScanNet rooms); the
no-neighbor version needs an empty stretch of wall and exists in
\RelNAccUnpairCubi{} and \RelNAccUnpairScan{} of them.

\paragraph{Results.}
Table~\ref{tab:relladder} gives the results. The neighbor effect is \RelPairingQwenCubi{} \RelPairingCIQwenCubi{} for Qwen and \RelPairingGemmaCubi{} \RelPairingCIGemmaCubi{} for Gemma on CubiCasa, and \RelPairingQwenScan{} \RelPairingCIQwenScan{} and \RelPairingGemmaScan{} \RelPairingCIGemmaScan{} on ScanNet. On CubiCasa, whose fixtures are placed according to the plumbing, the effect is positive for both models but small; on ScanNet, it is not distinguished from zero. For comparison, the difference between the normal and the tangential move of Sec.~\ref{sec:geometric}, which also compares two versions that move an object by the same distance, is \GeoGapKitchen{} in kitchens at a displacement of $0.10$~m, smaller than the median displacement used here (Table~\ref{tab:app-rel-pairs}). By room type, the CubiCasa effect comes from the kitchens, and the only ScanNet room type with an interval above zero is Qwen's bedrooms (\RelPairingQwenBed{} \RelPairingCIQwenBed; Table~\ref{tab:app-rel-ladder}); we do not interpret these subsets further. The proximity effect, which compares the no-neighbor version (beside nothing) with the wrong-neighbor version (beside an object), is \RelProximityQwenCubi{} \RelProximityCIQwenCubi{} for Qwen and \RelProximityGemmaCubi{} \RelProximityCIGemmaCubi{} for Gemma on CubiCasa, and \RelProximityQwenScan{} \RelProximityCIQwenScan{} and \RelProximityGemmaScan{} \RelProximityCIGemmaScan{} on ScanNet. None of the four intervals lies entirely above zero; the lower bounds of two of them, Qwen on CubiCasa and Gemma on ScanNet, are zero to three decimal places. No preference for an object being beside another object is therefore detected. In the no-neighbor version, the \texttt{wall\_contact} declarations agree least with the coordinates, because a target moved to an empty stretch of wall more often ends near a corner (Table~\ref{tab:app-rel-hygiene}); by the effect of Sec.~\ref{sec:selfconsistency}, this would raise its pairwise accuracy and thus the proximity effect, which is nevertheless not distinguished from zero. The pile result of Sec.~\ref{sec:reversal} is therefore not explained by a preference for objects beside other objects, which is consistent with the account given there by the diversity of the coordinate values. Two checks support the neighbor effect. First, the three versions have the same diversity of coordinate values, overlap and protrusion, and the wrong-neighbor and same-neighbor versions the same agreement of the declared walls (Table~\ref{tab:app-rel-hygiene}), so neither the numerals nor the contradiction of Sec.~\ref{sec:selfconsistency} separates them. Second, the serialization has no instance identifiers, so a target moved beside a second object of the anchor's category is, in the text, still beside an anchor. The construction therefore rejects a wrong-neighbor destination whose only neighbor is such an object. In \RelSameCatPct\% of rooms, however, the destination is beside both an object of another category, as required, and a second object of the anchor's category (for example, a chair moved beside a table that stands next to another desk). Excluding these rooms leaves the neighbor effect unchanged (Appendix~\ref{app:relations}). Sec.~\ref{sec:observed} examines the relation between an anchor and its target in generation.

\begin{table}[tbp]
  \centering
    \caption{Pair perturbations scored by likelihood: pairwise accuracy of the real layout against each version (symmetrized, ties excluded). The columns None, Wrong and Same are the no-neighbor, wrong-neighbor and same-neighbor versions. Wrong-neighbor and same-neighbor exist in every room; no-neighbor exists only where an empty stretch of wall is available, and the number of those rooms is given in parentheses. The neighbor effect is wrong-neighbor minus same-neighbor over all rooms, the effect of the target's neighbor with the displacement held fixed; the proximity effect is no-neighbor minus wrong-neighbor over the rooms in which no-neighbor exists, so it is not the difference between the no-neighbor and wrong-neighbor columns, which cover different rooms. Intervals resample rooms. The breakdown by room type, the quality checks of the three versions and the pairs used are in Appendix~\ref{app:relations}.}
  \label{tab:relladder}
  \footnotesize
  \setlength{\tabcolsep}{4pt}
    \begin{tabular}{llccccc}
    \toprule
    Data & Model & None & Wrong & Same & Neighbor effect & Proximity effect \\
    \midrule
    CubiCasa & Qwen & \RelAccUnpairQwenCubi{} (\RelNAccUnpairCubi) & \RelAccRepairQwenCubi{} & \RelAccCtrlQwenCubi
      & \RelPairingQwenCubi{} \RelPairingCIQwenCubi
      & \RelProximityQwenCubi{} \RelProximityCIQwenCubi \\
                          & Gemma & \RelAccUnpairGemmaCubi{} (\RelNAccUnpairCubi) & \RelAccRepairGemmaCubi{} & \RelAccCtrlGemmaCubi
      & \RelPairingGemmaCubi{} \RelPairingCIGemmaCubi
      & \RelProximityGemmaCubi{} \RelProximityCIGemmaCubi \\
        ScanNet & Qwen & \RelAccUnpairQwenScan{} (\RelNAccUnpairScan) & \RelAccRepairQwenScan{} & \RelAccCtrlQwenScan
      & \RelPairingQwenScan{} \RelPairingCIQwenScan
      & \RelProximityQwenScan{} \RelProximityCIQwenScan \\
                        & Gemma & \RelAccUnpairGemmaScan{} (\RelNAccUnpairScan) & \RelAccRepairGemmaScan{} & \RelAccCtrlGemmaScan
      & \RelPairingGemmaScan{} \RelPairingCIGemmaScan
      & \RelProximityGemmaScan{} \RelProximityCIGemmaScan \\
    \bottomrule
  \end{tabular}
\end{table}

%% file: sections/generation.tex
\section{Room Layout Generation and its Evaluation}
\label{sec:generation}

This section evaluates the layouts that the models generate. It measures which properties of the generated layouts change with the room type (Sec.~\ref{sec:hierarchy}), how the prompt changes them (Sec.~\ref{sec:overwrite}), and whether the placement of an object changes with an object given in the prompt (Sec.~\ref{sec:observed}).

A search prior over the position of an object of category $o$ in a room of type $c$ is a distribution $p(x, y \mid o, c)$. If the generated placements do not change with $c$, a prior built from them contains no information about the room type, whatever the results of the likelihood tests of Sec.~\ref{sec:discrimination}.

\subsection{Dependence of three properties on the room type}
\label{sec:hierarchy}

We measured how three properties of the generated rooms, the objects they contain, the room dimensions and the placement of the objects, depend on the room type, and compared each with the same property of the real ScanNet rooms of the four room types, over the closed vocabulary of \nVocab{} categories (Sec.~\ref{sec:vocab}). The layouts are those generated under condition~\condA{} (Sec.~\ref{sec:conditions}), \NGenPerRoom{} per room type and model. The results are as follows.

\begin{description}
  \item[Object identity]
For each room type we compute $p(o \mid c)$, the fraction of generated rooms
containing category $o$, and correlate it with the same quantity in the real
data. Spearman's $\rho$ is \IdentRhoQwenA{} for Qwen and \IdentRhoGemmaA{} for
Gemma under condition~\condA. A high correlation within a room type could also result from frequencies that are the same in every room type. As a control, we correlated each model's frequencies for a room type with the real frequencies of the other three room types; the mean correlation is \IdentMismatchQwen{} for Qwen and \IdentMismatchGemma{} for Gemma. The object frequencies therefore follow the room type.
Table~\ref{tab:app-identity} gives the underlying $p(o \mid c)$ for every
category and room type, for the real data and both models, so that the
correlation can be read off its inputs.

\item[Room dimensions]
The median floor area varies across room types in the real data with a sample
standard deviation of \RealAreaSD~m$^2$. The generated rooms vary by a
comparable amount: the ratio of the model's standard deviation to the real one is
\AreaSDRatioQwenA{} for Qwen and \AreaSDRatioGemmaA{} for Gemma. A similar spread does not by itself show that the areas follow the real ones, as the placement of Gemma shows below. We therefore also regressed the four generated medians on the four real medians, as for the placement: the slope is \AreaSlopeQwen{} for Qwen and \AreaSlopeGemma{} for Gemma, and the Pearson correlation is \AreaPearsonQwen{} and \AreaPearsonGemma; with four room types these two values are descriptive. The room areas therefore follow the room type. Table~\ref{tab:app-area}
gives the area distribution per room type, sample size, median and
interquartile range (IQR), for the real data and both models. The ratio compares only the spread of the four medians, not their values. The medians differ from the real ones: Qwen's median area is larger than the real median in all four room types, and Gemma's median bathroom is smaller than the real one (Table~\ref{tab:app-area}). The generated dimensions take few distinct values. Over the \NGenPerRoom{} layouts of a room type, Qwen uses \SpreadDimsMinQwen--\SpreadDimsMaxQwen{} distinct pairs of width and depth and Gemma \SpreadDimsMinGemma--\SpreadDimsMaxGemma, whereas almost every real room has its own dimensions (\SpreadDimsMinReal--\SpreadDimsMaxReal{} distinct pairs); the most frequent pair accounts for \SpreadDimsTopMinGemma--\SpreadDimsTopMaxGemma\% of Gemma's layouts. Sec.~\ref{sec:searchcost} analyzes the effect of this on the search prior.

\item[Placement]

The wall-proximity excess (Sec.~\ref{sec:excess}) of the real rooms ranges from \ExcessRealBath{} in bathrooms to \ExcessRealLiving{} in living rooms, with a sample standard deviation of \RealSD{} across the four room types. The excess of Qwen's layouts has a standard deviation of \ModelSDQwenA{} across room types, so the ratio of the real standard deviation to Qwen's is \SDRatioQwenA{} (95\% CI \SDRatioQwenACI, bootstrap over rooms). A permutation test that
shuffles room-type labels across rooms does not reject the null
hypothesis that all four room types have the same excess
($p$~\HomogPQwenA, against $p$~\RealHomogP{} for the real data).  Qwen's excess therefore does not differ detectably across the four room types. The excess of Gemma's layouts has a standard deviation of \ModelSDGemmaA{} across room types, a ratio of \SDRatioGemmaA{} (95\% CI \SDRatioGemmaACI), and the permutation test rejects equality ($p$~\HomogPGemmaA). Gemma's excess therefore differs across room types; the paragraph on Gemma below examines whether it follows the real differences.

\end{description}

\begin{figure}[tbp]
  \centering
  \includegraphics[width=\linewidth]{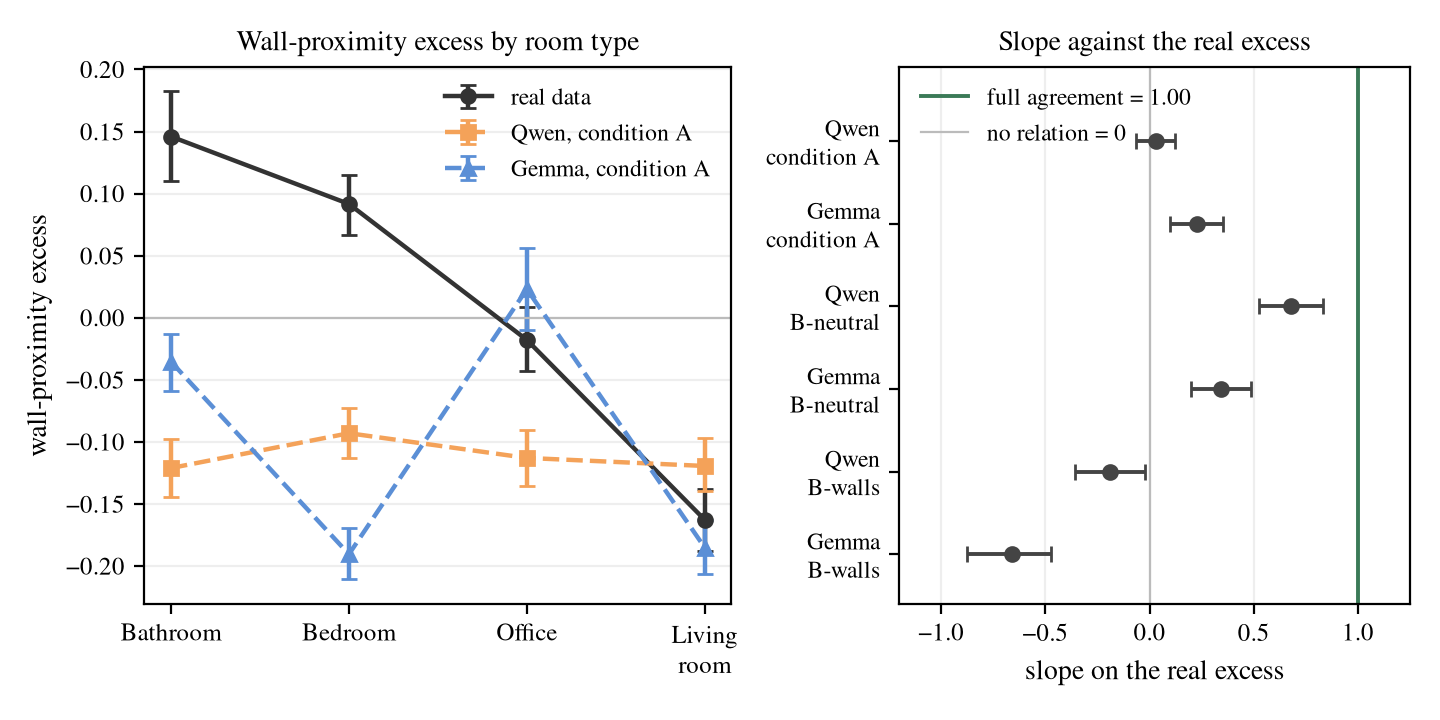}
  \caption{Left: wall-proximity excess by room type, ordered by the real
    data. The real excess varies with the room type; Qwen's is nearly constant; Gemma's varies, but not in the order of the real data. Right: the slope of the model's excess regressed on the real excess.
    Full agreement would be $1.00$; no interval reaches it. Bars are 95\%
    bootstrap intervals over rooms.}
  \label{fig:conditioning}
\end{figure}

Table~\ref{tab:hierarchy} summarizes the three properties, and Fig.~\ref{fig:conditioning} (left) shows the excess by room type: the real excess ranges from \ExcessRealBath{} to \ExcessRealLiving, Qwen's from \ExcessQwenAMinAll{} to \ExcessQwenAMaxAll, and Gemma's from \ExcessGemmaAMinAll{} to \ExcessGemmaAMaxAll. The following paragraphs analyze the placement further.

\begin{table}[tbp]
  \centering
  \caption{Three properties of conditioning, all measured on the same ScanNet
    set of real rooms and the same closed vocabulary, condition~\condA.
    The ratio of standard deviations is model over real for the room dimensions, so that a value above $1$ means that the model varies more than the real data, and real over model for the placement, so that a value above $1$ means that the model varies less than the real data.}
  \label{tab:hierarchy}
  \begin{tabular}{llcc}
    \toprule
    Level & Statistic & Qwen & Gemma \\
    \midrule
    Object identity & Spearman $\rho$ vs.\ real
      & \IdentRhoQwenA & \IdentRhoGemmaA \\
    Object identity & Spearman $\rho$ vs.\ real, other room types
      & \IdentMismatchQwen & \IdentMismatchGemma \\
    Room dimensions & SD ratio, model\,/\,real
      & \AreaSDRatioQwenA & \AreaSDRatioGemmaA \\
    Room dimensions & regression slope of the medians
      & \AreaSlopeQwen & \AreaSlopeGemma \\
    Placement & SD ratio, real\,/\,model
      & \SDRatioQwenA & \SDRatioGemmaA \\
    Placement & $p$, equality across room types
      & $p$~\HomogPQwenA & $p$~\HomogPGemmaA \\
    Placement & regression slope
      & \SlopeQwenA & \SlopeGemmaA \\
    \bottomrule
  \end{tabular}
\end{table}

\paragraph{Placement statistics of Gemma.}
Gemma's wall-proximity excess varies across room types with a standard deviation (\ModelSDGemmaA) close to the real one (\RealSD). However, the order of the room types by Gemma's excess is unrelated to the real order: the Spearman rank correlation between the two is \PlaceRhoGemmaA{} (Table~\ref{tab:app-slope}). The two models therefore differ: Qwen's excess varies little across room types, and Gemma's varies, in an order unrelated to the real one. What the two models have in common is that neither follows the real differences between room types, and we take only this common property as the result of this section. The form of the failure, no variation for Qwen and unrelated variation for Gemma, is a property of each model and may differ for other models.

\paragraph{Regression slope.}
The agreement between the model's excess and the real excess across the four room types is summarized by the slope of a regression of the former on the latter, which, unlike the rank correlation, uses the size of the differences between room types. Full agreement gives a slope of $1.00$; indifference to room
type gives $0$. We obtain \SlopeQwenA{} (95\% CI \SlopeQwenACI) for
Qwen and \SlopeGemmaA{} (\SlopeGemmaACI) for Gemma. The slope is an
unweighted least-squares fit with an intercept over the four room
types, and its interval resamples rooms; Appendix~\ref{app:slope}
gives both, and Appendix~\ref{app:permutation} explains the
permutation test. Fig.~\ref{fig:conditioning} (right) shows the slopes for the two models under condition~\condA{} and the two route conditions of Sec.~\ref{sec:overwrite}; none of the six intervals contains $1.00$.
The permutation test and the slope use
rooms as the unit of resampling ($n$ between \nRoomsMin{} and \nRoomsMax{} per
room type); with only four room types the rank correlations have almost no
power and we rest no claim on them.

\paragraph{A second placement statistic.}
The excess does not record which wall an object is near. We therefore repeated the measurement
on the \emph{orientation} of placement: in the canonical frame we assign each
object to its nearest wall and compare the distribution
$q(\mathrm{wall} \mid c)$ over $\{N, E, S, W\}$. Regressing the model's orientation profile on the real one, after
removing the mean level of each wall so that only room-type variation
remains, gives \OriAllQwenCentred{} (\OriAllQwenCentredCI) for Qwen and
\OriAllGemmaCentred{} (\OriAllGemmaCentredCI) for Gemma. Neither interval contains $1$. As for the excess, Qwen's orientation profile varies less across room types than the real one (ratio of real to model spread \OriSpreadAllQwen), and Gemma's varies more (ratio \OriSpreadAllGemma) with a negative slope.

The resolution of this statistic on ScanNet is limited, because the real orientation distributions there are close to uniform. In CubiCasa bathrooms, where the real distribution is far from uniform (only \CubiOriRealBathS{} of the objects on the door wall), Qwen's layouts under the three prompting conditions have between \CubiOriAMinS{} and \CubiOriAMaxS{} of their objects on the door wall, close to uniform. Appendix~\ref{app:orientation} gives both, with
Table~\ref{tab:orientation}, which places the four orientation slopes
alongside the wall-proximity slope.\footnote{The four orientation shares sum
to one, so the four slopes are not independent evidence. Rooms without an observed door cannot be canonicalized and are
excluded; the rate is \CanonRateBath--\CanonRateOffice{} for bathrooms,
bedrooms and offices but only \CanonRateLiving{} for living rooms.}

\subsection{Effect of choosing a prompt on the placements}
\label{sec:overwrite}

Condition~\condA{} asks for the whole layout in one response. We also generated layouts under the two route conditions of Sec.~\ref{sec:conditions}, in which the model reports one object per step: \condBw{} instructs it to walk along the walls clockwise and report the objects it passes, and \condBn{} instructs it to walk through the room without mentioning the walls. Comparing \condBw{} with \condBn{} separates the effect of the wall instruction from that of the step-by-step output.

Table~\ref{tab:conditions} summarizes the results. Under \condBw, the excess is positive in every room type for both models; in living rooms it is \ExcessQwenBwLiving{} for Qwen, against \ExcessRealLiving{} for the real rooms. The regression slope is negative (\SlopeQwenBw, CI \SlopeQwenBwCI{} for Qwen; \SlopeGemmaBw, CI \SlopeGemmaBwCI{} for Gemma), that is, the room types with a lower real excess receive a higher excess under \condBw. Under \condBn, the excess is negative in every room type for both models, and the slope is \SlopeQwenBn{} (CI \SlopeQwenBnCI) for Qwen and \SlopeGemmaBn{} (CI \SlopeGemmaBnCI) for Gemma, higher than under \condA{} but below $1$ (Table~\ref{tab:app-slope}).

\begin{table}[tbp]
  \centering
  \footnotesize
  \setlength{\tabcolsep}{4pt}
  \caption{What the prompt changes, condition~\condA{} against the two route
    conditions (ScanNet). The first four columns are the wall-proximity
    excess by room type; the real data runs from $+$ in bathrooms to $-$ in
    living rooms.
    Slope
    intervals are in Table~\ref{tab:app-slope}; the full excess table, with
    the permutation tests, is Table~\ref{tab:app-excess}, and the validity
    rates per room type are in Table~\ref{tab:app-validity}.}
  \label{tab:conditions}
  \begin{tabular}{lrrrrrrr}
    \toprule
    & \multicolumn{4}{c}{Wall-proximity excess} & & Outside & Overlap \\
    \cmidrule(lr){2-5}
    Population & Bath. & Bedr. & Office & Living & Slope & \% & rate \\
    \midrule
    Real data & \ExcessRealBath & \ExcessRealBed & \ExcessRealOffice
      & \ExcessRealLiving & --- & --- & --- \\
    \midrule
    Qwen \condA & \ExcessQwenABath & \ExcessQwenABed & \ExcessQwenAOffice
      & \ExcessQwenALiving & \SlopeQwenA
      & \OutsideQwenAMin--\OutsideQwenAMax
      & \OverlapQwenAMin--\OverlapQwenAMax \\
    Qwen \condBn & \ExcessQwenBnBath & \ExcessQwenBnBed & \ExcessQwenBnOffice
      & \ExcessQwenBnLiving & \SlopeQwenBn
      & \OutsideQwenBnMin--\OutsideQwenBnMax
      & \OverlapQwenBnMin--\OverlapQwenBnMax \\
    Qwen \condBw & \ExcessQwenBwBath & \ExcessQwenBwBed & \ExcessQwenBwOffice
      & \ExcessQwenBwLiving & \SlopeQwenBw
      & \OutsideQwenBwMin--\OutsideQwenBwMax
      & \OverlapQwenBwMin--\OverlapQwenBwMax \\
    \midrule
    Gemma \condA & \ExcessGemmaABath & \ExcessGemmaABed & \ExcessGemmaAOffice
      & \ExcessGemmaALiving & \SlopeGemmaA
      & \OutsideGemmaAMin--\OutsideGemmaAMax
      & \OverlapGemmaAMin--\OverlapGemmaAMax \\
    Gemma \condBn & \ExcessGemmaBnBath & \ExcessGemmaBnBed
      & \ExcessGemmaBnOffice & \ExcessGemmaBnLiving & \SlopeGemmaBn
      & \OutsideGemmaBnMin--\OutsideGemmaBnMax
      & \OverlapGemmaBnMin--\OverlapGemmaBnMax \\
    Gemma \condBw & \ExcessGemmaBwBath & \ExcessGemmaBwBed
      & \ExcessGemmaBwOffice & \ExcessGemmaBwLiving & \SlopeGemmaBw
      & \OutsideGemmaBwMin--\OutsideGemmaBwMax
      & \OverlapGemmaBwMin--\OverlapGemmaBwMax \\
    \bottomrule
  \end{tabular}
\end{table}

Under condition~\condBw{} between \OutsideQwenBwMin\% and
\OutsideQwenBwMax\% of Qwen's objects, and between
\OutsideGemmaBwMin\% and \OutsideGemmaBwMax\% of Gemma's, extend
beyond the rectangle of the room they were generated for, against
\OutsideQwenAMin--\OutsideQwenAMax\% and
\OutsideGemmaAMin--\OutsideGemmaAMax\% under \condA{}.
(Table~\ref{tab:conditions}). The fraction of overlapping object pairs does not increase under \condBw{} (\OverlapQwenBwMin--\OverlapQwenBwMax{} for Qwen, \OverlapGemmaBwMin--\OverlapGemmaBwMax{} for Gemma), so the objects outside the room are not the result of crowding. Under \condBw{}, the objects are therefore placed against the walls at the cost of extending beyond them.

The prompt thus changes the distribution of the wall-proximity excess over all room types, but under none of the three conditions does the excess follow the real differences between room types (slopes below $1$ for both models; Table~\ref{tab:app-slope}).

\subsection{Conditioning on an observed object}
\label{sec:observed}

The measurements above concern placement before any object of the room is known. A robot in a room observes some of its objects, and a search prior for the remaining objects can be conditioned on them; for example, the position of a chair can be predicted from that of an observed desk. This subsection measures whether the generated position of an object changes with the position of another object given in the prompt.

\paragraph{Method.}
We used ScanNet bedrooms that contain both objects of a pair (an \emph{anchor}
and a \emph{target}) and whose door can be recovered. Each room was presented to the model twice, with the room rectangle and the door, using the prompt of condition~C (the room is given and the model generates the objects; Appendix~\ref{app:prompts}) with one added line that lists the anchor. Under condition O-anchor the
prompt lists the anchor at its real position and footprint, with the
instruction not to move it or list it again; under condition O-move it lists
the anchor at a position on another wall, at the same distance from that
wall, chosen so that it overlaps no other object. The target is never
listed. Five layouts were generated per room and condition at temperature
$1.0$ (Appendix~\ref{app:prompts} gives the prompt). With $\Delta a$ the
displacement of the anchor between the two prompts and $\Delta t$ that of the
mean generated target position, the \emph{following slope} is
$\kappa = \sum \Delta t \cdot \Delta a / \sum \Delta a \cdot \Delta a$ over
rooms (Equation~\eqref{eq:beta}): $\kappa = 1$ means that the target moves by the same displacement as the anchor, $\kappa = 0$ that the target does not move, and a negative value that it moves in the opposite direction. Intervals resample rooms.

Three pairs and one reversed pair were compared. Desk~$\rightarrow$~chair
(\NANDeskChair{} rooms) and bed~$\rightarrow$~night stand
(\NANBedNightstand{} rooms) are pairs that real bedrooms keep adjacent: \RelPairNearBedDeskChair\% of the chairs and \RelPairNearBedBedNightstand\% of the night stands in bedrooms that also contain the anchor lie within \RelNear~m of one. Bed~$\rightarrow$~chair (\NANBedChair{} rooms) is the
negative control: in the real bedrooms, chairs are not closer to the bed than under the null placement of Appendix~\ref{app:relations}, which places each object at its real distance from a randomly chosen wall. Comparing the related pairs with the control distinguishes placement relative to a related object from placement near any object given in the prompt. Night stand~$\rightarrow$~bed (\NARevN{} rooms) reverses bed~$\rightarrow$~night stand, listing the night stand and measuring the bed. Rooms in which the anchor could not be moved to another wall without overlap were excluded: \NADropBedNightstand{} of \NABothBedNightstand{} rooms for bed~$\rightarrow$~night stand, \NADropBedChair{} of \NABothBedChair{} for bed~$\rightarrow$~chair and \NADropDeskChair{} of \NABothDeskChair{} for desk~$\rightarrow$~chair. Appendix~\ref{app:observed} gives the details.

\begin{figure}[tbp]
  \centering
  \includegraphics[width=\linewidth]{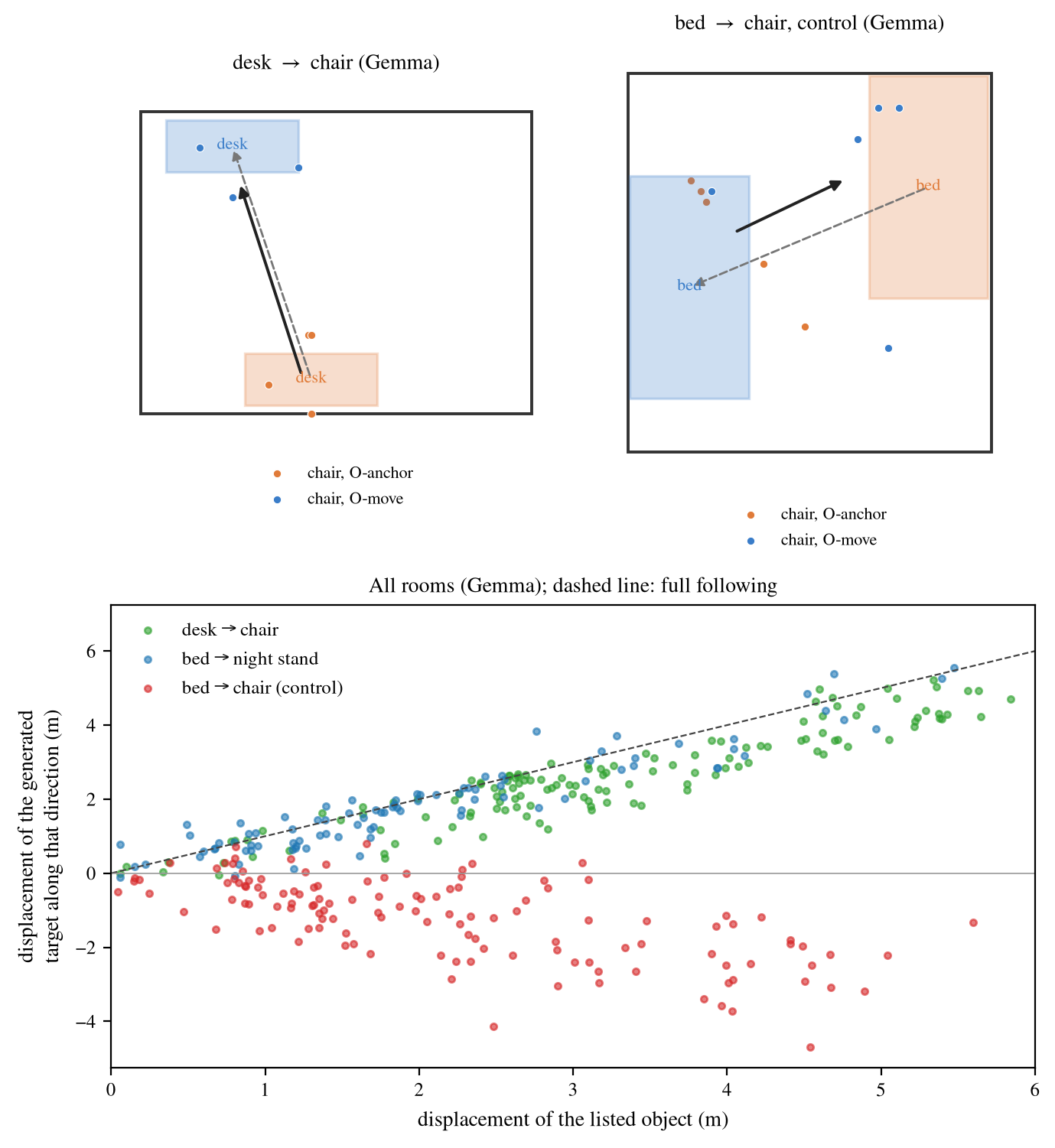}
  \caption{Generation with an observed object (Gemma; Qwen behaves alike, Table~\ref{tab:observed}). Top: one bedroom each. The listed object is drawn at its position under O-anchor (orange) and O-move (blue); dots are the generated target in the five samples of each prompt, the solid arrow joins the means of the target, and the dashed arrow the two positions of the listed object. The chair moves in the same direction as the desk (left) and in the opposite direction to the bed (right). Bottom: every room; the displacement of the generated target along the direction in which the listed object was moved, against that object's displacement. The dashed line is full following.}
  \label{fig:observed}
\end{figure}

\paragraph{Results.}
For the two related pairs, the target follows the anchor (Table~\ref{tab:observed}; Fig.~\ref{fig:observed}). When the
desk is moved, the generated chair moves with it by \NADeskChairQwen{} of
the desk's displacement for Qwen and \NADeskChairGemma{} for Gemma; when the
bed is moved, the night stand follows by \NABedNightstandQwen{} and
\NABedNightstandGemma. The negative control does the opposite: when the bed
is moved, the chair moves \emph{away} from it (\NABedChairQwen{} and
\NABedChairGemma). The generated objects of other categories also move away from the anchor: \NAOtherDeskChairQwen{} and \NAOtherDeskChairGemma{} in the desk~$\rightarrow$~chair runs, and \NAOtherBedNightstandQwen{} and \NAOtherBedNightstandGemma{} in the bed~$\rightarrow$~night stand runs. The difference between each related pair and the control is \NADiffDeskChairQwen{} \NADiffDeskChairCIQwen{} and \NADiffBedNightstandQwen{} \NADiffBedNightstandCIQwen{} for Qwen, and \NADiffDeskChairGemma{} \NADiffDeskChairCIGemma{} and \NADiffBedNightstandGemma{} \NADiffBedNightstandCIGemma{} for Gemma. The generated objects are therefore not moved toward every object given in the prompt: the target of a related pair moves with the anchor, and the other objects move away from it. The dependence is also directional. Listing the night stand and moving it leaves the bed where it was (\NARevQwen{} and \NARevGemma, against \NABedNightstandQwen{} and \NABedNightstandGemma{} in the forward direction; difference \NARevDiffQwen{} \NARevDiffCIQwen{} and \NARevDiffGemma{} \NARevDiffCIGemma): the dependent object is placed relative to the one it depends on, not the reverse.  

\begin{table}[tbp]
  \centering
  \caption{Following slope $\kappa$ in ScanNet bedrooms: how far the generated
    target moves when the listed anchor is moved to another wall, as a
    fraction of the anchor's displacement ($1$ = moves with it, negative =
    moves away). The difference rows subtract the negative control, with
    rooms present in both pairs resampled together. ``Other objects'' are the generated objects of neither the anchor's nor the target's category, in the runs of the pair named. The reversed pair lists the night stand and measures the bed. Room counts and intervals are in Table~\ref{tab:app-rel-nonadj}.}
  \label{tab:observed}
  \small
  \begin{tabular}{lcc}
    \toprule
    & Qwen & Gemma \\
    \midrule
    desk $\rightarrow$ chair & \NADeskChairQwen{} \NADeskChairCIQwen & \NADeskChairGemma{} \NADeskChairCIGemma \\
    bed $\rightarrow$ night stand & \NABedNightstandQwen{} \NABedNightstandCIQwen & \NABedNightstandGemma{} \NABedNightstandCIGemma \\
    bed $\rightarrow$ chair (control) & \NABedChairQwen{} \NABedChairCIQwen & \NABedChairGemma{} \NABedChairCIGemma \\
    \midrule
    desk $\rightarrow$ chair $-$ control & \NADiffDeskChairQwen{} \NADiffDeskChairCIQwen & \NADiffDeskChairGemma{} \NADiffDeskChairCIGemma \\
    bed $\rightarrow$ night stand $-$ control & \NADiffBedNightstandQwen{} \NADiffBedNightstandCIQwen & \NADiffBedNightstandGemma{} \NADiffBedNightstandCIGemma \\
    \midrule
        other objects, desk $\rightarrow$ chair runs & \NAOtherDeskChairQwen{} \NAOtherDeskChairCIQwen & \NAOtherDeskChairGemma{} \NAOtherDeskChairCIGemma \\
    other objects, bed $\rightarrow$ night stand runs & \NAOtherBedNightstandQwen{} \NAOtherBedNightstandCIQwen & \NAOtherBedNightstandGemma{} \NAOtherBedNightstandCIGemma \\
    night stand $\rightarrow$ bed (reversed) & \NARevQwen{} \NARevCIQwen & \NARevGemma{} \NARevCIGemma \\
    \bottomrule
  \end{tabular}
\end{table}

\paragraph{Range of validity of the result.}
In this experiment, the prompt gives the position of the anchor and nothing about the target, so the position of the target is chosen by the model. The result is limited in three respects. First, it was measured in bedrooms, for the three pairs and the reversed pair above; other room types and pairs were not measured with this design. Second, it holds for generation and not for the likelihood: in the pair perturbations of Sec.~\ref{sec:relladder}, which move a target beside a wrong neighbor or keep it beside its anchor in a real layout, the neighbor effect is \RelPairingQwenCubi{} (Qwen) and \RelPairingGemmaCubi{} (Gemma) on CubiCasa and is not distinguished from zero on ScanNet. Third, the result shows that the coordinates written for the target change with the coordinates of the anchor given in the prompt; it does not show how the target's position is computed; Sec.~\ref{sec:mechanism} analyzes the relation between the generated coordinates and the constraints given in the prompt further.

%% file: sections/mechanism.tex
\section{Analysis of Layout Generation Mechanisms}
\label{sec:mechanism}

On CubiCasa, the likelihood of a real kitchen is higher than that of a copy whose fixtures are moved away from the walls (Sec.~\ref{sec:ladder}). On ScanNet, the generated layouts have a lower wall-proximity excess than the real rooms in bathrooms and bedrooms (for Qwen, \ExcessQwenABath{} and \ExcessQwenABed{} against \ExcessRealBath{} and \ExcessRealBed; Table~\ref{tab:conditions}), and the excess does not follow the room type (Sec.~\ref{sec:hierarchy}). This section analyzes the generated coordinates: how much of their likelihood depends on wall contact (Sec.~\ref{sec:adsorption}), whether they agree with the walls that the model declares (Sec.~\ref{sec:symbols}), and whether they keep the objects inside the room (Sec.~\ref{sec:containment}). Sec.~\ref{sec:mechsummary} summarizes the results.

\subsection{Snapping generated objects to their declared walls}
\label{sec:adsorption}
\label{sec:entropy}

This subsection measures how much of the likelihood difference between a generated layout and an edited copy of it depends on the contact of the objects with the walls. We edited each model's own generations for CubiCasa kitchens and bathrooms under condition~\condA{} (\NGenPerRoom{} layouts per room type and model) and scored each edited version against the original with the symmetrized score of
Sec.~\ref{sec:setup}, using the model that generated the layouts. Appendix~\ref{app:adsorption} defines every version; Table~\ref{tab:adsorption} summarizes the comparisons, Table~\ref{tab:app-adsorption} gives them with intervals, and Table~\ref{tab:app-adsorption-diag} gives the diagnostics used below. The comparison with real rooms at the end of the subsection uses the CubiCasa set of real rooms.

\begin{table}[tbp]
  \centering
  \caption{Wall-snapping control on the models' own generations (Qwen and Gemma,
    condition~\condA, CubiCasa, \NGenPerRoom{} layouts per room type and model, symmetrized; each model scores its own generations).
    Each row compares two versions of the same layout: pairwise accuracy is the
    fraction of layouts in which the first-named version scores higher, and
    $\Delta$ its mean margin in nats. The random version moves each object by
    the same distance as snapping, in a random direction; the other-wall
    version snaps to a wall other than the declared one and rewrites the
    declaration (Appendix~\ref{app:adsorption}). Intervals and tie rates are
    in Table~\ref{tab:app-adsorption}.}
  \label{tab:adsorption}
  \begin{tabular}{llcccc}
    \toprule
    & & \multicolumn{2}{c}{Kitchens} & \multicolumn{2}{c}{Bathrooms} \\
    \cmidrule(lr){3-4} \cmidrule(lr){5-6}
    Model & Comparison & Pairwise acc. & $\Delta$ & Pairwise acc. & $\Delta$ \\
    \midrule
    Qwen & snapped $>$ original & \AdsAdsVsOrigKitchen & \AdsDAdsVsOrigKitchen
      & \AdsAdsVsOrigBath & \AdsDAdsVsOrigBath \\
    & random $>$ original & \AdsRandVsOrigKitchen & \AdsDRandVsOrigKitchen
      & \AdsRandVsOrigBath & \AdsDRandVsOrigBath \\
    & snapped $>$ random & \AdsAdsVsRandKitchen & \AdsDAdsVsRandKitchen
      & \AdsAdsVsRandBath & \AdsDAdsVsRandBath \\
    & declared wall $>$ other wall & \AdsAdsVsWrongWallKitchen & \AdsDAdsVsWrongWallKitchen
      & \AdsAdsVsWrongWallBath & \AdsDAdsVsWrongWallBath \\
    & other wall $>$ random & \AdsWrongWallVsRandKitchen & \AdsDWrongWallVsRandKitchen
      & \AdsWrongWallVsRandBath & \AdsDWrongWallVsRandBath \\
    \midrule
    Gemma & snapped $>$ original & \AdsAdsVsOrigKitchenGemma & \AdsDAdsVsOrigKitchenGemma
      & \AdsAdsVsOrigBathGemma & \AdsDAdsVsOrigBathGemma \\
    & random $>$ original & \AdsRandVsOrigKitchenGemma & \AdsDRandVsOrigKitchenGemma
      & \AdsRandVsOrigBathGemma & \AdsDRandVsOrigBathGemma \\
    & snapped $>$ random & \AdsAdsVsRandKitchenGemma & \AdsDAdsVsRandKitchenGemma
      & \AdsAdsVsRandBathGemma & \AdsDAdsVsRandBathGemma \\
    & declared wall $>$ other wall & \AdsAdsVsWrongWallKitchenGemma & \AdsDAdsVsWrongWallKitchenGemma
      & \AdsAdsVsWrongWallBathGemma & \AdsDAdsVsWrongWallBathGemma \\
    & other wall $>$ random & \AdsWrongWallVsRandKitchenGemma & \AdsDWrongWallVsRandKitchenGemma
      & \AdsWrongWallVsRandBathGemma & \AdsDWrongWallVsRandBathGemma \\
    \bottomrule
  \end{tabular}
\end{table}

\begin{figure}[tbp]
  \centering
  \includegraphics[width=\linewidth]{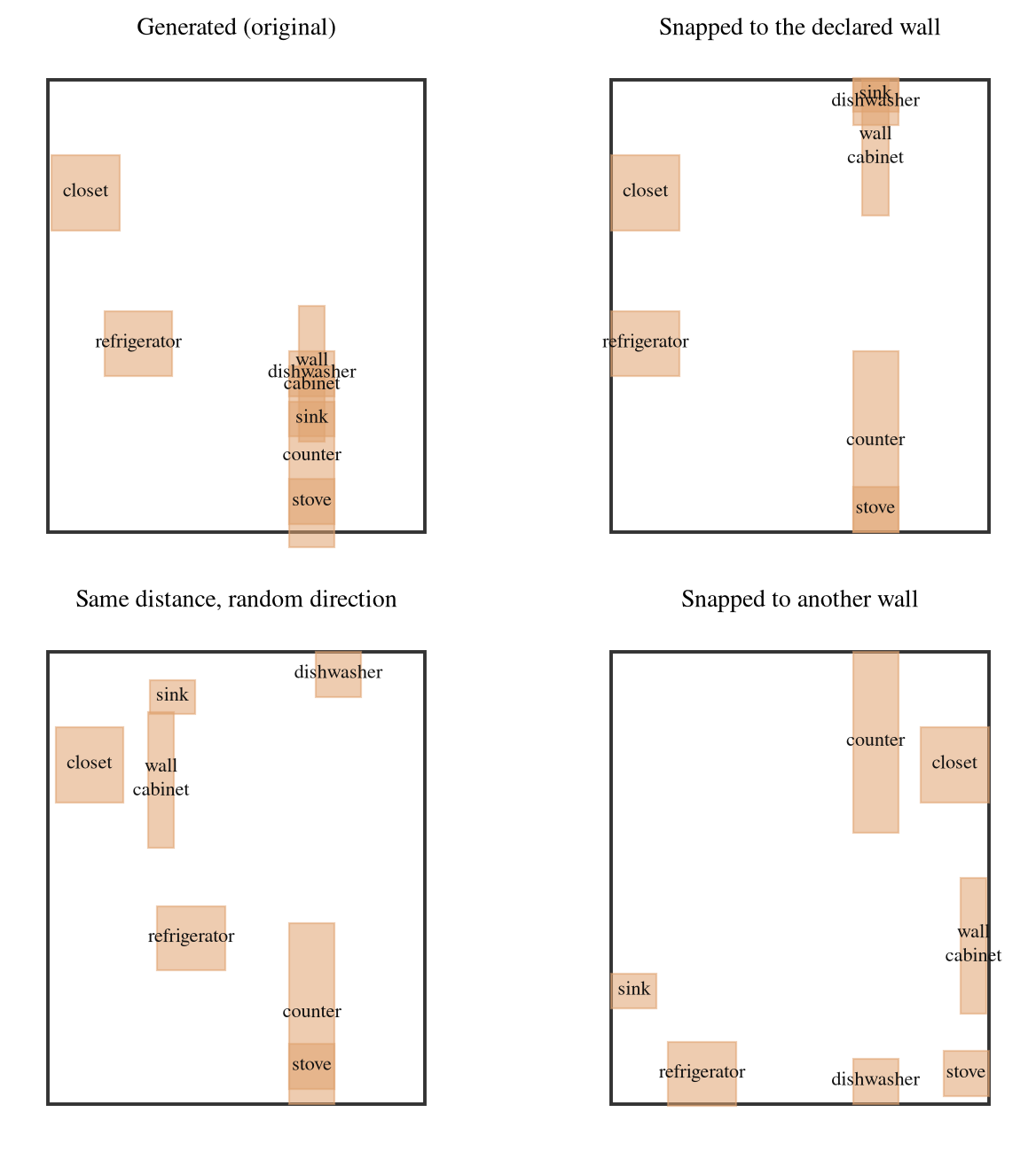}
  \caption{The four versions of one generated kitchen used in the wall-snapping control (Qwen, condition~\condA). Snapping moves each object that declares a wall perpendicularly onto that wall; the random version moves each object by the same distance in a random direction; the other-wall version snaps to a different wall and rewrites the declaration.}
  \label{fig:adsorb}
\end{figure}

\paragraph{Wall-snapping control.}
In the \emph{snapped} version (Fig.~\ref{fig:adsorb}), every object that declares a wall is moved
perpendicularly to that wall until its face touches it; the object list, the
sizes, the room and the declaration itself are unchanged, and objects
declared as touching no wall are not moved. Over the objects that move, the
displacement has a median of \AdsMoveMedianKitchen~m in kitchens (mean \AdsMoveMeanKitchen~m) for Qwen and \AdsMoveMedianKitchenGemma~m (mean \AdsMoveMeanKitchenGemma~m) for Gemma. After snapping, every object that declares a wall touches it, so the declarations agree with the coordinates. The likelihood of the snapped version is nevertheless lower than that of the original in most layouts: the fraction of layouts in which the snapped version scores higher is \AdsAdsVsOrigKitchen{} in kitchens and \AdsAdsVsOrigBath{} in bathrooms for Qwen and \AdsAdsVsOrigKitchenGemma{} and \AdsAdsVsOrigBathGemma{} for Gemma, and the mean difference in favor of the original is \AdsDeltaAdsVsOrigKitchen{} and \AdsDeltaAdsVsOrigBath{} nats for Qwen and \AdsDeltaAdsVsOrigKitchenGemma{} and \AdsDeltaAdsVsOrigBathGemma{} nats for Gemma.

\paragraph{Distance-matched random control.}
To separate the effect of wall contact from that of the displacement itself, the \emph{random} version
moves each object by the same distance as the snapped version would, in a
random direction. In kitchens, the random version costs \AdsDeltaRandVsOrigKitchen{} nats
relative to the original for Qwen, of which the snapped version recovers
\AdsDeltaAdsVsRandKitchen{} (pairwise accuracy \AdsAdsVsRandKitchen{} for snapped
over random); for Gemma, the random version costs \AdsDeltaRandVsOrigKitchenGemma{} nats, of which the snapped version recovers \AdsDeltaAdsVsRandKitchenGemma{} (pairwise accuracy \AdsAdsVsRandKitchenGemma). For both models, most of the likelihood lost by a displacement of this size is recovered when the displacement places the objects against their declared walls; a difference of \AdsDeltaAdsVsOrigKitchen{} nats (Qwen) and \AdsDeltaAdsVsOrigKitchenGemma{} nats (Gemma) in favor of the original remains.

\paragraph{Other-wall control.}
In the \emph{other-wall} version, each object is snapped to a wall other than the one it declared, and the declaration is rewritten to name the new wall, so that the version is consistent. Comparing it with the snapped version measures whether the remaining difference depends on which wall an object is snapped to. Snapping to the declared wall wins over snapping to another wall at
\AdsAdsVsWrongWallKitchen{} in kitchens and \AdsAdsVsWrongWallBath{} in
bathrooms for Qwen and at \AdsAdsVsWrongWallKitchenGemma{} and \AdsAdsVsWrongWallBathGemma{} for Gemma. The likelihood therefore depends on whether an object is placed against its declared wall, and the remaining difference between the snapped version and the original is not explained by the choice of wall. The two models differ in the comparison of the other-wall version with the random version: the other-wall version scores higher in a fraction \AdsWrongWallVsRandKitchen{} of the kitchens for Qwen and \AdsWrongWallVsRandKitchenGemma{} for Gemma (mean margins \AdsDWrongWallVsRandKitchen{} and \AdsDWrongWallVsRandKitchenGemma{} nats). For Gemma, moving an object onto a wall other than the declared one, with the declaration rewritten, lowers the likelihood more than moving it by the same distance in a random direction.

\paragraph{Repetition of coordinate values.}
All four versions are quantized to the same \AdsGrid~m lattice before
scoring, so lattice membership cannot separate them, and snapping does not
change the length of the layout: the median token count is \AdsNtokKitchen{} (Qwen) and \AdsNtokKitchenGemma{} (Gemma)
for every version in kitchens. What changes is the number of distinct numerical values. Across the \AdsNValuesKitchen{} coordinate and footprint values of Qwen's \NGenPerRoom{} generated kitchens, the model's own layouts use
\AdsDistinctOrigKitchen{} distinct values and their snapped copies
\AdsDistinctAdsKitchen{} (\AdsDistinctOrigBath{} against
\AdsDistinctAdsBath{} in bathrooms); the random version raises the count
again, to \AdsDistinctRandKitchen. For Gemma, the corresponding counts over \AdsNValuesKitchenGemma{} values are \AdsDistinctOrigKitchenGemma{}, \AdsDistinctAdsKitchenGemma{} and \AdsDistinctRandKitchenGemma{} in kitchens (\AdsDistinctOrigBathGemma, \AdsDistinctAdsBathGemma{} and \AdsDistinctRandBathGemma{} in bathrooms). For both models, the order of the three versions by likelihood (original, snapped, random) is the reverse of their order by the number of distinct values. A likely reason is that a coordinate determined by the room dimensions and the footprint, as produced by snapping, is often a value that does not occur elsewhere in the layout, whereas values that the model has already written receive a high probability.

The number of distinct values does not explain all differences. For Qwen, the snapped version has a higher likelihood than the other-wall version although it uses more distinct values (\AdsDistinctAdsKitchen{} against \AdsDistinctWrongKitchen). For Gemma, the other-wall version uses fewer distinct values than the random version (\AdsDistinctWrongKitchenGemma{} against \AdsDistinctRandKitchenGemma) but has a lower likelihood. In both cases, the wall to which an object is snapped also matters. Between the original and its snapped copy, the difference in the number of distinct values is the larger of the two effects.

\paragraph{Comparison with real rooms.}
The likelihood of Qwen's generated layouts is higher than that of real CubiCasa rooms with the same number of objects, by \EntGapRawKitchen{} nats in kitchens and \EntGapRawBath{} nats in bathrooms on average (Table~\ref{tab:app-entropy}); for Gemma, scored with Gemma, the differences are \EntGapRawKitchenGemma{} and \EntGapRawBathGemma{} nats. The two sets differ in the granularity of their coordinates: \CoordGridModelKitchen\% of Qwen's
kitchen coordinates are multiples of \EntGrid~m, and across the \NGenPerRoom{}
 generated kitchens it uses only \CoordDistinctModelKitchen{} distinct
coordinate values, where the real kitchens use \CoordDistinctRealKitchen{}
with only \CoordGridRealKitchen\% on the lattice
(Table~\ref{tab:app-coordvalues}). Gemma's kitchen coordinates are also all on the lattice (\CoordGridModelKitchenGemma\%) and take \CoordDistinctModelKitchenGemma{} distinct values. (This count is over $x$ and $y$ of the
generations as emitted; the counts in the preceding paragraph are over the
quantized layouts and include the footprints, which is why the two differ by
a few values.)

To measure the effect of this difference, both sets of layouts were quantized to a common grid
(Appendix~\ref{app:granularity} gives the matching and quantization rules).
The likelihood difference decreases as the grid becomes coarser. For Qwen, in kitchens
\EntGapRawKitchen{} raw, \EntGapQFiveKitchen{} at $0.05$~m,
\EntGapQTenKitchen{} at $0.10$~m and \EntGapQTwentyFiveKitchen{} at $0.25$~m;
in bathrooms \EntGapRawBath{}, \EntGapQFiveBath{}, \EntGapQTenBath{} and
\EntGapQTwentyFiveBath{}, where the ordering reverses at the coarsest grid.
For Gemma, the corresponding values are \EntGapRawKitchenGemma, \EntGapQFiveKitchenGemma, \EntGapQTenKitchenGemma{} and \EntGapQTwentyFiveKitchenGemma{} in kitchens and \EntGapRawBathGemma, \EntGapQFiveBathGemma, \EntGapQTenBathGemma{} and \EntGapQTwentyFiveBathGemma{} in bathrooms; the difference decreases to less than half of the raw value but does not vanish at the coarsest grid.
For both models, the raw comparison between generated and real layouts therefore depends strongly on the granularity of the numerals, and for Qwen it is almost entirely explained by it; we do not use it to support any claim. The other comparisons in this paper are between a layout and a perturbed copy of itself, which keeps the distribution of the numerical values approximately but not exactly fixed; the remaining difference of \AdsDeltaAdsVsOrigKitchen{} nats (Qwen) and \AdsDeltaAdsVsOrigKitchenGemma{} nats (Gemma) between the original and the snapped version shows that the effect of the numerical values can be of the same order as the effects measured.

In summary, in kitchens, snapping to the declared wall recovers \AdsDeltaAdsVsRandKitchen{} of the \AdsDeltaRandVsOrigKitchen{} nats that a random displacement of the same distance costs for Qwen, and \AdsDeltaAdsVsRandKitchenGemma{} of \AdsDeltaRandVsOrigKitchenGemma{} nats for Gemma. The remaining \AdsDeltaAdsVsOrigKitchen{} and \AdsDeltaAdsVsOrigKitchenGemma{} nats are not explained by the choice of wall and agree with the difference in the number of distinct values. The original is, however, a sample drawn from the model that scores it, so an edit of any kind is expected to lower its likelihood; the remaining difference therefore cannot be attributed to the numerical values alone. The granularity of the coordinates also has a large effect on the likelihood difference between generated and real layouts, so we do not use that difference as evidence.

\subsection{Consistency between declared walls and coordinates}
\label{sec:symbols}

This subsection examines whether the wall that the model declares for an object (the \texttt{wall\_contact} field) agrees with the wall against which the emitted coordinates place the object, and how the agreement changes with the order of the output fields and with walls given in the prompt. The data are generations for the four ScanNet room types under the four conditions defined below, for both models. Appendix~\ref{app:consistency} defines the measures, and Table~\ref{tab:app-symcoord} gives every value with its sample size.

\paragraph{Generation conditions.}
In the schema of condition~\condA{}, \texttt{center\_m} precedes \texttt{wall\_contact} (Appendix~\ref{app:schemas}), so the model writes the coordinates of an object before it declares the wall. Conditions~\condAp{} and~\condApp{} change this order of information, one factor at a time, and condition~\condApp-i is the control for~\condApp.
\begin{description}
  \item[\condA] The condition of Sec.~\ref{sec:conditions}: the model generates the whole layout, and each object's coordinates precede its wall. There are \NGenPerRoom{} layouts per room type.
  \item[\condAp] The field order of the schema is reversed: the wall is emitted first and the coordinates after it. The prompt, the vocabulary, the temperature and the model are unchanged. There are \NGenPerRoom{} layouts per room type.
  \item[\condApp] The room rectangle, the object list, each object's footprint and the wall that each object must touch are given in the prompt, and the model emits only the centers. The walls are those of the real objects, so the measurement does not depend on the model's own choice of walls. The inputs are \AppInUsedTotal{} real ScanNet rooms, selected as described in Appendix~\ref{app:consistency} (Table~\ref{tab:app-appinputs}).
  \item[\condApp-i] The same rooms, objects and footprints as~\condApp, without the walls. The sentence requiring each object to touch its wall is replaced by one requiring it to lie entirely within the room. Consistency is measured against the same real walls as under~\condApp.
\end{description}

\paragraph{Measures.}
\emph{Consistency} is the fraction of objects with a declared or given wall whose face lies within $0.15$~m of that wall (Equation~\eqref{eq:gap}). Its chance level is the consistency expected if the walls were assigned independently of the coordinates with the same distribution over walls (Equation~\eqref{eq:chance}). A second measure, without a distance threshold, compares each declared wall with the wall nearest to the object's coordinates (Appendix~\ref{app:declgeo}): \emph{agreement} is the fraction of objects whose declared wall is their nearest wall, and the Jensen--Shannon divergence compares the distribution of declared walls with that of nearest walls over all objects.

\paragraph{Results under condition~\condA.}
The declared walls carry information about the walls: snapping objects to their declared walls is preferred by the likelihood to snapping them to other walls (Sec.~\ref{sec:adsorption}). The coordinates emitted with the declarations, however, mostly do not satisfy them. The consistency is \ConsQwenA{} for Qwen and \ConsGemmaA{} for Gemma, against chance levels of \ConsRandQwenA{} and \ConsRandGemmaA, and the median distance between an object's face and its declared wall is \ConsGapMedQwenA~m for Qwen. The measure without a threshold shows the same separation between the aggregate and the individual object (Table~\ref{tab:app-declgeo}). Over all objects, the distribution of declared walls and that of nearest walls are almost identical (Jensen--Shannon divergence \DeclJSDQwenA{} nats for Qwen). For the individual object, the declared wall is the nearest wall for \DeclAgreeQwenA\% of objects, against \DeclChanceQwenA\% if the declarations were independent of the coordinates. The model thus reproduces how often each wall is used, both in its declarations and in its coordinates, but the two do not agree object by object; an evaluation based only on the distribution of walls would not detect this. Under the wall-following prompt~\condBw{} of Sec.~\ref{sec:conditions}, the agreement rises to \DeclAgreeGemmaBw\% for Gemma, with the objects outside the room reported in Sec.~\ref{sec:overwrite}.

\begin{figure}[tbp]
  \centering
  \includegraphics[width=\linewidth]{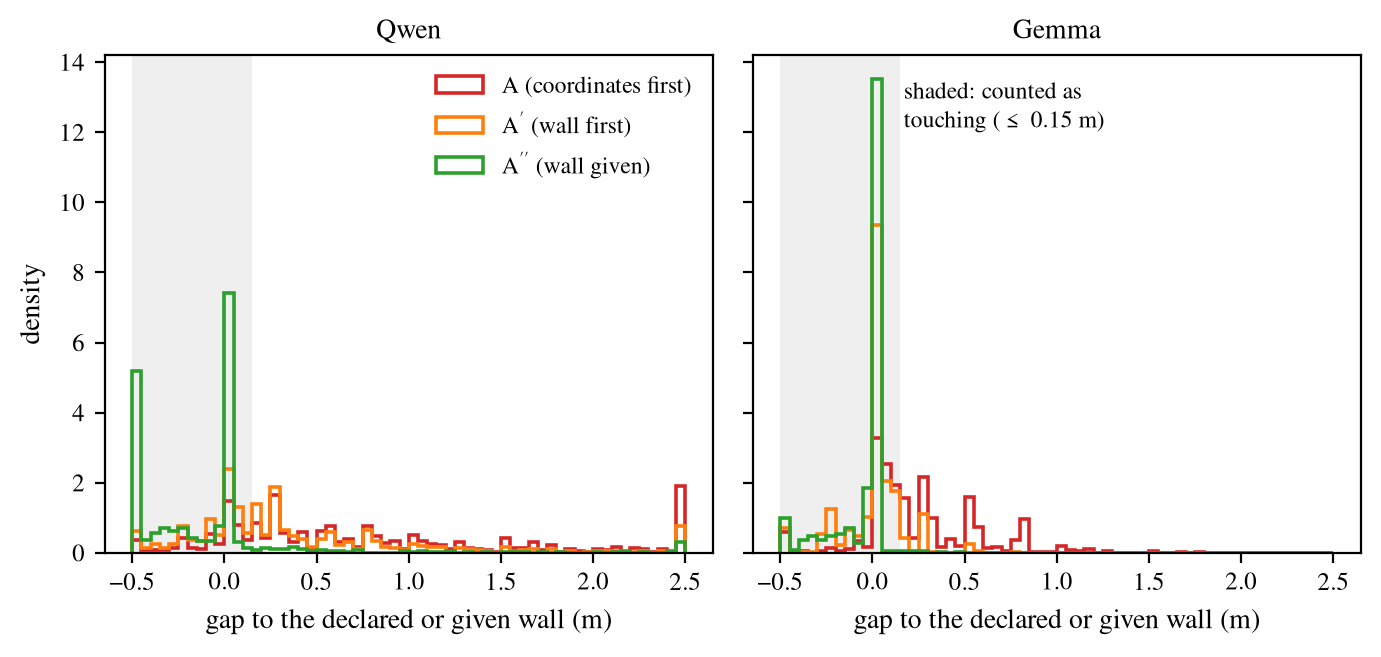}
  \caption{Distance between each object's face and the wall it declares (\condA, \condAp) or is given (\condApp), over the four ScanNet room types. The shaded band is the $0.15$~m within which an object counts as touching the wall. Emitting the wall first moves part of the mass into the band; supplying the wall moves most of it there.}
  \label{fig:symgap}
\end{figure}

\paragraph{Results under conditions~\condAp{} and~\condApp.}
Reversing the field order (\condAp) and giving the wall in the prompt (\condApp) both raise the consistency, and the second raises it almost to one (Table~\ref{tab:symcoord}; Fig.~\ref{fig:symgap}). Under~\condAp, the consistency is \ConsQwenAp{} for Qwen and \ConsGemmaAp{} for Gemma, against \ConsQwenA{} and \ConsGemmaA{} under~\condA. Under~\condApp, it is \ConsQwenApp{} and \ConsGemmaApp. The models therefore place an object against a wall that is stated before its coordinates are written, whether the statement comes from the prompt or from the model's own preceding field, but the walls they declare after writing the coordinates mostly do not describe those coordinates. Sec.~\ref{sec:design} uses this result in guideline~5.

\begin{table}[tbp]
  \centering
  \caption{Consistency between the wall an object is said to touch and the wall
    its coordinates place it against, within $0.15$~m. ScanNet, four room
    types. Under \condA{} and \condAp{} the wall is the model's own declaration; under \condApp{} and \condApp-i it is the wall of the real object, given to the model under \condApp{} and withheld under \condApp-i. Chance is computed from each condition's own distribution over walls, not as $1/4$ (Appendix~\ref{app:consistency}). Sample sizes and
    the remaining conditions are in Table~\ref{tab:app-symcoord}.}
  \label{tab:symcoord}
  \begin{tabular}{lcccc}
    \toprule
    & \condA & \condAp & \condApp & \condApp-i \\
    & \footnotesize coords first & \footnotesize wall first
      & \footnotesize wall given & \footnotesize wall withheld \\
    \midrule
    Qwen & \ConsQwenA & \ConsQwenAp & \ConsQwenApp & \ConsQwenAppi \\
    Gemma & \ConsGemmaA & \ConsGemmaAp & \ConsGemmaApp & \ConsGemmaAppi \\
    \midrule
    chance (Qwen) & \ConsRandQwenA & \ConsRandQwenAp & \ConsRandQwenApp & \ConsRandQwenAppi \\
    objects outside the room, Qwen
      & \ConsOutsideQwenA\% & \ConsOutsideQwenAp\% & \ConsOutsideQwenApp\% & \ConsOutsideQwenAppi\% \\
    \bottomrule
  \end{tabular}
\end{table}

\paragraph{Results under condition~\condApp-i.}
Condition~\condApp-i is the reference for \condApp{}; condition~\condA{} is not, because it measures agreement with the model's own declarations on rooms and objects that the model chose. Under \condApp-i, without the wall in the prompt, objects land on the wall the real object occupied in \ConsQwenAppi{} of cases for Qwen and \ConsGemmaAppi{} for Gemma, close to the chance levels of \ConsRandQwenAppi{} and \ConsRandGemmaAppi, against \ConsQwenApp{} and \ConsGemmaApp{} with the wall given (Table~\ref{tab:symcoord}). The consistency under \condApp{} is therefore produced by the wall instruction, not by the object list or the footprints.

\paragraph{Placement statistics under condition~\condAp.}
Condition~\condAp{} produces whole layouts, so the conditioning measurement
of Sec.~\ref{sec:hierarchy} can be repeated on them
(Table~\ref{tab:app-condap}). Agreement with the room type improves: the
regression slope goes from \CondSlopeQwenA{} to \CondSlopeQwenAp{}
(CI \CondSlopeQwenApCI) for Qwen and from \CondSlopeGemmaA{} to
\CondSlopeGemmaAp{} (\CondSlopeGemmaApCI) for Gemma. Neither interval reaches $1.00$. For Qwen the four room types now fall in the real order (rank correlation \CondRhoQwenAp{} with the real excesses), but the spread between them is only \CondSpanPctQwenAp\% of the real spread. Consistency with its own declarations does not imply that the declarations follow the room type: under \condAp{} Gemma's wall-proximity excess exceeds the real value in \CondOverRealGemmaAp{} of $4$ room types, as under the wall-following prompt \condBw, and the order of the room types is not preserved (rank correlation \CondRhoGemmaAp). The result of Sec.~\ref{sec:hierarchy}, that the placements do not follow the real differences between room types, therefore holds under \condAp{} as well; the value of the slope depends on the order of the output fields, and Table~\ref{tab:app-condap} gives both.

\subsection{Containment of objects within the room}
\label{sec:containment}

This subsection measures whether the generated objects lie inside the room when the room rectangle and the object footprints are given in the prompt. The inputs are the \AppInUsedTotal{} real ScanNet rooms of condition~\condApp{} (Sec.~\ref{sec:symbols}). Under~\condApp, Qwen places \ConsOutsideQwenApp\% of the objects partly outside the room, against \ConsOutsideQwenA\% under~\condA. Under~\condApp, each object must touch a given wall and lie inside the room, so the objects outside the room could result from the wall requirement. To test this, we removed the wall requirement (condition~\condApp-i) and varied the containment instruction (conditions~\condApp-c, \condApp-r and~\condApp-per).

\begin{figure}[tbp]
  \centering
  \includegraphics[width=\linewidth]{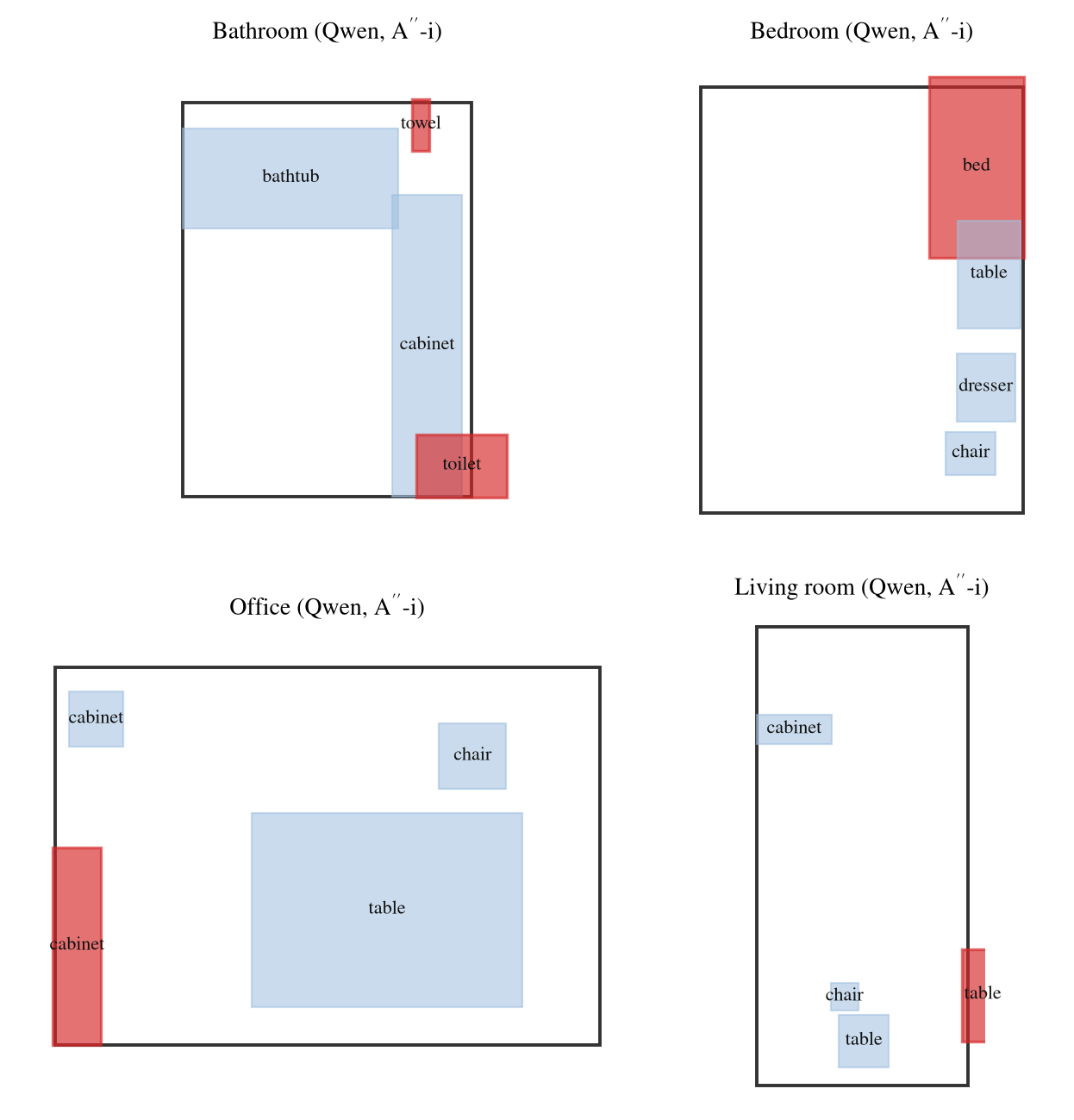}
  \caption{Layouts generated by Qwen under condition~\condApp-i, which gives the room, the objects and their footprints and asks only that every object lie inside the room. Objects that extend beyond the room are shown in red. One example per room type.}
  \label{fig:contain}
\end{figure}

\paragraph{Containment without the wall requirement.}
Under condition~\condApp-i of Sec.~\ref{sec:symbols}, the room, the object list and each object's footprint are given, the wall requirement is removed, and the only instruction on placement is that each object lie entirely within the room. Qwen places \ConsOutsideQwenAppi\% of objects partly outside the room, and Gemma \ConsOutsideGemmaAppi\% (Table~\ref{tab:containment}, last row; Fig.~\ref{fig:contain}). Paired by room, the rate changes by \PrioDAppiQwen{} \PrioDAppiQwenCI{} for Qwen and \PrioDAppiGemma{} \PrioDAppiGemmaCI{} for Gemma relative to \condApp{} (Table~\ref{tab:app-priority}). With the same room weighting the \condApp{} rate is \PrioBaseQwen\% for Qwen and \PrioBaseGemma\% for Gemma, so removing the wall requirement removes \PrioShareAppiQwen\% and \PrioShareAppiGemma\% of the objects outside the room, and the remaining ones occur without any other requirement.

\begin{table}[tbp]
  \centering
  \caption{Objects placed partly outside the room, on identical inputs
    (ScanNet, four room types). Differences are paired by room against
    \condApp{} with a bootstrap over rooms (Table~\ref{tab:app-priority}).
    The instruction added to \condApp{} is given for each condition (Appendix~\ref{app:prompts}).}
  \label{tab:containment}
  \begin{tabular}{llcc}
    \toprule
    Condition & Instruction added & Qwen & Gemma \\
    \midrule
    \condApp & wall given, nothing about containment
      & \ConsOutsideQwenApp\% & \ConsOutsideGemmaApp\% \\
    \condApp-c & ``keep the fixture inside the room''
      & \ConsOutsideQwenAppc\% & \ConsOutsideGemmaAppc\% \\
    \condApp-r & ``keep the fixture touching its wall''
      & \ConsOutsideQwenAppr\% & \ConsOutsideGemmaAppr\% \\
    \condApp-per & containment stated per object
      & \ConsOutsideQwenAppper\% & \ConsOutsideGemmaAppper\% \\
    \condApp-i & no wall requirement at all
      & \ConsOutsideQwenAppi\% & \ConsOutsideGemmaAppi\% \\
    \bottomrule
  \end{tabular}
\end{table}

The object lies inside the room under the following condition: for a room $W \times D$ and an object $w \times d$, the center $(x, y)$ must satisfy $w/2 \le x \le W - w/2$ and $d/2 \le y \le D - d/2$, and both rectangles are stated in the prompt. Of the \AppFeasObjectsQwen{} objects supplied,
\AppFeasInfeasibleQwen{} has a footprint larger than its room
(Appendix~\ref{app:consistency}), so infeasible inputs do not explain the
rate. The objects outside the room under~\condApp-i are therefore cases in which a single stated constraint is not satisfied.

\paragraph{Prompt interventions.}
Three further conditions add one instruction each to condition~\condApp, on the same inputs (Appendix~\ref{app:prompts} gives the prompts). Condition~\condApp-c adds that each object must touch its wall and lie entirely within the room and that, if both cannot be satisfied, it must stay inside the room. Condition~\condApp-r adds the same sentence with the priority reversed, so that the wall takes precedence. Condition~\condApp-per attaches the containment requirement to every item line of the object list, in the same way as the wall requirement. Table~\ref{tab:containment} gives the rates of objects outside the room, and Table~\ref{tab:app-priority} the differences from~\condApp{}, paired by room. For Qwen, \condApp-c changes the rate by \PrioDAppcQwen{} (\PrioDAppcQwenCI) and \condApp-r by \PrioDApprQwen{} (\PrioDApprQwenCI); both intervals contain zero, and the two priorities give the same result. \condApp-per changes the rate by \PrioDAppperQwen{} (\PrioDAppperQwenCI) and leaves \ConsOutsideQwenAppper\% of objects outside the room. For Gemma, whose rate under~\condApp{} is \ConsOutsideGemmaApp\%, the three conditions change it by \PrioDApprGemma{} to \PrioDAppperGemma{}, with intervals excluding zero (Table~\ref{tab:app-priority}). None of the instructions brings the rate close to zero for either model.

The wall instruction, in contrast, is followed also when the wall is atypical for the category. Under~\condApp, the consistency for categories that are rarely against a wall in the real data is \ConsUnnaturalQwen, and that for categories that usually are is \ConsNaturalQwen{} (Qwen). Objects given with ``none'' (no wall) are placed away from the walls: \NoneNearWallQwen{} of them end up within $0.15$~m of one for Qwen and \NoneNearWallGemma{} for Gemma, against \NoneBaselineQwen{} expected if the same objects were placed uniformly in their rooms (Equation~\eqref{eq:baseline}; Table~\ref{tab:app-symaux}; Appendix~\ref{app:consistency}). The models thus satisfy a given wall but not a given containment constraint.

\paragraph{Implications for coordinate-generating pipelines.}
This result concerns the computation of coordinates that satisfy a stated geometric constraint, not the positions at which objects are usually found. We measured it for room layouts only, but the same computation is required whenever a language model is asked for metric coordinates, for example in scene generation or in proposing a goal pose. In such a pipeline, the coordinates must be checked against the constraints and corrected after generation; none of the instructions tested here made the check unnecessary.

\subsection{Geometric information in the models' coordinates}
\label{sec:mechsummary}

This subsection summarizes the results of this section, together with the comparison of move directions of Sec.~\ref{sec:geometric}, in two parts: the geometric regularities that the coordinates and the likelihood reflect, and the computations that the models do not perform.

\paragraph{Geometric regularities.} The likelihood depends on wall contact: snapping objects to their declared walls recovers \AdsDeltaAdsVsRandKitchen{} of the \AdsDeltaRandVsOrigKitchen{} nats that a distance-matched random displacement costs for Qwen and \AdsDeltaAdsVsRandKitchenGemma{} of \AdsDeltaRandVsOrigKitchenGemma{} nats for Gemma (Sec.~\ref{sec:adsorption}), and in real kitchens the pairwise accuracy of a move away from a wall exceeds that of a move along it by \GeoGapKitchen{} in pairwise accuracy (Sec.~\ref{sec:geometric}); the likelihood also prefers snapping to the wall the model declared over snapping to another wall, with pairwise accuracy \AdsAdsVsWrongWallKitchen{} (Qwen) and \AdsAdsVsWrongWallKitchenGemma{} (Gemma) in kitchens (Sec.~\ref{sec:adsorption}). In generation, the declared wall is the wall nearest to the object's coordinates for \DeclAgreeQwenA\% of objects (Qwen, condition~\condA), against \DeclChanceQwenA\% if declarations were independent of the coordinates, and a wall stated in the prompt is turned into a coordinate near that wall with consistency \ConsQwenApp{} (condition~\condApp; Sec.~\ref{sec:symbols}).

\paragraph{Computations not performed.} The coordinates are not derived from the constraints that the models are given or have stated. The emitted coordinates take few distinct values: \CoordGridModelKitchen\% of the generated kitchen coordinates of both models are multiples of \EntGrid~m, with \CoordDistinctModelKitchen{} (Qwen) and \CoordDistinctModelKitchenGemma{} (Gemma) distinct values over \NGenPerRoom{} kitchens against \CoordDistinctRealKitchen{} in real ones (Sec.~\ref{sec:adsorption}). They are also not the values that the stated constraints determine. An object declared against a wall should have its center at half its depth from that wall, but under condition~\condA{} the declared walls agree with the coordinates emitted before them in only \ConsQwenA{} of cases, and the median object sits \ConsGapMedQwenA~m from the wall it claims to touch (Qwen, ScanNet; Sec.~\ref{sec:symbols}). Given a room rectangle and a footprint, the models do not keep all objects inside the rectangle, under any of the five instructions of Table~\ref{tab:containment} (\ConsOutsideQwenAppi\% of objects outside for Qwen under~\condApp-i; Sec.~\ref{sec:containment}). The likelihood of a generated layout is also higher than that of its snapped copy, whose objects touch their declared walls (Sec.~\ref{sec:adsorption}); because the generated layout is the model's own sample, this last result is weaker evidence than the others.

These results indicate that the generated coordinates follow regularities in how coordinates are written (an object declared against a wall tends to receive a coordinate near that wall) rather than being computed from constraints. This is consistent with the results of the preceding sections: the generated layouts reproduce a tendency to place objects near walls but not the dependence of placement on the room type (Sec.~\ref{sec:hierarchy}) or the exact position of an individual object.

%% file: sections/search.tex
\section{Evaluation of the generated placements as a search prior}
\label{sec:search}

This section evaluates the generated placements as a prior for object search: a distribution over the position of a target object in a room, used before any object of the room has been observed. Sec.~\ref{sec:searchmethod} describes the method of the evaluation, Sec.~\ref{sec:searchcost} gives the fraction of the room that a search guided by each prior examines before it finds the target, Sec.~\ref{sec:smoothing} evaluates the effect of smoothing the priors, and Sec.~\ref{sec:wallbound} derives a bound for priors that are uniform on a region near the walls. A prior conditioned on an observed object, whose generation Sec.~\ref{sec:observed} measured, is not evaluated in this section.

\subsection{Method of the search evaluation}
\label{sec:searchmethod}

\paragraph{Data.}
The targets are the real ScanNet rooms of the four room types, with the
floor-standing objects of the closed vocabulary. All rooms, real and
generated, are rotated and mirrored so that the door lies on the south wall,
left of center, as in the orientation analysis of
Sec.~\ref{sec:hierarchy}; real rooms without a recorded door cannot be
oriented and are left out, which keeps \SearchRealCanonBath{} of
\SearchRealAllBath{} bathrooms, \SearchRealCanonBed{} of \SearchRealAllBed{}
bedrooms, \SearchRealCanonOffice{} of \SearchRealAllOffice{} offices and
\SearchRealCanonLiving{} of \SearchRealAllLiving{} living rooms. For each room
type, the target categories are those present in at least \MinOcc{} of its
real rooms (\SearchCatsBath{}, \SearchCatsBed{}, \SearchCatsOffice{} and
\SearchCatsLiving{} categories), and every instance of a target category in a
real room is one search episode. The model priors were built from the \NGenPerRoom{} rooms per room type that each model generated under conditions~\condA{}
and~\condBw{} (Sec.~\ref{sec:conditions}); the real-data prior is built
from the other real rooms of the same type.

\paragraph{Priors.}
Every prior is a probability vector over a $G \times G$ grid ($G = \GridN$)
laid over the room in normalized coordinates $(x/W, y/D)$, so that a prior
built from rooms of one size applies to a room of another.
\textbf{U} is uniform. \textbf{W} is uniform over the cells within
$0.15$~m of a wall, the rule-based baseline. \textbf{M-A} and \textbf{M-Bw} (from conditions~\condA{} and~\condBw) are histograms of the generated centers of the target category, built from the generated rooms of the same type that contain it:
to search for a chair in a bedroom, M-A uses the chairs in the rooms the
model generated as bedrooms. \textbf{R} is the same histogram built from the real rooms of that type with the evaluated room held out; it is the prior obtained from real layouts of the room type. The main evaluation uses these histograms without smoothing, so that the generated distribution is evaluated as it is and the evaluation has no free parameter. When a prior is used as a probability map, it is commonly smoothed spatially; Sec.~\ref{sec:smoothing} evaluates the effect of this post-processing with the method given below.

\paragraph{Ablations.}
M-A uses two pieces of information, the room type and the target category.
Two ablations remove one each. \textbf{M-A/room} keeps the category and drops
the room type: for the chair in a bedroom it uses the chairs of all four
generated room types. \textbf{M-A/cat} keeps the room type and drops the
category: it uses all objects the model generated in bedrooms, whatever they
are. The change in cost caused by an ablation measures the contribution of the removed information to M-A. The same two ablations of R, \textbf{R/room} and \textbf{R/cat}, measure the contribution of each piece of information in the real layouts; an ablation of M-A can increase the cost substantially only where the corresponding ablation of R does. The ablations of M-A are
evaluated on the episodes for which M-A is defined. The pooled histograms of the ablations are built from more objects than M-A and are therefore smoother, which can lower their cost independently of the removed information.

\paragraph{Cost.}
The search policy opens cells in decreasing order of prior probability
$\pi$ and stops at the cell $t$ that contains the target's center. Cells of
equal probability are opened in uniformly random order, and the cost of an
episode is the expected fraction of the room opened,
\begin{equation}
  \mathrm{cost} = \frac{1}{G^2}\Bigl(
    \bigl|\{c : \pi_c > \pi_t\}\bigr|
    + \frac{\bigl|\{c : \pi_c = \pi_t\}\bigr| + 1}{2}\Bigr),
  \label{eq:cost}
\end{equation}
which is \CostUExpected{} for the uniform prior. Costs are averaged over
the episodes of a room and then over rooms, and intervals resample rooms.
Comparisons between two priors are paired: the difference is taken room by
room over the episodes for which both are defined.
Appendix~\ref{app:searchcost} gives the remaining details.

\paragraph{Smoothing.}
The cost of Equation~\eqref{eq:cost} depends only on the order of the cells by probability. Adding the same constant to every cell or mixing a prior with U therefore does not change the cost, and the cells without generated objects remain tied at the end of the order. Only a smoothing that depends on the position of the cells changes the order. We convolve each histogram on the $G \times G$ grid with a Gaussian kernel of standard deviation $\sigma$ cells, with reflection at the walls, and renormalize; a cell then receives probability according to its distance from the cells that contain objects. The smoothing is applied to M-A, M-A/room and R (R after the evaluated room has been held out), with $\sigma = 0.5$, $1$ and $2$ cells; one cell is $1/G$ of the room width and depth. U and W are not smoothed. All values of $\sigma$ are reported; none is selected.

\subsection{Search cost of the priors}
\label{sec:searchcost}

\begin{figure}[tbp]
  \centering
  \includegraphics[width=\linewidth]{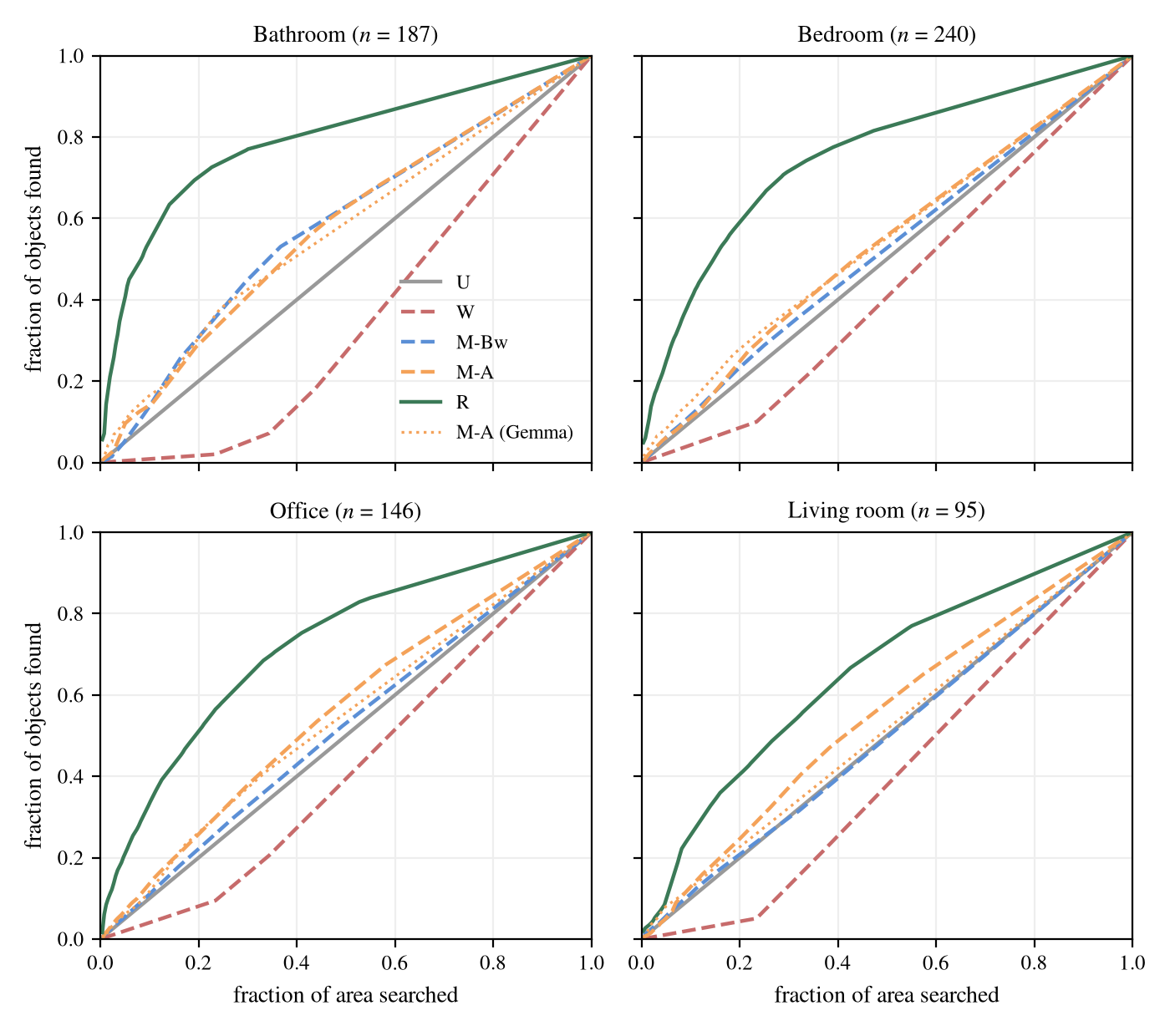}
  \caption{Search curves: the probability of having found the target against
    the fraction of the room opened, averaged over rooms (Qwen's priors; the
    dotted line is Gemma's M-A). In every room type, the curve of R lies above all others, the curve of W below that of U, and the curve of M-A slightly above that of U (Table~\ref{tab:search}).}
  \label{fig:search}
\end{figure}

Table~\ref{tab:search} gives the search cost of every prior for each room type, and Fig.~\ref{fig:search} gives the search curves, the probability of having found the target against the fraction of the room opened. We first compare the model priors with the uniform prior and with the prior built from real layouts, then examine the effect of the wall-following prompt, and finally use the ablations to identify the information in the priors that reduces the cost.

\begin{table}[tbp]
  \centering
  \caption{Search cost (expected fraction of the room opened before the
    target is found; lower is better) by prior and room type (ScanNet, rooms
    oriented with the door on the south wall). The uniform prior gives
    \CostUExpected{} in every room type. W and the R rows do not depend on the
    model. Intervals and episode counts are in
    Tables~\ref{tab:app-search-qwen} and~\ref{tab:app-search-gemma}, paired
    differences in Table~\ref{tab:app-search-diff}.}
  \label{tab:search}
  \small
  \begin{tabular}{llcccc}
    \toprule
    Prior & Model & Bathroom & Bedroom & Office & Living room \\
    \midrule
    W & --- & \CostQwenWBath & \CostQwenWBed & \CostQwenWOffice & \CostQwenWLiving \\
    M-A & Qwen & \CostQwenMABath & \CostQwenMABed & \CostQwenMAOffice & \CostQwenMALiving \\
        & Gemma & \CostGemmaMABath & \CostGemmaMABed & \CostGemmaMAOffice & \CostGemmaMALiving \\
    M-Bw & Qwen & \CostQwenMBwBath & \CostQwenMBwBed & \CostQwenMBwOffice & \CostQwenMBwLiving \\
         & Gemma & \CostGemmaMBwBath & \CostGemmaMBwBed & \CostGemmaMBwOffice & \CostGemmaMBwLiving \\
    M-A/room & Qwen & \CostQwenMAroomBath & \CostQwenMAroomBed & \CostQwenMAroomOffice & \CostQwenMAroomLiving \\
             & Gemma & \CostGemmaMAroomBath & \CostGemmaMAroomBed & \CostGemmaMAroomOffice & \CostGemmaMAroomLiving \\
    M-A/cat & Qwen & \CostQwenMAcatBath & \CostQwenMAcatBed & \CostQwenMAcatOffice & \CostQwenMAcatLiving \\
            & Gemma & \CostGemmaMAcatBath & \CostGemmaMAcatBed & \CostGemmaMAcatOffice & \CostGemmaMAcatLiving \\
    \midrule
    R & --- & \CostQwenRBath & \CostQwenRBed & \CostQwenROffice & \CostQwenRLiving \\
    R/room & --- & \CostQwenRroomBath & \CostQwenRroomBed & \CostQwenRroomOffice & \CostQwenRroomLiving \\
    R/cat & --- & \CostQwenRcatBath & \CostQwenRcatBed & \CostQwenRcatOffice & \CostQwenRcatLiving \\
    \bottomrule
  \end{tabular}
\end{table}

\paragraph{Comparison with the uniform and the real-data prior.}
The cost of M-A is lower than that of U, \CostUExpected, in every room type for both models, and every paired interval excludes zero (Table~\ref{tab:app-search-diff}): the reduction is \DiffMAUQwenMin--\DiffMAUQwenMax{} of the room area for Qwen and \DiffMAUGemmaMin--\DiffMAUGemmaMax{} for Gemma. The generated placements therefore contain information about the positions of the target categories, although their wall-proximity excess does not follow the room type (Sec.~\ref{sec:hierarchy}). The reduction is, however, small compared with that obtained with real layouts. The cost of R is \CostRMin--\CostRMax, and, as a fraction of the reduction that R achieves relative to U in the same room type, M-A achieves \ShareMAQwenMin--\ShareMAQwenMax\% for Qwen and \ShareMAGemmaMin--\ShareMAGemmaMax\% for Gemma.

The search curves of Fig.~\ref{fig:search} show where this difference arises. The curve of R rises steeply over the first cells opened: after $10\%$ of the room has been opened, the target has been found in \SAtTenRMin--\SAtTenRMax{} of the episodes with R, against \SAtTenMAQwenMin--\SAtTenMAQwenMax{} with Qwen's M-A, \SAtTenMAGemmaMin--\SAtTenMAGemmaMax{} with Gemma's and \SAtTenUMin{} with U (Tables~\ref{tab:app-search-qwen} and~\ref{tab:app-search-gemma}). The cells to which M-A gives the highest probability thus contain the target only slightly more often than cells chosen at random, and the curve of M-A stays close to that of U over the whole range. The curve of W lies below that of U, because W opens the cells near the walls first and these cells contain few target centers (Sec.~\ref{sec:wallbound}).

\paragraph{Wall-following prompt.}
Condition~\condBw{} places the generated objects closer to the walls (Sec.~\ref{sec:overwrite}), but the cost of M-Bw is not lower than that of M-A in any room type for either model, and it is higher in most: for Qwen in bedrooms (\DiffMBwMAQwenBed), offices (\DiffMBwMAQwenOffice) and living rooms (\DiffMBwMAQwenLiving), and for Gemma in all four room types (Table~\ref{tab:app-search-diff}). Placing the objects against the walls therefore does not make the prior more useful; this agrees with the cost of the wall-based prior W, which is higher than that of U (Sec.~\ref{sec:wallbound}).

\paragraph{Information used by the priors.}
The ablations show which of the two pieces of information, the target category and the room type, reduces the cost. In the real layouts, removing the category raises the cost of R by \DiffRcatRQwenBath{} to \DiffRcatRQwenBed{} across room types (R/cat against R), whereas removing the room type changes it by \DiffRroomRQwenBath{} to \DiffRroomRQwenBed{} (R/room against R; Table~\ref{tab:app-search-diff}). Once the category is known, the position of an object in a room oriented by its door therefore depends little on the room type.

\begin{figure}[tbp]
  \centering
  \includegraphics[width=\linewidth]{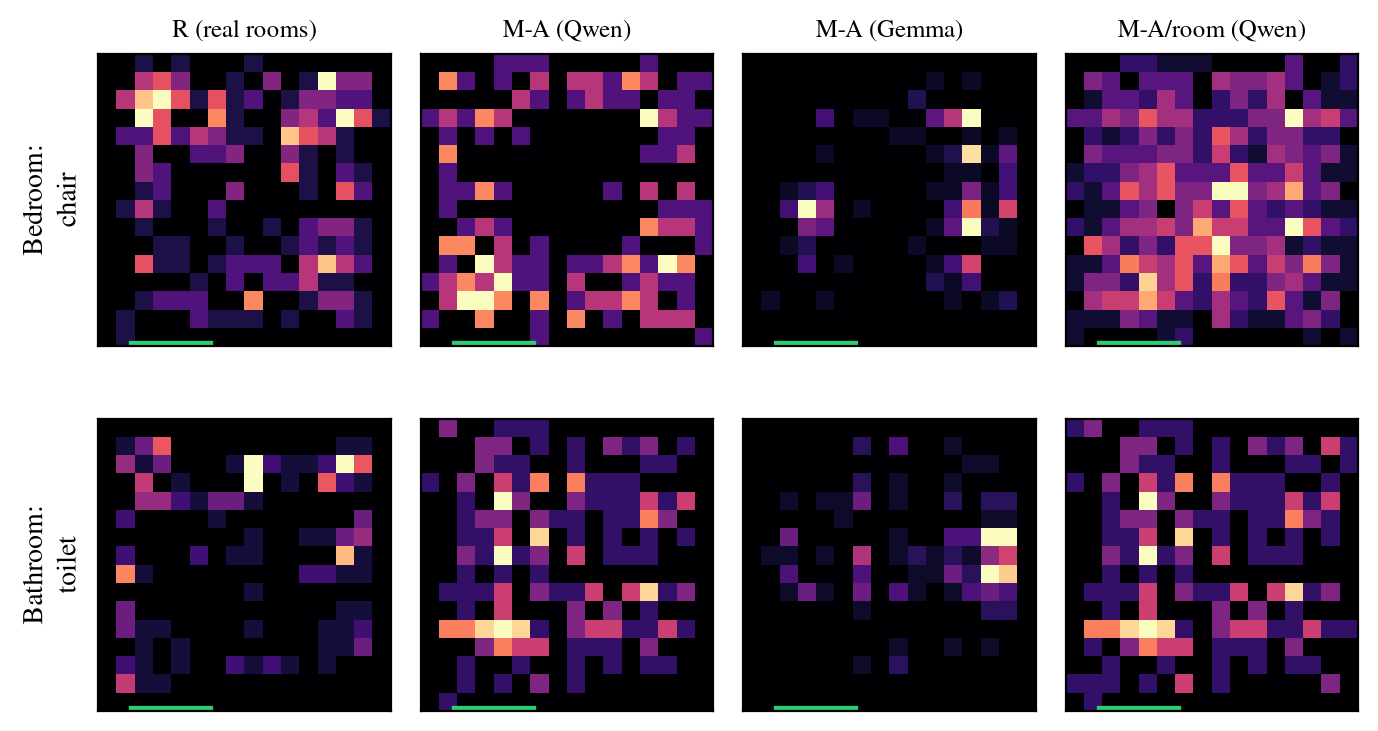}
  \caption{Priors over the position of a chair in a bedroom (top) and a toilet in a bathroom (bottom): the histogram of real rooms (R), of the generated rooms of the same type (M-A), and of the generated rooms of all four types (M-A/room). Rooms are oriented with the door (green) on the south wall, left of center, and normalized to the unit square; brighter cells are more probable. Appendix~\ref{app:searchcost} shows the priors of every target category.}
  \label{fig:priors}
\end{figure}

For the model priors, removing the room type changes the shape of the prior, as Fig.~\ref{fig:priors} shows. For a chair in a bedroom, Qwen's M-A gives non-zero probability to \NonzeroChairBedMA{} of the $256$ cells and leaves the center of the room mostly empty, whereas M-A/room, which adds the chairs that Qwen generated in offices and living rooms, gives non-zero probability to \NonzeroChairBedMAroom{} cells, including the center. For a toilet in a bathroom, which the models generate almost only in bathrooms (Table~\ref{tab:app-identity}), the two priors nearly coincide (\NonzeroToiletBathMA{} and \NonzeroToiletBathMAroom{} non-zero cells). The cost nevertheless does not increase when the room type is removed: the cost of M-A/room exceeds that of M-A by at most \DiffMAroomMAGemmaBath, and it is lower in offices (\DiffMAroomMAQwenOffice{} for Qwen, \DiffMAroomMAGemmaOffice{} for Gemma) and living rooms (\DiffMAroomMAQwenLiving, \DiffMAroomMAGemmaLiving). A likely reason is the number of empty cells. The priors are not smoothed, so a cell in which no generated object lies has zero probability, and a target in such a cell is found only after all cells of non-zero probability have been opened. Over all target categories and room types, the median number of non-zero cells is \NonzeroMAQwen{} for M-A and \NonzeroMAroomQwen{} for M-A/room (Qwen), and the fraction of episodes whose target lies in a zero-probability cell falls from \EmptyShareMAQwen{} to \EmptyShareMAroomQwen{} for Qwen and from \EmptyShareMAGemma{} to \EmptyShareMAroomGemma{} for Gemma. Pooling the four room types lowers the cost of the targets in these cells, and this reduction can exceed the increase caused by the loss of the room type. The reductions of M-A/room relative to M-A are larger than those of R/room relative to R. Sec.~\ref{sec:smoothing} examines this explanation further.

Removing the category raises the cost of M-A in offices (\DiffMAcatMAQwenOffice{} for Qwen, \DiffMAcatMAGemmaOffice{} for Gemma) and living rooms (\DiffMAcatMAQwenLiving, \DiffMAcatMAGemmaLiving), so the category-specific histograms of the models contain information there. It lowers the cost in bathrooms (\DiffMAcatMAQwenBath, \DiffMAcatMAGemmaBath) and bedrooms (\DiffMAcatMAQwenBed, \DiffMAcatMAGemmaBed), where the histogram of all generated objects gives a lower cost than the histograms of the individual categories, although the category lowers the cost of R in these room types. In bathrooms and bedrooms, the models therefore place the individual target categories at positions that differ from the real ones more than the positions of all generated objects together do.

\paragraph{Spread of the priors.}
The priors of the two models differ in shape: Qwen's M-A spreads over much of the room, whereas Gemma's concentrates on a few cells (Fig.~\ref{fig:priors} and Figs.~\ref{fig:priors-bathroom-scannet}, \ref{fig:priors-bedroom}, \ref{fig:priors-office} and~\ref{fig:priors-living-room}). We compared the spread over the \SpreadNPriors{} priors of M-A in which the real data and both models contain at least five objects of the target category (Table~\ref{tab:spread}; values are medians over the priors). Each model's prior is built from about \SpreadNQwen{} objects and the real prior from \SpreadNReal, so the number of non-zero cells is also given for an equal number of objects, drawn at random from each set down to the smallest of the three. At an equal number of objects, the real prior has non-zero probability in \SpreadCellsAtNReal{} cells, Qwen's in \SpreadCellsAtNQwen{} and Gemma's in \SpreadCellsAtNGemma; Qwen's prior is wider than the real one in \SpreadWiderQwen{} of the \SpreadNPriors{} priors, and Gemma's is narrower in \SpreadNarrowerGemma. The difference between the models is therefore not a difference in the number of generated objects.

\begin{table}[tbp]
  \centering
  \caption{Spread of the M-A priors and of the real-layout prior R (ScanNet, condition~\condA; medians over the \SpreadNPriors{} priors in which every set contains at least five objects of the target category). Cells: number of the $256$ cells with non-zero probability, with all objects and with an equal number of objects drawn from each set. Distinct positions: number of distinct positions, normalized by the room dimensions and rounded to $0.01$, divided by the number of objects. Most frequent position: share of the objects at the most frequent normalized position.}
  \label{tab:spread}
  \begin{tabular}{lccc}
    \toprule
    & R & Qwen & Gemma \\
    \midrule
    Objects & \SpreadNReal & \SpreadNQwen & \SpreadNGemma \\
    Cells, all objects & \SpreadCellsReal & \SpreadCellsQwen & \SpreadCellsGemma \\
    Cells, equal number of objects & \SpreadCellsAtNReal & \SpreadCellsAtNQwen & \SpreadCellsAtNGemma \\
    Distinct positions & \SpreadPosDistinctReal & \SpreadPosDistinctQwen & \SpreadPosDistinctGemma \\
    Most frequent position & \SpreadTopOneReal & \SpreadTopOneQwen & \SpreadTopOneGemma \\
    \bottomrule
  \end{tabular}
\end{table}

Gemma's prior is narrow because its generated positions repeat. The fraction of distinct normalized positions is \SpreadPosDistinctGemma{} for Gemma, against \SpreadPosDistinctQwen{} for Qwen and \SpreadPosDistinctReal{} for the real rooms, and the most frequent position holds \SpreadTopOneGemma{} of Gemma's objects. The repetition follows from the generated rooms. Over the \NGenPerRoom{} layouts of a room type, Gemma uses \SpreadDimsMinGemma--\SpreadDimsMaxGemma{} distinct room dimensions (Qwen \SpreadDimsMinQwen--\SpreadDimsMaxQwen; Sec.~\ref{sec:hierarchy}), \SpreadListsMinGemma--\SpreadListsMaxGemma{} distinct lists of object categories (Qwen \SpreadListsMinQwen--\SpreadListsMaxQwen) and \SpreadXYMinGemma--\SpreadXYMaxGemma{} distinct coordinate values (Qwen \SpreadXYMinQwen--\SpreadXYMaxQwen, real rooms \SpreadXYMinReal--\SpreadXYMaxReal). When the room dimensions and the coordinates are the same, the normalized position is the same, so Gemma's objects of a category fall into few cells. Both models generate at temperature $1.0$ with the same sampling settings, without top-$k$ or nucleus truncation (Sec.~\ref{sec:scoring}), so the difference is not caused by the sampling settings.

The two shapes give nearly the same cost for different reasons. Over all episodes, the target of a real room lies in a cell of zero probability in a fraction \EmptyShareMAQwen{} of the episodes for Qwen and \EmptyShareMAGemma{} for Gemma, as given above. When it lies in a cell of non-zero probability, the cost is \SmInQwenZero{} for Qwen and \SmInGemmaZero{} for Gemma; when it does not, the target is found only after all non-zero cells have been opened, and the cost is \SmOutQwenZero{} and \SmOutGemmaZero. Gemma's prior thus finds the target quickly when the target lies at one of the repeated positions, but most targets do not; Qwen's prior covers more targets, but because its probability is spread over more cells, a covered target is found later. Neither shape approaches R, whose cost is \CostQwenRBath--\CostQwenRLiving{} (Table~\ref{tab:search}).

\paragraph{Variation across room types.}
Sec.~\ref{sec:hierarchy} showed that the generated placement does not follow the room type. In this evaluation, this has little effect on the search cost: the target category is given, and given the category the real placements vary little with the room type (R/room against R above). The spread of the search cost across room types does not measure the dependence on the room type either.
The permutation test of equality across room types (the test of
Sec.~\ref{sec:hierarchy} applied to the per-room costs) gives
$p$~\CostHomogPQwenMA{} for Qwen's M-A and $p$~\CostHomogPGemmaMA{} for
Gemma's, but it also rejects for R/room ($p$~\CostHomogPQwenRroom), which
contains no room-type information by construction: the same prior meets
differently placed targets in different room types
(Table~\ref{tab:app-searchsd}). The difference between M-A and R therefore lies not in the room type but in the positions of the individual categories: removing the category raises the cost of R in every room type, but that of M-A only in offices and living rooms (Table~\ref{tab:app-search-diff}).

\subsection{Effect of smoothing}
\label{sec:smoothing}
This subsection evaluates the priors after the spatial smoothing defined in Sec.~\ref{sec:searchmethod}. Table~\ref{tab:smoothing} gives the cost of M-A and R for each $\sigma$, and Table~\ref{tab:app-smoothing} gives the paired differences with their intervals.

\begin{table}[tbp]
  \centering
  \caption{Search cost of the real-layout prior R and of M-A after spatial smoothing with a Gaussian kernel of standard deviation $\sigma$ cells (ScanNet; $\sigma = 0$ is the unsmoothed prior of Table~\ref{tab:search}). Lower is better; U costs \CostUExpected. Paired differences and intervals are in Table~\ref{tab:app-smoothing}.}
  \label{tab:smoothing}
  \begin{tabular}{llcccc}
    \toprule
    Room type & Prior & $\sigma = 0$ & $0.5$ & $1$ & $2$ \\
    \midrule
    Bathrooms & R & \SmCostRBathZero & \SmCostRBathHalf & \SmCostRBathOne & \SmCostRBathTwo \\
     & M-A, Qwen & \SmCostQwenBathZero & \SmCostQwenBathHalf & \SmCostQwenBathOne & \SmCostQwenBathTwo \\
     & M-A, Gemma & \SmCostGemmaBathZero & \SmCostGemmaBathHalf & \SmCostGemmaBathOne & \SmCostGemmaBathTwo \\
    \addlinespace
    Bedrooms & R & \SmCostRBedZero & \SmCostRBedHalf & \SmCostRBedOne & \SmCostRBedTwo \\
     & M-A, Qwen & \SmCostQwenBedZero & \SmCostQwenBedHalf & \SmCostQwenBedOne & \SmCostQwenBedTwo \\
     & M-A, Gemma & \SmCostGemmaBedZero & \SmCostGemmaBedHalf & \SmCostGemmaBedOne & \SmCostGemmaBedTwo \\
    \addlinespace
    Offices & R & \SmCostROfficeZero & \SmCostROfficeHalf & \SmCostROfficeOne & \SmCostROfficeTwo \\
     & M-A, Qwen & \SmCostQwenOfficeZero & \SmCostQwenOfficeHalf & \SmCostQwenOfficeOne & \SmCostQwenOfficeTwo \\
     & M-A, Gemma & \SmCostGemmaOfficeZero & \SmCostGemmaOfficeHalf & \SmCostGemmaOfficeOne & \SmCostGemmaOfficeTwo \\
    \addlinespace
    Living rooms & R & \SmCostRLivingZero & \SmCostRLivingHalf & \SmCostRLivingOne & \SmCostRLivingTwo \\
     & M-A, Qwen & \SmCostQwenLivingZero & \SmCostQwenLivingHalf & \SmCostQwenLivingOne & \SmCostQwenLivingTwo \\
     & M-A, Gemma & \SmCostGemmaLivingZero & \SmCostGemmaLivingHalf & \SmCostGemmaLivingOne & \SmCostGemmaLivingTwo \\
    \bottomrule
  \end{tabular}
\end{table}

\paragraph{Effect on each prior.}
Smoothing with $\sigma = 0.5$ lowers the cost of R and of both models' M-A in every room type, and every paired interval excludes zero (Table~\ref{tab:app-smoothing}). With larger $\sigma$, the priors differ. With $\sigma = 2$, the cost of R exceeds its unsmoothed value in bathrooms, bedrooms and offices (\SmGainRBathTwo, \SmGainRBedTwo{} and \SmGainROfficeTwo), whereas the cost of Gemma's M-A remains below its unsmoothed value in bedrooms, offices and living rooms (\SmGainGemmaBedTwo, \SmGainGemmaOfficeTwo{} and \SmGainGemmaLivingTwo) and exceeds it only in bathrooms (\SmGainGemmaBathTwo). For Qwen's M-A with $\sigma = 2$, the change is \SmGainQwenOfficeTwo{} in offices and \SmGainQwenLivingTwo{} in living rooms, and is not detectable in bathrooms (\SmGainQwenBathTwo, \SmGainQwenBathTwoCI) and bedrooms (\SmGainQwenBedTwo, \SmGainQwenBedTwoCI).

\paragraph{Targets inside and outside the unsmoothed support.}
The reduction comes from the targets that lie in cells of zero probability before smoothing. Over all room types, these targets cost \SmOutQwenZero{} for Qwen, \SmOutGemmaZero{} for Gemma and \SmOutRZero{} for R without smoothing, and \SmOutQwenOne, \SmOutGemmaOne{} and \SmOutROne{} with $\sigma = 1$. The targets in cells of non-zero probability cost more after smoothing: \SmInQwenZero{} and \SmInQwenOne{} for Qwen, \SmInGemmaZero{} and \SmInGemmaOne{} for Gemma, and \SmInRZero{} and \SmInROne{} for R. Smoothing therefore lowers the cost of the targets outside the support and raises that of the targets inside it. The fraction of targets outside the support is \EmptyShareMAQwen{} for Qwen and \EmptyShareMAGemma{} for Gemma (Sec.~\ref{sec:searchcost}), against \SmShareOutR{} for R. With $\sigma = 2$, at which this exchange is largest, the cost rises for R, which has the smallest fraction outside the support, and falls for Gemma in three room types. Relative to Qwen, the reduction of Gemma's cost with $\sigma = 1$ is larger by \SmGemQwenBedOne{} in bedrooms (95\% CI \SmGemQwenBedOneCI) and \SmGemQwenLivingOne{} in living rooms (\SmGemQwenLivingOneCI), not detectably different in offices (\SmGemQwenOfficeOne, \SmGemQwenOfficeOneCI), and smaller in bathrooms (\SmGemQwenBathOne, \SmGemQwenBathOneCI). In bedrooms and living rooms, the smoothed M-A of Gemma costs less than that of Qwen (Table~\ref{tab:smoothing}). Part of the cost of Gemma's unsmoothed M-A (Sec.~\ref{sec:searchcost}) is therefore due to the narrowness of its priors: positions near the positions that Gemma repeats contain real targets, and smoothing assigns probability to them.

\paragraph{Comparison with R.}
Compared at the same $\sigma$, the cost of M-A exceeds that of R in every room type for both models and every $\sigma$, and every paired interval excludes zero (Table~\ref{tab:app-smoothing}). If each prior is taken at the $\sigma$ that gives its lowest cost in a room type, the cost of M-A exceeds that of R by \SmBestGapQwenMin--\SmBestGapQwenMax{} for Qwen and \SmBestGapGemmaMin--\SmBestGapGemmaMax{} for Gemma across room types, against \SmGapZeroQwenMin--\SmGapZeroQwenMax{} and \SmGapZeroGemmaMin--\SmGapZeroGemmaMax{} without smoothing. The lowest cost of M-A over $\sigma$ and room types is \SmBestQwenMin{} for Qwen and \SmBestGemmaMin{} for Gemma; that of R is \SmBestRMin--\SmBestRMax{} across room types. The difference between M-A and R at $\sigma = 2$ is smaller than without smoothing mainly because the cost of R increases.

\paragraph{Room type after smoothing.}
Sec.~\ref{sec:searchcost} attributed the lower cost of M-A/room in offices and living rooms to the cells that are empty in M-A. After smoothing with $\sigma = 0.5$, the difference between M-A/room and M-A for Qwen is \SmRoomQwenOfficeHalf{} in offices (\SmRoomQwenOfficeHalfCI) and \SmRoomQwenLivingHalf{} in living rooms (\SmRoomQwenLivingHalfCI), against \SmRoomQwenOfficeZero{} and \SmRoomQwenLivingZero{} without smoothing, and in bedrooms M-A/room costs more than M-A (\SmRoomQwenBedHalf, \SmRoomQwenBedHalfCI). For Qwen, once the empty cells receive probability, removing the room type no longer lowers the cost, and in bedrooms it raises it. For Gemma, the difference in living rooms is no longer detectable (\SmRoomGemmaLivingHalf, \SmRoomGemmaLivingHalfCI), but M-A/room still costs less than M-A in offices (\SmRoomGemmaOfficeHalf, \SmRoomGemmaOfficeHalfCI) and bedrooms (\SmRoomGemmaBedHalf, \SmRoomGemmaBedHalfCI).

In summary, spatial smoothing lowers the cost of the generated priors, and lowers it more for Gemma, whose priors are narrower. It does not close the difference from the prior built from real layouts: with the best $\sigma$ for each prior, the difference for Qwen is of the same size as without smoothing, and that for Gemma decreases in bedrooms and living rooms but remains at least \SmBestGapGemmaMin.

\subsection{Upper bound of wall-based priors}
\label{sec:wallbound}

The rule-based prior W is \emph{worse} than uniform in every room type at the
band width used throughout the paper, by \BandDiffBedOneFive{} to
\BandDiffBathOneFive{} (Table~\ref{tab:search}). The following identity explains this and bounds the cost of any prior that is uniform on a region.

Let $S$ be a region occupying a fraction $c$ of the room, let the prior be
uniform on $S$ and zero elsewhere, and let $p$ be the probability that the
target's center lies in $S$. The policy opens the cells of $S$ first, in
random order because they are tied, and then the rest. If the target is in
$S$, its cell is equally likely to be at any position in that order, so on
average half of $S$ is opened before it is found: a cost of $c/2$. If the
target is outside $S$, all of $S$ is opened first and then, by the same
argument, half of the remainder: $c + (1-c)/2$. Hence
\begin{equation}
  \mathbb{E}[\text{cost}] = p\,\frac{c}{2} + (1-p)\Bigl(c + \frac{1-c}{2}\Bigr)
  = \frac{1 + c - p}{2},
  \qquad
  \text{cost}(S) - \text{cost}(U) = \frac{c - p}{2}.
  \label{eq:region}
\end{equation}
The derivation needs no assumption about where in $S$ the target lies, only
that tied cells are opened in random order (up to the half-cell terms that
cancel between $S$ and $U$). A region prior has a lower cost than U exactly when the target lies in the region more often than the region's share of the floor, $p > c$, and the reduction is half the difference. The band width is set in meters
rather than as a fraction of the room, because wall contact is a physical
distance; $c$ is therefore larger in small rooms, and enters the identity per
room. At $0.15$~m the band covers more of the floor than its share of target
centers in every room type. Widening the band raises $p$ but also $c$, and the cost of W is lower than that of U only at $0.50$~m in bathrooms (\BandDiffBathFiveZero) and bedrooms
(\BandDiffBedFiveZero); Appendix~\ref{app:searchcost} gives the band sweep
(Fig.~\ref{fig:band}, Table~\ref{tab:app-band}).

The identity bounds only priors that are uniform on a region: proximity to a wall is one such region, and any increase in $p$ comes with an increase in $c$. M-A and R assign unequal probabilities to the cells and are not bound by it; the cost of R, \CostRMin--\CostRMax, shows the reduction obtained by a prior that represents the positions of the objects in real layouts beyond their distance from the walls.

%% file: sections/design.tex
\section{Design implications}
\label{sec:design}

This section derives design guidelines for the use of LLMs in object-goal navigation from the results of Secs.~\ref{sec:discrimination}--\ref{sec:search}. Table~\ref{tab:twoaxes} summarizes these results along the two axes of the evaluation, by the level of description of a room. Sec.~\ref{sec:guidelines} states the guidelines, each with the measurement and the dataset on which it rests. Sec.~\ref{sec:guidescope} states, for three of the guidelines, the conditions under which they hold and what was not measured. Sec.~\ref{sec:likelihooddesign} gives guidelines for measuring spatial knowledge through the likelihood.

\begin{table}[!hb]
  \centering
  \caption{Summary of the results along the two axes of the evaluation, by level of description. Numbers are for ScanNet, four room types, and condition~\condA{} unless noted (Qwen\,/\,Gemma).}
  \label{tab:twoaxes}
  \small
  \setlength{\tabcolsep}{4pt}
  \begin{tabular}{>{\raggedright\arraybackslash}p{0.17\linewidth}>{\raggedright\arraybackslash}p{0.36\linewidth}>{\raggedright\arraybackslash}p{0.25\linewidth}l}
    \toprule
    Level & Agreement with reality & Usefulness for navigation & Sec. \\
    \midrule
    Object identity
      & high (Spearman \IdentRhoQwenA\,/\,\IdentRhoGemmaA)
      & used by existing systems as an affinity
      & \ref{sec:hierarchy} \\
    Room dimensions
      & moderate (SD ratio \AreaSDRatioQwenA\,/\,\AreaSDRatioGemmaA)
      & not evaluated
      & \ref{sec:hierarchy} \\
    Placement (generated)
      & low; does not follow the room type (slope \SlopeQwenA\,/\,\SlopeGemmaA)
      & small gain over a uniform prior
      & \ref{sec:hierarchy}, \ref{sec:searchcost} \\
    Placement (likelihood)
      & responds to geometry and to the consistency of the text
      & piled layout preferred to the real one on ScanNet
      & \ref{sec:ladder}, \ref{sec:reversal} \\
    Relations (likelihood)
      & a change of neighbor barely registered (neighbor effect on CubiCasa \RelPairingQwenCubi\,/\,\RelPairingGemmaCubi)
      & not evaluated
      & \ref{sec:relladder} \\
    Relations (generated)
      & target follows an object given in the prompt (bedrooms; desk$\rightarrow$chair \NADeskChairQwen\,/\,\NADeskChairGemma)
      & not evaluated
      & \ref{sec:observed} \\
    \bottomrule
  \end{tabular}
\end{table}

\subsection{Guidelines}
\label{sec:guidelines}

\begin{enumerate}
  \item \textbf{Object--room affinities can be obtained from the models.} The fraction of generated rooms of a type that contain a category, $p(o \mid c)$, agrees with that of real rooms at Spearman \IdentRhoQwenA{} (Qwen) and \IdentRhoGemmaA{} (Gemma) (ScanNet; Sec.~\ref{sec:hierarchy}). This agreement supports the use of scalar object--room affinities from LLMs in existing systems~\citep{zhou2023esc,yu2023l3mvn,ji2024diffusion,jia2026hro}.
  \item \textbf{Generated coordinates should not be used as a search prior where real layouts are available.} The prior built from generated placements reduces the search cost relative to a uniform prior by \DiffMAUQwenMin--\DiffMAUQwenMax{} of the room area for Qwen and \DiffMAUGemmaMin--\DiffMAUGemmaMax{} for Gemma, which is \ShareMAQwenMin--\ShareMAQwenMax\% (Qwen) and \ShareMAGemmaMin--\ShareMAGemmaMax\% (Gemma) of the reduction obtained with a prior built from real layouts (Sec.~\ref{sec:searchcost}). These values are for the priors as generated. With spatial smoothing, which is a common post-processing of a probability map, the fraction is \SmShareBestQwenMin--\SmShareBestQwenMax\% (Qwen) and \SmShareBestGemmaMin--\SmShareBestGemmaMax\% (Gemma) when each prior is taken at its best smoothing width, and the cost of the generated prior exceeds that of the real-layout prior at every width in every room type (Sec.~\ref{sec:smoothing}).
  \item \textbf{The likelihood should not be used to judge layouts of movable furniture.} On ScanNet, the likelihood of a layout with all objects piled within \PileRadius~m of the room center is higher than that of the real layout in most offices and living rooms (pairwise accuracy \ScanQwenPileOffice{} and \ScanQwenPileLiving{} for Qwen; Sec.~\ref{sec:reversal}). On the fixtures of CubiCasa, the same perturbation has an accuracy of \CubiPileKitchen{} (Qwen) and \CubiGemmaPileKitchen{} (Gemma) in kitchens (Sec.~\ref{sec:perturbresults}). The guideline therefore applies to rooms with movable furniture, which are the rooms in which a robot searches for objects; in ScanNet bathrooms, the accuracy of pile is close to $0.5$ for both models.
  \item \textbf{A wall-following prompt does not make the placement follow the room type.} Under the wall-following prompt, the wall-proximity excess is positive in every room type, the regression slope against the real excess is negative (\SlopeQwenBw{} for Qwen, \SlopeGemmaBw{} for Gemma), and \OutsideQwenBwMin--\OutsideQwenBwMax\% of Qwen's objects lie partly outside the room (Sec.~\ref{sec:overwrite}).
  \item \textbf{A wall given for each object is satisfied.} When the wall that each object must touch is given in the prompt, the fraction of objects placed within $0.15$~m of that wall is \ConsQwenApp{} for Qwen and \ConsGemmaApp{} for Gemma (Sec.~\ref{sec:symbols}). The models can therefore be used to write coordinates for placement constraints determined by another component, with the check of guideline~6.
  \item \textbf{Generated coordinates must be checked and corrected geometrically.} With the room and the object footprints given and containment as the only requirement, \ConsOutsideQwenAppi\% of Qwen's objects and \ConsOutsideGemmaAppi\% of Gemma's lie partly outside the room. Instructions that state the priority of containment, reverse it or attach it to every object change the rate by \PrioDApprQwen{} to \PrioDAppperQwen{} for Qwen (Sec.~\ref{sec:containment}).
  \item \textbf{Placement conditioned on an observed object requires further evaluation before it is used as a search prior.} In bedrooms, the generated target of a related pair moves with an object given in the prompt (following slope \NADeskChairQwen--\NABedNightstandQwen{} for Qwen and \NADeskChairGemma--\NABedNightstandGemma{} for Gemma), and unrelated objects move away from it (Sec.~\ref{sec:observed}). Whether this holds in other room types and whether it reduces the search cost was not measured.
\end{enumerate}

\subsection{Range of validity of the guidelines}
\label{sec:guidescope}

The guidelines of Sec.~\ref{sec:guidelines} state what the measurements support. This subsection gives, for guidelines~2, 5 and~7, the conditions under which they hold and the cases that were not measured.

\paragraph{Guideline~2: information in the generated placements.}
The generated placements reduce the search cost relative to a uniform prior, so they contain some information about the positions of objects. Their placement statistics do not follow the room type: the ratio of real to model variation of the wall-proximity excess across room types is \SDRatioQwenA{} for Qwen, with equality across room types not rejected ($p$~\HomogPQwenA), and the orientation profile shows the same (Sec.~\ref{sec:hierarchy}). The search evaluation cannot measure this dependence, because once the category is known the real placements depend little on the room type (Sec.~\ref{sec:searchcost}). The category is used only in part. In the orientation profile, Qwen's profile of a category is closer to the real profile of that category than to the pooled real profile for only \CatOriWinQwen{} of \CatOriNQwen{} categories (Table~\ref{tab:app-catori}). In search, the category-specific histograms lower the cost in offices and living rooms but raise it in bathrooms and bedrooms, relative to the histogram of all generated objects (Sec.~\ref{sec:searchcost}). The rule-based prior W, which uses only the distance to the walls, has a higher cost than the uniform prior in every room type (Sec.~\ref{sec:wallbound}).

\paragraph{Guideline~2: settings without real layouts.}
Guideline~2 applies where real layouts of the room type are available. With the ScanNet layouts, the prior built from real rooms has a cost of \CostRMin--\CostRMax, against \CostMAQwenMin--\CostMAQwenMax{} for Qwen's generated placements and \CostMAGemmaMin--\CostMAGemmaMax{} for Gemma's (Table~\ref{tab:search}). Settings without real layouts, such as an unusual room type, a target described only in words or a building whose conventions differ from those of the datasets, were not evaluated; Sec.~\ref{sec:limitations} discusses the sources of the rooms.

\paragraph{Guideline~5: role of the model.}
Condition~\condApp{} gives the room rectangle, the object list, each footprint and the wall that each object must touch. A component that supplies these already determines most of the spatial information of the layout, and the model only converts the stated constraints into coordinates. Guideline~5 therefore concerns the use of a language model in place of a simple geometric solver, not the extraction of spatial knowledge from it. Together with guideline~6, it states that, when the constraints are known, the model writes coordinates that mostly satisfy the wall constraint, and that the coordinates must be checked.

\paragraph{Guideline~7: conditional priors.}
Most measurements of Secs.~\ref{sec:discrimination}--\ref{sec:search} concern the distribution of placements before any object of the room is observed; only Sec.~\ref{sec:relladder}, with the likelihood, and Sec.~\ref{sec:observed}, with generation, concern relations between objects. A robot in a room observes objects, and a prior conditioned on them, $p(x, y \mid o, c, \text{observed objects})$, uses relations between objects such as a night stand beside a bed; scene-generation systems use such relations~\citep{feng2023layoutgpt,yang2024holodeck,celen2024idesign}. Sec.~\ref{sec:observed} shows that, in bedrooms, the generated target of a related pair moves with an object given in the prompt, whereas Sec.~\ref{sec:relladder} shows that the likelihood of a real layout changes little when the neighbor of a target is changed (neighbor effect \RelPairingQwenCubi{} for Qwen and \RelPairingGemmaCubi{} for Gemma on CubiCasa, not distinguished from zero on ScanNet). A conditional prior from the models may therefore be more informative than the prior evaluated in Sec.~\ref{sec:searchcost}, but this was measured neither in other room types nor in terms of search cost.

\subsection{Measuring spatial knowledge through the likelihood}
\label{sec:likelihooddesign}

The likelihood that a language model assigns to a layout is a property of one string, not of the layout. The same arrangement can be written in many ways (with the objects or the fields in another order, the coordinates with another precision, or in a mirrored frame), and each string has its own likelihood; the probability of the arrangement itself would be a sum over all of them, which cannot be computed. The same has been shown for multiple-choice answers, where strings with the same meaning share the probability mass \citep{holtzman2021surface}. A comparison between two layouts by likelihood is therefore a comparison between two strings, and every property in which the strings differ enters it: their length, the predictability of their numerals, the order of their entries, and whether their fields agree with one another. The same holds for likelihood-based generative models in general; image models trained on one dataset can assign a higher likelihood to another, simpler dataset, because the likelihood depends strongly on low-level statistics of the input~\citep{nalisnick2019know,serra2020input}. In our measurements, the granularity of the coordinate values has this role (Sec.~\ref{sec:adsorption}).

Table~\ref{tab:stringeffects} compares, for Qwen and CubiCasa kitchens, the effect of each such property with that of geometric perturbations measured on the same rooms. An edit that moves no object (reordering, mirroring, a one-character change of a field) changes the likelihood by an amount of the same order as, or larger than, the geometric perturbations, and a difference in the granularity of the coordinates changes it by tens of nats. For work that uses the likelihood to measure spatial knowledge, we therefore recommend the following controls, each of which corresponds to a row of Table~\ref{tab:stringeffects}.

\begin{table}[tbp]
  \centering
  \caption{Effects of properties of the text on likelihood comparisons, compared with geometric perturbations (Qwen, CubiCasa kitchens). Pairwise accuracy is the fraction of rooms in which the first layout has the higher likelihood (chance $0.5$); $\Delta$ is the mean log-likelihood difference, first minus second, in nats. Mirroring is not symmetrized here, because it is the transformation that the symmetrization removes.}
  \label{tab:stringeffects}
  \small
  \setlength{\tabcolsep}{4pt}
  \begin{tabular}{>{\raggedright\arraybackslash}p{0.27\linewidth}>{\raggedright\arraybackslash}p{0.27\linewidth}ccl}
    \toprule
    Property & Comparison & Pairwise acc. & $\Delta$ (nats) & Sec. \\
    \midrule
    \multicolumn{5}{l}{\emph{Geometric perturbations}} \\
    move along a wall & real vs.\ moved $0.10$~m & \GeoTangentKitchen & --- & \ref{sec:geometric} \\
    detach & real vs.\ detached & \CubiDetachKitchen & \CubiDeltaDetachKitchen & \ref{sec:perturbresults} \\
    \midrule
    \multicolumn{5}{l}{\emph{Properties of the text}} \\
    order of the entries & real vs.\ sorted list & \OrderLexAccKitchen & \OrderLexDelta & \ref{sec:perturbresults} \\
    frame & real vs.\ mirrored & \MirrorRawAccKitchen & \MirrorRawDeltaKitchen & \ref{sec:scoring} \\
    agreement between fields & real vs.\ one wall field changed & \WcClaimQwenKitchen & --- & \ref{sec:selfconsistency} \\
    repetition of values & generated vs.\ snapped copy & \AdsAdsVsOrigKitchenInv & \AdsDeltaAdsVsOrigKitchen & \ref{sec:adsorption} \\
    granularity of values & generated vs.\ real, as written & --- & \EntGapRawKitchen & \ref{sec:adsorption} \\
    & same, both on a $0.25$~m grid & --- & \EntGapQTwentyFiveKitchen & \ref{sec:adsorption} \\
    \bottomrule
  \end{tabular}
\end{table}

\begin{enumerate}
  \item \textbf{Comparison of a layout with a perturbed copy of itself.} Compare a layout with a copy that keeps the object list, the room, the order of the entries and the number of tokens. A comparison between two sets of layouts, such as generated and real rooms, includes every difference between their strings; in Sec.~\ref{sec:adsorption}, the likelihood difference between generated and real kitchens falls from \EntGapRawKitchen{} to \EntGapQTwentyFiveKitchen{} nats when both are quantized to the same grid.
  \item \textbf{A control that moves no object.} Report the effect of an edit that moves nothing, such as reordering the object list, as the scale for the geometric effects. In kitchens, sorting the object list gives a pairwise accuracy of \OrderLexAccKitchen, so the sorted copy has the higher likelihood in most rooms, and the mean likelihood difference (\OrderLexDelta{} nats) is larger in magnitude than that of detach (\CubiDeltaDetachKitchen{} nats; Sec.~\ref{sec:perturbresults}). For Gemma, the corresponding values are \OrderLexAccKitchenGemma{} and \OrderLexDeltaKitchenGemma{} nats, against \CubiGemmaDeltaDetachKitchen{} nats for detach.
  \item \textbf{The numerical values of both layouts.} Report how many distinct coordinate values each layout or set uses and what fraction lies on a coarse grid. An edit that changes the number of distinct values changes the likelihood independently of its geometric effect (Sec.~\ref{sec:adsorption}).
  \item \textbf{Agreement between redundant fields.} When a serialization states a geometric fact in addition to encoding it in the coordinates (here, the wall an object touches), a geometric perturbation makes the statement contradict the coordinates, and the contradiction changes the likelihood by itself. Separate the two effects by scoring the perturbation with the statement kept, with the statement alone changed and with the statement removed (Sec.~\ref{sec:selfconsistency}).
  \item \textbf{Transformations that preserve the arrangement, and ties.} Average the score over transformations that keep the arrangement, such as mirroring (Sec.~\ref{sec:scoring}): without averaging, the real layout has the higher likelihood than its mirror image in only \MirrorRawAccKitchen{} of kitchens. Report the fraction of pairs whose strings do not differ.
\end{enumerate}

Averaging the likelihood over several serializations of the same layout, which we did only for mirroring, would further reduce the dependence on one format. These controls do not exclude a property of the text that changes together with a perturbation and has not been identified; they make the known properties measurable.

%% file: sections/limitations.tex
\section{Limitations}
\label{sec:limitations}

\paragraph{Number of room types.}
Every result on the dependence of placement on the room type rests on four ScanNet room types. Rank correlations
over four points have almost no power, and we treat them as descriptive
throughout; the claims are carried by the permutation test and the regression
slope, both of which resample rooms
(\nRoomsMin--\nRoomsMax{} per room type). An agreement with the real differences that is not detectable with four room types could appear with more room types.

\paragraph{Number of models.}
We evaluated two open-weight models of similar size, Qwen3.5-27B and Gemma-4-31B-it. For both, the placement does not follow the room type; they differ in that Qwen's wall-proximity excess varies little across room types and Gemma's varies in an order unrelated to the real one (Sec.~\ref{sec:hierarchy}). We report only the common property as a result. Larger or proprietary models may behave differently, and the results do not show that the limitation holds for language models in general.

\paragraph{Sources of the rooms.}
CubiCasa5K is Finnish residential floor plans; ScanNet is scanned interiors,
largely North American academic and residential spaces. Both the real
distributions we compare against and, plausibly, the models' training data are
skewed toward those conventions. Our experiments cannot distinguish between placements that do not depend on the room type and placements that depend on the room types of other sets of rooms.

\paragraph{Room-type labels.}
ScanNet's categories are \emph{Living room / Lounge} and \emph{Bedroom /
Hotel}: each name covers two settings. Shared lounges arrange seating in
islands rather than against walls, so a model asked for a ``living room'' and the set of real living rooms may not describe the same kind of room. We tested whether this affects the result. Restricting the living rooms to those
that look domestic (at most one sofa, at most three chairs, and below the median area) moves the excess from \MixLivingAll{} (\MixLivingAllCI, $n = \MixLivingAllN$) to \MixLivingDom{} (\MixLivingDomCI, $n = \MixLivingDomN$; Table~\ref{tab:app-mixture}, Appendix~\ref{app:mixture}). The excess of the subset is still negative, with an interval below zero, the order of the four room types by the real excess is unchanged, and the regression slopes change from \MixSlopeQwenAll{} to \MixSlopeQwenDom{} for Qwen and from \MixSlopeGemmaAll{} to \MixSlopeGemmaDom{} for Gemma. The mixture of the two kinds of rooms therefore changes the living-room excess but not the result that the generated placement does not follow the room type.

\paragraph{Partial scans.}
The rectangle of a ScanNet room is the extent of the scanned floor and walls, and many scans do not cover all four walls. With the wall coverage of Appendix~\ref{app:scancoverage}, only \ScanCovShareBath\% of the bathrooms, \ScanCovShareBed\% of the bedrooms, \ScanCovShareOffice\% of the offices and \ScanCovShareLiving\% of the living rooms have a wall scanned along at least half of every side of the rectangle. In the other rooms, at least one side of the rectangle is largely the edge of the scan and not a wall, so the real wall-proximity excess and the positions normalized by the rectangle depend on the extent of the scan. We tested whether this affects the result of Sec.~\ref{sec:hierarchy}. Restricted to the \ScanCovNHalf{} rooms with a coverage of at least $0.50$, the real excess is \ScanCovExcessBathHalf{} in bathrooms, \ScanCovExcessBedHalf{} in bedrooms, \ScanCovExcessOfficeHalf{} in offices and \ScanCovExcessLivingHalf{} in living rooms, against \ExcessRealBath, \ExcessRealBed, \ExcessRealOffice{} and \ExcessRealLiving{} for all rooms, and the regression slopes change from \SlopeQwenA{} to \ScanCovSlopeQwenHalf{} for Qwen and from \SlopeGemmaA{} to \ScanCovSlopeGemmaHalf{} for Gemma (Table~\ref{tab:app-scancoverage}). The values of the real excess therefore depend on the coverage, in particular in offices and living rooms, but the result that the generated placement does not follow the real differences between room types is unchanged. The search evaluation of Sec.~\ref{sec:search} was not repeated on this subset; the real-layout prior R is built from the same scans as its targets and can therefore also represent regularities of the scanning, which a language model cannot know.

\paragraph{Room shape.}
Every room is represented by a rectangle. For CubiCasa we keep only rooms whose polygon fills at least \RectMin{} of its oriented bounding box; for ScanNet no such selection is made, and the rectangle is the axis-aligned extent of the scanned floor and walls, so a non-rectangular room is represented by its bounding rectangle. Real rooms have alcoves, islands and non-convex boundaries, all of which carry placement information that our representation discards.

\paragraph{Two-dimensional layouts.}
Objects are footprints on the floor plane; we keep only instances whose base is
below \FloorZMax~m. Shelving, wall-mounted fixtures and stacking are outside the representation, and so is the heading of an object: we model where objects are and which wall they are against, not which way they face.

\paragraph{Generation and serialization settings.}
Every generation uses temperature $1.0$ without top-$k$ or nucleus truncation, with structured output against a fixed JSON schema and English prompts; other temperatures, decoding schemes and prompt languages were not explored. Every likelihood is that of one serialization, JSON with two-decimal coordinates and a fixed field order. The format matters on both sides: the field order alone changes what the model writes (Sec.~\ref{sec:symbols}), and reordering the object list alone changes the likelihood (Sec.~\ref{sec:perturbresults}). The values we report hold for this format. The vocabulary is closed (\nVocab{} categories for ScanNet, \nVocabCubi{} for CubiCasa), which makes the sets of layouts comparable but excludes the long tail of objects a robot may be asked to find.

\paragraph{Likelihood-based measurement.}
Every likelihood comparison in this paper is a comparison between two strings
(Sec.~\ref{sec:likelihooddesign}). The pairs are built so that the strings
differ as little as possible beyond the property of interest, and the known confounds (length, the order of the entries, the granularity of the numerals and the agreement between redundant fields) are controlled or reported. A property of the text that changes together with a perturbation and that we have not identified would still bias the result. The contradiction between the \texttt{wall\_contact} field and the moved coordinates (Sec.~\ref{sec:selfconsistency}) is an example of such a property, which we identified and measured.

\paragraph{Marginal and conditional distributions.}
Most measurements in this paper concern the distribution of placements before any object of the room is observed. A robot in a room uses the objects it has observed, and the relevant distribution is $p(x, y \mid o, c, \text{observed objects})$, which depends on relations between objects such as a night stand beside a bed. We measured such conditional placement in two ways only. In generation, the target of a related pair follows an object given in the prompt, measured in bedrooms for three pairs and one reversed pair (Sec.~\ref{sec:observed}). With the likelihood, the neighbor effect of the pair perturbations is small on CubiCasa and not distinguished from zero on ScanNet (Sec.~\ref{sec:relladder}). Other room types, other pairs and the effect of a conditional prior on the search cost were not measured, so our conclusions about the prior before observation do not cover the conditional prior.

\paragraph{Search cost as a proxy for navigation.}
The search evaluation gives the search policy the room rectangle and the target category, and measures the fraction of the room area that it examines before it finds the target. An object-goal navigation system must also move to the examined regions, handle occlusion, recognize the object and choose which room to enter. The evaluation also has two limits of its own. It uses only the real rooms whose door is recorded, so that rooms can be oriented consistently, which keeps \SearchRealCanonLiving{} of \SearchRealAllLiving{} living rooms. And because the goal category is given, it is insensitive to the room type: given the category, the real placements themselves depend little on the room type (Sec.~\ref{sec:searchcost}), so the result of Sec.~\ref{sec:hierarchy}, that the placement does not follow the room type, cannot appear as a search cost. This paper evaluates the prior, not a navigation system; an evaluation in a navigation simulator such as Habitat is future work.

%% file: sections/conclusion.tex
\section{Conclusion}
\label{sec:conclusion}

This paper has evaluated whether large language models can provide a spatial
prior for object-goal navigation. Using two open-weight models and real rooms
from CubiCasa5K and ScanNet, we examined the likelihood the models assign to
real and perturbed layouts and the layouts they generate, and we used the
generated placements as a prior for object search. The findings are as
follows. First, the object frequencies (Spearman \IdentRhoQwenA{} for Qwen)
and room sizes of generated rooms follow the room type, but object placement
does not: the slope of the model's wall-proximity excess against the real one
is \SlopeQwenA{} for Qwen and \SlopeGemmaA{} for Gemma, against $1$ for full
agreement. Second, the likelihood distinguishes real layouts from perturbed
ones partly through their geometry and partly through the agreement between
coordinates and stated wall contacts, and for the movable furniture of ScanNet it is higher for a layout with all objects piled within \PileRadius~m of the room center than for the real layout in most rooms of three of the four room types. Third, the
generated coordinates reflect regularities in how coordinates are written
rather than the constraints that determine them: they take few distinct values, mostly multiples of \EntGrid~m, they often disagree with the walls that the models declare, and \ConsOutsideQwenAppi\%
of Qwen's objects are placed partly outside the room even when containment is
the only constraint. Fourth, in bedrooms, when an object is given in the prompt and moved, the generated object of a related pair moves with it (following slope \NADeskChairQwen{} and \NABedNightstandQwen{} for Qwen, \NADeskChairGemma{} and \NABedNightstandGemma{} for Gemma), objects of other categories move away from it, and moving the night stand does not move the bed (\NARevQwen{} and \NARevGemma). Fifth, as a search prior, the generated placements reduce the fraction of the room searched before the target is found, relative to a uniform prior, by \DiffMAUQwenMin--\DiffMAUQwenMax{} (Qwen) and \DiffMAUGemmaMin--\DiffMAUGemmaMax{} (Gemma); a prior built from real layouts reduces it by more in every room type, and the reduction obtained with the generated placements is \ShareMAQwenMin--\ShareMAQwenMax\% (Qwen) and \ShareMAGemmaMin--\ShareMAGemmaMax\% (Gemma) of that obtained with real layouts. The difference from the real layouts lies in the positions of the individual categories rather than in the room type. Spatial smoothing of the priors lowers the search cost of the generated placements, more for Gemma, whose priors concentrate on few positions, but with the best smoothing for each prior the cost still exceeds that of the real-layout prior by \SmBestGapQwenMin--\SmBestGapQwenMax{} (Qwen) and \SmBestGapGemmaMin--\SmBestGapGemmaMax{} (Gemma), and the reduction relative to a uniform prior is \SmShareBestQwenMin--\SmShareBestQwenMax\% and \SmShareBestGemmaMin--\SmShareBestGemmaMax\% of that of the real-layout prior.

These findings lead to the following practical conclusions (Sec.~\ref{sec:design}). LLM-generated coordinates should not be used as a search prior where real layouts are available, and the likelihood should not be used to judge layouts of movable furniture. Object--room affinities from the models agree with real rooms. When the wall of each object is given, the models place the objects against it (consistency \ConsQwenApp{} for Qwen and \ConsGemmaApp{} for Gemma), so they can write coordinates for constraints determined by another component; every coordinate they emit must, however, be checked and corrected geometrically. In addition, likelihood-based measurement of spatial knowledge requires the controls listed in Sec.~\ref{sec:likelihooddesign}, because every comparison is a comparison between strings.

We plan to evaluate the generated priors in a full object-goal navigation system in a simulator, to measure placement given observed objects in other room types together with its effect on search, and to extend the evaluation to more models, more room types and other serializations of the layouts (Sec.~\ref{sec:limitations}).

%% file: sections/appendix_measurement.tex
\section{Measurement details}
\label{app:measurement}

This appendix states every quantity used in the paper as a formula or a
procedure, including the aggregation unit and the resampling scheme. Section
numbers in the main text point here.

\subsection{Canonicalization}
\label{app:canon}

A layout is a room rectangle $W \times D$, a set of openings, and a set of
axis-aligned objects, each with a center $(x, y)$ and a footprint
$w \times d$, all in meters from the south-west corner.

The canonical frame is defined by a door. \textbf{The principal door is the
widest door in the layout.} Let its wall be $\omega \in \{N, E, S, W\}$ and its
offset along that wall be $f \in [0, 1]$. We rotate the room anticlockwise
$k$ times, where $k = 0, 1, 2, 3$ for $\omega = S, W, N, E$ respectively, so
that the door lies on the south wall. One anticlockwise quarter turn maps
\begin{equation}
  (x, y) \mapsto (D - y,\; x), \qquad (w, d) \mapsto (d, w),
  \qquad W \leftrightarrow D,
  \label{eq:rot}
\end{equation}
and relabels wall annotations $S \to E \to N \to W \to S$. If the door then
lies right of center ($x_{\text{door}} > W/2$) we mirror in $x$:
$x \mapsto W - x$, with $E \leftrightarrow W$ on wall annotations. The result
places the door on the south wall, left of center.

\textbf{Layouts with no door cannot be canonicalized and are excluded from any
measurement that uses orientation.} We report the rate: on ScanNet it is
\CanonRateBath{} (bathrooms), \CanonRateBed{} (bedrooms), \CanonRateOffice{}
(offices) and \CanonRateLiving{} (living rooms), and at least
\CanonRateGenMin{} for every generated set. ScanNet door annotations sometimes
lack a wall label; there we assign the wall nearest the door's center. Of the rooms with a door, this applies to \DoorInferredBath{} of \DoorFoundBath{} bathrooms, \DoorInferredBed{} of \DoorFoundBed{} bedrooms, \DoorInferredOffice{} of \DoorFoundOffice{} offices and \DoorInferredLiving{} of \DoorFoundLiving{} living rooms.
Measurements that do not use orientation --- wall-proximity excess, search
cost, consistency --- do not require canonicalization and use all rooms.

\subsection{Serialization and scoring}
\label{app:serial}

A layout is serialized as JSON with the object list in the order given by the
source, each object rendered as
\begin{quote}\ttfamily\footnotesize\raggedright
\{"category": "bed",\\
\hspace*{1em}"center\_m": \{"x": 2.35, "y": 0.76\},\\
\hspace*{1em}"size\_m": \{"w": 2.03, "d": 1.52\},\\
\hspace*{1em}"wall\_contact": "N", "stated\_probability": 0.90\}
\end{quote}
\textbf{All numbers are written with exactly two decimal places}, so that the
token count does not vary with the value. The room type is not part of the
scored string; it appears only in the prompt. Keys whose name begins with an
underscore (provenance, room identifiers) are stripped before scoring.

Log-likelihoods are obtained from the completions endpoint with
\texttt{echo=true}, which returns per-token log-probabilities for a supplied
string; the score of a layout is the sum over the tokens of the serialized
JSON body. \textbf{Requests are issued sequentially}: with continuous batching
the returned log-probabilities depend on what else is in the batch, and mixing
scoring with generation changes them further. No KV-cache quantization is
used. Both models are served in bfloat16 with eager execution; the sampling parameters shipped with the models (temperature, top-$k$ and top-$p$) are disabled on the server, and every generation request sets only the temperature.

\paragraph{Symmetrization.}
Mirror symmetry is a property of the room, not of the model, so every score
used for a \emph{discrimination} comparison is symmetrized,
\begin{equation}
  \mathrm{score}(L) = \tfrac{1}{2}\bigl[\log p(L)
    + \log p(\mathrm{mirror}(L))\bigr],
  \label{eq:symmetrise}
\end{equation}
where $\mathrm{mirror}$ reflects the layout in $x$ and swaps $E$ and $W$ wall
annotations. \textbf{Equation~\eqref{eq:symmetrise} is applied to every perturbation test in Secs.~\ref{sec:ladder}--\ref{sec:pilemech} and to the snapping
control of Sec.~\ref{sec:adsorption}. It is not applied to generation:} the
generated layouts are never scored against a mirrored counterpart, and the
placement statistics of Sec.~\ref{sec:generation} are computed on the
layouts as generated.

\subsection{Wall-proximity excess}
\label{app:excess}

Let object $i$ have footprint $w_i \times d_i$ and center $(x_i, y_i)$ in a
room $W_j \times D_j$, and let $\beta = \BandM$~m. The object is
\emph{near a wall} when its nearest face is within $\beta$:
\begin{equation}
  n_i = \mathbf{1}\Bigl[\min\bigl(x_i - \tfrac{w_i}{2},\;
    W_j - x_i - \tfrac{w_i}{2},\; y_i - \tfrac{d_i}{2},\;
    D_j - y_i - \tfrac{d_i}{2}\bigr) \le \beta\Bigr].
  \label{eq:near2}
\end{equation}
Its baseline is the probability that an object of that size in a room of that
size would be near a wall if its center were uniform over the admissible
region, which is Equation~\eqref{eq:baseline}, repeated here as $b_i$. The
excess for a set of objects is
\begin{equation}
  \mathrm{excess} = \frac{\sum_i n_i - \sum_i b_i}{\sum_i 1},
  \label{eq:excess2}
\end{equation}
that is, \textbf{objects are pooled, not averaged per room first}. Rooms
contribute in proportion to how many objects they contain.

\textbf{The unit of resampling is nevertheless the room.} For a confidence
interval we resample rooms with replacement, recompute the three sums over the
resampled multiset of rooms, and take the $2.5$ and $97.5$ percentiles over
$10\,000$ draws. This is what makes the intervals cluster-robust: objects in
the same room are not independent, and treating them as such understates the
interval.

Only floor-standing objects in the closed vocabulary enter the sums. For
ScanNet, floor-standing means the instance's lower bound lies below
\FloorZMax~m above the floor.

\subsection{Perturbations}
\label{app:perturbations}

Each perturbation maps a reference layout to a copy with one property broken.
The object list, the object footprints and the room rectangle are
preserved unless stated. All perturbations are deterministic given the layout:
where a choice is needed, the random number generator is seeded with a hash of
the layout, so results do not depend on the order in which rooms are
processed or on resuming an interrupted run.

\paragraph{Pile (radius $\rho$, default $\rho = \PileRadius$~m).}
Every object is moved to a golden-angle spiral about the room center
$(W/2, D/2)$: object $i$ of $n$ is placed at
\begin{equation}
  r_i = \rho \sqrt{\frac{i + 0.5}{n}}, \qquad
  \theta_i = i \cdot \pi(3 - \sqrt{5}) \approx i \cdot 2.39958~\text{rad},
  \qquad (x_i, y_i) = \Bigl(\tfrac{W}{2} + r_i \cos\theta_i,\;
    \tfrac{D}{2} + r_i \sin\theta_i\Bigr).
  \label{eq:pile}
\end{equation}
The square root spaces the points by equal area, and the golden angle makes
every coordinate distinct. \textbf{Distinctness is deliberate}: placing all
objects at the same point would make the repeated numerals easy to predict and
would raise the likelihood for a reason unrelated to space. Objects overlap
after piling, which is the intended physical implausibility. The radius sweep
of Sec.~\ref{sec:pilemech} varies $\rho$ over $0.2$, $0.35$, $0.6$ and
$1.0$~m; we also report $\rho$ normalized by half the room's shorter side.

\paragraph{Detach (fraction $\phi = \DetachFrac$).}
\emph{Every} object that declares wall contact is moved perpendicularly away
from the wall it declares, by the single distance
$d = \min(\max(\phi \cdot \min(W, D),\, 0.10),\, 0.40)$~m: a fraction $\phi$ of
the room's shorter side, bounded below and above. The distance is deterministic
and shared by all objects in the room; it is not sampled per object. Each move
is clamped so that the object stays inside the room with a $0.05$~m margin, and
objects whose clamped displacement falls below the margin are left in place.

\textbf{The \texttt{wall\_contact} field is left as declared.} The perturbation
rewrites coordinates only, so the \texttt{wall\_contact} field of a detached object names a wall that the object no longer touches: after the edit, \ClaimFalseDetachKitchen\% of declarations
in kitchens and \ClaimFalseDetachBath\% in bathrooms disagree with the
coordinates, against \ClaimFalseRefKitchen\% and \ClaimFalseRefBath\% in the
reference. A declaration counts as agreeing when the declared wall lies within
\ClaimTol~m of the object's face, with a face inside the wall counting as
agreement, matching the wall-proximity convention of Equation~\eqref{eq:near} (Sec.~\ref{sec:excess}). Piling likewise leaves the field in contradiction with the coordinates (\ClaimFalsePileKitchen\% and \ClaimFalsePileBath\% in disagreement), as do
swapping positions and sizes. Sec.~\ref{sec:selfconsistency} separates the effect of this contradiction from that of the geometric change.

\paragraph{Swap positions.}
Two objects are chosen and their centers exchanged; footprints and categories
stay with their objects. \textbf{The serialized order of the object list is
unchanged}, so this perturbation does not confound position with list order.

\paragraph{Swap sizes.}
Two objects are chosen and their footprints exchanged, centers unchanged. If
the two footprints round to the same pair of numbers at two decimals the
perturbation is a no-op and the room is excluded as a tie.

\paragraph{Tangential and normal moves (distance $\delta$).}
A single object that touches a wall is moved by $\delta$ either along that
wall (tangential) or away from it (normal, outward) or into it (normal,
inward). \textbf{Both edits rewrite exactly one coordinate pair}, so the
string is perturbed to the same degree and only the geometric consequence
differs. Displacements are clamped to the room; a clamped move that leaves the
two-decimal string unchanged is a tie.

\paragraph{Reordering.}
The object list is sorted lexicographically by category, or shuffled with the
seeded generator. \textbf{No coordinate changes.} This is the control that
shows likelihood responds to non-spatial properties of the string.

\paragraph{Ties.}
A perturbation that leaves the serialized string unchanged --- through
clamping, or through rounding at two decimals --- does not change the text, and the room is excluded from that comparison rather than counted as an error. We report the tie rate for every perturbation test; ignoring it would make larger displacements look less detectable than they are.

\paragraph{Pairwise accuracy.}
For a set of rooms, the pairwise accuracy of a perturbation is the fraction of
non-tied rooms in which the reference scores strictly above the perturbed
copy. Chance is $0.5$. Confidence intervals resample rooms, $10\,000$ draws.

\begin{figure}[htbp]
  \centering
  \includegraphics[width=\linewidth]{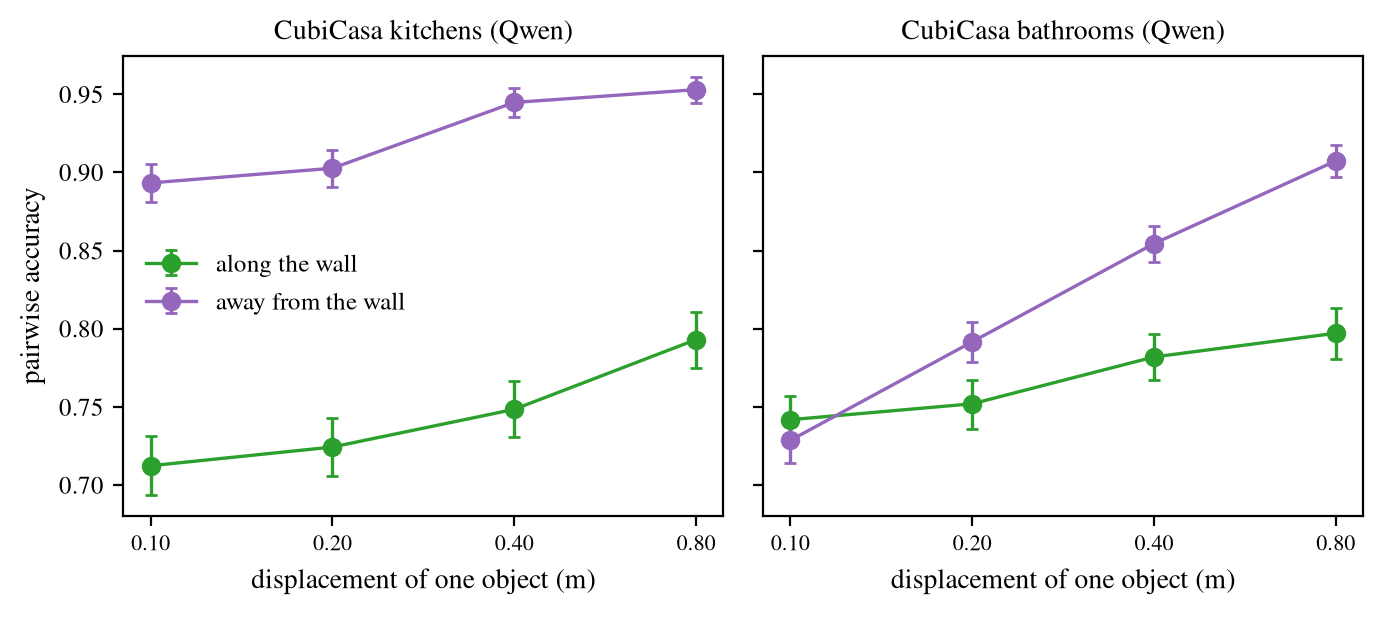}
  \caption{Pairwise accuracy when one object is moved along its nearest wall or away from it, against the distance moved (CubiCasa, Qwen; Table~\ref{tab:app-direction}).}
  \label{fig:dirsweep}
\end{figure}

\begin{figure}[htbp]
  \centering
  \includegraphics[width=\linewidth]{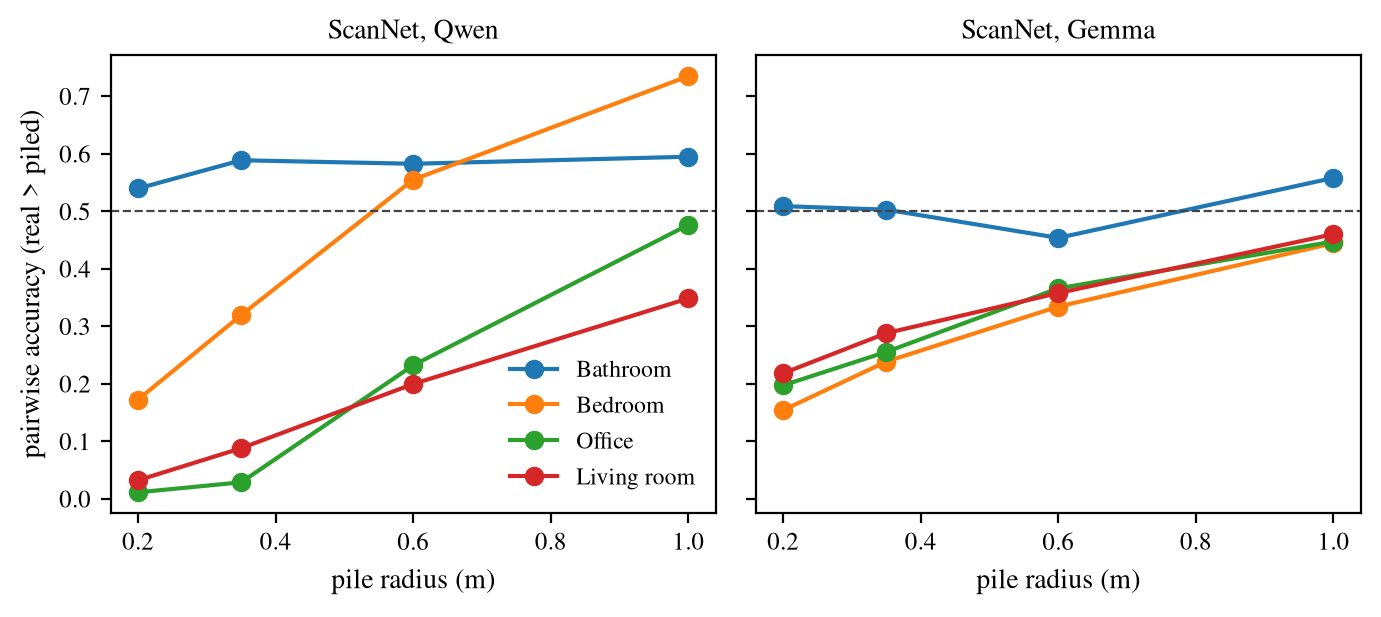}
  \caption{Pairwise accuracy of the real layout against the piled one as the pile radius varies (ScanNet; Table~\ref{tab:app-pile}). Below $0.5$ the piled layout is preferred.}
  \label{fig:pilesweep}
\end{figure}

Fig.~\ref{fig:dirsweep} shows the pairwise accuracy of the single-object moves against the distance moved, and Fig.~\ref{fig:pilesweep} the pile radius sweep.
\input{tables/app-ladder-cubi}

\input{tables/app-ladder-cubi-gemma}

\input{tables/app-ladder-qwen}

\input{tables/app-ladder-gemma}

\input{tables/app-direction}

\input{tables/app-wallclaim}

\input{tables/app-wallclaim-diff}

\input{tables/app-pile}

\FloatBarrier

\subsection{Relations between objects}
\label{app:relations}

This appendix gives the pairs used in the two relational measurements of the
main text and the details of the pair perturbations scored by likelihood
(Sec.~\ref{sec:relladder}). Appendix~\ref{app:observed} gives the details of
generation with an observed object (Sec.~\ref{sec:observed}).

\paragraph{Pairs.}
A pair is an ordered pair of categories (anchor, target). For every pair
occurring in at least \RelMinRooms{} rooms of a room type, the real data
give the median edge-to-edge distance from each target to its nearest
anchor (center distances cannot express ``beside'' for a large object). The
same median is computed under a band null that re-places each target at its
real distance from the nearest wall along a uniformly chosen wall
(\RelNullDraws{} draws), which removes crowding and the wall-proximity regularity measured by the excess of Sec.~\ref{sec:excess} while keeping everything else. The candidate list of a room type is the pairs whose null median
exceeds the real median by at least \RelMinGap~m, in decreasing order of that gap. Two objects are adjacent when
their edge-to-edge distance is at most \RelNear~m. The strongest pairs are
desk~$\rightarrow$~chair in bedrooms (gap \RelPairGapBedDeskChair~m, \RelPairNearBedDeskChair\% of chairs adjacent to a desk, against \RelPairNullBedDeskChair\% under the null), table~$\rightarrow$~chair in living rooms
and offices, bathtub~$\rightarrow$~towel in ScanNet bathrooms, and, in
CubiCasa, counter~$\rightarrow$~sink and sink~$\rightarrow$~dishwasher. The
CubiCasa pairs are of two kinds, containment (a sink inside a counter) and
adjacency (a dishwasher beside a sink); they are reported together because the analysis pools each dataset. The negative controls of the pair perturbations are bed~$\rightarrow$~bed (\RelPairNearBedBedBed\% of beds adjacent to another bed, against \RelPairNullBedBedBed\% under the null) and sofa~$\rightarrow$~chair in living rooms (\RelPairNearLivingSofaChair\% against \RelPairNullLivingSofaChair\%). Table~\ref{tab:app-rel-pairs} lists the pairs actually
used.

\paragraph{Pair perturbations (Sec.~\ref{sec:relladder}).}
For each room of both datasets we take the first pair of the candidate list
that the room contains with the two objects adjacent, and move the target.
Three versions are built from one set of candidate positions, obtained by
sliding the target along its nearest wall in steps of \RelStep~m while
keeping its distance to that wall (so its \texttt{wall\_contact} is
preserved), keeping it inside the room, not increasing the footprint overlap
of the layout, and keeping the two-decimal string the same length. In the
\emph{no-neighbor} version the destination is adjacent to nothing; in the
\emph{wrong-neighbor} version it is adjacent to an object of a category other than
the anchor's; in the \emph{same-neighbor} version it remains adjacent to its
anchor. All three move the same object by the same distance $d$ to within
$\pm\RelDTol$\%, with $d$ the largest displacement at which both the wrong-neighbor
and the same-neighbor version exist; rooms without such a $d$ are dropped
(Table~\ref{tab:app-rel-pairs} gives the yield). The no-neighbor version requires
an empty destination and exists in fewer rooms; it is not required for the
primary measure. Scoring and symmetrization are those of Sec.~\ref{sec:scoring}, and ties are treated as in Appendix~\ref{app:perturbations}.

Sec.~\ref{sec:relladder} defines the two measures and gives the pooled results; Table~\ref{tab:app-rel-ladder} adds the breakdown by room type. In that breakdown the kitchen carries the CubiCasa effect (\RelPairingQwenKitchen{} \RelPairingCIQwenKitchen{} for Qwen, \RelPairingGemmaKitchen{} \RelPairingCIGemmaKitchen{} for Gemma; the bathroom is at zero), and the bedroom is the one ScanNet room type with a positive interval for Qwen (\RelPairingQwenBed{} \RelPairingCIQwenBed{}; Gemma \RelPairingGemmaBed{} \RelPairingCIGemmaBed{}). Table~\ref{tab:app-rel-hygiene}
shows that the three versions do not differ in the measures of coordinate diversity, overlap or protrusion; the one quality-check quantity that separates them
is the declared-wall agreement of the no-neighbor version, which is lower because
a target slid to an empty stretch of wall more often ends in a corner where
its declaration names the other wall. The wrong-neighbor and same-neighbor versions, on
which the primary measure rests, have identical agreement, so the primary
measure is not affected by the self-consistency effect of
Sec.~\ref{sec:selfconsistency}; the secondary measure may be.

The layouts carry no instance identity, so a target re-attached beside a
second instance of its anchor's category has not had its pairing broken at
the level the model sees. The construction rejects destinations whose only
neighbor is such an instance; Table~\ref{tab:app-rel-samecat} counts the
remaining cases, \RelSameCatN{} of \RelSameCatTotal{} triples
(\RelSameCatPct\%). Excluding them moves the CubiCasa neighbor effect to
\RelPairingCleanQwenCubi{} \RelPairingCleanCIQwenCubi{} (Qwen) and
\RelPairingCleanGemmaCubi{} \RelPairingCleanCIGemmaCubi{} (Gemma), inside
the original intervals.

\paragraph{Negative controls of the pair perturbations.}
The pair perturbations were also run on two pairs without a relation,
bed~$\rightarrow$~bed and sofa~$\rightarrow$~chair. Because the construction
needs the two objects to be adjacent in the real room, and unrelated pairs
rarely are, only \RelNegNBed{} and \RelNegNLiving{} rooms qualified. The
neighbor effect on the pooled controls is \RelNegPairingQwen{}
\RelNegPairingCIQwen{} (Qwen) and \RelNegPairingGemma{}
\RelNegPairingCIGemma{} (Gemma), both with intervals containing zero but much wider than the CubiCasa effect. Since the main result is that the neighbor effect is small, the low resolution of this control does not affect it.

\begin{figure}[htbp]
  \centering
  \includegraphics[width=\linewidth]{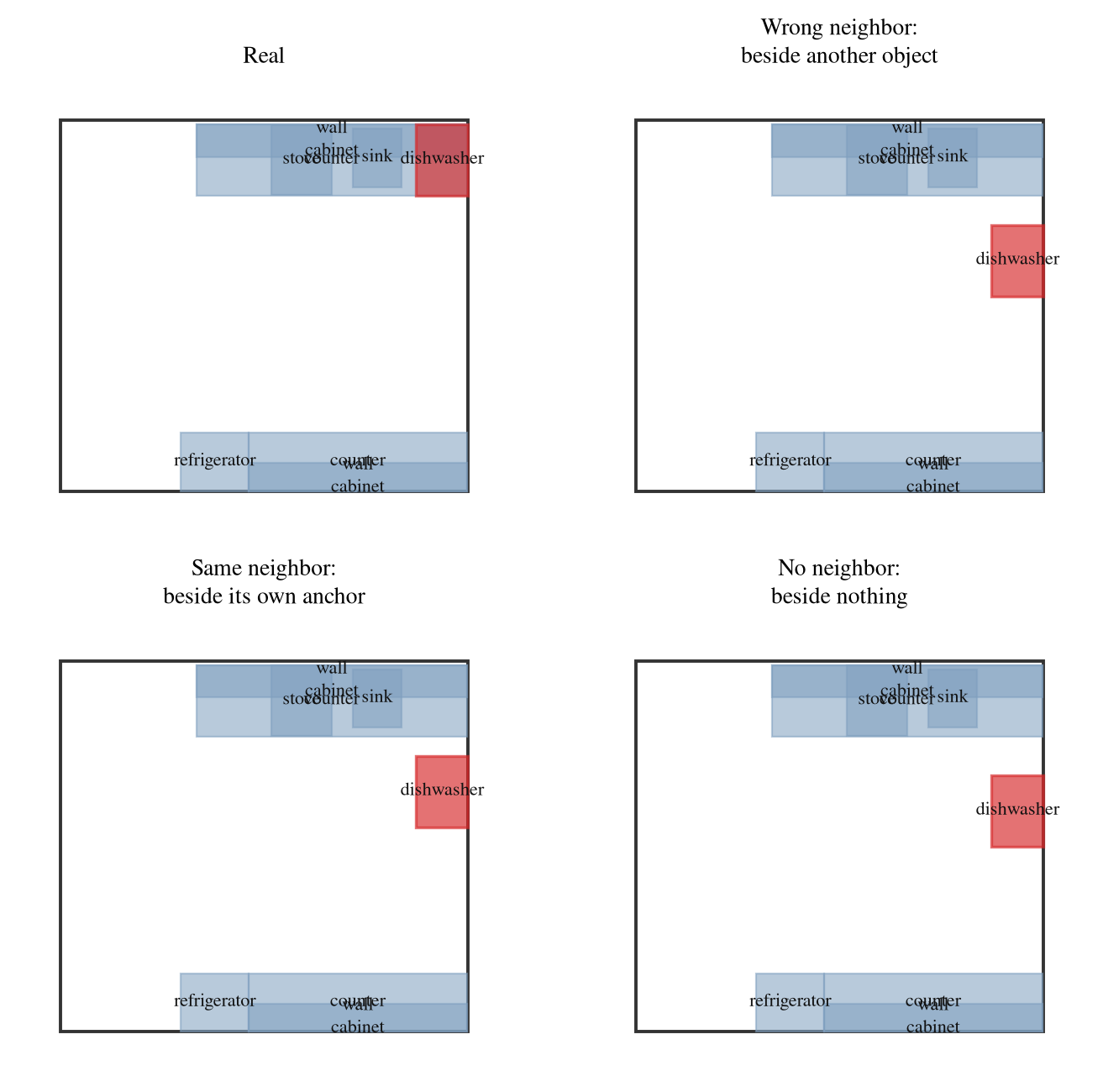}
  \caption{The three versions of the pair perturbation of Sec.~\ref{sec:relladder} for one CubiCasa kitchen (anchor: sink; target, in red: dishwasher). Top left: the real layout. All three versions slide the target along its wall by the same distance and differ in what it ends up beside.}
  \label{fig:relpert}
\end{figure}

Fig.~\ref{fig:relpert} shows the three versions of the pair perturbation for one kitchen.
\input{tables/app-rel-pairs}

\input{tables/app-rel-ladder}

\input{tables/app-rel-hygiene}

\input{tables/app-rel-samecat}

\FloatBarrier

\subsection{Conditioning measurements}
\label{app:conditioning}

The tables below give the inputs of Sec.~\ref{sec:hierarchy} and
Sec.~\ref{sec:overwrite}: the category frequencies behind the
Spearman correlation (Table~\ref{tab:app-identity}), the room areas behind
the dimension ratio (Table~\ref{tab:app-area}), the wall-proximity excess by
room type with the permutation test of equality across room types (Table~\ref{tab:app-excess}), the
regression slope (Table~\ref{tab:app-slope}) and the geometric validity of
the generated layouts (Table~\ref{tab:app-validity}). The excess is defined
in Appendix~\ref{app:excess}, the slope in Appendix~\ref{app:slope} and the
test in Appendix~\ref{app:permutation}.

Fig.~\ref{fig:gallery} shows example layouts for each room type and condition.
\input{tables/app-identity}

\input{tables/app-area}

\input{tables/app-slope}

\begin{table}[tbp]
  \centering
  \caption{Agreement with the real differences for two placement
    statistics (ScanNet, four room types), condition~\condA{}. Full agreement would be $1.00$ in every
    row. Orientation shares sum to one, so the four orientation rows are not
    independent.}
  \label{tab:orientation}
  \begin{tabular}{lcc}
    \toprule
    Statistic & Qwen & Gemma \\
    \midrule
    Wall-proximity excess & \SlopeQwenA & \SlopeGemmaA \\
    Orientation, $N$ & \OriAllQwenN & \OriAllGemmaN \\
    Orientation, $E$ & \OriAllQwenE & \OriAllGemmaE \\
    Orientation, $S$ & \OriAllQwenS & \OriAllGemmaS \\
    Orientation, $W$ & \OriAllQwenW & \OriAllGemmaW \\
    Orientation, combined & \OriAllQwenCentred & \OriAllGemmaCentred \\
    \bottomrule
  \end{tabular}
\end{table}

\input{tables/app-excess}

\input{tables/app-validity}

\begin{figure}[htbp]
  \centering
  \includegraphics[width=\linewidth]{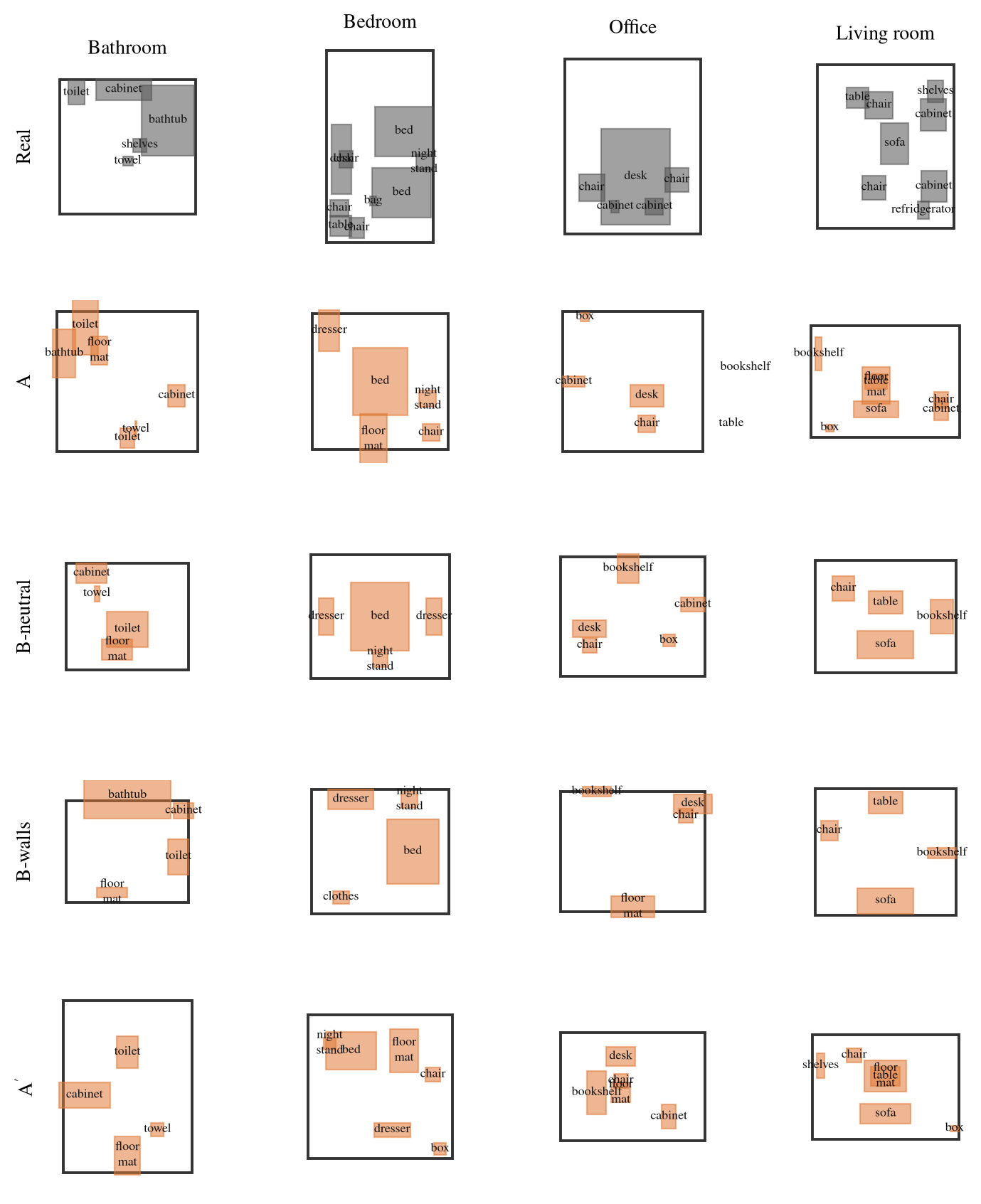}
  \caption{Example layouts by room type: a real ScanNet room and Qwen's generations under conditions~\condA, \condBn, \condBw{} and \condAp. Rows are the real rooms and the conditions, and columns are the room types. Each panel is the first layout of its set with four to nine objects; the panels are illustrative.}
  \label{fig:gallery}
\end{figure}
\FloatBarrier

\subsection{Resolution of the orientation measurement}
\label{app:orientation}

The resolution of this measurement on ScanNet is limited, because
\textbf{the reference signal itself is weak}: real orientation distributions there are close to uniform
(Jensen--Shannon divergence from uniform
\OriRealUnifMinAll--\OriRealUnifMaxAll, against a maximum of $\ln 2$),
whereas the real wall-proximity excess varied across room types with a
standard deviation of \RealSD.

In CubiCasa bathrooms, the real orientation distribution is far from uniform: the door wall holds \CubiOriRealBathS{} of the objects and the east wall \CubiOriRealBathE. Qwen's layouts under the three prompting conditions have \CubiOriAMinS--\CubiOriAMaxS{} of their objects on the door wall, close to uniform, and diverge from the real distribution by \CubiOriJSDMin--\CubiOriJSDMax{} nats, an order of magnitude more than in any ScanNet room type.

In the canonical frame, the door is on the south wall. In the real ScanNet rooms, the share of the south wall is the smallest of the four walls in three of the four room types (\OriAllRealBathS, \OriAllRealBedS{} and \OriAllRealOfficeS{} in bathrooms, bedrooms and offices). Gemma's share of the south wall is lower than the real one in bathrooms (\OriAllGemmaBathS), and Qwen's is not (\OriAllQwenBathS).

Tables~\ref{tab:app-orient-all} and~\ref{tab:app-orient-near-wall-only} give the orientation distributions by room type and condition, for all objects and for the objects within $0.15$~m of a wall.

\begin{figure}[htbp]
  \centering
  \includegraphics[width=\linewidth]{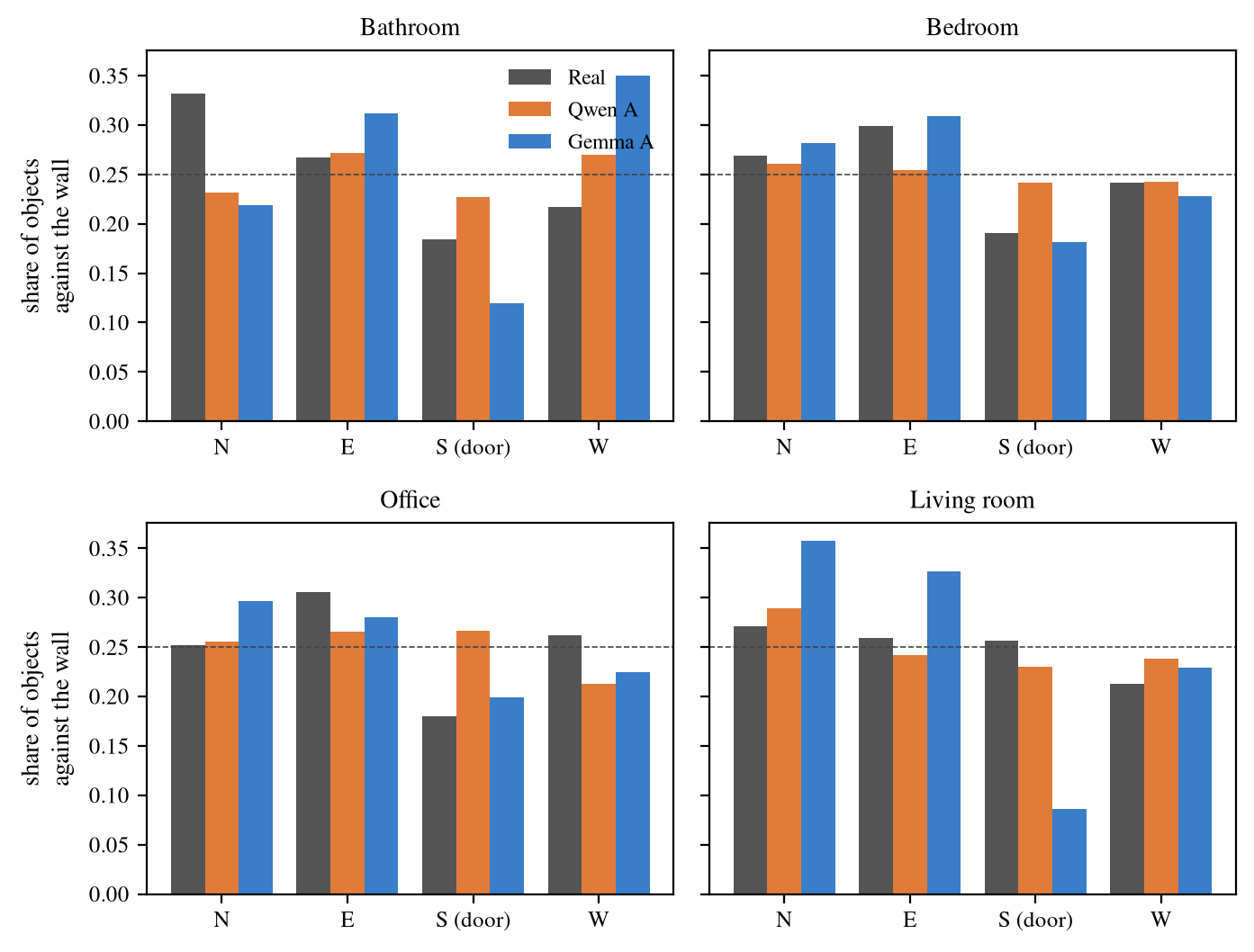}
  \caption{Share of the objects against each wall in the canonical frame (door on the south wall), for the real rooms and condition~\condA{} (Tables~\ref{tab:app-orient-all} and~\ref{tab:app-orient-near-wall-only}). The dashed line is a uniform share.}
  \label{fig:orient}
\end{figure}

Fig.~\ref{fig:orient} shows the orientation distributions of the real rooms and of condition~\condA{}.
\input{tables/app-orient-all}

\input{tables/app-orient-near-wall-only}

\FloatBarrier

\subsection{Generation with an observed object}
\label{app:observed}

The inputs are the ScanNet bedrooms whose rectangle and door can be recovered
(Appendix~\ref{app:canon}), which have \AppInMinObj{} to \AppInMaxObj{}
floor-standing objects of the vocabulary, and which contain both categories
of the pair; adjacency is not required. If the room has several instances of
the anchor's category, one is chosen with a fixed seed. Under O-move the
anchor is placed on a wall other than its own, at its real distance from
that wall and at a uniformly chosen position along it, with its footprint
unchanged; a position that overlaps another object is redrawn, and a room in
which \ObsMoveTries{} attempts per wall fail is dropped. Beds are large, so
this drops \NADropBedNightstand{} of the \NABothBedNightstand{} rooms for
bed~$\rightarrow$~night stand and \NADropBedChair{} of the \NABothBedChair{}
for bed~$\rightarrow$~chair, against \NADropDeskChair{} of
\NABothDeskChair{} for desk~$\rightarrow$~chair. The prompt is that of
condition~C with one line listing the anchor (Appendix~\ref{app:prompts}).
Five layouts are generated per room and condition at temperature $1.0$ with
structured output. The following slope is
\begin{equation}
  \kappa = \frac{\sum_{\text{rooms}} \Delta t \cdot \Delta a}
               {\sum_{\text{rooms}} \Delta a \cdot \Delta a},
  \label{eq:beta}
\end{equation}
with $\Delta a$ the displacement of the anchor between the two prompts and
$\Delta t$ the displacement of the mean generated target position, the five
samples of a room averaged before regression; rooms in which the target
category is not generated under one of the two prompts are left out. For the
other objects, the same slope is computed for every generated category that
is neither the anchor's nor the target's, category by category within a room;
it is reported for the desk~$\rightarrow$~chair and bed~$\rightarrow$~night
stand runs, because in the bed~$\rightarrow$~chair runs the other objects
include the night stand, the target of bed~$\rightarrow$~night stand. The difference between two pairs
resamples the union of their rooms, a room present in both being drawn
together. Table~\ref{tab:app-rel-nonadj} gives every value.

The control pair bed~$\rightarrow$~chair is not adjacent in real bedrooms more often than under the null placement of Appendix~\ref{app:relations}, and it is taken from the same room type as the related pairs. Comparing it with the related pairs distinguishes placement relative to a related object from placement near any object given in the prompt. Bed~$\rightarrow$~night stand is the pair that real bedrooms keep adjacent most consistently. Night stand~$\rightarrow$~bed reverses it and measures whether the dependence has a direction. The inputs do not require the pair to be adjacent in the real layout, so that the control pair has a number of rooms comparable to the related pairs.

\begin{figure}[htbp]
  \centering
  \includegraphics[width=\linewidth]{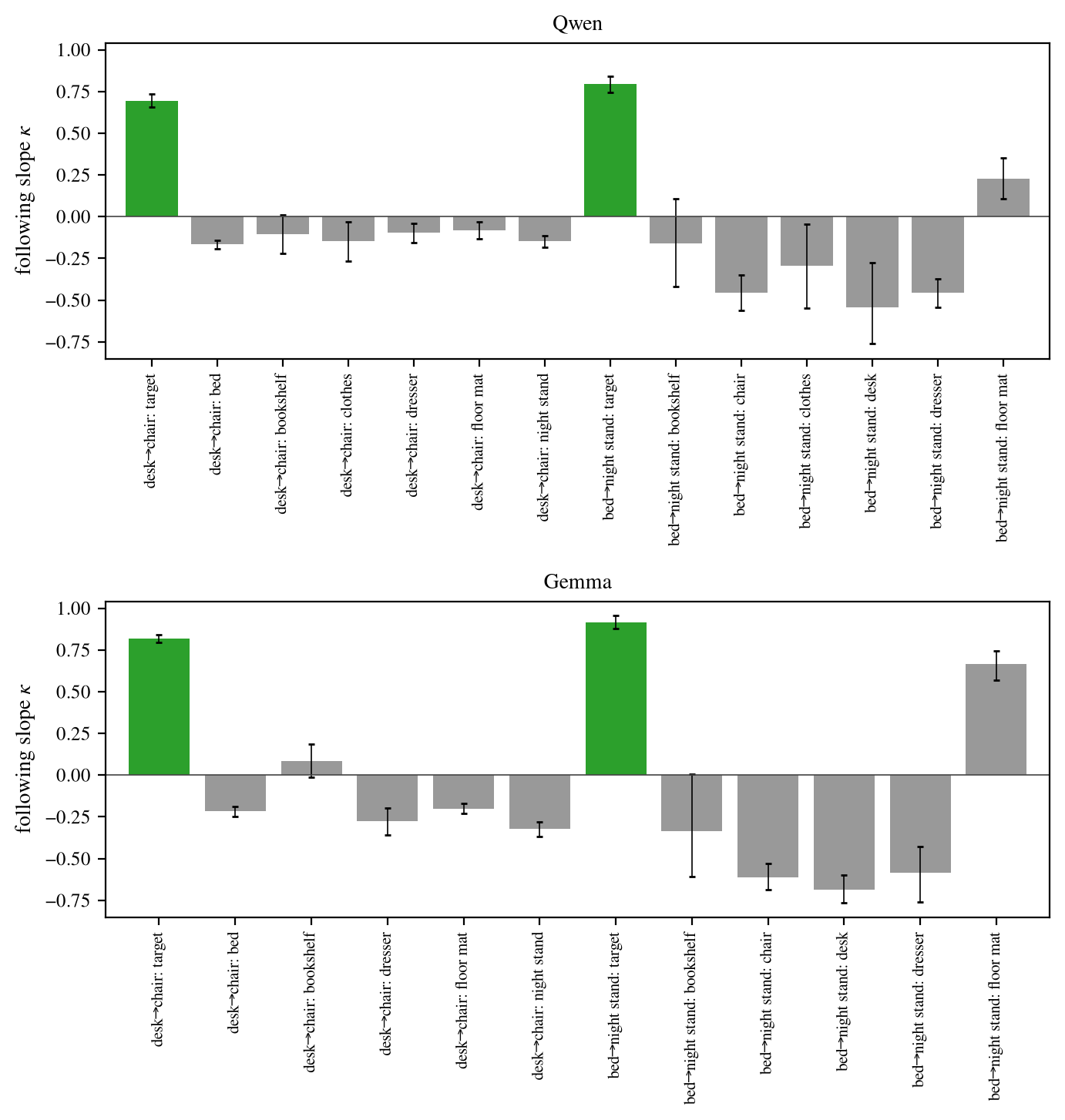}
  \caption{Following slopes in the generation experiment of Sec.~\ref{sec:observed}: the target of the listed object (green) and every other generated category (gray), in the desk~$\rightarrow$~chair and bed~$\rightarrow$~night stand runs, with intervals over rooms. Top: Qwen; bottom: Gemma. Besides the target, only the floor mat follows the bed, in both models.}
  \label{fig:nonpartner}
\end{figure}

Fig.~\ref{fig:nonpartner} shows the following slopes of every generated category.
\input{tables/app-rel-nonadj}

\FloatBarrier

\subsection{Regression slope and its interval}
\label{app:slope}

For each of the $T = 4$ ScanNet room types we compute the real excess $r_t$
and the model excess $m_t$ by Equation~\eqref{eq:excess}. The regression slope
is the ordinary least-squares slope of $m$ on $r$ over those $T$ points, with
an intercept, unweighted:
\begin{equation}
  \hat\beta = \frac{\sum_t (r_t - \bar r)(m_t - \bar m)}
                   {\sum_t (r_t - \bar r)^2}.
  \label{eq:slope}
\end{equation}
Full agreement gives $\hat\beta = 1$; indifference to room type gives
$\hat\beta = 0$. The intercept absorbs any constant offset, so the slope
measures only whether the model moves \emph{with} the real data.

\textbf{The interval resamples rooms, not room types.} In each of $5\,000$
draws we resample the rooms of every room type with replacement,
\emph{independently on the real and the model side}, recompute all $2T$
excesses, and refit Equation~\eqref{eq:slope}; the reported interval is the
$2.5$--$97.5$ percentile range of those slopes. Draws in which the real
excesses have zero variance are discarded. With $T = 4$ the slope is a
four-point fit and we treat it as descriptive, alongside the permutation test
below.

The orientation version of this measurement (Sec.~\ref{sec:hierarchy})
fits the same equation to the four wall shares. Because the shares sum to one
we also report a version in which each wall's mean across room types is
subtracted before fitting, which removes the profile shared by the model and
the real data and leaves only room-type variation.

\subsection{Permutation tests}
\label{app:permutation}

The null hypothesis of this test is that all $T$ room types have the same placement
statistic. \textbf{The unit shuffled is the room.} We pool the per-room
aggregate rows across room types, permute the room-type labels, split the
permuted pool into groups of the original sizes, and recompute the statistic.
The reported $p$ is $(\#\{\text{permuted} \ge \text{observed}\} + 1) /
(N + 1)$ with $N = 10\,000$.

Two statistics are used. For wall-proximity excess it is the sample standard
deviation of the $T$ group excesses (we also report the range, which gives the
same verdicts). For orientation it is the mean Jensen--Shannon divergence over
the $\binom{T}{2}$ pairs of group distributions,
\begin{equation}
  \mathrm{JSD}(p, q) = \tfrac{1}{2} D_{\mathrm{KL}}(p \,\|\, \tfrac{p+q}{2})
    + \tfrac{1}{2} D_{\mathrm{KL}}(q \,\|\, \tfrac{p+q}{2}),
  \label{eq:jsd}
\end{equation}
in nats, so that the maximum is $\ln 2$.

\textbf{Power differs between sets of layouts}, because the number of rooms
differs; a $p$ value is therefore not comparable across conditions on its own.
Where we compare the size of the variation we report the ratio of real to
model spread with a bootstrap interval, which is comparable.

\subsection{Wall-snapping control}
\label{app:adsorption}

The control of Sec.~\ref{sec:adsorption} starts from the model's own
generations under condition~\condA{} (Qwen and Gemma, CubiCasa kitchens and bathrooms,
\NGenPerRoom{} layouts per room type and model; each model scores its own generations) and builds four versions of each layout. Before
anything else, every coordinate and footprint value of the original is
rounded to the $0.05$~m lattice; all versions are built from this rounded
original, so that membership of the lattice cannot distinguish them.

\begin{description}
  \item[Snapped] Every object whose \texttt{wall\_contact} is one of
    N, E, S, W is moved perpendicularly to that wall until its face touches
    it; the other coordinate is unchanged. The target is rounded to the
    lattice and, if it would put the object outside the room, clamped so
    that the object lies inside; the number of clamped objects is reported.
    Objects declared as touching no wall are not moved. The declaration is
    left as it is, which the move makes true.
  \item[Random] Each object that the snapped version moves is moved by
    \emph{the same distance} in a random direction, trying sixteen equally
    spaced directions from a seeded starting angle and taking the first that
    keeps the object inside the room; if none does, the move is clamped. This
    is the distance-matched control: it differs from the snapped version
    only in direction.
  \item[Other wall] Each object is snapped, by the same rule, to a wall
    other than the one it declared, and the declaration is rewritten to the
    new wall, so that this version is self-consistent and differs from the
    snapped version only in which wall the object touches.
\end{description}

Every comparison scores two versions of the same layout with the
symmetrized score of Equation~\eqref{eq:symmetrise}; pairwise accuracy is the
fraction of layouts in which the first-named version scores higher, ties
excluded, with a bootstrap over layouts. The displacement statistics are
taken over the objects that the snapped version moves (a declared wall, and
a non-zero move after rounding), as the Euclidean distance between the
original and the snapped center. For each version we record the number of
distinct values among all coordinate and footprint values after rounding to
two decimals, the share on the lattice, and the median token count of the
serialized layout. Table~\ref{tab:app-adsorption} gives the comparisons and
Table~\ref{tab:app-adsorption-diag} the diagnostics.

\input{tables/app-adsorption}

\input{tables/app-adsorption-diag}

\FloatBarrier

\subsection{Coordinate granularity}
\label{app:granularity}

For the comparison between generated and real layouts in
Sec.~\ref{sec:entropy}, layouts are matched on object count: within each
count bin we take the difference of mean log-likelihoods, and average the bins
weighted by the number of real rooms. Quantization to a grid $\gamma$ replaces
every coordinate by $\gamma \lfloor v/\gamma + 1/2 \rfloor$ before
serialization, on both sides.

Distinct coordinate values are counted after rounding to two decimals, over
all objects of a set of layouts, pooling $x$ and $y$. The same count for the
wall-snapping control of Sec.~\ref{sec:adsorption} is taken after every version
has been quantized to the \AdsGrid~m lattice, which the control applies to all
four versions so that lattice membership cannot separate them, and it
additionally pools the two footprint dimensions, which snapping leaves
untouched; the difference between versions therefore comes from the
coordinates alone. It is recorded alongside the median token count of each
version, which snapping also leaves unchanged. These two conventions ---
quantized against as emitted, with footprints against without --- are why the
counts in Sec.~\ref{sec:adsorption} differ by a few values from those in
the same section's comparison with real rooms. The spread measures used in
Sec.~\ref{sec:pilemech} are the standard deviation of those values within a
room, the number of distinct values after rounding to one decimal, and the
number of distinct integer parts; each is reported as a reduction relative to
the same measure on that room type's real layouts.

\begin{figure}[htbp]
  \centering
  \includegraphics[width=\linewidth]{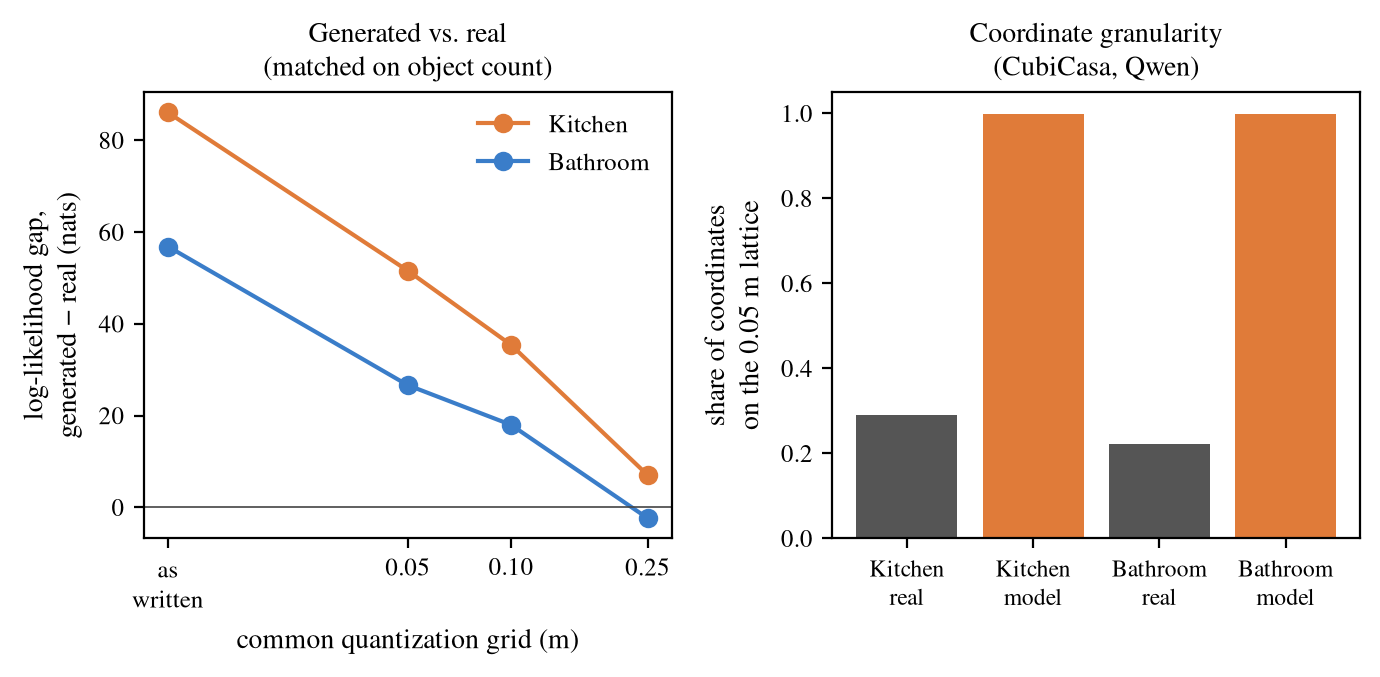}
  \caption{Left: likelihood gap between Qwen's generated kitchens and bathrooms and real ones with the same object count, as both are quantized to coarser grids (Table~\ref{tab:app-entropy}). Right: share of coordinates on the $0.05$~m lattice (Table~\ref{tab:app-coordvalues}).}
  \label{fig:granular}
\end{figure}

Fig.~\ref{fig:granular} shows the likelihood gap as a function of the quantization grid and the share of coordinates on the $0.05$~m grid for Qwen; Table~\ref{tab:app-entropy} gives the values for both models.
\input{tables/app-entropy}

\input{tables/app-coordvalues}

\FloatBarrier

\subsection{Consistency between declared and emitted walls}
\label{app:consistency}

Let $\mathcal{D}$ be the set of objects for which a wall $\omega_i$ is
declared by the model (conditions~\condA, \condAp) or supplied to it
(condition~\condApp); objects marked as touching no wall are excluded. Write
$g_i$ for the gap between the object's face and that wall,
\begin{equation}
  g_i = \begin{cases}
    x_i - w_i/2 & \omega_i = W \\
    W_j - x_i - w_i/2 & \omega_i = E \\
    y_i - d_i/2 & \omega_i = S \\
    D_j - y_i - d_i/2 & \omega_i = N.
  \end{cases}
  \label{eq:gap}
\end{equation}
Consistency is $|\mathcal{D}|^{-1} \sum_{i \in \mathcal{D}}
\mathbf{1}[g_i \le \beta]$ with $\beta = \BandM$~m.

\paragraph{Chance level.}
The declared walls are not distributed uniformly (Table~\ref{tab:app-declgeo}), so the chance level is not $1/4$. Let $q(\omega)$ be the
distribution of declared walls within $\mathcal{D}$, and let $p(\omega)$ be
the fraction of the same objects whose face lies within $\beta$ of wall
$\omega$, whatever they declared. The chance level is
\begin{equation}
  \mathrm{chance} = \sum_{\omega} q(\omega)\, p(\omega),
  \label{eq:chance}
\end{equation}
the consistency expected if declarations were drawn from $q$ independently of
the coordinates. We report it alongside every consistency figure.

\paragraph{Inputs of condition~\condApp.}
The inputs of condition~\condApp{} are built from the real ScanNet rooms, one input per room. A room keeps its floor-standing objects that belong to the closed vocabulary; it is canonicalized so that its door lies on the south wall (Appendix~\ref{app:canon}), and rooms whose door cannot be recovered are dropped; rooms with fewer than \AppInMinObj{} or more than \AppInMaxObj{} remaining objects are dropped as well. For every object the input gives its category, its footprint $w \times d$ and a wall: the wall nearest to the real object if the real object's face lies within \AppInTol~m of it, and ``none'' otherwise. Of the \AppInEligibleTotal{} eligible rooms, at most \AppInMax{} per room type are drawn after a shuffle with a fixed seed. Only bedrooms exceed the limit (\AppInUsedBed{} of \AppInEligibleBed{} are used); the other room types use every eligible room, which gives \AppInUsedTotal{} rooms in total. Table~\ref{tab:app-appinputs} gives the count after each rule; most living rooms are lost because their door cannot be recovered. About \AppInNoWallPct\% of the supplied objects carry ``none'', and these are excluded from the consistency measure.

\paragraph{Feasibility of the inputs.}
An object of footprint $w \times d$ can be placed inside a room $W \times D$
while touching any given wall if and only if $w \le W$ and $d \le D$. Of the
\AppFeasObjectsQwen{} objects supplied under condition~\condApp{}
(\AppFeasRoomsQwen{} rooms, identical for both models),
\AppFeasInfeasibleQwen{} fails this test; the rate of objects placed outside
the room is therefore not explained by infeasible inputs.

\paragraph{Natural and unnatural categories.}
For the split reported in Table~\ref{tab:app-symaux}, a category is
\emph{natural} if, in the real ScanNet rooms, at least half of its instances
are within $\beta$ of a wall, and \emph{unnatural} otherwise. The supplied
wall for an object of an unnatural category is still the wall of the real
object, so the split compares the consistency for walls that are atypical for the category with that for typical walls.

\paragraph{Paired comparisons.}
Conditions \condApp{} and its variants share their inputs by construction
(same seed, same rooms, same objects, same walls), so differences are paired
by source room and the bootstrap resamples those paired rooms.

\input{tables/app-symcoord}

\input{tables/app-appinputs}

\input{tables/app-declgeo}

\input{tables/app-condap}

\input{tables/app-priority}

\input{tables/app-symaux}

\FloatBarrier

\subsection{Declared wall against nearest wall}
\label{app:declgeo}

For each generated object that declares a wall, the \emph{nearest} wall is
the wall with the smallest gap $g_i$ of Equation~\eqref{eq:gap}, computed for
all four walls. Agreement is the fraction of objects whose declared wall is their nearest wall; its chance level is $\sum_\omega q(\omega)\,p(\omega)$ with $q$ the declared and $p$ the nearest distribution, the agreement expected if declarations were independent of the coordinates. The declared distribution is the distribution of
\texttt{wall\_contact} values over those objects, the nearest distribution
is the distribution of their nearest walls, and the divergence between the
two is the Jensen--Shannon divergence of Equation~\eqref{eq:jsd}. Unlike the
consistency of Appendix~\ref{app:consistency}, this measure uses no distance
threshold. Table~\ref{tab:app-declgeo} gives the values for every model and
condition.

\subsection{Search-cost evaluation}
\label{app:searchcost}

\paragraph{Grid and priors.}
The room is divided into a $G \times G$ grid, $G = \GridN$, in normalized room
coordinates; cell $(i, j)$ covers
$[jW/G, (j{+}1)W/G] \times [iD/G, (i{+}1)D/G]$. A prior is a probability vector
over the $G^2$ cells. \textbf{U} is uniform. \textbf{W} places uniform mass on
every cell whose center lies within $\beta$ of a wall, plus half a cell
diagonal to avoid a resolution artifact. \textbf{M-A} and \textbf{M-Bw} are
histograms of the generated object centers for that category, in normalized
coordinates, \emph{built only from generated rooms that contain the category},
with no smoothing. \textbf{R} is the same histogram built from the real rooms
of that room type, \emph{excluding the room being evaluated} (leave-one-out). \textbf{M-A/cat} is the histogram of all generated object centers in the room type, whatever their category; \textbf{M-A/room} is the histogram of the target category's generated centers pooled over the four room types. Both are evaluated only on episodes for which M-A is defined. \textbf{R/cat} and \textbf{R/room} are the same two ablations of R, built from the real rooms with the evaluated room held out. The main evaluation applies no smoothing (Sec.~\ref{sec:searchmethod} gives the reason); Sec.~\ref{sec:smoothing} evaluates spatial smoothing, defined below. Before any histogram is built, every room, real or generated, is canonicalized as in Appendix~\ref{app:canon} (door on the south wall, left of center); real rooms without a door are excluded (Sec.~\ref{sec:searchmethod} gives the counts), and generated rooms almost all have one.

\paragraph{Smoothing.}
For Sec.~\ref{sec:smoothing}, each histogram (M-A, M-A/room and R, the last after the evaluated room has been held out) is reshaped to the $G \times G$ grid, convolved with an isotropic Gaussian kernel of standard deviation $\sigma \in \{0.5, 1, 2\}$ cells with reflection at the grid boundary, and renormalized to sum to one. The episodes, the averaging over rooms and the paired comparisons are those of the unsmoothed evaluation; M-A/room is evaluated on the episodes for which M-A is defined. Table~\ref{tab:app-smoothing} gives the paired differences. The split of the targets by the cell's probability before smoothing (Sec.~\ref{sec:smoothing}) pools the episodes of the four room types.

\begin{table}[htbp]
  \centering
  \footnotesize
  \setlength{\tabcolsep}{4pt}
  \caption{Paired differences in search cost after spatial smoothing (ScanNet), with 95\% intervals that resample rooms. ``Smoothed $-$ unsmoothed'' compares a prior with itself at $\sigma = 0$; ``M-A $-$ R'' and ``M-A/room $-$ M-A'' compare two priors at the same $\sigma$; ``Gemma's change $-$ Qwen's change'' is the difference between the two models in the change of M-A from $\sigma = 0$. Negative values of ``smoothed $-$ unsmoothed'' are reductions. The $\sigma = 0$ values of ``M-A/room $-$ M-A'' are those of Table~\ref{tab:app-search-diff}.}
  \label{tab:app-smoothing}
  \begin{tabular}{llccc}
    \toprule
    Room type & Comparison & $\sigma = 0.5$ & $\sigma = 1$ & $\sigma = 2$ \\
    \midrule
    Bathrooms & smoothed $-$ unsmoothed, R & \SmGainRBathHalf & \SmGainRBathOne & \SmGainRBathTwo \\
     &  & \SmGainRBathHalfCI & \SmGainRBathOneCI & \SmGainRBathTwoCI \\
     & smoothed $-$ unsmoothed, Qwen & \SmGainQwenBathHalf & \SmGainQwenBathOne & \SmGainQwenBathTwo \\
     &  & \SmGainQwenBathHalfCI & \SmGainQwenBathOneCI & \SmGainQwenBathTwoCI \\
     & smoothed $-$ unsmoothed, Gemma & \SmGainGemmaBathHalf & \SmGainGemmaBathOne & \SmGainGemmaBathTwo \\
     &  & \SmGainGemmaBathHalfCI & \SmGainGemmaBathOneCI & \SmGainGemmaBathTwoCI \\
     & M-A $-$ R, Qwen & \SmVsRQwenBathHalf & \SmVsRQwenBathOne & \SmVsRQwenBathTwo \\
     &  & \SmVsRQwenBathHalfCI & \SmVsRQwenBathOneCI & \SmVsRQwenBathTwoCI \\
     & M-A $-$ R, Gemma & \SmVsRGemmaBathHalf & \SmVsRGemmaBathOne & \SmVsRGemmaBathTwo \\
     &  & \SmVsRGemmaBathHalfCI & \SmVsRGemmaBathOneCI & \SmVsRGemmaBathTwoCI \\
     & M-A/room $-$ M-A, Qwen & \SmRoomQwenBathHalf & \SmRoomQwenBathOne & \SmRoomQwenBathTwo \\
     &  & \SmRoomQwenBathHalfCI & \SmRoomQwenBathOneCI & \SmRoomQwenBathTwoCI \\
     & M-A/room $-$ M-A, Gemma & \SmRoomGemmaBathHalf & \SmRoomGemmaBathOne & \SmRoomGemmaBathTwo \\
     &  & \SmRoomGemmaBathHalfCI & \SmRoomGemmaBathOneCI & \SmRoomGemmaBathTwoCI \\
     & Gemma's change $-$ Qwen's change & \SmGemQwenBathHalf & \SmGemQwenBathOne & \SmGemQwenBathTwo \\
     &  & \SmGemQwenBathHalfCI & \SmGemQwenBathOneCI & \SmGemQwenBathTwoCI \\
    \addlinespace
    Bedrooms & smoothed $-$ unsmoothed, R & \SmGainRBedHalf & \SmGainRBedOne & \SmGainRBedTwo \\
     &  & \SmGainRBedHalfCI & \SmGainRBedOneCI & \SmGainRBedTwoCI \\
     & smoothed $-$ unsmoothed, Qwen & \SmGainQwenBedHalf & \SmGainQwenBedOne & \SmGainQwenBedTwo \\
     &  & \SmGainQwenBedHalfCI & \SmGainQwenBedOneCI & \SmGainQwenBedTwoCI \\
     & smoothed $-$ unsmoothed, Gemma & \SmGainGemmaBedHalf & \SmGainGemmaBedOne & \SmGainGemmaBedTwo \\
     &  & \SmGainGemmaBedHalfCI & \SmGainGemmaBedOneCI & \SmGainGemmaBedTwoCI \\
     & M-A $-$ R, Qwen & \SmVsRQwenBedHalf & \SmVsRQwenBedOne & \SmVsRQwenBedTwo \\
     &  & \SmVsRQwenBedHalfCI & \SmVsRQwenBedOneCI & \SmVsRQwenBedTwoCI \\
     & M-A $-$ R, Gemma & \SmVsRGemmaBedHalf & \SmVsRGemmaBedOne & \SmVsRGemmaBedTwo \\
     &  & \SmVsRGemmaBedHalfCI & \SmVsRGemmaBedOneCI & \SmVsRGemmaBedTwoCI \\
     & M-A/room $-$ M-A, Qwen & \SmRoomQwenBedHalf & \SmRoomQwenBedOne & \SmRoomQwenBedTwo \\
     &  & \SmRoomQwenBedHalfCI & \SmRoomQwenBedOneCI & \SmRoomQwenBedTwoCI \\
     & M-A/room $-$ M-A, Gemma & \SmRoomGemmaBedHalf & \SmRoomGemmaBedOne & \SmRoomGemmaBedTwo \\
     &  & \SmRoomGemmaBedHalfCI & \SmRoomGemmaBedOneCI & \SmRoomGemmaBedTwoCI \\
     & Gemma's change $-$ Qwen's change & \SmGemQwenBedHalf & \SmGemQwenBedOne & \SmGemQwenBedTwo \\
     &  & \SmGemQwenBedHalfCI & \SmGemQwenBedOneCI & \SmGemQwenBedTwoCI \\
    \addlinespace
    Offices & smoothed $-$ unsmoothed, R & \SmGainROfficeHalf & \SmGainROfficeOne & \SmGainROfficeTwo \\
     &  & \SmGainROfficeHalfCI & \SmGainROfficeOneCI & \SmGainROfficeTwoCI \\
     & smoothed $-$ unsmoothed, Qwen & \SmGainQwenOfficeHalf & \SmGainQwenOfficeOne & \SmGainQwenOfficeTwo \\
     &  & \SmGainQwenOfficeHalfCI & \SmGainQwenOfficeOneCI & \SmGainQwenOfficeTwoCI \\
     & smoothed $-$ unsmoothed, Gemma & \SmGainGemmaOfficeHalf & \SmGainGemmaOfficeOne & \SmGainGemmaOfficeTwo \\
     &  & \SmGainGemmaOfficeHalfCI & \SmGainGemmaOfficeOneCI & \SmGainGemmaOfficeTwoCI \\
     & M-A $-$ R, Qwen & \SmVsRQwenOfficeHalf & \SmVsRQwenOfficeOne & \SmVsRQwenOfficeTwo \\
     &  & \SmVsRQwenOfficeHalfCI & \SmVsRQwenOfficeOneCI & \SmVsRQwenOfficeTwoCI \\
     & M-A $-$ R, Gemma & \SmVsRGemmaOfficeHalf & \SmVsRGemmaOfficeOne & \SmVsRGemmaOfficeTwo \\
     &  & \SmVsRGemmaOfficeHalfCI & \SmVsRGemmaOfficeOneCI & \SmVsRGemmaOfficeTwoCI \\
     & M-A/room $-$ M-A, Qwen & \SmRoomQwenOfficeHalf & \SmRoomQwenOfficeOne & \SmRoomQwenOfficeTwo \\
     &  & \SmRoomQwenOfficeHalfCI & \SmRoomQwenOfficeOneCI & \SmRoomQwenOfficeTwoCI \\
     & M-A/room $-$ M-A, Gemma & \SmRoomGemmaOfficeHalf & \SmRoomGemmaOfficeOne & \SmRoomGemmaOfficeTwo \\
     &  & \SmRoomGemmaOfficeHalfCI & \SmRoomGemmaOfficeOneCI & \SmRoomGemmaOfficeTwoCI \\
     & Gemma's change $-$ Qwen's change & \SmGemQwenOfficeHalf & \SmGemQwenOfficeOne & \SmGemQwenOfficeTwo \\
     &  & \SmGemQwenOfficeHalfCI & \SmGemQwenOfficeOneCI & \SmGemQwenOfficeTwoCI \\
    \addlinespace
    Living rooms & smoothed $-$ unsmoothed, R & \SmGainRLivingHalf & \SmGainRLivingOne & \SmGainRLivingTwo \\
     &  & \SmGainRLivingHalfCI & \SmGainRLivingOneCI & \SmGainRLivingTwoCI \\
     & smoothed $-$ unsmoothed, Qwen & \SmGainQwenLivingHalf & \SmGainQwenLivingOne & \SmGainQwenLivingTwo \\
     &  & \SmGainQwenLivingHalfCI & \SmGainQwenLivingOneCI & \SmGainQwenLivingTwoCI \\
     & smoothed $-$ unsmoothed, Gemma & \SmGainGemmaLivingHalf & \SmGainGemmaLivingOne & \SmGainGemmaLivingTwo \\
     &  & \SmGainGemmaLivingHalfCI & \SmGainGemmaLivingOneCI & \SmGainGemmaLivingTwoCI \\
     & M-A $-$ R, Qwen & \SmVsRQwenLivingHalf & \SmVsRQwenLivingOne & \SmVsRQwenLivingTwo \\
     &  & \SmVsRQwenLivingHalfCI & \SmVsRQwenLivingOneCI & \SmVsRQwenLivingTwoCI \\
     & M-A $-$ R, Gemma & \SmVsRGemmaLivingHalf & \SmVsRGemmaLivingOne & \SmVsRGemmaLivingTwo \\
     &  & \SmVsRGemmaLivingHalfCI & \SmVsRGemmaLivingOneCI & \SmVsRGemmaLivingTwoCI \\
     & M-A/room $-$ M-A, Qwen & \SmRoomQwenLivingHalf & \SmRoomQwenLivingOne & \SmRoomQwenLivingTwo \\
     &  & \SmRoomQwenLivingHalfCI & \SmRoomQwenLivingOneCI & \SmRoomQwenLivingTwoCI \\
     & M-A/room $-$ M-A, Gemma & \SmRoomGemmaLivingHalf & \SmRoomGemmaLivingOne & \SmRoomGemmaLivingTwo \\
     &  & \SmRoomGemmaLivingHalfCI & \SmRoomGemmaLivingOneCI & \SmRoomGemmaLivingTwoCI \\
     & Gemma's change $-$ Qwen's change & \SmGemQwenLivingHalf & \SmGemQwenLivingOne & \SmGemQwenLivingTwo \\
     &  & \SmGemQwenLivingHalfCI & \SmGemQwenLivingOneCI & \SmGemQwenLivingTwoCI \\
    \bottomrule
  \end{tabular}
\end{table}

\paragraph{Targets.}
For each room type, the target categories are those whose occurrence rate
across the real rooms of that type is at least \MinOcc. Every instance of
every target category in every real room is a separate search episode. If a
generated heat map for a (room type, category) pair is empty, that pair is
dropped from the model prior's evaluation and the number of dropped pairs is reported: \SearchSkippedQwen{} (room type, category) pairs for Qwen and \SearchSkippedGemma{} for Gemma, over M-A and M-Bw together.

\paragraph{Cost.}
The policy opens cells in decreasing order of prior probability. The cost of an episode is Equation~\eqref{eq:cost}: the expected fraction of the room opened when cells of equal probability are opened in uniformly random order, computed exactly rather than by sampling orders. Opening tied cells in a fixed order instead would make the uniform prior's cost depend on where the targets lie relative to the first cell. The search curves of Fig.~\ref{fig:search} give, for each number of opened cells, the probability that the target's cell is among them under the same random order. \textbf{The object occupies exactly one cell --- the one
containing its center --- even when its footprint spans several.}

\paragraph{Aggregation.}
Episodes are averaged within a room first, then across rooms, so that a room
with many target objects does not dominate. Confidence intervals resample
rooms with replacement, $2\,000$ draws for the per-condition means and the paired differences and $10\,000$ for the spread ratios. A paired difference between two priors is computed room by room over the episodes for which both are defined, so that a prior with missing heat maps is compared on the same targets.

\paragraph{Variation across room types.}
For each prior, the per-room mean cost is averaged within each room type;
the spread across room types is the sample standard deviation of the four
means, with an interval from resampling rooms within each room type
(10\,000 draws). The test of equality across room types is the permutation test of
Appendix~\ref{app:permutation} with the per-room mean cost as the row and the
variance of the four group means as the statistic. Table~\ref{tab:app-searchsd} gives all values.

\begin{figure}[htbp]
  \centering
  \includegraphics[width=\linewidth]{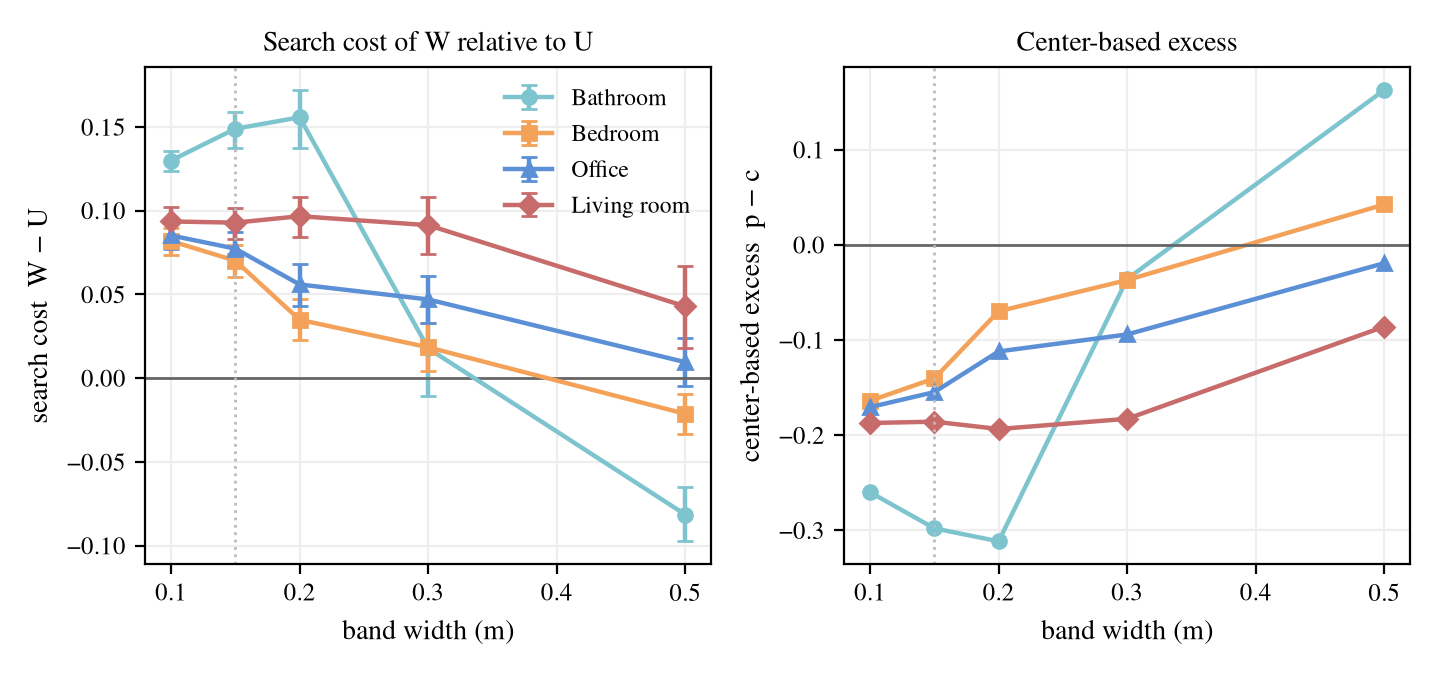}
  \caption{Left: the wall rule's search cost relative to uniform, as the band
    width varies. Right: the center-based excess $p - c$ for the same bands.
    The two panels are mirror images, as Equation~\eqref{eq:region} requires.}
  \label{fig:band}
\end{figure}

\paragraph{Band sweep and the region identity.}
Equation~\eqref{eq:region} is exact per room when tied cells are opened in random order: the expected rank of the target's cell within $S$ is $(|S|+1)/2$ wherever the target lies, and the half-cell terms cancel against U. Sweeping the band width of W from $0.10$ to $0.50$~m therefore reproduces it to within \BandMaxErr{} in all twenty (room type, band width) cells (Table~\ref{tab:app-band}). Fig.~\ref{fig:band} shows the sweep. W beats uniform only where the center-based excess $p - c$ turns positive, which happens at $0.50$~m in bathrooms (\BandExcessBathFiveZero) and bedrooms (\BandExcessBedFiveZero) and never in living rooms (\BandExcessLivingFiveZero{} at $0.50$~m); at $0.50$~m the band already covers \BandCovBathFiveZero{} of the median bathroom floor.

\begin{figure}[htbp]
  \centering
  \includegraphics[width=\linewidth]{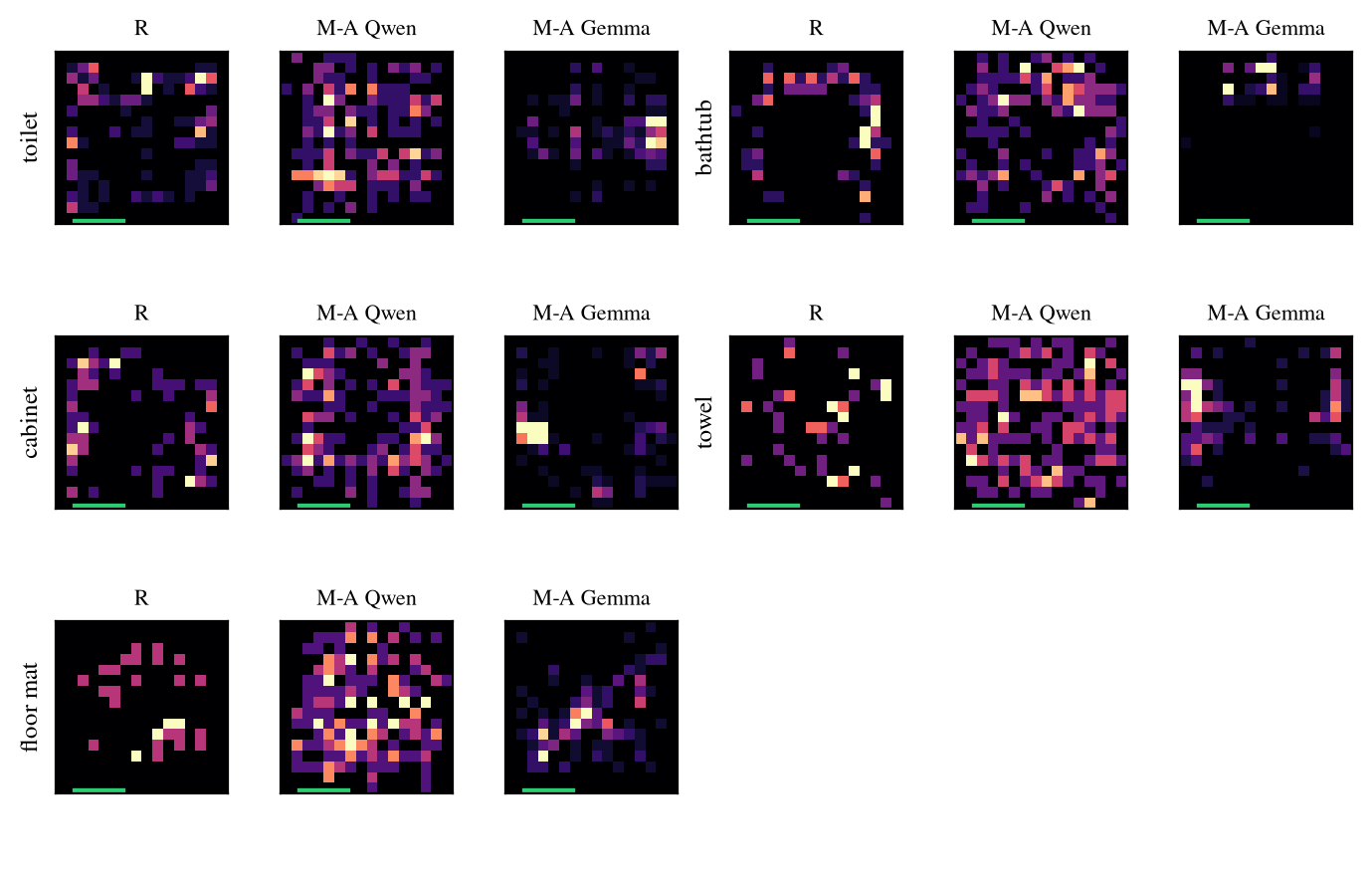}
  \caption{Priors of the most frequent target categories in bathrooms: real rooms (R) and condition~\condA{} generations (M-A), in the oriented and normalized frame of Fig.~\ref{fig:priors}. Each row shows two categories, each with R and with M-A for Qwen and Gemma. Empty panels are categories the model did not generate.}
  \label{fig:priors-bathroom-scannet}
\end{figure}

\begin{figure}[htbp]
  \centering
  \includegraphics[width=\linewidth]{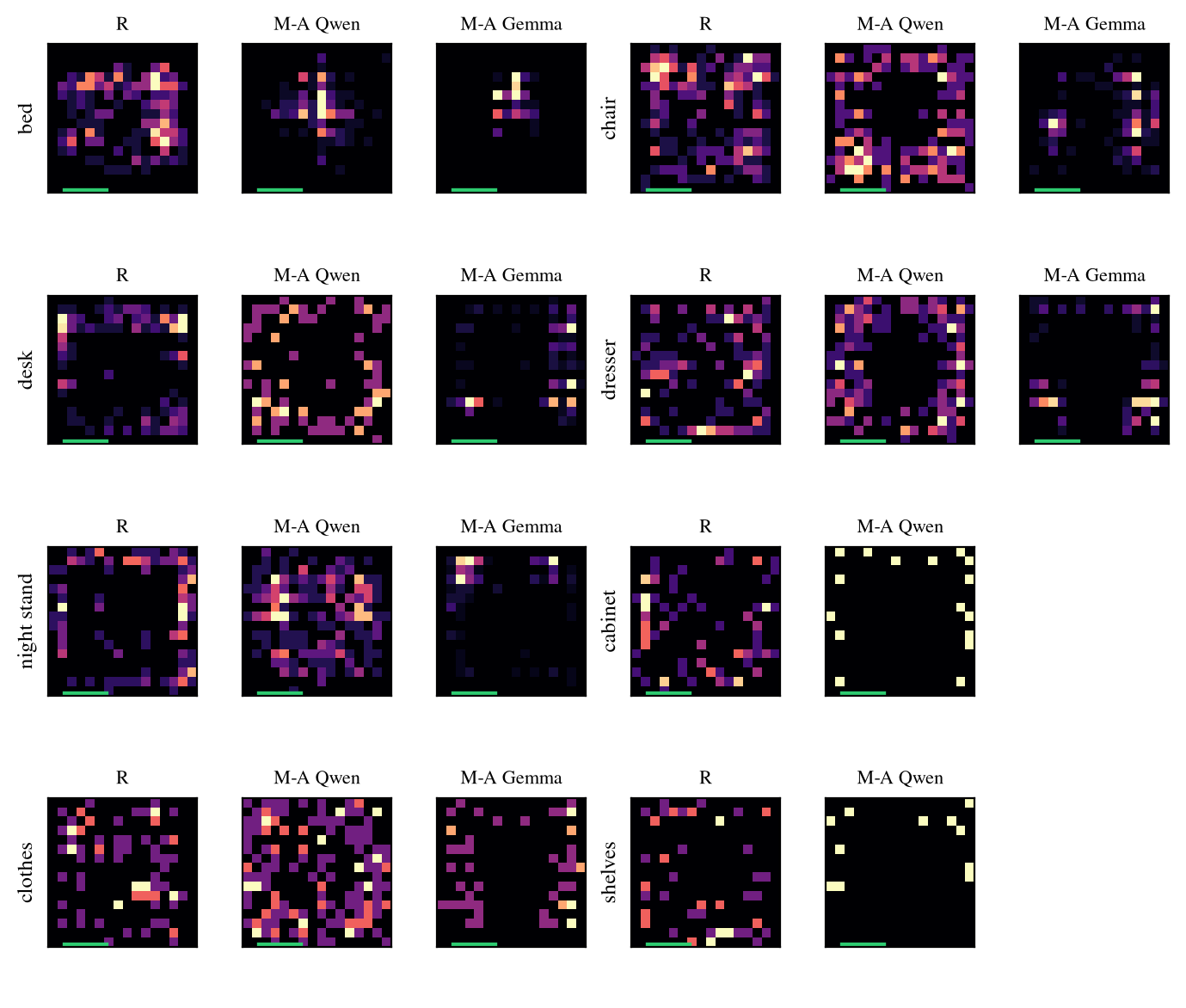}
  \caption{Priors of the most frequent target categories in bedrooms: real rooms (R) and condition~\condA{} generations (M-A), in the oriented and normalized frame of Fig.~\ref{fig:priors}. Each row shows two categories, each with R and with M-A for Qwen and Gemma. Empty panels are categories the model did not generate.}
  \label{fig:priors-bedroom}
\end{figure}

\begin{figure}[htbp]
  \centering
  \includegraphics[width=\linewidth]{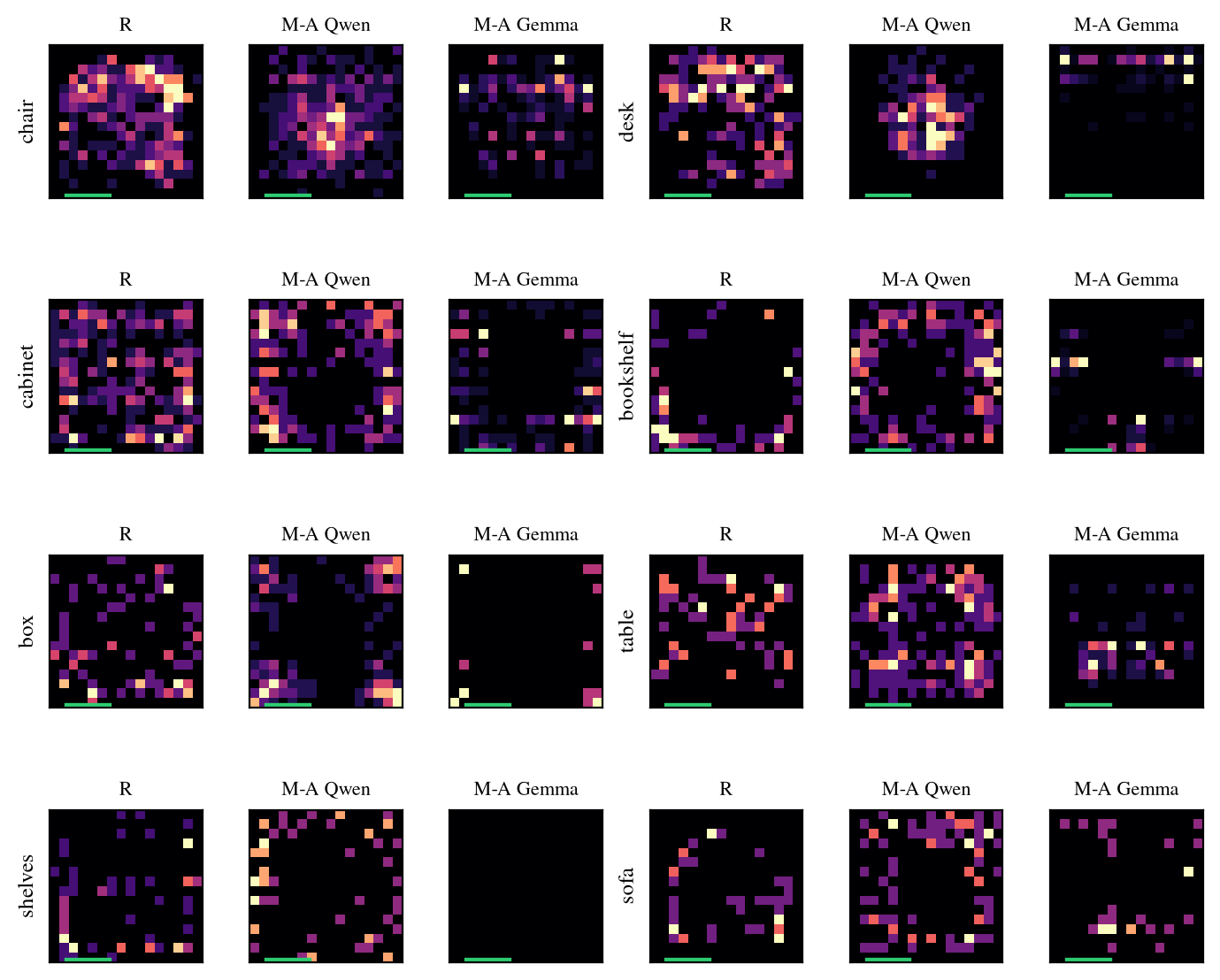}
  \caption{Priors of the most frequent target categories in offices: real rooms (R) and condition~\condA{} generations (M-A), in the oriented and normalized frame of Fig.~\ref{fig:priors}. Each row shows two categories, each with R and with M-A for Qwen and Gemma. Empty panels are categories the model did not generate.}
  \label{fig:priors-office}
\end{figure}

\begin{figure}[htbp]
  \centering
  \includegraphics[width=\linewidth]{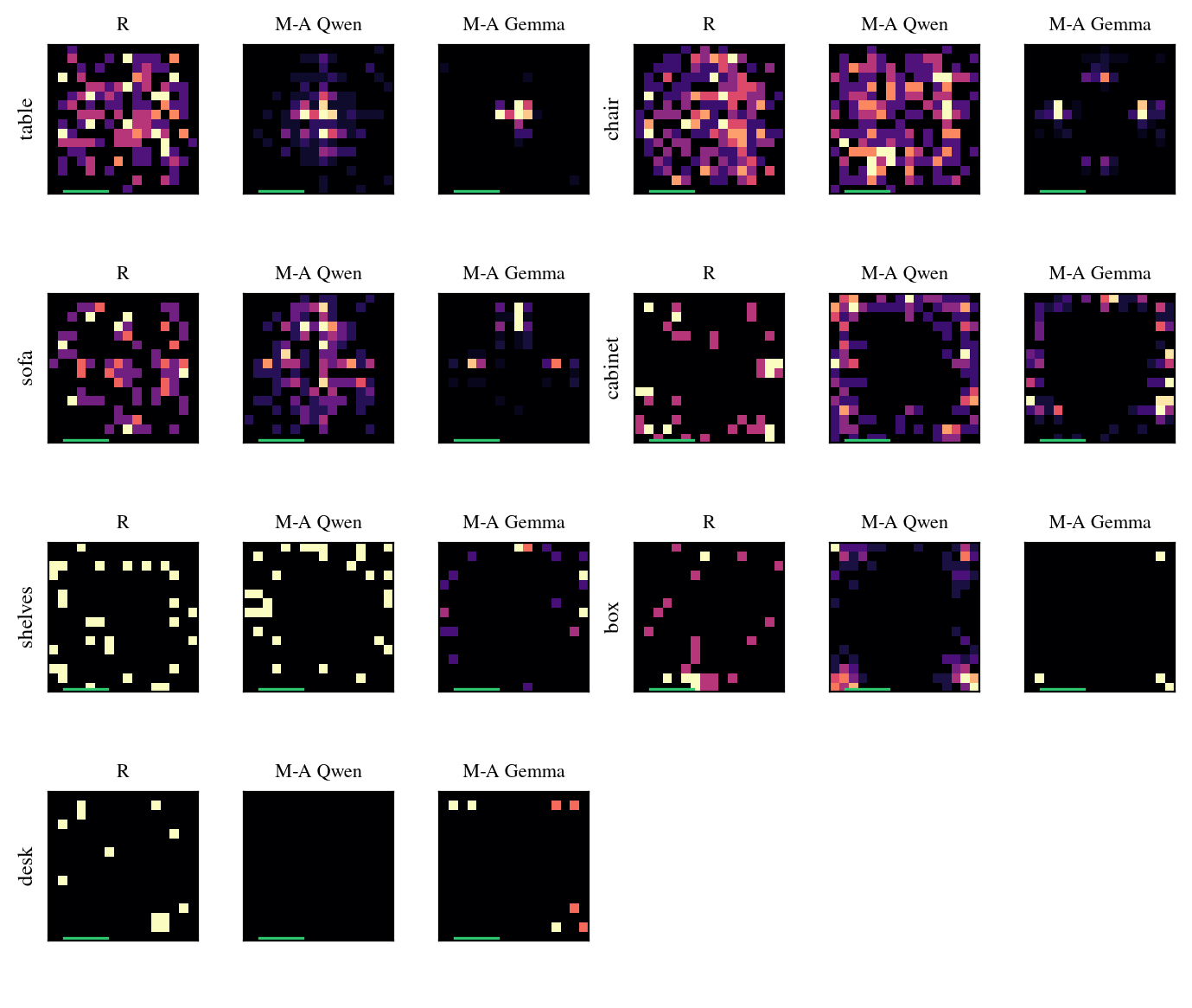}
  \caption{Priors of the most frequent target categories in living rooms: real rooms (R) and condition~\condA{} generations (M-A), in the oriented and normalized frame of Fig.~\ref{fig:priors}. Each row shows two categories, each with R and with M-A for Qwen and Gemma. Empty panels are categories the model did not generate.}
  \label{fig:priors-living-room}
\end{figure}

Figs.~\ref{fig:priors-bathroom-scannet}, \ref{fig:priors-bedroom}, \ref{fig:priors-office}, and~\ref{fig:priors-living-room} show the priors R and M-A of the most frequent target categories of each room type.

\paragraph{Two definitions of excess.}
The $p$ in Equation~\eqref{eq:region} is the fraction of object \emph{centers} in the band. The wall-proximity excess of Sec.~\ref{sec:excess} counts an object as near a wall when its \emph{nearest face} is within the band, which is the right definition for describing how furniture is arranged but not the one that governs search cost. The two differ substantially: in bathrooms at $0.15$~m the face-based excess is \ExcessRealBath{} while the center-based excess is \BandExcessBathOneFive{}, because a cabinet flush against a wall has its center half its depth away from it.

\input{tables/app-search-qwen}

\input{tables/app-search-gemma}

\input{tables/app-search-diff}

\input{tables/app-searchsd}

\input{tables/app-band}

\FloatBarrier

\subsection{Orientation by category}
\label{app:catori}

The orientation profile $q(\mathrm{wall} \mid o)$ of a category $o$ is the
distribution of nearest walls over its objects in canonicalized rooms, pooled
over the four room types; categories with fewer than
\CatOriMinObjects{} real objects are excluded. For each remaining category and
each model, we compute the Jensen--Shannon divergence between the model's
profile for the category (condition~\condA) and (i) the real profile for the
same category, (ii) the real profile pooled over all categories, and record which is smaller. We count the categories for which (i) is smaller. Table~\ref{tab:app-catori} lists every
category.

\input{tables/app-catori}

\FloatBarrier

\subsection{Living rooms by subset}
\label{app:mixture}

The ScanNet label \emph{Living room / Lounge} covers domestic living rooms
and shared lounges. To test whether the mixture carries the living-room
result, the living rooms are split by sofa count (at most one; two or more),
by chair count (at most three; four or more) and by floor area (below the
median; at or above it), and the intersection ``at most one sofa, at most
three chairs, below the median area'' is taken as the domestic subset. The
wall-proximity excess of Equation~\eqref{eq:excess} is recomputed on each
subset with its room-resampled interval, and the regression slope of
Appendix~\ref{app:slope} is refitted with the living-room point taken from
the domestic subset. Table~\ref{tab:app-mixture} gives the values.

\input{tables/app-mixture}

\FloatBarrier\subsection{Coverage of the walls by the scans}
\label{app:scancoverage}
The room rectangle of a ScanNet room is the extent of the scanned floor and wall geometry (Sec.~\ref{sec:data}). Where a wall was not scanned, the side of the rectangle is the edge of the scan and not a wall. For each scan and each of the four sides of the rectangle, we measured the fraction of the side's length, in bins of $0.10$~m, that has a vertex labeled as wall within \ScanCovTol~m of the side, and took the minimum over the four sides as the wall coverage of the scan. The median coverage is \ScanCovMedianBath{} in bathrooms, \ScanCovMedianBed{} in bedrooms, \ScanCovMedianOffice{} in offices and \ScanCovMedianLiving{} in living rooms. Table~\ref{tab:app-scancoverage} gives the wall-proximity excess of the real rooms and the regression slopes of Sec.~\ref{sec:hierarchy} when the real rooms are restricted to those with a coverage of at least $0.25$, $0.50$ and $0.75$; the generated layouts are unchanged.

\begin{table}[htbp]
  \centering
  \footnotesize
  \caption{Wall-proximity excess of the real ScanNet rooms restricted by the wall coverage of the scan (minimum over the four sides of the rectangle), with the number of rooms in parentheses, and the regression slope of each model's excess (condition~\condA) on the restricted real excess. The first row is the set of real rooms used in the main text.}
  \label{tab:app-scancoverage}
  \begin{tabular}{lrrrrrrr}
    \toprule
    Coverage & Rooms & Bathroom & Bedroom & Office & Living room & Slope, Qwen & Slope, Gemma \\
    \midrule
    all rooms & \ScanCovNAll & \ScanCovExcessBathAll & \ScanCovExcessBedAll & \ScanCovExcessOfficeAll & \ScanCovExcessLivingAll & \ScanCovSlopeQwenAll & \ScanCovSlopeGemmaAll \\
     & & (\ScanCovRoomsBathAll) & (\ScanCovRoomsBedAll) & (\ScanCovRoomsOfficeAll) & (\ScanCovRoomsLivingAll) & & \\
    $\ge 0.25$ & \ScanCovNQuarter & \ScanCovExcessBathQuarter & \ScanCovExcessBedQuarter & \ScanCovExcessOfficeQuarter & \ScanCovExcessLivingQuarter & \ScanCovSlopeQwenQuarter & \ScanCovSlopeGemmaQuarter \\
     & & (\ScanCovRoomsBathQuarter) & (\ScanCovRoomsBedQuarter) & (\ScanCovRoomsOfficeQuarter) & (\ScanCovRoomsLivingQuarter) & & \\
    $\ge 0.50$ & \ScanCovNHalf & \ScanCovExcessBathHalf & \ScanCovExcessBedHalf & \ScanCovExcessOfficeHalf & \ScanCovExcessLivingHalf & \ScanCovSlopeQwenHalf & \ScanCovSlopeGemmaHalf \\
     & & (\ScanCovRoomsBathHalf) & (\ScanCovRoomsBedHalf) & (\ScanCovRoomsOfficeHalf) & (\ScanCovRoomsLivingHalf) & & \\
    $\ge 0.75$ & \ScanCovNThreeQ & \ScanCovExcessBathThreeQ & \ScanCovExcessBedThreeQ & \ScanCovExcessOfficeThreeQ & \ScanCovExcessLivingThreeQ & \ScanCovSlopeQwenThreeQ & \ScanCovSlopeGemmaThreeQ \\
     & & (\ScanCovRoomsBathThreeQ) & (\ScanCovRoomsBedThreeQ) & (\ScanCovRoomsOfficeThreeQ) & (\ScanCovRoomsLivingThreeQ) & & \\
    \bottomrule
  \end{tabular}
\end{table}

%% file: tables/app-ladder-cubi.tex
\begin{table}[htbp]
  \centering
  \footnotesize
  \setlength{\tabcolsep}{4pt}
  \caption{Perturbation tests on CubiCasa, Qwen. Pairwise accuracy is the fraction of rooms in which the reference outscores the perturbed copy; $\Delta$ is the mean log-likelihood difference in nats. Symmetrized, ties excluded. $^\dagger$Detach leaves the \texttt{wall\_contact} field unchanged, so the field contradicts the moved coordinates; these rows combine the contradiction with the geometric change, and Sec.~\ref{sec:selfconsistency} separates the two. Piling and the two swaps leave the same contradiction.}
  \label{tab:app-ladder-cubi}
  \begin{tabular}{llrrrr}
    \toprule
    Perturbation & Room & $n$ & Pairwise acc. & 95\% CI & $\Delta$ \\
    \midrule
    transplant & Kitchen & 2,848 & 0.141 & [0.128, 0.153] & $-$52.30 \\
    transplant & Bathroom & 4,258 & 0.954 & [0.948, 0.960] & 82.24 \\
    pile & Kitchen & 2,848 & 0.955 & [0.948, 0.962] & 17.36 \\
    pile & Bathroom & 4,258 & 0.955 & [0.948, 0.961] & 10.61 \\
    detach$^\dagger$ & Kitchen & 2,830 & 0.945 & [0.936, 0.953] & 6.47 \\
    detach$^\dagger$ & Bathroom & 4,258 & 0.813 & [0.802, 0.825] & 2.35 \\
    swap\_pos & Kitchen & 2,848 & 0.994 & [0.991, 0.996] & 18.85 \\
    swap\_pos & Bathroom & 4,258 & 0.977 & [0.973, 0.982] & 9.99 \\
    swap\_size & Kitchen & 2,848 & 0.996 & [0.994, 0.999] & 17.54 \\
    swap\_size & Bathroom & 4,258 & 0.945 & [0.938, 0.951] & 7.23 \\
    mirror & Kitchen & 2,848 & 0.000 & [0.000, 0.000] & 0.00 \\
    mirror & Bathroom & 4,258 & 0.000 & [0.000, 0.000] & 0.00 \\
    \bottomrule
  \end{tabular}
\end{table}

%% file: tables/app-ladder-cubi-gemma.tex
\begin{table}[htbp]
  \centering
  \footnotesize
  \setlength{\tabcolsep}{4pt}
  \caption{Perturbation tests on CubiCasa, Gemma, with the same rooms and perturbations as Table~\ref{tab:app-ladder-cubi}. $^\dagger$As in Table~\ref{tab:app-ladder-cubi}, the detach rows combine the geometric change with the contradiction between the \texttt{wall\_contact} field and the coordinates.}
  \label{tab:app-ladder-cubi-gemma}
  \begin{tabular}{llrrrr}
    \toprule
    Perturbation & Room & $n$ & Pairwise acc. & 95\% CI & $\Delta$ \\
    \midrule
    pile & Kitchen & 2,848 & 0.928 & [0.919, 0.938] & 36.58 \\
    pile & Bathroom & 4,258 & 0.862 & [0.852, 0.872] & 18.82 \\
    detach$^\dagger$ & Kitchen & 2,830 & 0.892 & [0.880, 0.903] & 15.54 \\
    detach$^\dagger$ & Bathroom & 4,258 & 0.671 & [0.657, 0.685] & 4.26 \\
    swap\_pos & Kitchen & 2,848 & 0.982 & [0.978, 0.987] & 40.74 \\
    swap\_pos & Bathroom & 4,258 & 0.950 & [0.943, 0.956] & 21.33 \\
    swap\_size & Kitchen & 2,848 & 0.976 & [0.970, 0.981] & 32.18 \\
    swap\_size & Bathroom & 4,258 & 0.879 & [0.869, 0.889] & 14.53 \\
    \bottomrule
  \end{tabular}
\end{table}

%% file: tables/app-ladder-qwen.tex
\begin{table}[htbp]
  \centering
  \footnotesize
  \setlength{\tabcolsep}{4pt}
  \caption{Perturbation tests on ScanNet, Qwen. $^\dagger$As in Table~\ref{tab:app-ladder-cubi}, the detach rows combine the geometric change with the contradiction between the \texttt{wall\_contact} field and the coordinates.}
  \label{tab:app-ladder-qwen}
  \begin{tabular}{llrrrr}
    \toprule
    Perturbation & Room & $n$ & Pairwise acc. & 95\% CI & $\Delta$ \\
    \midrule
    pile & Bathroom & 163 & 0.589 & [0.509, 0.663] & 1.00 \\
    pile & Bedroom & 272 & 0.320 & [0.265, 0.375] & $-$4.80 \\
    pile & Office & 172 & 0.029 & [0.006, 0.058] & $-$13.60 \\
    pile & Living room & 215 & 0.088 & [0.051, 0.130] & $-$14.19 \\
    detach$^\dagger$ & Bathroom & 163 & 0.785 & [0.724, 0.847] & 2.08 \\
    detach$^\dagger$ & Bedroom & 272 & 0.853 & [0.809, 0.893] & 2.12 \\
    detach$^\dagger$ & Office & 172 & 0.756 & [0.692, 0.820] & 1.85 \\
    detach$^\dagger$ & Living room & 215 & 0.442 & [0.377, 0.507] & 0.61 \\
    swap\_pos & Bathroom & 163 & 0.785 & [0.724, 0.847] & 2.75 \\
    swap\_pos & Bedroom & 272 & 0.963 & [0.938, 0.985] & 10.61 \\
    swap\_pos & Office & 172 & 0.948 & [0.913, 0.977] & 9.77 \\
    swap\_pos & Living room & 215 & 0.870 & [0.823, 0.912] & 5.65 \\
    swap\_size & Bathroom & 163 & 0.926 & [0.883, 0.963] & 8.32 \\
    swap\_size & Bedroom & 272 & 0.993 & [0.982, 1.000] & 15.18 \\
    swap\_size & Office & 172 & 0.977 & [0.953, 0.994] & 11.72 \\
    swap\_size & Living room & 215 & 0.935 & [0.902, 0.967] & 7.60 \\
    \bottomrule
  \end{tabular}
\end{table}

%% file: tables/app-ladder-gemma.tex
\begin{table}[htbp]
  \centering
  \footnotesize
  \setlength{\tabcolsep}{4pt}
  \caption{Perturbation tests on ScanNet, Gemma. $^\dagger$As in Table~\ref{tab:app-ladder-cubi}, the detach rows combine the geometric change with the contradiction between the \texttt{wall\_contact} field and the coordinates.}
  \label{tab:app-ladder-gemma}
  \begin{tabular}{llrrrr}
    \toprule
    Perturbation & Room & $n$ & Pairwise acc. & 95\% CI & $\Delta$ \\
    \midrule
    pile & Bathroom & 163 & 0.503 & [0.429, 0.583] & $-$0.43 \\
    pile & Bedroom & 272 & 0.239 & [0.188, 0.290] & $-$16.68 \\
    pile & Office & 172 & 0.256 & [0.192, 0.320] & $-$19.12 \\
    pile & Living room & 215 & 0.288 & [0.228, 0.353] & $-$19.89 \\
    detach$^\dagger$ & Bathroom & 163 & 0.742 & [0.675, 0.810] & 4.70 \\
    detach$^\dagger$ & Bedroom & 272 & 0.702 & [0.647, 0.757] & 4.43 \\
    detach$^\dagger$ & Office & 172 & 0.605 & [0.529, 0.674] & 4.57 \\
    detach$^\dagger$ & Living room & 215 & 0.428 & [0.363, 0.498] & 2.33 \\
    swap\_pos & Bathroom & 163 & 0.773 & [0.706, 0.834] & 8.99 \\
    swap\_pos & Bedroom & 272 & 0.831 & [0.787, 0.875] & 16.83 \\
    swap\_pos & Office & 172 & 0.785 & [0.721, 0.843] & 16.17 \\
    swap\_pos & Living room & 215 & 0.707 & [0.647, 0.767] & 10.11 \\
    swap\_size & Bathroom & 163 & 0.742 & [0.675, 0.810] & 8.80 \\
    swap\_size & Bedroom & 272 & 0.676 & [0.621, 0.732] & 8.39 \\
    swap\_size & Office & 172 & 0.791 & [0.727, 0.849] & 13.75 \\
    swap\_size & Living room & 215 & 0.749 & [0.688, 0.805] & 7.79 \\
    \bottomrule
  \end{tabular}
\end{table}

%% file: tables/app-direction.tex
\begin{table}[htbp]
  \centering
  \footnotesize
  \setlength{\tabcolsep}{4pt}
  \caption{Single-object displacement controls and the reordering control on CubiCasa (Qwen, symmetrized, ties excluded). $n$ is the number of non-tied rooms; Ties is the tie rate over all rooms; $\Delta$ is the mean log-likelihood difference in nats (reference minus perturbed). ``Isolated objects'' restricts to rooms whose moved object has no neighbor within $0.30$~m, so that a tangential move cannot create an overlap. The reordering rows move no object.}
  \label{tab:app-direction}
  \begin{tabular}{llrrrrr}
    \toprule
    Room & Edit & $n$ & Pairwise acc. & 95\% CI & Ties & $\Delta$ \\
    \midrule
    Kitchen & tangential, $0.10$~m & 2,264 & 0.712 & [0.694, 0.731] & 0.20 & 1.73 \\
    Kitchen & normal outward, $0.10$~m & 2,390 & 0.893 & [0.881, 0.905] & 0.16 & 3.46 \\
    Kitchen & normal inward, $0.10$~m & 2,843 & 0.910 & [0.899, 0.920] & 0.00 & 3.56 \\
    Kitchen & tangential, $0.20$~m & 2,239 & 0.724 & [0.706, 0.743] & 0.21 & 1.96 \\
    Kitchen & normal outward, $0.20$~m & 2,374 & 0.903 & [0.890, 0.914] & 0.17 & 3.91 \\
    Kitchen & tangential, $0.40$~m & 2,215 & 0.749 & [0.731, 0.767] & 0.22 & 2.37 \\
    Kitchen & normal outward, $0.40$~m & 2,372 & 0.945 & [0.935, 0.954] & 0.17 & 4.79 \\
    Kitchen & tangential, $0.80$~m & 2,006 & 0.793 & [0.775, 0.811] & 0.29 & 2.71 \\
    Kitchen & normal outward, $0.80$~m & 2,316 & 0.953 & [0.944, 0.961] & 0.19 & 6.15 \\
    Kitchen & tangential, isolated objects & 1,268 & 0.656 & [0.630, 0.682] & --- & --- \\
    Kitchen & normal outward, isolated objects & 1,268 & 0.907 & [0.891, 0.923] & --- & --- \\
    Kitchen & one object, $0.01$~m & 2,790 & 0.729 & [0.712, 0.745] & 0.02 & 1.72 \\
    Kitchen & reorder, lexicographic & 2,826 & 0.191 & [0.177, 0.206] & 0.01 & $-$9.09 \\
    Kitchen & reorder, random & 2,839 & 0.574 & [0.557, 0.592] & 0.00 & 1.68 \\
    Bathroom & tangential, $0.10$~m & 3,057 & 0.742 & [0.726, 0.757] & 0.28 & 2.31 \\
    Bathroom & normal outward, $0.10$~m & 3,837 & 0.729 & [0.714, 0.743] & 0.10 & 1.90 \\
    Bathroom & normal inward, $0.10$~m & 4,152 & 0.738 & [0.725, 0.752] & 0.02 & 2.16 \\
    Bathroom & tangential, $0.20$~m & 3,037 & 0.752 & [0.736, 0.768] & 0.29 & 2.33 \\
    Bathroom & normal outward, $0.20$~m & 3,845 & 0.792 & [0.779, 0.804] & 0.10 & 2.07 \\
    Bathroom & tangential, $0.40$~m & 2,927 & 0.782 & [0.767, 0.797] & 0.31 & 2.50 \\
    Bathroom & normal outward, $0.40$~m & 3,821 & 0.854 & [0.843, 0.865] & 0.10 & 3.13 \\
    Bathroom & tangential, $0.80$~m & 2,204 & 0.797 & [0.780, 0.814] & 0.48 & 3.24 \\
    Bathroom & normal outward, $0.80$~m & 3,185 & 0.907 & [0.897, 0.917] & 0.25 & 4.64 \\
    Bathroom & tangential, isolated objects & 1,069 & 0.812 & [0.788, 0.835] & --- & --- \\
    Bathroom & normal outward, isolated objects & 1,069 & 0.832 & [0.809, 0.854] & --- & --- \\
    Bathroom & one object, $0.01$~m & 4,176 & 0.668 & [0.653, 0.682] & 0.02 & 1.77 \\
    Bathroom & reorder, lexicographic & 4,005 & 0.516 & [0.501, 0.531] & 0.06 & 0.52 \\
    Bathroom & reorder, random & 3,927 & 0.610 & [0.594, 0.625] & 0.08 & 2.18 \\
    \bottomrule
  \end{tabular}
\end{table}

%% file: tables/app-wallclaim.tex
\begin{table}[htbp]
  \centering
  \footnotesize
  \setlength{\tabcolsep}{4pt}
  \caption{The three versions of the detach comparison of Sec.~\ref{sec:selfconsistency}, with intervals (CubiCasa, symmetrized, no ties). Keep is the detach perturbation as run; Claim only changes one character of \texttt{wall\_contact} and no coordinate; Drop field deletes \texttt{wall\_contact} from both the reference and the detached copy. All three are scored over the same rooms, which are the rooms of the detach row of Table~\ref{tab:app-ladder-cubi} minus the rooms in which no declaration could be falsified (1 kitchen, 3 bathrooms: every declaring object there lies within $0.15$~m of the opposite wall and of the farthest wall, so the claim-only edit would leave the string unchanged).}
  \label{tab:app-wallclaim}
  \begin{tabular}{lllrrr}
    \toprule
    Model & Room & Version & $n$ & Pairwise acc. & 95\% CI \\
    \midrule
    Qwen & Kitchen & Keep & 2,829 & 0.945 & [0.936, 0.953] \\
    Qwen & Kitchen & Claim only & 2,829 & 0.866 & [0.854, 0.879] \\
    Qwen & Kitchen & Drop field & 2,829 & 0.875 & [0.863, 0.887] \\
    Qwen & Bathroom & Keep & 4,255 & 0.814 & [0.802, 0.825] \\
    Qwen & Bathroom & Claim only & 4,255 & 0.857 & [0.846, 0.867] \\
    Qwen & Bathroom & Drop field & 4,255 & 0.669 & [0.655, 0.682] \\
    Gemma & Kitchen & Keep & 2,829 & 0.892 & [0.881, 0.903] \\
    Gemma & Kitchen & Claim only & 2,829 & 0.828 & [0.814, 0.842] \\
    Gemma & Kitchen & Drop field & 2,829 & 0.849 & [0.836, 0.862] \\
    Gemma & Bathroom & Keep & 4,255 & 0.671 & [0.657, 0.685] \\
    Gemma & Bathroom & Claim only & 4,255 & 0.855 & [0.844, 0.866] \\
    Gemma & Bathroom & Drop field & 4,255 & 0.581 & [0.567, 0.596] \\
    \bottomrule
  \end{tabular}
\end{table}

%% file: tables/app-wallclaim-diff.tex
\begin{table}[htbp]
  \centering
  \footnotesize
  \setlength{\tabcolsep}{4pt}
  \caption{Paired differences between the versions of Table~\ref{tab:app-wallclaim}: Keep minus Drop field is the part of the pairwise accuracy of detach carried by the contradiction between the field and the coordinates; Keep minus Claim only compares the full perturbation with the one-character edit. Both resample rooms, 10\,000 draws.}
  \label{tab:app-wallclaim-diff}
  \begin{tabular}{llrrrrr}
    \toprule
    Model & Room & $n$ & Keep $-$ Drop & 95\% CI & Keep $-$ Claim & 95\% CI \\
    \midrule
    Qwen & Kitchen & 2,829 & 0.070 & [0.059, 0.080] & 0.078 & [0.063, 0.093] \\
    Qwen & Bathroom & 4,255 & 0.145 & [0.133, 0.157] & $-$0.043 & [$-$0.058, $-$0.027] \\
    Gemma & Kitchen & 2,829 & 0.042 & [0.031, 0.054] & 0.064 & [0.046, 0.081] \\
    Gemma & Bathroom & 4,255 & 0.090 & [0.077, 0.103] & $-$0.183 & [$-$0.200, $-$0.166] \\
    \bottomrule
  \end{tabular}
\end{table}

%% file: tables/app-pile.tex
\begin{table}[htbp]
  \centering
  \scriptsize
  \setlength{\tabcolsep}{4pt}
  \caption{Pile radius sweep on ScanNet. Normalized radius is $\rho$ divided by half the room's shorter side; spread is the standard deviation of the coordinate values after piling.}
  \label{tab:app-pile}
  \begin{tabular}{llrrrrrr}
    \toprule
    Room & Model & $\rho$ (m) & $n$ & Pairwise acc. & 95\% CI & Norm.\ $\rho$ & Spread \\
    \midrule
    Bathroom & Qwen & 0.2 & 163 & 0.540 & [0.466, 0.614] & 0.17 & 0.23 \\
    Bathroom & Qwen & 0.35 & 163 & 0.589 & [0.509, 0.663] & 0.30 & 0.28 \\
    Bathroom & Qwen & 0.6 & 163 & 0.583 & [0.509, 0.656] & 0.52 & 0.38 \\
    Bathroom & Qwen & 1.0 & 163 & 0.595 & [0.515, 0.669] & 0.86 & 0.55 \\
    Bathroom & Gemma & 0.2 & 163 & 0.509 & [0.436, 0.589] & 0.17 & 0.23 \\
    Bathroom & Gemma & 0.35 & 163 & 0.503 & [0.429, 0.583] & 0.30 & 0.28 \\
    Bathroom & Gemma & 0.6 & 163 & 0.454 & [0.380, 0.528] & 0.52 & 0.38 \\
    Bathroom & Gemma & 1.0 & 163 & 0.558 & [0.485, 0.632] & 0.86 & 0.55 \\
    Bedroom & Qwen & 0.2 & 272 & 0.173 & [0.129, 0.217] & 0.11 & 0.43 \\
    Bedroom & Qwen & 0.35 & 272 & 0.320 & [0.265, 0.375] & 0.19 & 0.46 \\
    Bedroom & Qwen & 0.6 & 272 & 0.555 & [0.496, 0.614] & 0.32 & 0.54 \\
    Bedroom & Qwen & 1.0 & 272 & 0.735 & [0.684, 0.787] & 0.53 & 0.68 \\
    Bedroom & Gemma & 0.2 & 272 & 0.154 & [0.110, 0.199] & 0.11 & 0.43 \\
    Bedroom & Gemma & 0.35 & 272 & 0.239 & [0.188, 0.290] & 0.19 & 0.46 \\
    Bedroom & Gemma & 0.6 & 272 & 0.335 & [0.279, 0.390] & 0.32 & 0.54 \\
    Bedroom & Gemma & 1.0 & 272 & 0.445 & [0.386, 0.504] & 0.53 & 0.68 \\
    Office & Qwen & 0.2 & 172 & 0.012 & [0.000, 0.029] & 0.11 & 0.42 \\
    Office & Qwen & 0.35 & 172 & 0.029 & [0.006, 0.058] & 0.20 & 0.45 \\
    Office & Qwen & 0.6 & 172 & 0.233 & [0.169, 0.297] & 0.34 & 0.52 \\
    Office & Qwen & 1.0 & 172 & 0.477 & [0.401, 0.552] & 0.57 & 0.66 \\
    Office & Gemma & 0.2 & 172 & 0.198 & [0.140, 0.256] & 0.11 & 0.42 \\
    Office & Gemma & 0.35 & 172 & 0.256 & [0.192, 0.320] & 0.20 & 0.45 \\
    Office & Gemma & 0.6 & 172 & 0.366 & [0.297, 0.436] & 0.34 & 0.52 \\
    Office & Gemma & 1.0 & 172 & 0.448 & [0.372, 0.523] & 0.57 & 0.66 \\
    Living room & Qwen & 0.2 & 215 & 0.033 & [0.009, 0.056] & 0.08 & 0.40 \\
    Living room & Qwen & 0.35 & 215 & 0.088 & [0.051, 0.130] & 0.15 & 0.43 \\
    Living room & Qwen & 0.6 & 215 & 0.200 & [0.149, 0.256] & 0.25 & 0.51 \\
    Living room & Qwen & 1.0 & 215 & 0.349 & [0.288, 0.414] & 0.42 & 0.66 \\
    Living room & Gemma & 0.2 & 215 & 0.219 & [0.163, 0.279] & 0.08 & 0.40 \\
    Living room & Gemma & 0.35 & 215 & 0.288 & [0.228, 0.353] & 0.15 & 0.43 \\
    Living room & Gemma & 0.6 & 215 & 0.358 & [0.293, 0.423] & 0.25 & 0.51 \\
    Living room & Gemma & 1.0 & 215 & 0.460 & [0.395, 0.530] & 0.42 & 0.66 \\
    \bottomrule
  \end{tabular}
\end{table}

%% file: tables/app-rel-pairs.tex
\begin{table}[htbp]
  \centering
  \footnotesize
  \setlength{\tabcolsep}{4pt}
  \caption{Pairs used in the pair perturbations of Appendix~\ref{app:relations}. Each room contributes one triple, built for the first pair of the candidate list that the room contains at edge-to-edge distance at most \RelNear~m. Pairs with fewer than 20 rooms are pooled. The last column is the median displacement $d$ of the target in that room type, shared by the three versions to within $\pm\RelDTol$\%. The room sets are identical for both models.}
  \label{tab:app-rel-pairs}
  \begin{tabular}{lllrr}
    \toprule
    Data & Room & anchor $\rightarrow$ target & Rooms & median $d$ (m) \\
    \midrule
    CubiCasa & Kitchen & sink $\rightarrow$ dishwasher & 627 & 0.96 \\
     &  & sink $\rightarrow$ wall cabinet & 620 &  \\
     &  & stove $\rightarrow$ wall cabinet & 281 &  \\
     &  & sink $\rightarrow$ stove & 227 &  \\
     &  & dishwasher $\rightarrow$ sink & 165 &  \\
     &  & sink $\rightarrow$ counter & 100 &  \\
     &  & wall cabinet $\rightarrow$ sink & 73 &  \\
     &  & closet $\rightarrow$ refrigerator & 67 &  \\
     &  & dishwasher $\rightarrow$ wall cabinet & 60 &  \\
     &  & wall cabinet $\rightarrow$ stove & 41 &  \\
     &  & refrigerator $\rightarrow$ closet & 40 &  \\
     &  & refrigerator $\rightarrow$ counter & 39 &  \\
     &  & stove $\rightarrow$ counter & 37 &  \\
     &  & refrigerator $\rightarrow$ refrigerator & 36 &  \\
     &  & stove $\rightarrow$ sink & 35 &  \\
     &  & closet $\rightarrow$ closet & 29 &  \\
     &  & wall cabinet $\rightarrow$ refrigerator & 27 &  \\
     &  & stove $\rightarrow$ dishwasher & 22 &  \\
     &  & other (3 pairs) & 6 &  \\
    CubiCasa & Bathroom & counter $\rightarrow$ sink & 1,461 & 0.91 \\
     &  & wall cabinet $\rightarrow$ sink & 309 &  \\
     &  & sink $\rightarrow$ counter & 216 &  \\
     &  & wall cabinet $\rightarrow$ counter & 198 &  \\
     &  & sink $\rightarrow$ wall cabinet & 62 &  \\
     &  & bathtub $\rightarrow$ shower & 52 &  \\
     &  & shower $\rightarrow$ shower & 50 &  \\
     &  & dryer $\rightarrow$ washing machine & 26 &  \\
     &  & closet $\rightarrow$ dryer & 25 &  \\
     &  & washing machine $\rightarrow$ wall cabinet & 20 &  \\
     &  & other (7 pairs) & 29 &  \\
    ScanNet & Bedroom & desk $\rightarrow$ chair & 73 & 1.13 \\
     &  & other (13 pairs) & 83 &  \\
    ScanNet & Living room & table $\rightarrow$ chair & 69 & 1.25 \\
     &  & sofa $\rightarrow$ table & 23 &  \\
     &  & other (1 pairs) & 8 &  \\
    ScanNet & Bathroom & bathtub $\rightarrow$ towel & 22 & 0.57 \\
     &  & other (1 pairs) & 8 &  \\
    ScanNet & Office & table $\rightarrow$ chair & 20 & 1.46 \\
     &  & other (1 pairs) & 6 &  \\
    negative control & Bedroom & bed $\rightarrow$ bed & 3 & 0.45 \\
    negative control & Living room & sofa $\rightarrow$ chair & 17 & 0.48 \\
    \bottomrule
  \end{tabular}
\end{table}

%% file: tables/app-rel-ladder.tex
\begin{table}[htbp]
  \centering
  \scriptsize
  \setlength{\tabcolsep}{4pt}
  \caption{Pair perturbations of Appendix~\ref{app:relations} (None, Wrong, Same: the no-neighbor, wrong-neighbor and same-neighbor versions): pairwise accuracy of the reference against each version of the triple (symmetrized, ties excluded; for no-neighbor, the number of rooms in which that version exists is in parentheses, the other two exist in every room), and the two paired differences with room-resampled 95\% intervals (10\,000 draws). The no-neighbor version moves the target where it is adjacent to nothing, the wrong-neighbor version where it is adjacent to an object of another category, and the same-neighbor version where it stays adjacent to its anchor; all three slide the same target along the same wall by the same distance. The neighbor effect is wrong-neighbor minus same-neighbor (the primary measure); the proximity effect is no-neighbor minus wrong-neighbor, computed over the rooms in which the no-neighbor version exists. The two all-dataset rows are the main results; $^\ddagger$ marks sets below 40 rooms, reported for completeness and not interpreted.}
  \label{tab:app-rel-ladder}
  \begin{tabular}{llrrrrrr}
    \toprule
    Model & Rooms & $n$ & None & Wrong & Same & Neighbor eff. & Proximity eff. \\
    \midrule
    Qwen & \textbf{CubiCasa (all)} & 4,980 & 0.925 (506) & 0.844 & 0.832 & 0.012 [0.003, 0.020] & 0.016 [0.000, 0.034] \\
    Qwen & \textbf{ScanNet (all)} & 312 & 0.712 (66) & 0.638 & 0.625 & 0.013 [$-$0.038, 0.064] & 0.000 [$-$0.076, 0.076] \\
    Qwen & CubiCasa kitchen & 2,532 & 0.920 (312) & 0.795 & 0.768 & 0.027 [0.013, 0.041] & 0.029 [0.006, 0.051] \\
    Qwen & CubiCasa bathroom & 2,448 & 0.933 (194) & 0.895 & 0.898 & $-$0.004 [$-$0.013, 0.006] & $-$0.005 [$-$0.031, 0.015] \\
    Qwen & ScanNet bedroom & 156 & 0.600 (30) & 0.660 & 0.583 & 0.077 [0.006, 0.147] & $-$0.100 [$-$0.233, 0.000] \\
    Qwen & ScanNet living room & 100 & 0.824 (34) & 0.660 & 0.650 & 0.010 [$-$0.080, 0.100] & 0.088 [0.000, 0.206] \\
    Qwen & ScanNet bathroom$^\ddagger$ & 30 & 1.000 (1) & 0.500 & 0.700 & $-$0.200 [$-$0.400, 0.000] & 0.000 [0.000, 0.000] \\
    Qwen & ScanNet office$^\ddagger$ & 26 & 0.000 (1) & 0.577 & 0.692 & $-$0.115 [$-$0.269, 0.038] & 0.000 [0.000, 0.000] \\
    Qwen & negative controls$^\ddagger$ & 20 & 0.750 (4) & 0.500 & 0.550 & $-$0.050 [$-$0.150, 0.000] & $-$0.250 [$-$0.750, 0.000] \\
    Qwen & \quad bed $\rightarrow$ bed$^\ddagger$ & 3 & --- & 0.333 & 0.333 & 0.000 [0.000, 0.000] & --- \\
    Qwen & \quad sofa $\rightarrow$ chair$^\ddagger$ & 17 & 0.750 (4) & 0.529 & 0.588 & $-$0.059 [$-$0.176, 0.000] & $-$0.250 [$-$0.750, 0.000] \\
    Gemma & \textbf{CubiCasa (all)} & 4,980 & 0.866 (506) & 0.811 & 0.798 & 0.013 [0.004, 0.022] & $-$0.012 [$-$0.032, 0.008] \\
    Gemma & \textbf{ScanNet (all)} & 312 & 0.758 (66) & 0.593 & 0.599 & $-$0.006 [$-$0.061, 0.051] & 0.091 [0.000, 0.182] \\
    Gemma & CubiCasa kitchen & 2,532 & 0.843 (312) & 0.755 & 0.738 & 0.017 [0.003, 0.031] & $-$0.003 [$-$0.032, 0.022] \\
    Gemma & CubiCasa bathroom & 2,448 & 0.902 (194) & 0.869 & 0.861 & 0.008 [$-$0.002, 0.019] & $-$0.026 [$-$0.057, 0.005] \\
    Gemma & ScanNet bedroom & 156 & 0.667 (30) & 0.590 & 0.558 & 0.032 [$-$0.045, 0.109] & 0.033 [$-$0.067, 0.133] \\
    Gemma & ScanNet living room & 100 & 0.853 (34) & 0.640 & 0.640 & 0.000 [$-$0.110, 0.110] & 0.147 [0.000, 0.294] \\
    Gemma & ScanNet bathroom$^\ddagger$ & 30 & 1.000 (1) & 0.467 & 0.767 & $-$0.300 [$-$0.467, $-$0.133] & 0.000 [0.000, 0.000] \\
    Gemma & ScanNet office$^\ddagger$ & 26 & 0.000 (1) & 0.577 & 0.500 & 0.077 [0.000, 0.192] & 0.000 [0.000, 0.000] \\
    Gemma & negative controls$^\ddagger$ & 20 & 0.500 (4) & 0.400 & 0.400 & 0.000 [$-$0.150, 0.150] & $-$0.250 [$-$0.750, 0.000] \\
    Gemma & \quad bed $\rightarrow$ bed$^\ddagger$ & 3 & --- & 0.000 & 0.000 & 0.000 [0.000, 0.000] & --- \\
    Gemma & \quad sofa $\rightarrow$ chair$^\ddagger$ & 17 & 0.500 (4) & 0.471 & 0.471 & 0.000 [$-$0.176, 0.176] & $-$0.250 [$-$0.750, 0.000] \\
    \bottomrule
  \end{tabular}
\end{table}

%% file: tables/app-rel-hygiene.tex
\begin{table}[htbp]
  \centering
  \scriptsize
  \setlength{\tabcolsep}{2pt}
  \caption{Quality checks of the pair perturbations (room means; the strings are the same for both models). Distinct integer parts and distinct first decimals are counted over all coordinates of the layout, spread is the standard deviation of the coordinates, overlap is the summed pairwise footprint overlap in m$^2$, outside the summed protrusion beyond the room in m, and declared-wall agreement is the share of \texttt{wall\_contact} declarations consistent with the geometry at $\ClaimTol$~m. The no-neighbor subset is the rooms in which all three versions exist; $d$ is the mean displacement of the target there. Versions: the real layout and the None, Wrong and Same versions of Table~\ref{tab:app-rel-ladder}. Token counts are preserved by construction.}
  \label{tab:app-rel-hygiene}
  \begin{tabular}{lllrrrrrrrr}
    \toprule
    Rooms & Set & Version & $n$ & Int.\ parts & First dec. & Spread & Overlap & Outside & Wall agr. & $d$ (m) \\
    \midrule
    CubiCasa kitchen & all & Real & 2,532 & 3.59 & 8.85 & 0.96 & 3.54 & 0.02 & 1.000 & --- \\
     & all & Wrong & 2,532 & 3.65 & 9.05 & 0.97 & 3.01 & 0.02 & 0.979 & --- \\
     & all & Same & 2,532 & 3.64 & 9.05 & 0.96 & 3.05 & 0.02 & 0.979 & --- \\
     & no-neighbor subset & None & 312 & 3.65 & 8.65 & 0.99 & 1.73 & 0.02 & 0.953 & 1.21 \\
     & no-neighbor subset & Wrong & 312 & 3.59 & 8.64 & 0.97 & 1.76 & 0.02 & 0.953 & 1.13 \\
     & no-neighbor subset & Same & 312 & 3.57 & 8.62 & 0.97 & 1.79 & 0.02 & 0.953 & 1.07 \\
    CubiCasa bathroom & all & Real & 2,448 & 2.48 & 5.98 & 0.53 & 0.32 & 0.01 & 1.000 & --- \\
     & all & Wrong & 2,448 & 2.50 & 6.58 & 0.55 & 0.15 & 0.01 & 0.979 & --- \\
     & all & Same & 2,448 & 2.50 & 6.52 & 0.54 & 0.15 & 0.01 & 0.979 & --- \\
     & no-neighbor subset & None & 194 & 3.05 & 6.92 & 0.73 & 0.07 & 0.01 & 0.992 & 0.98 \\
     & no-neighbor subset & Wrong & 194 & 2.86 & 6.91 & 0.71 & 0.12 & 0.01 & 0.992 & 0.94 \\
     & no-neighbor subset & Same & 194 & 2.93 & 6.93 & 0.71 & 0.10 & 0.01 & 0.992 & 0.86 \\
    ScanNet bedroom & all & Real & 156 & 5.07 & 12.59 & 1.34 & 0.49 & 0.01 & 0.997 & --- \\
     & all & Wrong & 156 & 5.12 & 12.62 & 1.34 & 0.40 & 0.01 & 0.997 & --- \\
     & all & Same & 156 & 5.15 & 12.70 & 1.36 & 0.38 & 0.01 & 0.997 & --- \\
     & no-neighbor subset & None & 30 & 5.67 & 13.37 & 1.53 & 0.27 & 0.02 & 1.000 & 1.75 \\
     & no-neighbor subset & Wrong & 30 & 5.50 & 13.23 & 1.46 & 0.33 & 0.02 & 1.000 & 1.63 \\
     & no-neighbor subset & Same & 30 & 5.67 & 13.33 & 1.51 & 0.29 & 0.02 & 1.000 & 1.49 \\
    ScanNet living room & all & Real & 100 & 5.40 & 13.81 & 1.43 & 0.50 & 0.04 & 0.990 & --- \\
     & all & Wrong & 100 & 5.42 & 13.93 & 1.46 & 0.39 & 0.04 & 0.990 & --- \\
     & all & Same & 100 & 5.48 & 13.96 & 1.46 & 0.40 & 0.04 & 0.990 & --- \\
     & no-neighbor subset & None & 34 & 5.71 & 14.62 & 1.62 & 0.51 & 0.08 & 0.985 & 1.80 \\
     & no-neighbor subset & Wrong & 34 & 5.59 & 14.53 & 1.60 & 0.53 & 0.08 & 0.985 & 1.67 \\
     & no-neighbor subset & Same & 34 & 5.68 & 14.56 & 1.61 & 0.52 & 0.08 & 0.985 & 1.54 \\
    \bottomrule
  \end{tabular}
\end{table}

%% file: tables/app-rel-samecat.tex
\begin{table}[htbp]
  \centering
  \footnotesize
  \setlength{\tabcolsep}{4pt}
  \caption{Same-category neighbors at the wrong-neighbor destination. The layouts carry categories but no instance identity, so a target re-attached next to a second instance of its anchor's category (a chair moved to another desk) has not had its pairing broken at the category level. The construction already rejects destinations whose only neighbor is such an instance; the rows count the triples in which the destination is adjacent to both an object of another category and a second instance of the anchor's category. The last column is the number of rooms with at least two instances of the anchor's category.}
  \label{tab:app-rel-samecat}
  \begin{tabular}{llrrrr}
    \toprule
    Data & Room & Triples & Same-cat.\ neighbor & \% & Rooms with 2+ anchors \\
    \midrule
    CubiCasa & Kitchen & 2,532 & 25 & 1.0 & 163 \\
    CubiCasa & Bathroom & 2,448 & 25 & 1.0 & 145 \\
    ScanNet & Bedroom & 156 & 6 & 3.8 & 27 \\
    ScanNet & Living room & 100 & 12 & 12.0 & 61 \\
    ScanNet & Bathroom & 30 & 0 & 0.0 & 3 \\
    ScanNet & Office & 26 & 6 & 23.1 & 14 \\
    \textbf{Total} &  & 5,292 & 74 & 1.4 &  \\
    \bottomrule
  \end{tabular}
\end{table}

%% file: tables/app-identity.tex
\begin{table}[htbp]
  \centering
  \scriptsize
  \setlength{\tabcolsep}{2pt}
  \caption{Object identity, condition~A: $p(o \mid c)$, the fraction of rooms containing each category, for the real data (R), Qwen (Q) and Gemma (G). \textbf{These are the columns the Spearman $\rho$ of Sec.~\ref{sec:hierarchy} is computed from}, one correlation per room type over the \nVocab{} vocabulary words; the mean of the four is the figure quoted in the main text. Generated categories are mapped to NYU40 before counting. The real $n$ counts every room of that type in the set of real rooms; the placement statistics of Sec.~\ref{sec:hierarchy} use the subset for which an excess can be computed (\nRoomsMin--\nRoomsMax{} per room type), so their $n$ differs. }
  \label{tab:app-identity}
  \begin{tabular}{lrrrrrrrrrrrr}
    \toprule
    Category & Bath.\ R & Bath.\ Q & Bath.\ G & Bedr.\ R & Bedr.\ Q & Bedr.\ G & Offi.\ R & Offi.\ Q & Offi.\ G & Livi.\ R & Livi.\ Q & Livi.\ G \\
    \midrule
    bag & 0.11 & 0.42 & 0.00 & 0.23 & 0.52 & 0.00 & 0.21 & 0.72 & 0.60 & 0.08 & 0.52 & 0.00 \\
    bathtub & 0.55 & 0.95 & 1.00 & 0.04 & 0.01 & 0.00 & 0.00 & 0.01 & 0.00 & 0.00 & 0.00 & 0.00 \\
    bed & 0.00 & 0.01 & 0.00 & 0.99 & 1.00 & 1.00 & 0.00 & 0.02 & 0.00 & 0.01 & 0.00 & 0.00 \\
    bookshelf & 0.00 & 0.00 & 0.00 & 0.19 & 0.72 & 0.86 & 0.43 & 0.92 & 1.00 & 0.08 & 0.97 & 1.00 \\
    box & 0.05 & 0.46 & 0.03 & 0.19 & 0.56 & 0.04 & 0.52 & 0.94 & 0.10 & 0.16 & 0.78 & 0.02 \\
    cabinet & 0.63 & 1.00 & 1.00 & 0.35 & 0.09 & 0.00 & 0.75 & 0.99 & 1.00 & 0.24 & 0.90 & 1.00 \\
    chair & 0.01 & 0.03 & 0.00 & 0.84 & 0.95 & 1.00 & 0.96 & 1.00 & 1.00 & 0.86 & 1.00 & 1.00 \\
    clothes & 0.06 & 0.26 & 0.00 & 0.40 & 0.89 & 0.30 & 0.13 & 0.06 & 0.00 & 0.04 & 0.04 & 0.00 \\
    desk & 0.00 & 0.00 & 0.00 & 0.66 & 0.46 & 1.00 & 0.82 & 1.00 & 1.00 & 0.07 & 0.01 & 0.05 \\
    dresser & 0.00 & 0.00 & 0.00 & 0.54 & 0.99 & 1.00 & 0.01 & 0.01 & 0.00 & 0.02 & 0.00 & 0.00 \\
    floor mat & 0.12 & 1.00 & 1.00 & 0.02 & 0.95 & 1.00 & 0.01 & 0.95 & 1.00 & 0.00 & 0.99 & 1.00 \\
    night stand & 0.00 & 0.01 & 0.00 & 0.56 & 1.00 & 1.00 & 0.00 & 0.01 & 0.00 & 0.03 & 0.00 & 0.00 \\
    refridgerator & 0.01 & 0.01 & 0.00 & 0.21 & 0.01 & 0.00 & 0.03 & 0.17 & 1.00 & 0.04 & 0.00 & 0.00 \\
    shelves & 0.16 & 0.60 & 0.90 & 0.47 & 0.06 & 0.00 & 0.44 & 0.31 & 0.01 & 0.17 & 0.14 & 0.27 \\
    sofa & 0.01 & 0.01 & 0.00 & 0.11 & 0.01 & 0.00 & 0.28 & 0.45 & 0.16 & 0.68 & 1.00 & 1.00 \\
    table & 0.02 & 0.00 & 0.00 & 0.23 & 0.10 & 0.00 & 0.44 & 0.85 & 0.56 & 0.96 & 1.00 & 1.00 \\
    toilet & 0.92 & 1.00 & 1.00 & 0.07 & 0.01 & 0.00 & 0.00 & 0.06 & 0.00 & 0.00 & 0.00 & 0.00 \\
    towel & 0.57 & 0.99 & 1.00 & 0.16 & 0.01 & 0.00 & 0.01 & 0.03 & 0.00 & 0.01 & 0.01 & 0.00 \\
    \midrule Spearman $\rho$ & --- & 0.89 & 0.84 & --- & 0.57 & 0.52 & --- & 0.86 & 0.71 & --- & 0.70 & 0.63 \\
    $n$ rooms & 210 & 200 & 200 & 273 & 200 & 200 & 173 & 200 & 200 & 224 & 200 & 200 \\
    \bottomrule
  \end{tabular}
\end{table}

%% file: tables/app-area.tex
\begin{table}[htbp]
  \centering
  \scriptsize
  \setlength{\tabcolsep}{4pt}
  \caption{Room dimensions, condition~A: floor area in m$^2$ by room type --- sample size, median and interquartile range. \textbf{The SD ratio quoted in Sec.~\ref{sec:hierarchy} is the standard deviation of the four medians in a column, divided by that of the real column} (real 8.23, Qwen 12.37, Gemma 8.95). The quartiles are not used by any statistic in the main text; they are here to show the spread the medians come from. Each model generated 200 rooms per room type. The real $n$ counts every room of that type in the set of real rooms; the placement statistics of Sec.~\ref{sec:hierarchy} use the subset for which an excess can be computed (\nRoomsMin--\nRoomsMax{} per room type), so their $n$ differs. }
  \label{tab:app-area}
  \begin{tabular}{lrrrrrrr}
    \toprule
    Room type & Real $n$ & Real med. & Real IQR & Qwen med. & Qwen IQR & Gemma med. & Gemma IQR \\
    \midrule
    Bathroom & 210 & 7.0 & 5.0--12.1 & 16.0 & 12.0--20.0 & 6.2 & 6.2--7.5 \\
    Bedroom & 273 & 19.5 & 13.1--25.5 & 30.0 & 25.0--36.0 & 16.0 & 16.0--20.0 \\
    Office & 173 & 18.1 & 14.5--21.6 & 42.0 & 30.0--56.0 & 25.0 & 25.0--30.0 \\
    Living room & 224 & 27.0 & 20.7--40.5 & 42.0 & 30.0--54.0 & 25.0 & 25.0--25.0 \\
    \midrule SD of the medians &  & 8.23 &  & 12.37 &  & 8.95 &  \\
    Ratio to real &  & 1.00 &  & 1.50 &  & 1.09 &  \\
    \bottomrule
  \end{tabular}
\end{table}

%% file: tables/app-slope.tex
\begin{table}[htbp]
  \centering
  \footnotesize
  \setlength{\tabcolsep}{4pt}
  \caption{Agreement with the real room-type differences. Slope is the regression of the model's excess on the real excess over the four room types (Appendix~\ref{app:slope}); the SD ratio is real over model.}
  \label{tab:app-slope}
  \begin{tabular}{lrrrrr}
    \toprule
    Layouts & Slope & 95\% CI & Pearson $r$ & SD ratio & 95\% CI \\
    \midrule
    Qwen A & 0.029 & [$-$0.065, 0.120] & 0.31 & 10.58 & [4.88, 23.59] \\
    Qwen B-neutral & 0.677 & [0.524, 0.832] & 0.92 & 1.36 & [1.10, 1.71] \\
    Qwen B-walls & $-$0.188 & [$-$0.358, $-$0.022] & $-$0.42 & 2.24 & [1.56, 3.43] \\
    Gemma A & 0.227 & [0.097, 0.354] & 0.29 & 1.27 & [1.05, 1.53] \\
    Gemma B-neutral & 0.345 & [0.201, 0.489] & 0.41 & 1.19 & [1.01, 1.39] \\
    Gemma B-walls & $-$0.661 & [$-$0.878, $-$0.473] & $-$0.53 & 0.80 & [0.67, 0.95] \\
    \bottomrule
  \end{tabular}
\end{table}

%% file: tables/app-excess.tex
\begin{table}[htbp]
  \centering
  \footnotesize
  \setlength{\tabcolsep}{4pt}
  \caption{Wall-proximity excess by room type (ScanNet), the standard deviation across the four room types, and the permutation test of equality across room types.}
  \label{tab:app-excess}
  \begin{tabular}{lrrrrrr}
    \toprule
    Layouts & Bathroom & Bedroom & Office & Living room & SD & $p$ (equality) \\
    \midrule
    Real data & 0.146 & 0.092 & $-$0.018 & $-$0.163 & 0.1363 & $<$0.001 \\
    Qwen A & $-$0.121 & $-$0.093 & $-$0.113 & $-$0.119 & 0.0129 & 0.274 \\
    Qwen B-neutral & $-$0.017 & $-$0.145 & $-$0.153 & $-$0.261 & 0.0999 & $<$0.001 \\
    Qwen B-walls & 0.182 & 0.059 & 0.148 & 0.193 & 0.0607 & $<$0.001 \\
    Gemma A & $-$0.035 & $-$0.190 & 0.023 & $-$0.185 & 0.1075 & $<$0.001 \\
    Gemma B-neutral & $-$0.053 & $-$0.313 & $-$0.155 & $-$0.257 & 0.1148 & $<$0.001 \\
    Gemma B-walls & 0.365 & 0.051 & 0.326 & 0.443 & 0.1705 & $<$0.001 \\
    \bottomrule
  \end{tabular}
\end{table}

%% file: tables/app-validity.tex
\begin{table}[htbp]
  \centering
  \footnotesize
  \setlength{\tabcolsep}{4pt}
  \caption{Geometric validity of the generated layouts (ScanNet): the fraction of objects extending outside the room rectangle, and the fraction of object pairs that overlap.}
  \label{tab:app-validity}
  \begin{tabular}{lllrrr}
    \toprule
    Model & Condition & Room & Objects & Outside & Overlap \\
    \midrule
    Qwen & A & Bathroom & 1,403 & 0.102 & 0.075 \\
    Qwen & A & Bedroom & 1,736 & 0.124 & 0.051 \\
    Qwen & A & Office & 1,872 & 0.085 & 0.060 \\
    Qwen & A & Living room & 1,554 & 0.104 & 0.074 \\
    Qwen & B-neutral & Bathroom & 680 & 0.269 & 0.147 \\
    Qwen & B-neutral & Bedroom & 924 & 0.177 & 0.105 \\
    Qwen & B-neutral & Office & 857 & 0.083 & 0.046 \\
    Qwen & B-neutral & Living room & 738 & 0.111 & 0.092 \\
    Qwen & B-walls & Bathroom & 647 & 0.431 & 0.073 \\
    Qwen & B-walls & Bedroom & 886 & 0.360 & 0.077 \\
    Qwen & B-walls & Office & 943 & 0.339 & 0.043 \\
    Qwen & B-walls & Living room & 786 & 0.411 & 0.046 \\
    Gemma & A & Bathroom & 1,206 & 0.165 & 0.057 \\
    Gemma & A & Bedroom & 1,455 & 0.091 & 0.082 \\
    Gemma & A & Office & 1,551 & 0.152 & 0.083 \\
    Gemma & A & Living room & 1,294 & 0.021 & 0.135 \\
    Gemma & B-neutral & Bathroom & 769 & 0.138 & 0.021 \\
    Gemma & B-neutral & Bedroom & 770 & 0.012 & 0.112 \\
    Gemma & B-neutral & Office & 711 & 0.056 & 0.065 \\
    Gemma & B-neutral & Living room & 870 & 0.054 & 0.065 \\
    Gemma & B-walls & Bathroom & 741 & 0.816 & 0.004 \\
    Gemma & B-walls & Bedroom & 815 & 0.364 & 0.020 \\
    Gemma & B-walls & Office & 798 & 0.494 & 0.010 \\
    Gemma & B-walls & Living room & 814 & 0.548 & 0.000 \\
    \bottomrule
  \end{tabular}
\end{table}

%% file: tables/app-orient-all.tex
\begin{table}[htbp]
  \centering
  \scriptsize
  \setlength{\tabcolsep}{4pt}
  \caption{Orientation distribution $q(\mathrm{wall} \mid c)$, all objects (ScanNet, canonicalized rooms only).}
  \label{tab:app-orient-all}
  \begin{tabular}{llrrrrrr}
    \toprule
    Room & Layouts & Objects & N & E & S & W & JSD vs real \\
    \midrule
    Bathroom & Real & 531 & 0.332 & 0.267 & 0.185 & 0.217 & --- \\
    Bathroom & qwen A & 1,397 & 0.232 & 0.271 & 0.227 & 0.270 & 0.0070 \\
    Bathroom & qwen Bn & 680 & 0.260 & 0.222 & 0.281 & 0.237 & 0.0084 \\
    Bathroom & qwen Bw & 647 & 0.278 & 0.261 & 0.238 & 0.223 & 0.0029 \\
    Bathroom & gemma A & 1,206 & 0.219 & 0.312 & 0.119 & 0.350 & 0.0181 \\
    Bathroom & gemma Bn & 769 & 0.273 & 0.202 & 0.261 & 0.264 & 0.0082 \\
    Bathroom & gemma Bw & 741 & 0.263 & 0.270 & 0.198 & 0.269 & 0.0035 \\
    Bedroom & Real & 1,751 & 0.269 & 0.299 & 0.190 & 0.242 & --- \\
    Bedroom & qwen A & 1,736 & 0.261 & 0.255 & 0.242 & 0.242 & 0.0025 \\
    Bedroom & qwen Bn & 924 & 0.276 & 0.239 & 0.235 & 0.250 & 0.0029 \\
    Bedroom & qwen Bw & 886 & 0.275 & 0.228 & 0.259 & 0.238 & 0.0050 \\
    Bedroom & gemma A & 1,455 & 0.282 & 0.309 & 0.181 & 0.228 & 0.0003 \\
    Bedroom & gemma Bn & 770 & 0.377 & 0.303 & 0.125 & 0.196 & 0.0091 \\
    Bedroom & gemma Bw & 815 & 0.355 & 0.280 & 0.137 & 0.228 & 0.0053 \\
    Office & Real & 1,377 & 0.252 & 0.306 & 0.180 & 0.262 & --- \\
    Office & qwen A & 1,872 & 0.255 & 0.266 & 0.267 & 0.213 & 0.0062 \\
    Office & qwen Bn & 857 & 0.256 & 0.236 & 0.275 & 0.233 & 0.0077 \\
    Office & qwen Bw & 943 & 0.280 & 0.215 & 0.273 & 0.232 & 0.0095 \\
    Office & gemma A & 1,551 & 0.297 & 0.281 & 0.199 & 0.224 & 0.0021 \\
    Office & gemma Bn & 711 & 0.240 & 0.247 & 0.242 & 0.270 & 0.0039 \\
    Office & gemma Bw & 798 & 0.348 & 0.259 & 0.130 & 0.262 & 0.0068 \\
    Living room & Real & 756 & 0.271 & 0.259 & 0.257 & 0.213 & --- \\
    Living room & qwen A & 1,554 & 0.290 & 0.242 & 0.230 & 0.238 & 0.0010 \\
    Living room & qwen Bn & 738 & 0.295 & 0.234 & 0.258 & 0.213 & 0.0006 \\
    Living room & qwen Bw & 786 & 0.272 & 0.228 & 0.267 & 0.233 & 0.0008 \\
    Living room & gemma A & 1,294 & 0.358 & 0.327 & 0.086 & 0.230 & 0.0274 \\
    Living room & gemma Bn & 870 & 0.178 & 0.259 & 0.323 & 0.240 & 0.0072 \\
    Living room & gemma Bw & 814 & 0.248 & 0.252 & 0.236 & 0.264 & 0.0019 \\
    \bottomrule
  \end{tabular}
\end{table}

%% file: tables/app-orient-near-wall-only.tex
\begin{table}[htbp]
  \centering
  \scriptsize
  \setlength{\tabcolsep}{4pt}
  \caption{Orientation distribution $q(\mathrm{wall} \mid c)$, objects within 0.15 m of a wall (ScanNet, canonicalized rooms only).}
  \label{tab:app-orient-near-wall-only}
  \begin{tabular}{llrrrrrr}
    \toprule
    Room & Layouts & Objects & N & E & S & W & JSD vs real \\
    \midrule
    Bathroom & Real & 411 & 0.331 & 0.280 & 0.153 & 0.236 & --- \\
    Bathroom & qwen A & 371 & 0.275 & 0.291 & 0.181 & 0.253 & 0.0021 \\
    Bathroom & qwen Bn & 395 & 0.266 & 0.225 & 0.306 & 0.203 & 0.0169 \\
    Bathroom & qwen Bw & 500 & 0.296 & 0.262 & 0.228 & 0.214 & 0.0046 \\
    Bathroom & gemma A & 624 & 0.345 & 0.314 & 0.048 & 0.293 & 0.0166 \\
    Bathroom & gemma Bn & 422 & 0.355 & 0.258 & 0.344 & 0.043 & 0.0561 \\
    Bathroom & gemma Bw & 711 & 0.273 & 0.276 & 0.191 & 0.260 & 0.0027 \\
    Bedroom & Real & 982 & 0.253 & 0.328 & 0.182 & 0.237 & --- \\
    Bedroom & qwen A & 466 & 0.294 & 0.281 & 0.191 & 0.234 & 0.0017 \\
    Bedroom & qwen Bn & 314 & 0.198 & 0.245 & 0.274 & 0.283 & 0.0103 \\
    Bedroom & qwen Bw & 495 & 0.246 & 0.242 & 0.263 & 0.248 & 0.0069 \\
    Bedroom & gemma A & 410 & 0.205 & 0.468 & 0.059 & 0.268 & 0.0246 \\
    Bedroom & gemma Bn & 172 & 0.599 & 0.267 & 0.058 & 0.076 & 0.0766 \\
    Bedroom & gemma Bw & 473 & 0.378 & 0.319 & 0.099 & 0.203 & 0.0132 \\
    Office & Real & 630 & 0.209 & 0.282 & 0.187 & 0.321 & --- \\
    Office & qwen A & 372 & 0.309 & 0.280 & 0.194 & 0.218 & 0.0098 \\
    Office & qwen Bn & 186 & 0.194 & 0.210 & 0.306 & 0.290 & 0.0105 \\
    Office & qwen Bw & 493 & 0.304 & 0.183 & 0.278 & 0.235 & 0.0175 \\
    Office & gemma A & 609 & 0.268 & 0.332 & 0.105 & 0.296 & 0.0089 \\
    Office & gemma Bn & 202 & 0.173 & 0.317 & 0.253 & 0.257 & 0.0055 \\
    Office & gemma Bw & 583 & 0.305 & 0.298 & 0.117 & 0.280 & 0.0094 \\
    Living room & Real & 166 & 0.229 & 0.289 & 0.283 & 0.199 & --- \\
    Living room & qwen A & 372 & 0.328 & 0.218 & 0.177 & 0.277 & 0.0163 \\
    Living room & qwen Bn & 167 & 0.216 & 0.335 & 0.198 & 0.252 & 0.0063 \\
    Living room & qwen Bw & 527 & 0.269 & 0.220 & 0.279 & 0.232 & 0.0038 \\
    Living room & gemma A & 368 & 0.359 & 0.291 & 0.106 & 0.245 & 0.0293 \\
    Living room & gemma Bn & 168 & 0.083 & 0.232 & 0.357 & 0.327 & 0.0293 \\
    Living room & gemma Bw & 711 & 0.256 & 0.255 & 0.217 & 0.273 & 0.0061 \\
    \bottomrule
  \end{tabular}
\end{table}

%% file: tables/app-rel-nonadj.tex
\begin{table}[htbp]
  \centering
  \scriptsize
  \setlength{\tabcolsep}{4pt}
  \caption{Generation with an observed object (Sec.~\ref{sec:observed}; ScanNet bedrooms, O-anchor against O-move, five samples per room and condition). Rooms is the number of bedrooms that contain both categories; dropped are those in which the anchor could not be moved to another wall without overlap; used are those in which the target was generated under both prompts. $\kappa$ is the following slope of Equation~\eqref{eq:beta} with a room-resampled interval; $|\Delta a|$ is the median displacement of the anchor. Other objects is the same slope for the generated objects of neither category, given for the two related pairs (intervals in Table~\ref{tab:observed}).}
  \label{tab:app-rel-nonadj}
  \begin{tabular}{llrrrrrrr}
    \toprule
    Model & Pair & Rooms & Dropped & Used & $\kappa$ & 95\% CI & $|\Delta a|$ (m) & Other objects \\
    \midrule
    Qwen & desk $\rightarrow$ chair & 146 & 10 & 136 & 0.694 & [0.655, 0.733] & 3.10 & $-$0.134 \\
    Qwen & bed $\rightarrow$ night stand & 129 & 37 & 92 & 0.794 & [0.742, 0.842] & 1.76 & $-$0.275 \\
    Qwen & bed $\rightarrow$ chair (control) & 183 & 55 & 128 & $-$0.410 & [$-$0.489, $-$0.326] & 1.98 & --- \\
    Qwen & night stand $\rightarrow$ bed (reversed) & 129 & 3 & 126 & $-$0.027 & [$-$0.048, $-$0.006] & 3.67 & --- \\
    Gemma & desk $\rightarrow$ chair & 146 & 10 & 136 & 0.817 & [0.795, 0.840] & 3.10 & $-$0.196 \\
    Gemma & bed $\rightarrow$ night stand & 129 & 37 & 92 & 0.915 & [0.876, 0.955] & 1.76 & $-$0.310 \\
    Gemma & bed $\rightarrow$ chair (control) & 183 & 55 & 128 & $-$0.544 & [$-$0.612, $-$0.480] & 1.98 & --- \\
    Gemma & night stand $\rightarrow$ bed (reversed) & 129 & 3 & 126 & 0.035 & [0.009, 0.059] & 3.67 & --- \\
    \bottomrule
  \end{tabular}
\end{table}

%% file: tables/app-adsorption.tex
\begin{table}[htbp]
  \centering
  \scriptsize
  \setlength{\tabcolsep}{4pt}
  \caption{Wall-snapping control on the models' own generations (condition~A, CubiCasa, \NGenPerRoom{} layouts per room type and model, symmetrized). Each row is a pairwise comparison of two versions of the same layout; pairwise accuracy is the fraction of layouts in which the first-named version scores higher, $\Delta$ its mean margin in nats. Appendix~\ref{app:adsorption} defines the versions.}
  \label{tab:app-adsorption}
  \begin{tabular}{lllrrrrr}
    \toprule
    Model & Room & Comparison & $n$ & Pairwise acc. & 95\% CI & $\Delta$ & Ties \\
    \midrule
    Qwen & Kitchen & snapped $>$ original & 200 & 0.085 & [0.050, 0.125] & $-$13.2 & 0.00 \\
    Qwen & Kitchen & snapped $>$ random & 200 & 1.000 & [1.000, 1.000] & 38.8 & 0.00 \\
    Qwen & Kitchen & random $>$ original & 200 & 0.000 & [0.000, 0.000] & $-$51.9 & 0.00 \\
    Qwen & Kitchen & declared wall $>$ other wall & 200 & 0.940 & [0.905, 0.970] & 22.0 & 0.00 \\
    Qwen & Kitchen & other wall $>$ random & 200 & 0.870 & [0.820, 0.915] & 16.8 & 0.00 \\
    Qwen & Bathroom & snapped $>$ original & 200 & 0.015 & [0.000, 0.035] & $-$14.3 & 0.00 \\
    Qwen & Bathroom & snapped $>$ random & 200 & 0.995 & [0.985, 1.000] & 26.3 & 0.00 \\
    Qwen & Bathroom & random $>$ original & 200 & 0.000 & [0.000, 0.000] & $-$40.6 & 0.00 \\
    Qwen & Bathroom & declared wall $>$ other wall & 200 & 0.910 & [0.870, 0.950] & 13.8 & 0.00 \\
    Qwen & Bathroom & other wall $>$ random & 200 & 0.860 & [0.810, 0.905] & 12.4 & 0.00 \\
    Gemma & Kitchen & snapped $>$ original & 200 & 0.025 & [0.005, 0.050] & $-$22.0 & 0.00 \\
    Gemma & Kitchen & snapped $>$ random & 200 & 1.000 & [1.000, 1.000] & 66.2 & 0.00 \\
    Gemma & Kitchen & random $>$ original & 200 & 0.000 & [0.000, 0.000] & $-$88.2 & 0.00 \\
    Gemma & Kitchen & declared wall $>$ other wall & 200 & 1.000 & [1.000, 1.000] & 102.1 & 0.00 \\
    Gemma & Kitchen & other wall $>$ random & 200 & 0.160 & [0.110, 0.210] & $-$35.9 & 0.00 \\
    Gemma & Bathroom & snapped $>$ original & 200 & 0.075 & [0.040, 0.115] & $-$23.9 & 0.00 \\
    Gemma & Bathroom & snapped $>$ random & 200 & 0.990 & [0.975, 1.000] & 45.7 & 0.00 \\
    Gemma & Bathroom & random $>$ original & 200 & 0.000 & [0.000, 0.000] & $-$69.6 & 0.00 \\
    Gemma & Bathroom & declared wall $>$ other wall & 200 & 0.980 & [0.960, 0.995] & 62.4 & 0.00 \\
    Gemma & Bathroom & other wall $>$ random & 200 & 0.305 & [0.245, 0.370] & $-$16.7 & 0.00 \\
    \bottomrule
  \end{tabular}
\end{table}

%% file: tables/app-adsorption-diag.tex
\begin{table}[htbp]
  \centering
  \footnotesize
  \setlength{\tabcolsep}{4pt}
  \caption{Diagnostics for the wall-snapping control. For each version: the number of coordinate and footprint values ($x$, $y$, $w$, $d$ over all objects), the number of distinct values among them after rounding to two decimals, the share on the $0.05$~m lattice, and the median token count of the serialized layout. The last two rows per room give the snapping displacement over the objects that moved, and the number of objects whose target had to be clamped inside the room.}
  \label{tab:app-adsorption-diag}
  \begin{tabular}{lllrrrr}
    \toprule
    Model & Room & Version / quantity & Values & Distinct & On grid & Tokens (median) \\
    \midrule
    Qwen & Kitchen & original & 7,036 & 107 & 1.00 & 372.5 \\
    Qwen & Kitchen & snapped & 7,036 & 154 & 1.00 & 372.5 \\
    Qwen & Kitchen & other wall & 7,036 & 149 & 1.00 & 372.5 \\
    Qwen & Kitchen & random & 7,036 & 180 & 1.00 & 372.5 \\
    Qwen & Kitchen & moved objects / median / mean / p90 (m) & 1,520 & 0.75 & 1.36 & 3.50 \\
    Qwen & Kitchen & clamped (snapped / random) & 87 & 9 &  &  \\
    Qwen & Bathroom & original & 5,956 & 90 & 1.00 & 327.0 \\
    Qwen & Bathroom & snapped & 5,956 & 126 & 1.00 & 327.0 \\
    Qwen & Bathroom & other wall & 5,956 & 125 & 1.00 & 327.0 \\
    Qwen & Bathroom & random & 5,956 & 135 & 1.00 & 327.0 \\
    Qwen & Bathroom & moved objects / median / mean / p90 (m) & 1,321 & 0.75 & 1.03 & 2.35 \\
    Qwen & Bathroom & clamped (snapped / random) & 42 & 3 &  &  \\
    Gemma & Kitchen & original & 6,304 & 51 & 1.00 & 393.0 \\
    Gemma & Kitchen & snapped & 6,304 & 64 & 1.00 & 393.0 \\
    Gemma & Kitchen & other wall & 6,304 & 70 & 1.00 & 393.0 \\
    Gemma & Kitchen & random & 6,304 & 110 & 1.00 & 393.0 \\
    Gemma & Kitchen & moved objects / median / mean / p90 (m) & 1,386 & 0.20 & 0.19 & 0.30 \\
    Gemma & Kitchen & clamped (snapped / random) & 52 & 4 &  &  \\
    Gemma & Bathroom & original & 5,180 & 37 & 1.00 & 300.0 \\
    Gemma & Bathroom & snapped & 5,180 & 49 & 1.00 & 300.0 \\
    Gemma & Bathroom & other wall & 5,180 & 51 & 1.00 & 300.0 \\
    Gemma & Bathroom & random & 5,180 & 60 & 1.00 & 300.0 \\
    Gemma & Bathroom & moved objects / median / mean / p90 (m) & 1,162 & 0.20 & 0.28 & 0.50 \\
    Gemma & Bathroom & clamped (snapped / random) & 63 & 4 &  &  \\
    \bottomrule
  \end{tabular}
\end{table}

%% file: tables/app-entropy.tex
\begin{table}[htbp]
  \centering
  \scriptsize
  \setlength{\tabcolsep}{4pt}
  \caption{Likelihood gap between the model's own generations and real rooms, matched on object count (condition~A, CubiCasa). Gap is the object-count-weighted mean over bins of (mean log-likelihood of generated minus mean log-likelihood of real), in nats, after quantizing both sets of layouts to the grid in the second column; the per-bin columns give the gap with the number of real and generated layouts in the bin in parentheses. Appendix~\ref{app:granularity} gives the matching rule.}
  \label{tab:app-entropy}
  \begin{tabular}{lllrrrrrr}
    \toprule
    Model & Room & Grid & Gap & 5 objects & 6 objects & 7 objects & 8 objects & 9+ objects \\
    \midrule
    Qwen & Kitchen & as emitted & 86.1 &  &  & 87.1 (52/39) & 81.3 (32/59) & 89.8 (27/99) \\
    Qwen & Kitchen & $0.05$~m & 51.6 &  &  & 55.4 (52/39) & 46.4 (32/59) & 50.4 (27/99) \\
    Qwen & Kitchen & $0.10$~m & 35.4 &  &  & 38.7 (52/39) & 30.7 (32/59) & 34.6 (27/99) \\
    Qwen & Kitchen & $0.25$~m & 7.1 &  &  & 12.3 (52/39) & 2.7 (32/59) & 2.1 (27/99) \\
    Qwen & Bathroom & as emitted & 56.9 &  & 56.9 (17/47) &  &  &  \\
    Qwen & Bathroom & $0.05$~m & 26.6 &  & 26.6 (17/47) &  &  &  \\
    Qwen & Bathroom & $0.10$~m & 18.0 &  & 18.0 (17/47) &  &  &  \\
    Qwen & Bathroom & $0.25$~m & -2.3 &  & -2.3 (17/47) &  &  &  \\
    Gemma & Kitchen & as emitted & 252.8 &  &  & 238.6 (52/60) & 247.1 (32/107) & 287.1 (27/33) \\
    Gemma & Kitchen & $0.05$~m & 217.4 &  &  & 203.2 (52/60) & 209.5 (32/107) & 254.1 (27/33) \\
    Gemma & Kitchen & $0.10$~m & 174.4 &  &  & 161.5 (52/60) & 160.9 (32/107) & 215.5 (27/33) \\
    Gemma & Kitchen & $0.25$~m & 113.8 &  &  & 103.1 (52/60) & 107.6 (32/107) & 141.6 (27/33) \\
    Gemma & Bathroom & as emitted & 170.1 & 168.3 (40/24) & 174.5 (17/100) &  &  &  \\
    Gemma & Bathroom & $0.05$~m & 132.5 & 129.1 (40/24) & 140.3 (17/100) &  &  &  \\
    Gemma & Bathroom & $0.10$~m & 97.4 & 94.5 (40/24) & 104.3 (17/100) &  &  &  \\
    Gemma & Bathroom & $0.25$~m & 65.1 & 61.1 (40/24) & 74.4 (17/100) &  &  &  \\
    \bottomrule
  \end{tabular}
\end{table}

%% file: tables/app-coordvalues.tex
\begin{table}[htbp]
  \centering
  \footnotesize
  \setlength{\tabcolsep}{4pt}
  \caption{Coordinate values of the real and the generated layouts (CubiCasa, condition~A): number of $x$ and $y$ values, number of distinct values after rounding to two decimals, and the share that are multiples of 0.05~m.}
  \label{tab:app-coordvalues}
  \begin{tabular}{lllrrr}
    \toprule
    Model & Room & Layouts & Values & Distinct & On grid \\
    \midrule
    --- & Kitchen & real & 39,172 & 605 & 0.289 \\
    Qwen & Kitchen & generated & 3,518 & 109 & 0.998 \\
    --- & Bathroom & real & 35,392 & 396 & 0.223 \\
    Qwen & Bathroom & generated & 2,978 & 89 & 0.999 \\
    Gemma & Kitchen & generated & 3,152 & 49 & 1.000 \\
    Gemma & Bathroom & generated & 2,590 & 36 & 1.000 \\
    \bottomrule
  \end{tabular}
\end{table}

%% file: tables/app-symcoord.tex
\begin{table}[htbp]
  \centering
  \scriptsize
  \setlength{\tabcolsep}{4pt}
  \caption{Consistency between the declared or supplied wall and the emitted coordinates (ScanNet). Chance is computed from each condition's own distribution over walls (Equation~\eqref{eq:chance}).}
  \label{tab:app-symcoord}
  \begin{tabular}{llrrrrrrrr}
    \toprule
    Model & Cond. & Rooms & Declared & Consistency & Chance & Gap med. & Gap p90 & Outside & Overlap \\
    \midrule
    Qwen & A & 799 & 3,711 & 0.244 & 0.093 & 0.55 & 2.40 & 0.102 & 0.063 \\
    Qwen & A$'$ & 800 & 3,626 & 0.429 & 0.125 & 0.20 & 1.25 & 0.151 & 0.060 \\
    Qwen & A$''$ & 565 & 1,829 & 0.896 & 0.300 & $-$0.00 & 0.21 & 0.321 & 0.091 \\
    Qwen & A$''$-c & 565 & 1,829 & 0.890 & 0.293 & $-$0.00 & 0.22 & 0.305 & 0.099 \\
    Qwen & A$''$-r & 565 & 1,829 & 0.894 & 0.292 & 0.00 & 0.19 & 0.307 & 0.110 \\
    Qwen & A$''$-i & 565 & 1,829 & 0.225 & 0.193 & 1.15 & 3.23 & 0.265 & 0.091 \\
    Qwen & A$''$-per & 565 & 1,829 & 0.877 & 0.290 & 0.00 & 0.28 & 0.285 & 0.093 \\
    Gemma & A & 800 & 3,589 & 0.467 & 0.161 & 0.20 & 0.75 & 0.108 & 0.089 \\
    Gemma & A$'$ & 800 & 3,511 & 0.905 & 0.312 & 0.00 & 0.15 & 0.168 & 0.126 \\
    Gemma & A$''$ & 565 & 1,829 & 0.989 & 0.308 & 0.00 & 0.00 & 0.132 & 0.076 \\
    Gemma & A$''$-c & 565 & 1,829 & 0.987 & 0.310 & 0.00 & 0.00 & 0.121 & 0.076 \\
    Gemma & A$''$-r & 565 & 1,829 & 0.986 & 0.310 & 0.00 & 0.00 & 0.123 & 0.075 \\
    Gemma & A$''$-i & 565 & 1,829 & 0.349 & 0.298 & 0.92 & 3.25 & 0.113 & 0.077 \\
    Gemma & A$''$-per & 565 & 1,829 & 0.990 & 0.308 & 0.00 & 0.00 & 0.119 & 0.072 \\
    \bottomrule
  \end{tabular}
\end{table}

%% file: tables/app-appinputs.tex
\begin{table}[htbp]
  \centering
  \scriptsize
  \setlength{\tabcolsep}{4pt}
  \caption{Selection of the input rooms of condition~\condApp{} and its variants (ScanNet). Each column applies one more rule: rooms with at least one floor-standing object in the closed vocabulary; rooms whose rectangle and door can be recovered (Appendix~\ref{app:canon}); rooms with \AppInMinObj{} to \AppInMaxObj{} such objects. At most \AppInMax{} eligible rooms per room type are drawn after a seeded shuffle; only bedrooms exceed the limit. The last two columns count the objects supplied and, among them, those supplied with no wall (``none'') because the real object is farther than \AppInTol~m from every wall. Inputs are identical for both models and all variants.}
  \label{tab:app-appinputs}
  \begin{tabular}{lrrrrrrr}
    \toprule
    Room & ScanNet & Vocabulary & Door & Size & Used & Objects & No wall \\
    \midrule
    Bathroom & 210 & 197 & 189 & 155 & 155 & 497 & 116 \\
    Bedroom & 273 & 273 & 240 & 237 & 200 & 1,431 & 618 \\
    Living room & 224 & 221 & 95 & 85 & 85 & 581 & 441 \\
    Office & 173 & 173 & 146 & 125 & 125 & 1,040 & 545 \\
    \textbf{Total} & 880 & 864 & 670 & 602 & 565 & 3,549 & 1,720 \\
    \bottomrule
  \end{tabular}
\end{table}

%% file: tables/app-declgeo.tex
\begin{table}[htbp]
  \centering
  \scriptsize
  \setlength{\tabcolsep}{2pt}
  \caption{Declared wall against nearest wall, per object (ScanNet, all room types pooled, objects that declare a wall). ``Declared'' is the distribution of \texttt{wall\_contact} values; ``Nearest'' is the distribution of the wall nearest to each of the same objects by its coordinates; JSD is the Jensen--Shannon divergence between the two distributions in nats; Agreement is the fraction of objects whose declared wall is their nearest wall, and Chance the agreement expected if declarations were independent of the coordinates, $\sum_\omega q(\omega)\,p(\omega)$ over the two distributions (Appendix~\ref{app:declgeo}).}
  \label{tab:app-declgeo}
  \begin{tabular}{llrrrrrrrrrrrr}
    \toprule
    Model & Cond. & $n$ & Decl.\ N & Decl.\ E & Decl.\ S & Decl.\ W & Near.\ N & Near.\ E & Near.\ S & Near.\ W & JSD & Agree. & Chance \\
    \midrule
    Qwen & A & 3,711 & 0.270 & 0.283 & 0.178 & 0.269 & 0.275 & 0.301 & 0.182 & 0.241 & 0.0005 & 0.642 & 0.257 \\
    Qwen & B-neutral & 2,057 & 0.352 & 0.327 & 0.112 & 0.208 & 0.365 & 0.314 & 0.136 & 0.185 & 0.0010 & 0.732 & 0.285 \\
    Qwen & B-walls & 2,932 & 0.299 & 0.278 & 0.193 & 0.231 & 0.311 & 0.268 & 0.233 & 0.188 & 0.0022 & 0.803 & 0.256 \\
    Gemma & A & 3,589 & 0.296 & 0.270 & 0.127 & 0.307 & 0.259 & 0.328 & 0.113 & 0.301 & 0.0022 & 0.827 & 0.272 \\
    Gemma & B-neutral & 2,086 & 0.235 & 0.339 & 0.105 & 0.320 & 0.245 & 0.321 & 0.113 & 0.321 & 0.0002 & 0.948 & 0.281 \\
    Gemma & B-walls & 2,998 & 0.291 & 0.268 & 0.155 & 0.286 & 0.288 & 0.274 & 0.158 & 0.280 & 0.0001 & 0.972 & 0.262 \\
    \bottomrule
  \end{tabular}
\end{table}

%% file: tables/app-condap.tex
\begin{table}[htbp]
  \centering
  \scriptsize
  \setlength{\tabcolsep}{4pt}
  \caption{The conditioning measurement of Sec.~\ref{sec:hierarchy} repeated on the layouts generated with the field order reversed (condition~A$'$), alongside condition~A. Columns as in Tables~\ref{tab:app-excess} and~\ref{tab:app-slope}; the A rows repeat those tables.}
  \label{tab:app-condap}
  \begin{tabular}{lrrrrrrr}
    \toprule
    Layouts & Bathroom & Bedroom & Office & Living room & SD & Slope & 95\% CI \\
    \midrule
    Real data & 0.146 & 0.092 & $-$0.018 & $-$0.163 &  &  &  \\
    Qwen A & $-$0.121 & $-$0.093 & $-$0.113 & $-$0.119 & 0.0129 & 0.029 & [$-$0.065, 0.120] \\
    Qwen A$'$ & $-$0.019 & $-$0.038 & $-$0.051 & $-$0.107 & 0.0379 & 0.272 & [0.167, 0.381] \\
    Gemma A & $-$0.035 & $-$0.190 & 0.023 & $-$0.185 & 0.1075 & 0.227 & [0.103, 0.352] \\
    Gemma A$'$ & 0.224 & 0.104 & 0.226 & 0.004 & 0.1067 & 0.531 & [0.435, 0.636] \\
    \bottomrule
  \end{tabular}
\end{table}

%% file: tables/app-priority.tex
\begin{table}[htbp]
  \centering
  \footnotesize
  \setlength{\tabcolsep}{4pt}
  \caption{Containment instructions: change in the fraction of objects placed partly outside the room relative to condition~A$''$, paired by source room (same inputs), with a bootstrap over rooms. Negative is fewer objects outside.}
  \label{tab:app-priority}
  \begin{tabular}{llrrr}
    \toprule
    Model & Condition & Rooms & $\Delta$ outside & 95\% CI \\
    \midrule
    Qwen & A$''$-c & 565 & $-$0.010 & [$-$0.032, 0.011] \\
    Qwen & A$''$-r & 565 & $-$0.008 & [$-$0.030, 0.014] \\
    Qwen & A$''$-per & 565 & $-$0.038 & [$-$0.060, $-$0.015] \\
    Qwen & A$''$-i & 565 & $-$0.087 & [$-$0.113, $-$0.061] \\
    Gemma & A$''$-c & 565 & $-$0.017 & [$-$0.026, $-$0.008] \\
    Gemma & A$''$-r & 565 & $-$0.013 & [$-$0.022, $-$0.003] \\
    Gemma & A$''$-per & 565 & $-$0.018 & [$-$0.028, $-$0.008] \\
    Gemma & A$''$-i & 565 & $-$0.066 & [$-$0.089, $-$0.044] \\
    \bottomrule
  \end{tabular}
\end{table}

%% file: tables/app-symaux.tex
\begin{table}[htbp]
  \centering
  \scriptsize
  \setlength{\tabcolsep}{4pt}
  \caption{Condition~A$''$, auxiliary quantities (ScanNet). Natural / unnatural: consistency with the supplied wall for objects of categories that are usually against a wall in the real data and for those that rarely are, with the number of objects in each group. None: objects supplied with ``none'' and the fraction of them that nonetheless end up within $0.15$~m of a wall; Uniform is the same fraction expected if each of these objects were placed uniformly in its room (Equation~\eqref{eq:baseline}), while the real objects are by construction all farther than $0.15$~m from every wall. Appendix~\ref{app:consistency} gives the definitions.}
  \label{tab:app-symaux}
  \begin{tabular}{lrrrrrrr}
    \toprule
    Model & Natural $n$ & Consist. & Unnatural $n$ & Consist. & None $n$ & Near wall & Uniform \\
    \midrule
    Qwen & 1,437 & 0.891 & 392 & 0.911 & 1,720 & 0.163 & 0.406 \\
    Gemma & 1,437 & 0.986 & 392 & 0.997 & 1,720 & 0.056 & 0.406 \\
    \bottomrule
  \end{tabular}
\end{table}

%% file: tables/app-search-qwen.tex
\begin{table}[htbp]
  \centering
  \scriptsize
  \setlength{\tabcolsep}{4pt}
  \caption{Search cost by room type and prior, Qwen generations (ScanNet; Sec.~\ref{sec:searchcost}). Cost is the expected fraction of the room opened before the target's cell, with tied cells opened in uniformly random order; lower is better and an uninformative prior gives \CostUExpected. Objects is the number of search episodes; a model prior has fewer when its heat map for a category is empty. M-A/cat and R/cat ignore the category, M-A/room and R/room the room type. Top-1 dist.\ is the median distance from the center of the most probable cell to the target; S@10\% the probability of having found the target after opening 10\% of the room.}
  \label{tab:app-search-qwen}
  \begin{tabular}{llrrrrrr}
    \toprule
    Room & Prior & Rooms & Objects & Cost & 95\% CI & Top-1 dist.\ (m) & S@10\% \\
    \midrule
    Bathroom & U & 187 & 493 & 0.502 & [0.502, 0.502] & 1.38 & 0.098 \\
    Bathroom & W & 187 & 493 & 0.651 & [0.640, 0.660] & 1.60 & 0.008 \\
    Bathroom & M-A & 187 & 493 & 0.425 & [0.399, 0.451] & 1.09 & 0.139 \\
    Bathroom & M-Bw & 187 & 493 & 0.418 & [0.394, 0.443] & 1.25 & 0.134 \\
    Bathroom & M-A/cat & 187 & 493 & 0.390 & [0.364, 0.415] & 1.31 & 0.178 \\
    Bathroom & M-A/room & 187 & 493 & 0.417 & [0.390, 0.444] & 1.14 & 0.136 \\
    Bathroom & R & 187 & 493 & 0.209 & [0.183, 0.236] & 1.02 & 0.539 \\
    Bathroom & R/cat & 187 & 493 & 0.228 & [0.206, 0.252] & 1.01 & 0.443 \\
    Bathroom & R/room & 187 & 493 & 0.208 & [0.183, 0.234] & 0.99 & 0.524 \\
    Bedroom & U & 240 & 1,718 & 0.502 & [0.502, 0.502] & 2.44 & 0.098 \\
    Bedroom & W & 240 & 1,718 & 0.572 & [0.563, 0.582] & 2.93 & 0.042 \\
    Bedroom & M-A & 240 & 1,718 & 0.466 & [0.451, 0.479] & 2.45 & 0.107 \\
    Bedroom & M-Bw & 240 & 1,669 & 0.482 & [0.469, 0.493] & 2.29 & 0.113 \\
    Bedroom & M-A/cat & 240 & 1,718 & 0.454 & [0.439, 0.469] & 1.73 & 0.102 \\
    Bedroom & M-A/room & 240 & 1,718 & 0.465 & [0.451, 0.479] & 2.16 & 0.087 \\
    Bedroom & R & 240 & 1,718 & 0.250 & [0.234, 0.267] & 2.19 & 0.394 \\
    Bedroom & R/cat & 240 & 1,718 & 0.318 & [0.303, 0.332] & 2.14 & 0.254 \\
    Bedroom & R/room & 240 & 1,718 & 0.263 & [0.246, 0.280] & 2.06 & 0.348 \\
    Office & U & 146 & 1,363 & 0.502 & [0.502, 0.502] & 2.25 & 0.098 \\
    Office & W & 146 & 1,363 & 0.579 & [0.569, 0.589] & 2.69 & 0.039 \\
    Office & M-A & 146 & 1,363 & 0.444 & [0.429, 0.459] & 1.96 & 0.133 \\
    Office & M-Bw & 146 & 1,363 & 0.484 & [0.470, 0.498] & 2.61 & 0.108 \\
    Office & M-A/cat & 146 & 1,363 & 0.476 & [0.458, 0.494] & 1.74 & 0.102 \\
    Office & M-A/room & 146 & 1,363 & 0.419 & [0.404, 0.434] & 1.99 & 0.130 \\
    Office & R & 146 & 1,363 & 0.278 & [0.259, 0.298] & 1.94 & 0.327 \\
    Office & R/cat & 146 & 1,363 & 0.321 & [0.305, 0.337] & 2.94 & 0.214 \\
    Office & R/room & 146 & 1,363 & 0.286 & [0.268, 0.303] & 2.05 & 0.287 \\
    Living room & U & 95 & 720 & 0.502 & [0.502, 0.502] & 2.85 & 0.098 \\
    Living room & W & 95 & 720 & 0.595 & [0.585, 0.604] & 3.57 & 0.021 \\
    Living room & M-A & 95 & 720 & 0.453 & [0.430, 0.475] & 2.46 & 0.125 \\
    Living room & M-Bw & 95 & 720 & 0.502 & [0.481, 0.521] & 2.90 & 0.108 \\
    Living room & M-A/cat & 95 & 720 & 0.506 & [0.482, 0.530] & 1.93 & 0.098 \\
    Living room & M-A/room & 95 & 720 & 0.424 & [0.405, 0.443] & 2.25 & 0.121 \\
    Living room & R & 95 & 720 & 0.347 & [0.320, 0.375] & 2.60 & 0.251 \\
    Living room & R/cat & 95 & 720 & 0.388 & [0.366, 0.409] & 2.95 & 0.164 \\
    Living room & R/room & 95 & 720 & 0.352 & [0.325, 0.379] & 2.54 & 0.198 \\
    \bottomrule
  \end{tabular}
\end{table}

%% file: tables/app-search-gemma.tex
\begin{table}[htbp]
  \centering
  \scriptsize
  \setlength{\tabcolsep}{4pt}
  \caption{Search cost by room type and prior, Gemma generations (ScanNet; Sec.~\ref{sec:searchcost}). Cost is the expected fraction of the room opened before the target's cell, with tied cells opened in uniformly random order; lower is better and an uninformative prior gives \CostUExpected. Objects is the number of search episodes; a model prior has fewer when its heat map for a category is empty. M-A/cat and R/cat ignore the category, M-A/room and R/room the room type. Top-1 dist.\ is the median distance from the center of the most probable cell to the target; S@10\% the probability of having found the target after opening 10\% of the room.}
  \label{tab:app-search-gemma}
  \begin{tabular}{llrrrrrr}
    \toprule
    Room & Prior & Rooms & Objects & Cost & 95\% CI & Top-1 dist.\ (m) & S@10\% \\
    \midrule
    Bathroom & U & 187 & 493 & 0.502 & [0.502, 0.502] & 1.38 & 0.098 \\
    Bathroom & W & 187 & 493 & 0.651 & [0.640, 0.660] & 1.60 & 0.008 \\
    Bathroom & M-A & 187 & 493 & 0.433 & [0.410, 0.456] & 1.10 & 0.162 \\
    Bathroom & M-Bw & 187 & 493 & 0.489 & [0.471, 0.506] & 1.23 & 0.126 \\
    Bathroom & M-A/cat & 187 & 493 & 0.396 & [0.373, 0.420] & 1.15 & 0.190 \\
    Bathroom & M-A/room & 187 & 493 & 0.437 & [0.413, 0.460] & 1.16 & 0.146 \\
    Bathroom & R & 187 & 493 & 0.209 & [0.183, 0.236] & 1.02 & 0.539 \\
    Bathroom & R/cat & 187 & 493 & 0.228 & [0.206, 0.252] & 1.01 & 0.443 \\
    Bathroom & R/room & 187 & 493 & 0.208 & [0.183, 0.234] & 0.99 & 0.524 \\
    Bedroom & U & 240 & 1,718 & 0.502 & [0.502, 0.502] & 2.44 & 0.098 \\
    Bedroom & W & 240 & 1,718 & 0.572 & [0.563, 0.582] & 2.93 & 0.042 \\
    Bedroom & M-A & 240 & 1,343 & 0.459 & [0.446, 0.472] & 2.10 & 0.146 \\
    Bedroom & M-Bw & 240 & 1,186 & 0.493 & [0.483, 0.502] & 2.00 & 0.098 \\
    Bedroom & M-A/cat & 240 & 1,343 & 0.429 & [0.411, 0.446] & 1.68 & 0.144 \\
    Bedroom & M-A/room & 240 & 1,343 & 0.442 & [0.429, 0.456] & 2.05 & 0.161 \\
    Bedroom & R & 240 & 1,718 & 0.250 & [0.234, 0.267] & 2.19 & 0.394 \\
    Bedroom & R/cat & 240 & 1,718 & 0.318 & [0.303, 0.332] & 2.14 & 0.254 \\
    Bedroom & R/room & 240 & 1,718 & 0.263 & [0.246, 0.280] & 2.06 & 0.348 \\
    Office & U & 146 & 1,363 & 0.502 & [0.502, 0.502] & 2.25 & 0.098 \\
    Office & W & 146 & 1,363 & 0.579 & [0.569, 0.589] & 2.69 & 0.039 \\
    Office & M-A & 146 & 1,363 & 0.463 & [0.449, 0.477] & 2.40 & 0.114 \\
    Office & M-Bw & 146 & 1,207 & 0.485 & [0.473, 0.497] & 1.98 & 0.131 \\
    Office & M-A/cat & 146 & 1,363 & 0.501 & [0.485, 0.518] & 2.83 & 0.079 \\
    Office & M-A/room & 146 & 1,363 & 0.425 & [0.409, 0.440] & 2.09 & 0.145 \\
    Office & R & 146 & 1,363 & 0.278 & [0.259, 0.298] & 1.94 & 0.327 \\
    Office & R/cat & 146 & 1,363 & 0.321 & [0.305, 0.337] & 2.94 & 0.214 \\
    Office & R/room & 146 & 1,363 & 0.286 & [0.268, 0.303] & 2.05 & 0.287 \\
    Living room & U & 95 & 720 & 0.502 & [0.502, 0.502] & 2.85 & 0.098 \\
    Living room & W & 95 & 720 & 0.595 & [0.585, 0.604] & 3.57 & 0.021 \\
    Living room & M-A & 95 & 720 & 0.487 & [0.472, 0.502] & 2.35 & 0.123 \\
    Living room & M-Bw & 95 & 689 & 0.512 & [0.500, 0.523] & 3.13 & 0.091 \\
    Living room & M-A/cat & 95 & 720 & 0.503 & [0.481, 0.525] & 1.93 & 0.116 \\
    Living room & M-A/room & 95 & 720 & 0.453 & [0.427, 0.477] & 2.37 & 0.122 \\
    Living room & R & 95 & 720 & 0.347 & [0.320, 0.375] & 2.60 & 0.251 \\
    Living room & R/cat & 95 & 720 & 0.388 & [0.366, 0.409] & 2.95 & 0.164 \\
    Living room & R/room & 95 & 720 & 0.352 & [0.325, 0.379] & 2.54 & 0.198 \\
    \bottomrule
  \end{tabular}
\end{table}

%% file: tables/app-search-diff.tex
\begin{table}[htbp]
  \centering
  \footnotesize
  \setlength{\tabcolsep}{4pt}
  \caption{Paired differences in search cost (Sec.~\ref{sec:searchcost}). Each difference is taken room by room over the episodes for which both priors are defined, then averaged over rooms; intervals resample rooms (2\,000 draws), shown on the line below each difference. Negative means the first prior is better.}
  \label{tab:app-search-diff}
  \begin{tabular}{llrrrr}
    \toprule
    Model & Difference & Bathroom & Bedroom & Office & Living room \\
    \midrule
    Qwen & M-A $-$ U & $-$0.077 & $-$0.036 & $-$0.058 & $-$0.049 \\
     &  & [$-$0.102, $-$0.048] & [$-$0.050, $-$0.022] & [$-$0.074, $-$0.044] & [$-$0.071, $-$0.029] \\
    Qwen & M-Bw $-$ U & $-$0.084 & $-$0.020 & $-$0.018 & $-$0.000 \\
     &  & [$-$0.107, $-$0.060] & [$-$0.033, $-$0.008] & [$-$0.033, $-$0.004] & [$-$0.021, 0.019] \\
    Qwen & M-Bw $-$ M-A & $-$0.007 & 0.017 & 0.040 & 0.049 \\
     &  & [$-$0.040, 0.024] & [0.002, 0.032] & [0.021, 0.060] & [0.020, 0.077] \\
    Qwen & M-A/room $-$ M-A & $-$0.008 & $-$0.000 & $-$0.024 & $-$0.028 \\
     &  & [$-$0.022, 0.005] & [$-$0.010, 0.010] & [$-$0.039, $-$0.011] & [$-$0.049, $-$0.007] \\
    Qwen & M-A/cat $-$ M-A & $-$0.035 & $-$0.012 & 0.032 & 0.053 \\
     &  & [$-$0.065, $-$0.005] & [$-$0.027, 0.004] & [0.014, 0.051] & [0.028, 0.079] \\
    Qwen & R $-$ U & $-$0.293 & $-$0.252 & $-$0.224 & $-$0.155 \\
     &  & [$-$0.320, $-$0.267] & [$-$0.268, $-$0.235] & [$-$0.242, $-$0.205] & [$-$0.182, $-$0.127] \\
    Qwen & R/room $-$ R & $-$0.001 & 0.013 & 0.008 & 0.004 \\
     &  & [$-$0.012, 0.009] & [0.006, 0.019] & [$-$0.001, 0.016] & [$-$0.015, 0.023] \\
    Qwen & R/cat $-$ R & 0.019 & 0.068 & 0.043 & 0.040 \\
     &  & [0.002, 0.034] & [0.054, 0.081] & [0.029, 0.056] & [0.021, 0.059] \\
    Gemma & M-A $-$ U & $-$0.069 & $-$0.043 & $-$0.039 & $-$0.015 \\
     &  & [$-$0.093, $-$0.046] & [$-$0.056, $-$0.030] & [$-$0.053, $-$0.025] & [$-$0.030, $-$0.000] \\
    Gemma & M-Bw $-$ U & $-$0.013 & $-$0.009 & $-$0.017 & 0.010 \\
     &  & [$-$0.031, 0.003] & [$-$0.020, $-$0.000] & [$-$0.029, $-$0.005] & [$-$0.001, 0.021] \\
    Gemma & M-Bw $-$ M-A & 0.056 & 0.038 & 0.029 & 0.027 \\
     &  & [0.024, 0.087] & [0.021, 0.053] & [0.012, 0.046] & [0.008, 0.046] \\
    Gemma & M-A/room $-$ M-A & 0.003 & $-$0.017 & $-$0.037 & $-$0.034 \\
     &  & [$-$0.005, 0.011] & [$-$0.023, $-$0.010] & [$-$0.051, $-$0.025] & [$-$0.052, $-$0.017] \\
    Gemma & M-A/cat $-$ M-A & $-$0.037 & $-$0.030 & 0.039 & 0.016 \\
     &  & [$-$0.063, $-$0.011] & [$-$0.048, $-$0.012] & [0.025, 0.052] & [$-$0.001, 0.034] \\
    Gemma & R $-$ U & $-$0.293 & $-$0.252 & $-$0.224 & $-$0.155 \\
     &  & [$-$0.320, $-$0.267] & [$-$0.268, $-$0.235] & [$-$0.242, $-$0.205] & [$-$0.182, $-$0.127] \\
    Gemma & R/room $-$ R & $-$0.001 & 0.013 & 0.008 & 0.004 \\
     &  & [$-$0.012, 0.009] & [0.006, 0.019] & [$-$0.001, 0.016] & [$-$0.015, 0.023] \\
    Gemma & R/cat $-$ R & 0.019 & 0.068 & 0.043 & 0.040 \\
     &  & [0.002, 0.034] & [0.054, 0.081] & [0.029, 0.056] & [0.021, 0.059] \\
    \bottomrule
  \end{tabular}
\end{table}

%% file: tables/app-searchsd.tex
\begin{table}[htbp]
  \centering
  \footnotesize
  \setlength{\tabcolsep}{4pt}
  \caption{Variation of the search cost across the four room types. SD is the sample standard deviation of the four room-type means of the per-room mean cost; its interval resamples rooms within each room type (10\,000 draws). The ratio column is the real-data prior's SD over this prior's, with its bootstrap interval. $p$ is the permutation test of Appendix~\ref{app:permutation} applied to the per-room mean costs, with the variance of the four group means as the statistic. U is omitted: with random tie-breaking its cost is \CostUExpected{} in every room. W and R do not depend on the model; they are repeated because the bootstrap seeds differ between the two runs. M-A/room carries no room-type information by construction, yet its cost varies across room types, because the same prior meets differently placed targets.}
  \label{tab:app-searchsd}
  \begin{tabular}{llrrrr}
    \toprule
    Model & Prior & SD & 95\% CI & R / this & $p$ (equality) \\
    \midrule
    Qwen & W & 0.0357 & [0.0297, 0.0416] & 1.63 [1.19, 2.15] & $<$0.001 \\
    Qwen & M-A & 0.0170 & [0.0081, 0.0311] & 3.15 [1.78, 7.32] & 0.063 \\
    Qwen & M-Bw & 0.0366 & [0.0241, 0.0503] & 1.58 [1.04, 2.53] & $<$0.001 \\
    Qwen & M-A/cat & 0.0491 & [0.0357, 0.0643] & 1.18 [0.80, 1.76] & $<$0.001 \\
    Qwen & M-A/room & 0.0226 & [0.0151, 0.0338] & 2.43 [1.54, 4.13] & 0.007 \\
    Qwen & R & 0.0581 & [0.0434, 0.0742] & 1.00 [1.00, 1.00] & $<$0.001 \\
    Qwen & R/cat & 0.0654 & [0.0522, 0.0789] & 0.89 [0.73, 1.06] & $<$0.001 \\
    Qwen & R/room & 0.0593 & [0.0449, 0.0749] & 0.98 [0.84, 1.14] & $<$0.001 \\
    Gemma & W & 0.0357 & [0.0297, 0.0416] & 1.63 [1.19, 2.15] & $<$0.001 \\
    Gemma & M-A & 0.0220 & [0.0119, 0.0341] & 2.56 [1.53, 5.12] & 0.002 \\
    Gemma & M-Bw & 0.0120 & [0.0067, 0.0198] & 4.42 [2.63, 9.34] & 0.042 \\
    Gemma & M-A/cat & 0.0534 & [0.0416, 0.0661] & 1.09 [0.76, 1.52] & $<$0.001 \\
    Gemma & M-A/room & 0.0114 & [0.0050, 0.0250] & 4.17 [2.20, 11.84] & 0.283 \\
    Gemma & R & 0.0581 & [0.0434, 0.0742] & 1.00 [1.00, 1.00] & $<$0.001 \\
    Gemma & R/cat & 0.0654 & [0.0522, 0.0789] & 0.89 [0.73, 1.06] & $<$0.001 \\
    Gemma & R/room & 0.0593 & [0.0449, 0.0749] & 0.98 [0.84, 1.14] & $<$0.001 \\
    \bottomrule
  \end{tabular}
\end{table}

%% file: tables/app-band.tex
\begin{table}[htbp]
  \centering
  \scriptsize
  \setlength{\tabcolsep}{4pt}
  \caption{Band-width sweep for the wall rule (Sec.~\ref{sec:wallbound}). $c$ is the fraction of floor area the band covers, $p$ the fraction of object centers inside it, both averaged over rooms; the prediction is $(c-p)/2$ from Equation~\eqref{eq:region}, which holds exactly per room because tied cells are opened in random order.}
  \label{tab:app-band}
  \begin{tabular}{lrrrrrrr}
    \toprule
    Room & Band (m) & $c$ & $p$ & $p-c$ & Predicted & Measured & 95\% CI \\
    \midrule
    Bathroom & 0.10 & 0.272 & 0.012 & $-$0.260 & 0.130 & 0.130 & [0.123, 0.136] \\
    Bathroom & 0.15 & 0.329 & 0.031 & $-$0.298 & 0.149 & 0.149 & [0.138, 0.159] \\
    Bathroom & 0.20 & 0.402 & 0.091 & $-$0.311 & 0.156 & 0.156 & [0.138, 0.172] \\
    Bathroom & 0.30 & 0.516 & 0.481 & $-$0.036 & 0.018 & 0.018 & [$-$0.010, 0.046] \\
    Bathroom & 0.50 & 0.708 & 0.871 & 0.163 & $-$0.081 & $-$0.081 & [$-$0.097, $-$0.065] \\
    Bedroom & 0.10 & 0.240 & 0.077 & $-$0.164 & 0.082 & 0.082 & [0.074, 0.089] \\
    Bedroom & 0.15 & 0.263 & 0.122 & $-$0.140 & 0.070 & 0.070 & [0.060, 0.080] \\
    Bedroom & 0.20 & 0.312 & 0.242 & $-$0.070 & 0.035 & 0.035 & [0.023, 0.047] \\
    Bedroom & 0.30 & 0.384 & 0.347 & $-$0.037 & 0.018 & 0.018 & [0.004, 0.033] \\
    Bedroom & 0.50 & 0.534 & 0.577 & 0.043 & $-$0.021 & $-$0.021 & [$-$0.034, $-$0.009] \\
    Office & 0.10 & 0.247 & 0.077 & $-$0.170 & 0.085 & 0.085 & [0.077, 0.093] \\
    Office & 0.15 & 0.273 & 0.118 & $-$0.155 & 0.077 & 0.077 & [0.067, 0.087] \\
    Office & 0.20 & 0.315 & 0.203 & $-$0.112 & 0.056 & 0.056 & [0.043, 0.068] \\
    Office & 0.30 & 0.386 & 0.292 & $-$0.094 & 0.047 & 0.047 & [0.033, 0.061] \\
    Office & 0.50 & 0.541 & 0.522 & $-$0.019 & 0.010 & 0.010 & [$-$0.005, 0.024] \\
    Living room & 0.10 & 0.236 & 0.048 & $-$0.187 & 0.094 & 0.094 & [0.084, 0.102] \\
    Living room & 0.15 & 0.238 & 0.052 & $-$0.186 & 0.093 & 0.093 & [0.083, 0.102] \\
    Living room & 0.20 & 0.264 & 0.071 & $-$0.193 & 0.097 & 0.097 & [0.084, 0.108] \\
    Living room & 0.30 & 0.331 & 0.148 & $-$0.183 & 0.091 & 0.091 & [0.074, 0.108] \\
    Living room & 0.50 & 0.466 & 0.381 & $-$0.086 & 0.043 & 0.043 & [0.018, 0.067] \\
    \bottomrule
  \end{tabular}
\end{table}

%% file: tables/app-catori.tex
\begin{table}[htbp]
  \centering
  \scriptsize
  \setlength{\tabcolsep}{2pt}
  \caption{Orientation profile per category (ScanNet, condition~A, categories with at least 60 real objects). For each category, the JSD (nats) between the model's wall distribution for that category and (i) the real distribution for the same category, (ii) the real distribution pooled over all categories; ``Closer'' says which of the two is nearer. The third column is the JSD between the category's own real profile and the pooled real profile, which bounds what there is to track. Summary: the model is closer to its own category for 3 of 16 categories (Qwen) and 4 of 15 (Gemma). Appendix~\ref{app:catori} gives the procedure.}
  \label{tab:app-catori}
  \begin{tabular}{lrrrrrlrrrl}
    \toprule
    Category & $n$ real & Own vs pooled & Qwen $n$ & vs own & vs pooled & Closer & Gemma $n$ & vs own & vs pooled & Closer \\
    \midrule
    chair & 1,000 & 0.0006 & 800 & 0.0099 & 0.0082 & pooled & 688 & 0.0083 & 0.0092 & own \\
    cabinet & 591 & 0.0022 & 603 & 0.0020 & 0.0043 & own & 621 & 0.0091 & 0.0182 & own \\
    desk & 437 & 0.0066 & 296 & 0.0161 & 0.0057 & pooled & 410 & 0.0320 & 0.0401 & own \\
    table & 360 & 0.0025 & 412 & 0.0094 & 0.0032 & pooled & 314 & 0.0198 & 0.0130 & pooled \\
    bed & 307 & 0.0109 & 209 & 0.0348 & 0.0213 & pooled & 200 & 0.2196 & 0.2058 & pooled \\
    toilet & 210 & 0.0126 & 230 & 0.0233 & 0.0036 & pooled & 200 & 0.0829 & 0.0780 & pooled \\
    box & 192 & 0.0171 & 562 & 0.0068 & 0.0071 & own & --- & --- & --- & --- \\
    shelves & 181 & 0.0117 & 222 & 0.0071 & 0.0028 & pooled & 235 & 0.1086 & 0.0727 & pooled \\
    dresser & 175 & 0.0035 & 201 & 0.0038 & 0.0023 & pooled & 200 & 0.0560 & 0.0534 & pooled \\
    sofa & 172 & 0.0020 & 296 & 0.0129 & 0.0076 & pooled & 232 & 0.0803 & 0.0661 & pooled \\
    night stand & 168 & 0.0143 & 265 & 0.0239 & 0.0028 & pooled & 215 & 0.1516 & 0.1148 & pooled \\
    bookshelf & 137 & 0.0155 & 524 & 0.0234 & 0.0035 & pooled & 573 & 0.0208 & 0.0417 & own \\
    bathtub & 122 & 0.0006 & 195 & 0.0069 & 0.0061 & pooled & 200 & 0.2055 & 0.2046 & pooled \\
    clothes & 105 & 0.0159 & 248 & 0.0109 & 0.0029 & pooled & 60 & 0.0800 & 0.0264 & pooled \\
    bag & 98 & 0.0140 & 437 & 0.0084 & 0.0144 & own & 120 & 0.1348 & 0.0746 & pooled \\
    towel & 62 & 0.0161 & 214 & 0.0170 & 0.0006 & pooled & 201 & 0.1916 & 0.1298 & pooled \\
    refridgerator & 61 & 0.0154 & --- & --- & --- & --- & --- & --- & --- & --- \\
    \bottomrule
  \end{tabular}
\end{table}

%% file: tables/app-mixture.tex
\begin{table}[htbp]
  \centering
  \footnotesize
  \setlength{\tabcolsep}{4pt}
  \caption{Wall-proximity excess of ScanNet living rooms by subset (Sec.~\ref{sec:limitations}). The subsets split the 221 living rooms by sofa count, chair count and floor area (median 26.9~m$^2$); ``domestic'' is their intersection. The last two rows give the regression slope of Appendix~\ref{app:slope} when the living-room point is taken from the full set and from the domestic subset.}
  \label{tab:app-mixture}
  \begin{tabular}{lrrrr}
    \toprule
    Subset & Rooms & Excess & 95\% CI & Median area \\
    \midrule
    all living rooms & 221 & $-$0.163 & [$-$0.188, $-$0.137] & 26.9 \\
    at most one sofa & 160 & $-$0.176 & [$-$0.205, $-$0.146] & 26.1 \\
    two or more sofas & 61 & $-$0.139 & [$-$0.186, $-$0.092] & 33.9 \\
    at most three chairs & 153 & $-$0.153 & [$-$0.185, $-$0.122] & 26.9 \\
    four or more chairs & 68 & $-$0.175 & [$-$0.216, $-$0.132] & 27.5 \\
    below median area & 110 & $-$0.156 & [$-$0.198, $-$0.114] & 20.5 \\
    median area or above & 111 & $-$0.167 & [$-$0.199, $-$0.135] & 40.0 \\
    domestic (all three) & 54 & $-$0.095 & [$-$0.161, $-$0.033] & 19.9 \\
    slopes, all (real SD 0.1363) &  & Qwen 0.029 & Gemma 0.227 &  \\
    slopes, domestic (real SD 0.1081) &  & Qwen 0.031 & Gemma 0.189 &  \\
    \bottomrule
  \end{tabular}
\end{table}

%% file: prompts.tex
\section{Prompts}
\label{app:prompts}

All prompts are reproduced verbatim from the implementation. Braces mark the
fields substituted per request: \{room\} is the room word
(\texttt{kitchen}, \texttt{bathroom}, \texttt{bedroom}, \texttt{office},
\texttt{living room}), and the vocabulary list shown is the ScanNet one; the
CubiCasa runs use their own \nVocabCubi-word list. Generation always uses
structured output against the JSON schema of
Appendix~\ref{app:schemas}, temperature $1.0$, and no top-$k$ or nucleus truncation: the server ignores the sampling parameters shipped with each model.

\paragraph{Condition A (survey).}
The one-shot condition. The model chooses the room's dimensions and must include a door.
\begin{quote}\ttfamily\footnotesize\raggedright
Describe the layout of a typical \{room\} as JSON. Decide the room's dimensions yourself, and include at least one door. Coordinates are in meters from the room's southwest corner. x runs east, y runs north. wall\_contact is the wall a fixture touches (N/S/E/W), or "none" if it does not touch any wall. stated\_probability is how likely you think it is that a real \{room\} contains this item, between 0 and 1. Use only these categories: bag, bathtub, bed, bookshelf, box, cabinet, chair, clothes, desk, dresser, floor mat, night stand, refrigerator, shelves, sofa, table, toilet, towel.
\end{quote}

\paragraph{Condition A$'$ (survey, wall field first).}
Identical to condition~A except that the sentence describing \texttt{wall\_contact} precedes the sentence describing the coordinates, matching the reordered schema.
\begin{quote}\ttfamily\footnotesize\raggedright
Describe the layout of a typical \{room\} as JSON. Decide the room's dimensions yourself, and include at least one door. wall\_contact is the wall a fixture touches (N/S/E/W), or "none" if it does not touch any wall. Coordinates are in meters from the room's southwest corner. x runs east, y runs north. stated\_probability is how likely you think it is that a real \{room\} contains this item, between 0 and 1. Use only these categories: bag, bathtub, bed, bookshelf, box, cabinet, chair, clothes, desk, dresser, floor mat, night stand, refrigerator, shelves, sofa, table, toilet, towel.
\end{quote}

\paragraph{Condition B-neutral (route, first step).}
Preceded by a first call that fixes the room geometry (below). The model then emits one object per turn.
\begin{quote}\ttfamily\footnotesize\raggedright
You are standing in the doorway of this \{room\}, looking in:
\{geom\}
Walk through the room. Report the fixtures you come across **one at a time, in the order you reach them**. Coordinates are in meters from the room's southwest corner. x runs east, y runs north. wall\_contact is the wall a fixture touches (N/S/E/W), or "none" if it does not touch any wall. stated\_probability is how likely you think it is that a real \{room\} contains this item, between 0 and 1. Use only these categories: bag, bathtub, bed, bookshelf, box, cabinet, chair, clothes, desk, dresser, floor mat, night stand, refrigerator, shelves, sofa, table, toilet, towel.
Reply with the first fixture you reach.
\end{quote}

\paragraph{Condition B-neutral (subsequent steps).}
\begin{quote}\ttfamily\footnotesize\raggedright
What do you reach next? If there is nothing left, set done to true.
\end{quote}

\paragraph{Condition B-walls (route, first step).}
The original route prompt. \textbf{It names the walls and a direction of travel}, which is why the neutral variant exists.
\begin{quote}\ttfamily\footnotesize\raggedright
You are standing in the doorway of this \{room\}, looking in:
\{geom\}
Walk along the walls, starting to your left and going clockwise. Report the fixtures you meet **one at a time, in the order you reach them**. Coordinates are in meters from the room's southwest corner. x runs east, y runs north. wall\_contact is the wall a fixture touches (N/S/E/W), or "none" if it does not touch any wall. stated\_probability is how likely you think it is that a real \{room\} contains this item, between 0 and 1. Use only these categories: bag, bathtub, bed, bookshelf, box, cabinet, chair, clothes, desk, dresser, floor mat, night stand, refrigerator, shelves, sofa, table, toilet, towel.
Reply with the first fixture you reach.
\end{quote}

\paragraph{Condition B-walls (subsequent steps).}
\begin{quote}\ttfamily\footnotesize\raggedright
What do you reach next? If you have come back to the doorway and there is nothing left, set done to true and object to null.
\end{quote}

\paragraph{Route conditions, geometry step.}
Both route conditions first ask for the room rectangle and the door, then walk through the room in a second call.
\begin{quote}\ttfamily\footnotesize\raggedright
Think of a typical \{room\}. Decide its dimensions and where its door is, as JSON. Coordinates are in meters from the room's southwest corner. x runs east, y runs north. Use only these categories: bag, bathtub, bed, bookshelf, box, cabinet, chair, clothes, desk, dresser, floor mat, night stand, refrigerator, shelves, sofa, table, toilet, towel.
\end{quote}

\paragraph{Condition C (conditioned survey).}
The room rectangle and openings are given; only the objects are generated.
\begin{quote}\ttfamily\footnotesize\raggedright
Here is the shape of a real \{room\} and where its openings are:
\{geom\}
Place the fixtures you would expect in this room, as JSON. Keep the room dimensions and openings exactly as given. Coordinates are in meters from the room's southwest corner. x runs east, y runs north. wall\_contact is the wall a fixture touches (N/S/E/W), or "none" if it does not touch any wall. stated\_probability is how likely you think it is that a real \{room\} contains this item, between 0 and 1. Use only these categories: bag, bathtub, bed, bookshelf, box, cabinet, chair, clothes, desk, dresser, floor mat, night stand, refrigerator, shelves, sofa, table, toilet, towel.
\end{quote}

\paragraph{Condition E (sequential, no walking).}
Sequential like the route conditions but without the narrative of walking.
\begin{quote}\ttfamily\footnotesize\raggedright
Here is a \{room\}:
\{geom\}
Place the fixtures you would expect in it, **one at a time**. Coordinates are in meters from the room's southwest corner. x runs east, y runs north. wall\_contact is the wall a fixture touches (N/S/E/W), or "none" if it does not touch any wall. stated\_probability is how likely you think it is that a real \{room\} contains this item, between 0 and 1. Use only these categories: bag, bathtub, bed, bookshelf, box, cabinet, chair, clothes, desk, dresser, floor mat, night stand, refrigerator, shelves, sofa, table, toilet, towel.
Reply with the first fixture.
\end{quote}

\paragraph{Condition F (survey with wall wording).}
One-shot generation with the wall wording of B-walls added, completing the $2 \times 2$ of \{wall wording\} $\times$ \{sequential\}.
\begin{quote}\ttfamily\footnotesize\raggedright
Describe the layout of a typical \{room\} as JSON. Decide the room's dimensions yourself, and include at least one door. Go along the walls, starting from the door and moving clockwise, and place the fixtures you would expect. Coordinates are in meters from the room's southwest corner. x runs east, y runs north. wall\_contact is the wall a fixture touches (N/S/E/W), or "none" if it does not touch any wall. stated\_probability is how likely you think it is that a real \{room\} contains this item, between 0 and 1. Use only these categories: bag, bathtub, bed, bookshelf, box, cabinet, chair, clothes, desk, dresser, floor mat, night stand, refrigerator, shelves, sofa, table, toilet, towel.
\end{quote}

\paragraph{Condition A$''$ (coordinates only).}
The room, the object list, each footprint and the wall each object must touch are supplied from the real layout; the model emits centers only. The item list has one line per object, of the form \texttt{bed (2.03 x 1.52 m) -- N wall}.
\begin{quote}\ttfamily\footnotesize\raggedright
Here is a \{room\}, \{W\} x \{D\} m, with a door on the south wall at x = \{door\_x\} m.
Place each of the following fixtures. Each fixture must touch the wall given for it; a fixture marked "none" touches no wall.

\{items\}

Output the centre of each fixture, in the same order, as JSON. Coordinates are in meters from the room's southwest corner. x runs east, y runs north.
\end{quote}

\paragraph{Condition A$''$-c (containment given priority).}
Condition A$''$ with two sentences appended.
\begin{quote}\ttfamily\footnotesize\raggedright
Here is a \{room\}, \{W\} x \{D\} m, with a door on the south wall at x = \{door\_x\} m.
Place each of the following fixtures. Each fixture must touch the wall given for it; a fixture marked "none" touches no wall.

\{items\}

Output the centre of each fixture, in the same order, as JSON. Coordinates are in meters from the room's southwest corner. x runs east, y runs north. Each fixture must touch the wall given AND lie entirely within the room. If both cannot be satisfied, keep the fixture inside the room.
\end{quote}

\paragraph{Condition A$''$-r (wall given priority).}
The same two sentences with the priority reversed.
\begin{quote}\ttfamily\footnotesize\raggedright
Here is a \{room\}, \{W\} x \{D\} m, with a door on the south wall at x = \{door\_x\} m.
Place each of the following fixtures. Each fixture must touch the wall given for it; a fixture marked "none" touches no wall.

\{items\}

Output the centre of each fixture, in the same order, as JSON. Coordinates are in meters from the room's southwest corner. x runs east, y runs north. Each fixture must touch the wall given AND lie entirely within the room. If both cannot be satisfied, keep the fixture touching its wall.
\end{quote}

\paragraph{Condition A$''$-i (containment only).}
The wall requirement is removed entirely, and the item list carries no wall: \texttt{bed (2.03 x 1.52 m)}.
\begin{quote}\ttfamily\footnotesize\raggedright
Here is a \{room\}, \{W\} x \{D\} m, with a door on the south wall at x = \{door\_x\} m.
Place each of the following fixtures. Each fixture must lie entirely within the room.

\{items\}

Output the centre of each fixture, in the same order, as JSON. Coordinates are in meters from the room's southwest corner. x runs east, y runs north.
\end{quote}

\paragraph{Conditions O-anchor and O-move (observed object).}
Condition~C with one line listing an object already in the room, used in Sec.~\ref{sec:observed}. O-anchor lists the anchor at its real position, O-move at a position on another wall. The partner is never listed.
\begin{quote}\ttfamily\footnotesize\raggedright
Here is the shape of a real \{room\} and where its door is:
\{geom\}
The following furniture is already in the room. Do not move it and do not list it again:
  desk at (1.20, 3.40) m, 1.50 x 0.70 m
Place the furniture you would expect in this room, as JSON. Keep the room dimensions and the door exactly as given. Coordinates are in meters from the room's southwest corner. x runs east, y runs north. wall\_contact is the wall a fixture touches (N/S/E/W), or "none" if it does not touch any wall. stated\_probability is how likely you think it is that a real \{room\} contains this item, between 0 and 1. Use only these categories: bag, bathtub, bed, bookshelf, box, cabinet, chair, clothes, desk, dresser, floor mat, night stand, refrigerator, shelves, sofa, table, toilet, towel.
\end{quote}

\paragraph{Condition A$''$-per (containment per object).}
Condition A$''$ with the containment requirement attached to every item line,
which then reads \texttt{bed (2.03 x 1.52 m) -- N wall, entirely inside the
room}.

%% file: sections/appendix_schemas.tex
\section{Output schemas}
\label{app:schemas}

Generation uses structured output; the model is constrained to the JSON schema
below. \textbf{Field order matters}: the model emits the properties in the
order given, and condition~\condAp{} differs from condition~\condA{} only in
that order.

\paragraph{Object (condition~\condA).}
\begin{quote}\ttfamily\footnotesize\raggedright
category (enum over the closed vocabulary), center\_m \{x, y\},
size\_m \{w, d\}, wall\_contact (enum N/S/E/W/none),
stated\_probability (number)
\end{quote}

\paragraph{Object (condition~\condAp).}
\begin{quote}\ttfamily\footnotesize\raggedright
category, \textbf{wall\_contact}, center\_m, size\_m, stated\_probability
\end{quote}

\paragraph{Layout.}
\texttt{room\_type}, \texttt{extent\_m \{width, depth\}}, \texttt{openings}
(at least one, each \texttt{\{kind, wall, offset\_frac, width\_m\}}), and
\texttt{objects} (between 2 and 14). Under the conditioned variants the
rectangle and openings are supplied rather than generated and are omitted from
the schema.

\paragraph{Route conditions.}
Each step returns the fields of one object and a boolean \texttt{done}, which terminates the walk. The object is not nullable, because the server's grammar compilation of a schema with \texttt{anyOf} and a nullable enum did not terminate reliably.

\paragraph{Condition~\condApp.}
\texttt{objects}, an array of exactly $n$ entries, each
\texttt{\{category, center\_m \{x, y\}\}}. Footprints and walls are given in
the prompt and are not generated.

\paragraph{On ``none''.}
The enum for \texttt{wall\_contact} contains the string \texttt{"none"} rather than JSON \texttt{null}, and the prompts of Appendix~\ref{app:prompts} describe an object that touches no wall as \texttt{"none"}, so that the prompt and the schema use the same value.